\documentclass{article} %
\newif\ifpreprint
\preprinttrue
\ifdefined\preprintmode\preprinttrue\fi
\ifpreprint
\usepackage[preprint]{neurips_2026}
\setcitestyle{round} %
\newif\ificlrfinal\iclrfinaltrue
\else
\usepackage{iclr2027_conference,times}
\fi

\usepackage{microtype}
\usepackage{graphicx}
\usepackage{subcaption}
\usepackage{appendix}
\usepackage[dvipsnames]{xcolor} %
\usepackage{algorithm}
\usepackage{algorithmic}
\usepackage{inconsolata}
\usepackage{xspace}     %

\usepackage{booktabs}   %
\usepackage{multirow}   %
\usepackage{placeins}  %
\usepackage{needspace} %

\newcommand{\apptablesize}{\fontsize{8pt}{9.5pt}\selectfont}
\newcommand{\apptable}{\apptablesize\setlength{\tabcolsep}{3pt}}
\newcommand{\apptablewide}{\apptablesize\setlength{\tabcolsep}{2pt}}

\usepackage{titletoc}   %
\usepackage{hyperref}
\hypersetup{
  colorlinks=true,          %
  allcolors={RoyalPurple},
  pdftitle={Prior-Amortized In-Context Bayesian Inference for Generalized Linear Mixed-Effects Models},
}
\ificlrfinal\hypersetup{pdfauthor={Alex Kipnis, Marcel Binz, Eric Schulz}}\fi
\newcommand{\myfigref}[1]{\hyperref[#1]{Figure~\ref*{#1}}}
\newcommand{\mytabref}[1]{\hyperref[#1]{Table~\ref*{#1}}}
\newcommand{\mysecref}[1]{\hyperref[#1]{Section~\ref*{#1}}}
\newcommand{\myappref}[1]{\hyperref[#1]{Appendix~\ref*{#1}}}
\newcommand{\myeqref}[1]{\hyperref[#1]{Equation~\ref*{#1}}}
\newcommand{\mb}{\texttt{MetaBeta}\xspace} %
\newcommand{\MBz}{\ensuremath{\texttt{MB}^{0}}}  %
\newcommand{\rev}[1]{\ificlrfinal#1\else\textcolor{ForestGreen}{#1}\fi}

\usepackage{amsmath}
\usepackage{amssymb}
\usepackage{mathtools}
\usepackage{amsthm}
\usepackage{bbm}

\theoremstyle{plain}

\theoremstyle{definition}

\theoremstyle{remark}

\ifpreprint
\title{Prior-Amortized In-Context Bayesian Inference\\for Generalized Linear Mixed-Effects Models}
\author{
   Alex Kipnis\textsuperscript{1}\thanks{
      Correspondence to \href{mailto:adkipnis@mailbox.org}{adkipnis@mailbox.org};
      $\dagger$ These authors jointly directed this project.
   }
   \And Marcel Binz\textsuperscript{1 $\dagger$}
   \And Eric Schulz\textsuperscript{1 $\dagger$}
  \AND
  \textsuperscript{1}Institute for Human-Centered AI, Helmholtz Munich, Germany
}
\else
\title{Prior-Amortized In-Context Bayesian\\Inference for Generalized Linear\\Mixed-Effects Models}
\author{Alex Kipnis\thanks{Correspondence to \href{mailto:adkipnis@mailbox.org}{adkipnis@mailbox.org}.}%
   , Marcel Binz\textsuperscript{$\dagger$} \& Eric Schulz\thanks{These authors jointly directed this project.} \\
   Institute for Human-Centered AI, Helmholtz Munich, Germany
}
\fi

\begin{document}
\maketitle
\begin{abstract}
Hierarchical data is ubiquitous in the empirical sciences and most commonly analyzed with generalized linear mixed-effects models (GLMMs).
Bayesian inference for GLMMs yields calibrated uncertainty estimates but requires Markov Chain Monte Carlo (MCMC)\rev{. The} gold-standard No-U-Turn Sampler (NUTS) is slow and must restart from scratch for every new dataset, model and prior specification.
We introduce \mb, a pretrained \rev{amortized model for} in-context Bayesian inference over GLMMs. \rev{The full prior and model specification are supplied as context at test time, and the model returns an asymptotically exact posterior at the speed of a forward pass.} Two set transformers and two conditional neural spline flows mirror the GLMM's hierarchical structure (global parameters shared across groups, local parameters per group) and are trained jointly on millions of realistic simulated datasets with continuous, binary and count outcomes.
The flow posterior is refined by Independence Metropolis--Hastings against the unnormalized posterior\rev{. This yields} tuning-free asymptotically correct inference.
Alternatively, \mb \rev{can warm-start NUTS. This gives} nearly identical inference $3.5$--$13\times$ faster than \rev{cold-started NUTS and removes} over $98\%$ of NUTS' divergent transitions.
On controlled benchmarks with ground-truth parameters, \mb matches NUTS in parameter recovery, posterior calibration and out-of-sample prediction\rev{. On} real datasets with unknown parameters, priors and model structure, its posteriors closely agree with those of NUTS\rev{. Furthermore,} it remains faithful where amortized models typically struggle: under misspecified likelihoods, misspecified priors, out-of-distribution predictors, collinear designs and data-poor regimes.
On real data, a full prior sensitivity analysis including model comparison takes a few seconds with \mb against $14$--$18$ core-hours for NUTS, tracing its posteriors throughout.
Our model is open-source and open-weights and thus immediately deployable.
\end{abstract}

\section{Introduction}
\label{sec:intro}
Hierarchical structure is a defining feature of many real-world tabular datasets, from longitudinal clinical trials with site-specific effects to ecological studies where the same intervention performs differently across locations.
Generalized linear mixed-effects models (GLMMs) provide the standard framework for interpretable inference on such multi-level data, and are widely adopted across the empirical sciences \citep{Gelman.2007, Harrison.2018, Gordon.2019, Yu.2022}.
Popular libraries such as \texttt{lme4}, \texttt{nlme}, and \texttt{brms} \citep{Bates.2015, Pinheiro.1999, Burkner.2018} are downloaded nearly 10 million times per year\footnote{CRAN download counts for \texttt{nlme}, \texttt{lme4} and \texttt{brms};
retrieved 2026-09-25 via \href{https://cranlogs.r-pkg.org}{\texttt{cranlogs.r-pkg.org}}.}, and GLMM analyses appear in about half of recent quantitative empirical papers (survey over $1{,}133$ papers in \myappref{app:sur}).

In many such applications, Bayesian inference would be the method of choice: it yields full posterior uncertainty over interpretable parameters and enables domain knowledge to enter inference through priors \citep{Figueroa-Zuniga.2013, Gelman.2013}.
However, the GLMM posterior over fixed effects, per-group random effects and their covariance is intractable, grows in dimension with the number of groups and resists simple Gaussian approximations.
Markov Chain Monte Carlo \citep[MCMC;][]{Metropolis.1953, Hastings.1970, Gelfand.1990} is the standard numerical approximation, with the No-U-Turn Sampler \citep[NUTS;][]{Hoffman.2014, Burkner.2018, Capretto.2022} being the current gold standard for GLMMs.
However, NUTS is poorly suited to workflows with repeated inference across datasets, model and prior specifications, for three compounding reasons.
(1) Its wall time is governed by posterior geometry and cannot be bounded in advance \citep{Betancourt.2013, Hoffman.2021, Hoffman.2022}, complicating integration into latency-sensitive workflows.
(2) Correct use demands monitoring of chain-mixing diagnostics (cross-chain variance ratios, effective sample sizes, divergent transitions), which require additional statistical expertise \citep{Roy.2020, Vehtari.2021}.
(3) It provides no amortization:
every dataset, every prior change, and every model variant requires restarting the sampler from scratch \citep{Papaspiliopoulos.2007, Betancourt.2013, Gelman.2026}.
\rev{This makes simulation studies, power analyses and prior sensitivity analyses prohibitively slow \citep{Kumle.2021, Kallioinen.2024, Zhang.2022}.
In practice, researchers report trimming such analyses due to the costs of NUTS, by running fewer replications \citep{Oberauer.2022}, fewer simulated datasets \citep{Vasishth.2023}, or only checking a single alternative prior \citep{Wang.2021}.
Some even report that they had to simplify their main analysis by fitting their models to a subsample of the data \citep{Schrimpf.2021, Arslan.2017} or by dropping a random effect \citep{Wang.2021}.}

In this work, we introduce \mb, a pretrained model for in-context Bayesian inference over GLMMs.
Trained once on millions of realistic hierarchical datasets, it \rev{takes a new dataset together with its model and prior specification as context and returns full posteriors over all GLMM parameters at the speed of a forward pass. A built-in MCMC step then makes these posteriors asymptotically exact.}
Our key contributions are:
\begin{enumerate}
\ifpreprint\setlength{\itemsep}{0pt}\fi
   \item \rev{\textbf{One pretrained network for the GLMM class}:
   the model structure (number of fixed effects, random effects, groups and observations per group) and the complete prior specification (families and hyperparameters of the fixed-effect priors, of the random-effect scales and of their correlation prior) are inputs at inference time rather than fixed at training time. Two conditional flows return a joint posterior whose dimension follows the model structure: one trained network serves every GLMM within a verified envelope,} enabling zero-shot prior sensitivity analyses \rev{(\mysecref{sec:usecase}), simulation-based power analyses (\myappref{app:power})} and inference over many datasets in a single batched pass.
\item \textbf{Exactness of MCMC \rev{from an amortized proposal}}: the flow's tractable density serves as a proposal rather than the final answer. By default, \mb refines it by Independence Metropolis--Hastings \citep[IMH;][]{Tierney.1994} against the unnormalized posterior\rev{, so every dataset receives unweighted, asymptotically exact draws}. IMH acceptance rate and importance weight diagnostics flag unreliable posteriors per dataset, and one further importance-sampling pass yields an unbiased marginal-likelihood estimate for Bayes factors. Alternatively, the flow warm-starts NUTS, cutting its wall time $3.5$--$13\times$ and removing over $98\%$ of its divergent transitions.
\item \textbf{A realistic tabular data prior for hierarchical regression}: a simulator anchored in applied practice. A survey of $1{,}133$ publications sets the ranges of model structure and likelihoods. Predictors come from freshly sampled structural causal models with mixed-type features and from $596$ real tabular datasets. Group sizes and signal-to-noise levels mirror real hierarchical data. The simulator and the survey are released with the code.
\item \textbf{A verified operating envelope}: on oracle, real-data and stress benchmarks, \mb matches the posteriors of NUTS at a small fraction of its runtime, and it matches or outperforms the standard fast alternatives, Automatic Differentiation Variational Inference \citep[ADVI;][]{Kucukelbir.2017} and Pathfinder \citep{Zhang.2022}. The stress tests cover misspecified likelihoods and priors, out-of-distribution predictors, collinear designs and data-poor groups, so users know where the model can be trusted.
\end{enumerate}
All code is implemented in \texttt{PyTorch} \citep{Paszke.2019a}. Source code and weights are available on
\begin{center}
\ifpreprint\vspace{-0.3cm}\fi
\ificlrfinal
\href{https://github.com/adkipnis/metabeta/}{\texttt{github.com/adkipnis/metabeta}}
\else
\href{https://anonymous.4open.science/r/metabeta}{\texttt{anonymous.4open.science/r/metabeta}}
\fi
\end{center}

\subsection{Related Work}

\textbf{Amortized simulation-based inference.}
Neural Posterior Estimation (NPE) approximates Bayesian posteriors with a neural network \citep{Papamakarios.2016, Lueckmann.2017, Greenberg.2019, Cranmer.2020}.
\rev{The original BayesFlow method \citep{Radev.2022} pairs a summary network \citep{Zaheer.2017, Lee.2019} with a conditional normalizing flow \citep{Rezende.2015, Papamakarios.2021} trained by simulation under a fixed prior}\footnote{The BayesFlow library has since grown into a general framework for simulation-based inference \citep{Kuhmichel.2026}.}.
Following \citet{Rodrigues.2021}, NPE was adapted to hierarchical models \citep{Habermann.2025} and multi-level pharmacology models \citep{Arruda.2024}. Both fix the prior at training time \rev{and train one network per model. In contrast, \mb is released pretrained and optimized for GLMMs.
Standard NPE also fixes the dimension of the parameter vector at training time. Amortization over parameters of variable dimension has been reported for non-hierarchical regression \citep{Mittal.2025} and across cognitive model families \citep{Huang.2026}. \mb's flows apply it to the hierarchical case, where the numbers of fixed and random effects vary with the model (\myappref{app:nf}).}

\textbf{Prior-conditioned amortized inference.}
\rev{Test-time prior conditioning takes several forms.
Sensitivity-aware ABI \citep{Elsemuller.2024} feeds power-scaling exponents of a fixed prior
as context. Distribution Transformers \citep{Whittle.2026} map any prior embedded as a Gaussian mixture to a Gaussian-mixture posterior.
ACE \citep[][]{Chang.2025} takes a fixed set of histogram priors as context.
PriorGuide \citep{Yang.2026} steers a trained diffusion sampler toward a new prior at sampling time.
In \mb, each parameter's prior enters as a family indicator together with continuous hyperparameters. Unlike in previous work, these prior inputs shape a hierarchical posterior with per-group random effects and are combined with a flexible model structure as context.}

\textbf{Transformer-based amortized inference.}
TabPFN \citep{Muller.2021, Hollmann.2025} uses transformers \citep{Vaswani.2017} for tabular predictive distributions rather than posteriors over interpretable parameters.
\rev{The Simformer \citep{Gloeckler.2024} and OneFlowSBI \citep{Nautiyal.2026} amortize over conditioning queries of a single simulator under a fixed prior. In-context estimators also retrain per prior \citep{Mittal.2025, Reuter.2025}.}
CogFormer \citep{Huang.2026} amortizes across model families and design configurations of sequential-sampling models, but under fixed priors and for non-hierarchical models.
None of these expresses the hierarchical structure of GLMMs: nested exchangeability, structured random effects, and likelihood heterogeneity across varying numbers of groups and predictors.

\textbf{Amortized inference meets MCMC.}
Normalizing flows have served as reparameterizations or proposals for MCMC \citep{Hoffman.2019a, Gabrie.2022, Grenioux.2023}, fitted to one target and refitted for the next.
\rev{\cite{Li.2026} combines an amortized posterior with Pareto-smoothed importance sampling and many-chain HMC.
\mb instead keeps the amortized flow inside the transition kernel as the proposal of an Independence Metropolis--Hastings chain, so every dataset receives unweighted, asymptotically exact draws with the acceptance rate as the quality metric.
Alternatively, its draws seed the chains and set the diagonal mass matrix of NUTS, reaching the same NUTS posterior with half the tuning budget and almost none of the divergent transitions.}

\section{Methods}

\subsection{The Inference Problem} \label{sec:gen}
Consider $m$ groups, each contributing $n_i$ observations $(\mathbf{X}_i, \mathbf{y}_i)$.
GLMMs link predictors $\mathbf{X}_i$ to outcomes $\mathbf{y}_i$ through \textbf{fixed effects} $\boldsymbol{\beta} \in \mathbb{R}^d$ (global, shared across groups) and \textbf{random effects} $\boldsymbol{\alpha}_i \in \mathbb{R}^q$ (local, specific to each group).
The outcome $y_{ij}$ follows an exponential-family distribution $\mathcal{P}_y$, whose expectation is a function of observations and latent parameters:
\begin{equation} \label{eq:lp}
    y_{ij} \mid \eta_{ij} \sim \mathcal{P}_y\!\left(g^{-1}(\eta_{ij}),\, \phi\right),
    \qquad
    \eta_{ij} = \mathbf{x}_{ij}^\top \boldsymbol{\beta} + \mathbf{z}_{ij}^\top \boldsymbol{\alpha}_i,
\end{equation}
where $\eta_{ij}$ is the linear predictor,
$\mathbf{z}_{ij}$ is a subset of $\mathbf{x}_{ij}$ \rev{determined by the model specification},
$g$ is a canonical link function, and $\phi$ is a dispersion parameter (the residual variance $\phi = \sigma_\varepsilon^2$ for the Gaussian family, and $\phi \equiv 1$ for Bernoulli and Poisson).
The random effects are viewed as samples from $\mathcal{N}_q(\mathbf{0}, \mathbf{S})$ with $\mathbf{S} = \boldsymbol{\sigma\sigma}^\top \odot \mathbf{R}$, where $\mathbf{R}$ is the correlation matrix of the random effects.%

We collect the global parameters as $\boldsymbol{\vartheta} = (\boldsymbol{\beta},
\boldsymbol{\sigma}, \mathbf R, \phi)$, and denote group data as
$\mathbf{D}_i = [\mathbf{y}_i, \mathbf{X}_i, \mathbf{Z}_i] \in \mathbb{R}^{n_i \times (1 + d + q)}.$
The joint posterior $p(\boldsymbol{\vartheta}, \boldsymbol{\alpha}_{1:m} \mid \mathbf{D})$ is analytically intractable.
Crucially, groups are conditionally independent given $\boldsymbol{\vartheta}$, so the posterior factorizes as
\begin{equation} \label{eq:factorization}
   p(\boldsymbol{\vartheta}, \boldsymbol{\alpha}_{1:m} \mid \mathbf{D}) =
    \underbrace{p(\boldsymbol{\vartheta} \mid \mathbf{D})}_{\text{global}}
    \prod_{i=1}^m
    \underbrace{p(\boldsymbol{\alpha}_i \mid \boldsymbol{\vartheta}, \mathbf{D}_i)}_{\text{local}}.
\end{equation}

\subsection{Model Architecture and Training} \label{sec:arc}
\begin{figure*}[h!]
	\centering
	\includegraphics[width=0.924\textwidth]{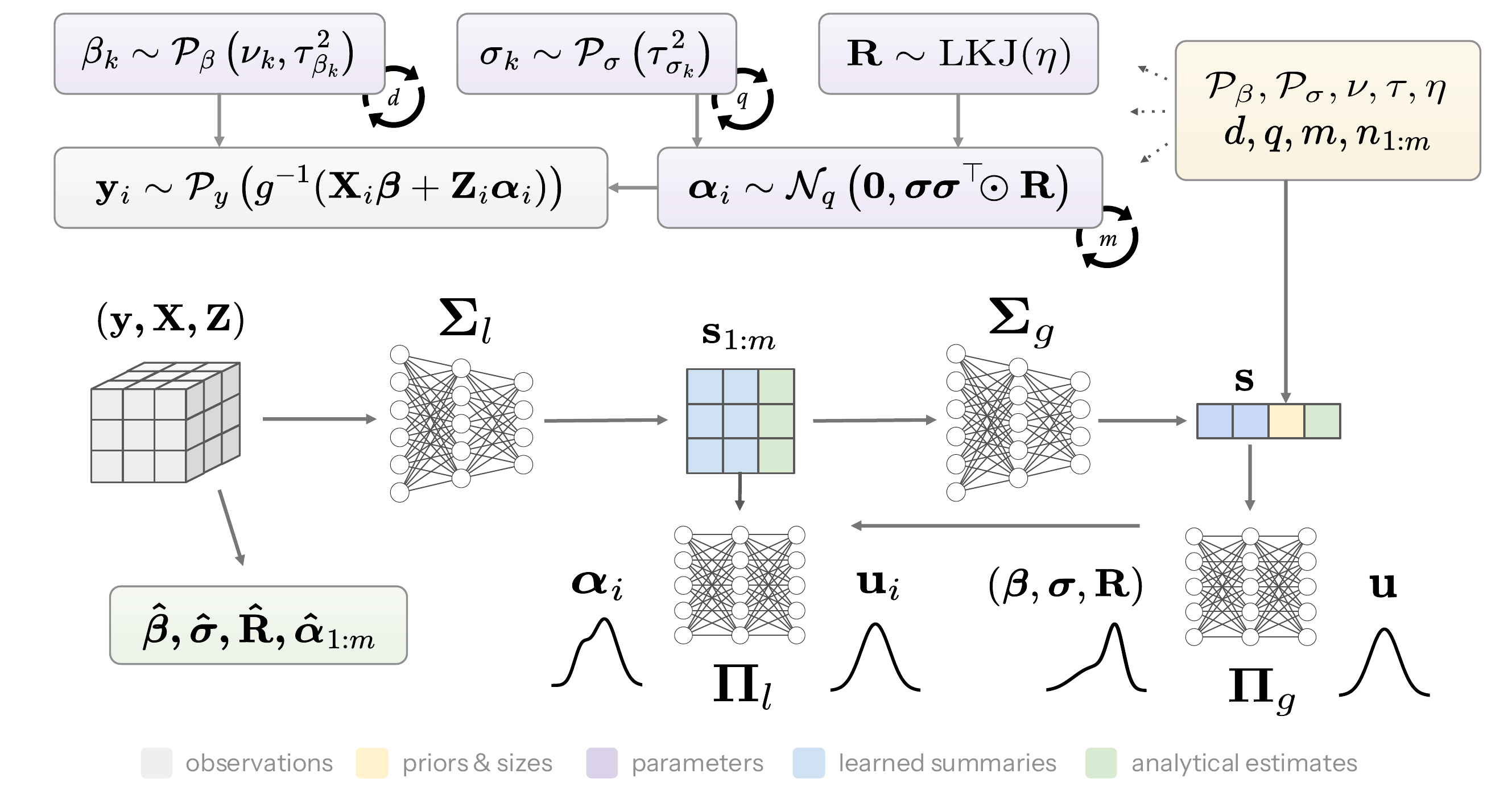}
	\caption{
    (Upper half) \textit{Dataset Simulation}.
    Structural dimensions and prior hyperparameters are sampled; regression parameters are drawn from the chosen priors and outcomes are generated via \myeqref{eq:lp}.
    Predictors come from structural causal models or real tabular datasets.
    (Lower half) \textit{Model Pipeline}.
    $\mathbf{D}_i = (\mathbf{y}_i, \mathbf{X}_i, \mathbf{Z}_i)$ are separately summarized per group $i$.
    Local summaries are pooled over groups. The resulting global summary and the prior specification condition $\boldsymbol\Pi_g \approx p( \boldsymbol \vartheta \mid \mathbf{D})$.
    Global posterior samples condition the local posterior $\boldsymbol\Pi_l \approx p(\boldsymbol{\alpha}_i \mid \boldsymbol{\vartheta}, \mathbf{D}_i)$.
    }
	\label{fig:1}
\end{figure*}

\textbf{Architecture overview.} Our model realizes the factorization in \myeqref{eq:factorization} through a four-stage pipeline (see \myfigref{fig:1}).
\textbf{(1) Local summarization.}
Each group's observations $[\mathbf{y}_i, \mathbf{X}_i, \mathbf{Z}_i]$ of variable length $n_i$ are compressed independently by a local summary network $\boldsymbol\Sigma_l$ into a fixed-size, row-permutation-invariant summary $\mathbf{s}_i \in \mathbb{R}^{h_l}$.
\textbf{(2) Global summarization.}
The local summaries $\mathbf{s}_{1:m}$ are stacked and processed by a global summary network $\boldsymbol\Sigma_g$ into a single dataset summary $\mathbf{s} \in \mathbb{R}^{h_g}$.
\textbf{(3) Global posterior.}
The summary $\mathbf{s}$ conditions a posterior network $\boldsymbol\Pi_g$, which efficiently produces
samples $\hat{\boldsymbol{\vartheta}} \sim p_{\boldsymbol\Pi_g}(\boldsymbol{\vartheta} \mid \mathbf{s})$
and evaluates the log-density for arbitrary $\boldsymbol \vartheta$.
\textbf{(4) Local posterior.}
For each group, the local summary $\mathbf{s}_i$ and global samples $\hat{\boldsymbol{\vartheta}}$ are passed to a shared local posterior network $\boldsymbol\Pi_l$,
which learns to sample $\hat{\boldsymbol{\alpha}}_i \sim p_{\boldsymbol\Pi_l}(\boldsymbol{\alpha}_i \mid
\hat{\boldsymbol{\vartheta}}, \mathbf{s}_i)$.
All networks are trained jointly (\myappref{app:arch}).

\textbf{Summary networks.} Both $n_i$ and $m$ vary across datasets, and observations within each level are exchangeable, so the summary networks must handle variable-length inputs and be permutation-invariant.
We use Set Transformers \citep{Lee.2019}, which aggregate variable-length, unordered inputs into a fixed-size vector via learned attention \rev{(\myappref{app:set})}.
A learned CLS token prepended to the input sequence serves as the pooled output.
The local summarizer pools over observations per group; the global summarizer then pools over group summaries.

\textbf{Summary augmentation.}
Before entering the posterior networks, each summary is augmented with the prior specification: each prior family (Normal, Student-$t$, half-Normal, half-Student-$t$, Exponential) is mapped to a one-hot embedding, and the corresponding hyperparameters are appended as continuous inputs to the global summary $\mathbf{s}$.
This makes the posterior networks prior-aware and enables zero-shot generalization to novel prior specifications (\myappref{app:prior_grid}).
In addition, we append cheap analytical summary statistics to each learned summary as an inductive head start, freeing capacity for posterior geometry, tails and parameter correlations (\myappref{app:suf}).

\textbf{Posterior networks.} We realize both posterior networks, $\boldsymbol\Pi_g$ and $\boldsymbol\Pi_l$, as conditional normalizing flows \rev{(\myappref{app:nf})}: learned invertible maps between a simple base distribution and an arbitrarily complex target, which enable both exact log-density evaluation and fast sampling \citep{Papamakarios.2021}.
We use coupling-based neural spline flows \citep{Durkan.2019} with residual multi-layer perceptron couplings \citep{He.2016}.
We condition the location and scale of the base distribution on the summary, so the flow need only model the residual around a data-dependent Gaussian.

\textbf{Training objective.} \label{sec:ler}
We train the network by minimizing the expected negative log-likelihood (NLL) of the neural posterior $p_{\boldsymbol \Pi}$ over simulated pairs $(\boldsymbol{\vartheta}, \mathbf{D})$, which is equivalent to minimizing the expected forward KL divergence to the true posterior, the mass-covering direction standard in NPE.
Applied to each factor in \myeqref{eq:factorization}, the loss reads
\begin{equation} \label{eq:loss}
   \ell(f_{\boldsymbol\Sigma}, p_{\boldsymbol\Pi}) = -\mathbb{E}_{\boldsymbol{\vartheta}, \boldsymbol{\alpha}, \mathbf{D}} \!\left[
       \log p_{\boldsymbol\Pi_g}\!\left(\boldsymbol{\vartheta} \mid
       f_{\boldsymbol\Sigma_g}([f_{\boldsymbol\Sigma_l}(\mathbf{D}_i)]_{i=1}^m)\right) +
       \sum_{i=1}^m \log p_{\boldsymbol\Pi_l}\!\left(\boldsymbol{\alpha}_i \mid
       \boldsymbol{\vartheta},\, f_{\boldsymbol\Sigma_l}(\mathbf{D}_i)\right)
    \right].
\end{equation}
At inference, the true $\boldsymbol{\vartheta}$ that conditions the local factor is replaced by global posterior samples.
\subsection{Amortized Inference meets MCMC} \label{sec:hyb}
The raw flow returns an approximate posterior in milliseconds but offers no exactness guarantee. NUTS, in contrast, is asymptotically exact but serial, non-amortized and restarted from scratch for every dataset.
Because the trained flow supplies fast samples and a tractable proposal density, it can be scored against the exact unnormalized posterior and coupled to MCMC in two ways.

\textbf{Independence Metropolis--Hastings}.
By default, \mb refines the flow posterior with Independence Metropolis--Hastings \citep[IMH;][]{Tierney.1994}.
The neural posterior $q(\boldsymbol\vartheta) \coloneqq p_{\boldsymbol\Pi_g}(\boldsymbol\vartheta \mid \mathbf{s})$ serves as a fixed proposal, and a candidate $\boldsymbol\vartheta'$ is accepted over the current state $\boldsymbol\vartheta$ with probability $\min(1,\, w(\boldsymbol\vartheta')/w(\boldsymbol\vartheta))$, where $w(\boldsymbol\vartheta) = \tilde p(\boldsymbol\vartheta \mid \mathbf{D})\,/\,q(\boldsymbol\vartheta)$ is the importance weight against the unnormalized posterior.
To keep the chain well-mixing, we collapse it onto the global parameters by marginalizing the random effects out of the target, so the chain explores the lower-dimensional global space alone (the same marginalization that stabilizes NUTS for latent Gaussian models; \citealp{Margossian.2020}).
This marginalization is analytical for Gaussian models; an unbiased marginal-likelihood estimate keeps it exact for Bernoulli and Poisson \citep{Andrieu.2009}.
The random effects are then reimputed for each retained global draw from the corresponding conditional $p(\boldsymbol\alpha_i \mid \boldsymbol\vartheta, \mathbf{D}_i)$ (Rao--Blackwellization).
The acceptance rate directly measures the flow--posterior overlap and doubles as a diagnostic: \mb automatically flags insufficient sampling budgets and suggests one projected to be adequate (\myappref{app:imh}).
The chains' effective sample size \citep[ESS;][]{Vehtari.2021} is an additional diagnostic: it flags chains that revisit relatively few distinct states, indicating that the refined posterior is likely too narrow.

\textbf{Warm-started NUTS.}
Where gradient-based sampling is preferred, \mb initializes the NUTS chains and their diagonal mass matrix from its draws \citep{Zhang.2022}, which halves the tuning budget and removes most divergent transitions at unchanged accuracy (\myappref{app:wn}).
\rev{Where the default pipeline raises a diagnostic flag, its warning recommends warm-started NUTS (\myappref{app:safe}).}

\subsection{Data Simulation} \label{sec:dat}
Our model is trained entirely on simulated data (\myfigref{fig:1}) whose structure, prior specifications and size ranges follow those reported in applied Bayesian mixed-effects analyses (\myappref{app:sur}--\ref{app:sizes}).

\textbf{Structural dimensions.} For each dataset, the number of fixed effects $d$, the number of random effects $q$, the number of groups $m$, and the per-group observation counts $n_{1:m}$ are drawn from a meta-prior that favors smaller values consistent with real-world datasets and usage.

\textbf{Priors and parameters.} Prior family and hyperparameters are sampled per dataset and passed as context to the network (\mysecref{sec:arc}).
Fixed-effect prior families are drawn from \{Normal, Student-$t$\}; scale parameter prior families from \{half-Normal, half-Student-$t$, Exponential\}.
Their hyperparameters are sampled from skewed distributions, with tighter ranges for Bernoulli and Poisson to prevent extreme linear predictors (\myappref{app:prior}).
The random-effect correlation matrix follows an LKJ prior; GLMM parameters are then drawn from these priors.

\textbf{Predictor sources.} Fixed-effect predictors $\mathbf{X}$ are drawn from two sources that balance structural diversity and empirical realism, mitigating simulation-to-real mismatch.
\textit{(i) Structural causal model generator}: predictor matrices are sampled from structural causal models implemented by sparse multi-layer perceptrons or directed acyclic graphs, with root causes propagated through a rich pool of nonlinear activations (including randomly sampled Gaussian Process kernels). Random post-hoc transforms then yield binary, ordinal, count and censored predictors. This follows the synthetic tabular prior of TabPFN \citep{Hollmann.2025} and TabICL \citep{Qu.2025}; details are in \myappref{app:scm}.
\textit{(ii) Real tabular data}: predictor matrices are subsampled from real tabular dataset collections (PMLB, \citealp{Romano.2022}; SRM, \citealp{Lichtenberg.2017}; UCI, \citealp{Dua.2019}; 596 datasets in total; see \myappref{app:emu}). A separate curated collection of 40 real hierarchical datasets, mostly from \texttt{R} mixed-effects packages (\href{https://CRAN.R-project.org/package=lme4}{\texttt{lme4}}, \href{https://CRAN.R-project.org/package=nlme}{\texttt{nlme}}, \href{https://CRAN.R-project.org/package=mlmRev}{\texttt{mlmRev}}, \href{https://CRAN.R-project.org/package=MEMSS}{\texttt{MEMSS}}) is held out for evaluation (\myappref{app:tes}).

\textbf{Outcomes.}
Outcomes are sampled from \myeqref{eq:lp}. Continuous outcomes are standardized to unit variance, and parameters and hyperparameters are rescaled analytically
(\myappref{app:sta}).

\textbf{Test sets.} Two complementary test set types (both built from the held-out real-data collection) probe orthogonal aspects of generalization:
\textit{Oracle test sets} pair real predictors with sampled parameters and outcomes. They enable direct evaluation of posterior quality given ground-truth parameters.
\textit{Real-world test sets} do not sample parameters and keep original outcomes intact.
As the true parameters, priors and model structure are unknown, we measure agreement with NUTS posteriors, i.e.\ whether \mb remains competitive with MCMC under realistic misspecification.
Each test set contains $512$ hierarchical datasets with varying sizes and signal-to-noise ratios. We use four versions that differ only in parameter count: \textit{small} ($d \in 1$--$4$, $q \leq 2$), \textit{medium} ($d \in 5$--$8$, $q \leq 3$), \textit{large} ($d \in 9$--$12$, $q \leq 4$) and \textit{huge} ($d \in 13$--$16$, $q \leq 5$).
All four share the same data budget: at most $200$ groups, at most $150$ observations per group and at most $3{,}000$ observations in total (\myappref{app:sizes}).

\section{Results} \label{sec:res}
We organize the evaluation of \mb around six claims (\mysecref{sec:cb}--\ref{sec:rob}):
(1) on oracle test sets with known ground truth, \mb matches NUTS in parameter recovery, calibration and predictive accuracy across all three likelihood families, and outperforms Automatic Differentiation Variational Inference \citep[ADVI;][]{Kucukelbir.2017} and Pathfinder \citep[PF;][]{Zhang.2022};
(2) on real-world test sets its posteriors are nearly indistinguishable from those of NUTS;
(3) the IMH head closes the raw flow's predictive gap, and the flow makes NUTS $3.5$--$13\times$ faster while removing over $98\%$ of its divergent transitions;
(4) importance weights yield Bayes factors that agree with bridge sampling;
(5) prior conditioning turns a prior sensitivity analysis on real data into one forward pass that traces NUTS; a fixed-prior network reports unreliable uncertainty under prior shift;
(6) \mb stays in high agreement with NUTS in stress tests including misspecification and out-of-distribution data.
Throughout, \mb takes $0.09$--$0.12$\,s per dataset on GPU and $0.5$--$2.3$\,s on CPU ($410$--$5{,}500\times$ resp. $21$--$540\times$ faster than NUTS on the same datasets; \myappref{app:rt}).

\subsection{Oracle Benchmark} \label{sec:cb}
On oracle test sets, we evaluate three posterior qualities.
\emph{Parameter recovery} (do posterior means track the ground truth?), measured by Pearson's $r$ (${\to}1$) and the normalized root mean square error (NRMSE, ${\to}0$).
\emph{Calibration} (do credible intervals cover the ground truth at their nominal rate?), measured by the expected coverage error and its absolute counterpart (ECE, EACE, both ${\to}0$; negative ECE shows overconfidence).
\emph{Predictive accuracy} (does the inferred model explain the observed outcomes?), measured by the leave-one-out negative log likelihood \citep[LOO-NLL, smaller is better;][]{Vehtari.2015}; formal definitions are in \myappref{app:met}.

\begin{figure*}[h!]
    \centering
    \includegraphics[width=0.92\textwidth]{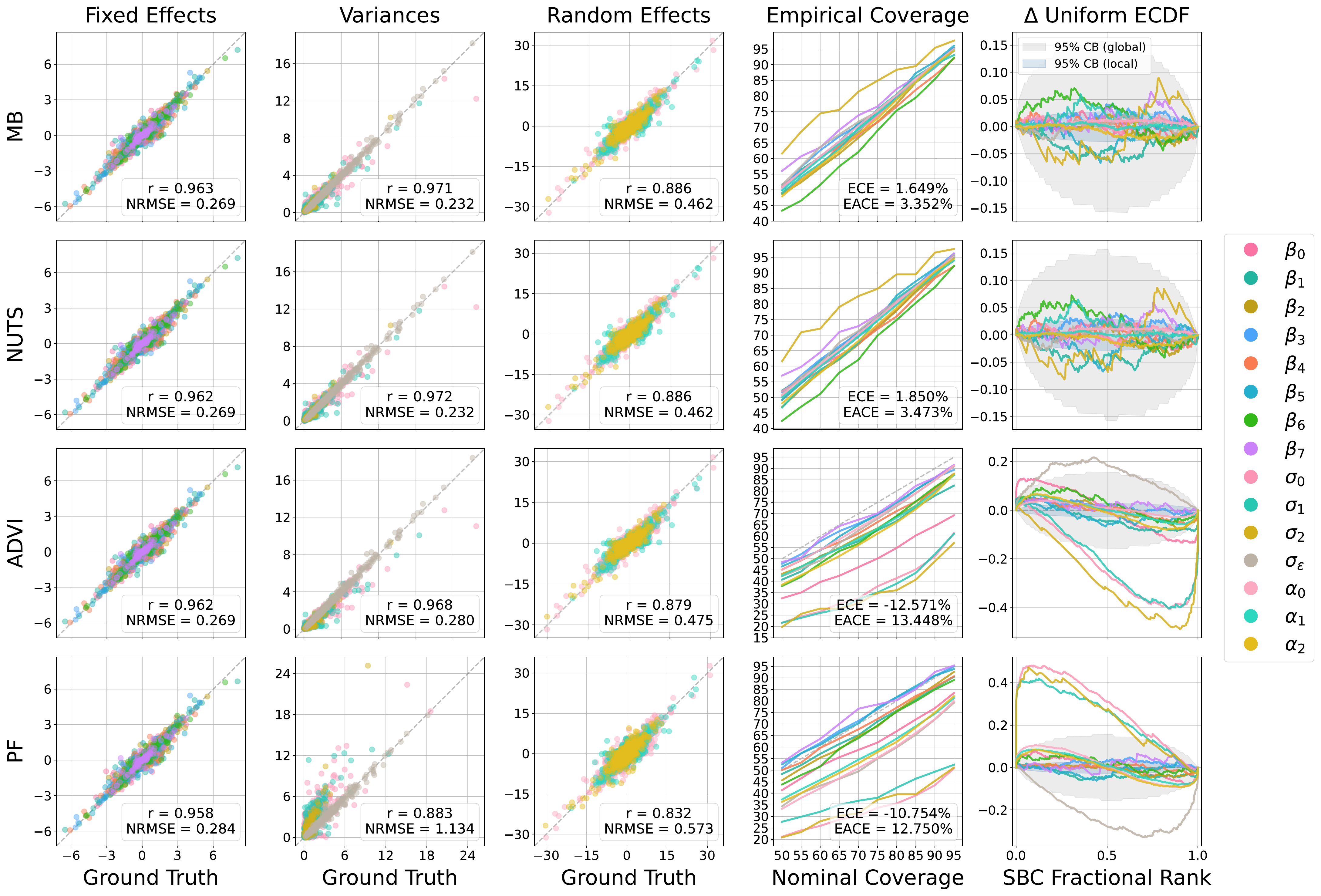}
    \caption{
       Performance on the \textit{medium} Gaussian oracle benchmark.
       Rows compare
       \texttt{MB}, \texttt{NUTS}, \texttt{ADVI} and \texttt{PF}.
       The first three columns show recovery for different groups of parameters. The last two show calibration by coverage analysis and simulation-based calibration checks \citep[][]{Talts.2020, Sailynoja.2022}.
       \rev{\myappref{app:oracle} shows the other regimes and families.}
    }
    \label{fig:2}
\end{figure*}

\mytabref{tab:oracle} summarizes performance across Gaussian models; \myfigref{fig:2} visualizes the \textit{medium} regime.
Overall, \mb closely matches NUTS in parameter recovery ($r=0.94$--$0.95$, NRMSE within $0.01$ of NUTS) across all regimes.
Calibration is near-nominal for both \mb and NUTS (EACE $\leq 0.02$),
whereas ADVI and PF are consistently overconfident (ECE $-0.12$ to $-0.17$ resp. $-0.10$ to $-0.16$). Additionally, PF has poorer recovery ($r = 0.89$--$0.91$, NRMSE $0.52$--$0.63$).
A Laplace approximation, included only as a failure check, recovers the fixed effects but not the variance components (\myappref{app:la}).
NUTS' predictive accuracy is matched best by \mb at every size (LOO-NLL within $0.01$).
On Bernoulli and Poisson data (\myappref{app:oracle}--\ref{app:corr}), \mb matches NUTS at every size in recovery ($r$ and NRMSE within $0.01$), calibration (EACE $\leq 0.03$) and predictive accuracy (LOO-NLL within $0.01$).
ADVI remains overconfident in both families, most strongly on count data (ECE $-0.12$ to $-0.14$ Bernoulli, $-0.19$ to $-0.25$ Poisson);
PF is less overconfident (ECE $-0.06$ to $-0.15$) but again recovers worst, most visibly on Bernoulli data ($r = 0.78$--$0.84$ vs.\ $0.86$--$0.89$ for NUTS).
\rev{In a worst-case analysis (\myappref{app:agree}), \mb agrees with NUTS on $99.3\%$ of the $5{,}744$ converged datasets. Our safeguards catch $39$ of $41$ disagreements.}

\begin{table}[h]
  \caption{
     Performance \rev{of \mb (\texttt{MB}) and reference methods} on the Gaussian oracle test sets (regimes as in \mysecref{sec:dat}). %
     $r$, NRMSE, ECE and EACE are mean $\pm$ SD over parameters; LOO-NLL and time (seconds per dataset, $4{,}000$ posterior draws) are median $\pm$ median absolute deviation (MAD) over NUTS-converged datasets (\myappref{app:tra}). Timing details in \myappref{app:rt}.
     }
  \label{tab:oracle}
  \centering
  \apptable
  \begin{tabular}{cc|cccccc}
    \toprule
    $\mathrm{regime}$ & $\mathrm{model}$ & $r$ & $\mathrm{NRMSE}$ & $\mathrm{ECE}$ & $\mathrm{EACE}$ & $\mathrm{LOO\text{-}NLL}$ & $\mathrm{time}$ \\
    \midrule
      \texttt{small} & \texttt{MB} & $0.94 \pm 0.05$ & $0.31 \pm 0.13$ & $\phantom{-}0.00 \pm 0.02$ & $0.02 \pm 0.01$ & $2.22 \pm 0.53$ & $\phantom{0}0.09 \pm \phantom{0}0.00$ \\
          & \texttt{NUTS} & $0.94 \pm 0.05$ & $0.31 \pm 0.13$ & $\phantom{-}0.00 \pm 0.02$ & $0.02 \pm 0.01$ & $2.22 \pm 0.53$ & $79.14 \pm 33.84$ \\
          & \texttt{ADVI} & $0.94 \pm 0.05$ & $0.33 \pm 0.13$ & $-0.12 \pm 0.14$ & $0.13 \pm 0.13$ & $2.24 \pm 0.54$ & $27.62 \pm \phantom{0}1.91$ \\
          & \texttt{PF} & $0.91 \pm 0.08$ & $0.52 \pm 0.42$ & $-0.10 \pm 0.13$ & $0.12 \pm 0.11$ & $2.23 \pm 0.54$ & $18.19 \pm \phantom{0}2.54$ \\
    \midrule
      \texttt{medium} & \texttt{MB} & $0.95 \pm 0.04$ & $0.30 \pm 0.11$ & $\phantom{-}0.00 \pm 0.03$ & $0.02 \pm 0.03$ & $2.26 \pm 0.50$ & $\phantom{0}0.09 \pm \phantom{0}0.00$ \\
          & \texttt{NUTS} & $0.95 \pm 0.04$ & $0.30 \pm 0.11$ & $\phantom{-}0.00 \pm 0.03$ & $0.02 \pm 0.03$ & $2.26 \pm 0.50$ & $82.31 \pm 32.77$ \\
          & \texttt{ADVI} & $0.95 \pm 0.04$ & $0.31 \pm 0.11$ & $-0.15 \pm 0.12$ & $0.15 \pm 0.12$ & $2.30 \pm 0.49$ & $33.15 \pm \phantom{0}4.58$ \\
          & \texttt{PF} & $0.91 \pm 0.07$ & $0.57 \pm 0.52$ & $-0.13 \pm 0.14$ & $0.14 \pm 0.13$ & $2.28 \pm 0.50$ & $28.01 \pm \phantom{0}6.33$ \\
    \midrule
      \texttt{large} & \texttt{MB} & $0.95 \pm 0.04$ & $0.30 \pm 0.10$ & $\phantom{-}0.00 \pm 0.02$ & $0.02 \pm 0.01$ & $2.31 \pm 0.45$ & $\phantom{0}0.10 \pm \phantom{0}0.00$ \\
          & \texttt{NUTS} & $0.95 \pm 0.04$ & $0.30 \pm 0.10$ & $\phantom{-}0.00 \pm 0.02$ & $0.02 \pm 0.01$ & $2.31 \pm 0.45$ & $83.50 \pm 31.19$ \\
          & \texttt{ADVI} & $0.94 \pm 0.04$ & $0.32 \pm 0.11$ & $-0.16 \pm 0.13$ & $0.16 \pm 0.13$ & $2.39 \pm 0.46$ & $36.76 \pm \phantom{0}5.13$ \\
          & \texttt{PF} & $0.91 \pm 0.08$ & $0.61 \pm 0.59$ & $-0.15 \pm 0.15$ & $0.15 \pm 0.15$ & $2.35 \pm 0.44$ & $36.15 \pm \phantom{0}7.85$ \\
    \midrule
      \texttt{huge} & \texttt{MB} & $0.95 \pm 0.03$ & $0.29 \pm 0.09$ & $-0.01 \pm 0.02$ & $0.02 \pm 0.01$ & $2.28 \pm 0.45$ & $\phantom{0}0.10 \pm \phantom{0}0.00$ \\
          & \texttt{NUTS} & $0.95 \pm 0.03$ & $0.29 \pm 0.09$ & $\phantom{-}0.00 \pm 0.02$ & $0.02 \pm 0.01$ & $2.28 \pm 0.45$ & $88.96 \pm 31.09$ \\
          & \texttt{ADVI} & $0.95 \pm 0.04$ & $0.32 \pm 0.09$ & $-0.17 \pm 0.13$ & $0.17 \pm 0.13$ & $2.40 \pm 0.44$ & $40.31 \pm \phantom{0}5.19$ \\
          & \texttt{PF} & $0.89 \pm 0.09$ & $0.63 \pm 0.59$ & $-0.16 \pm 0.16$ & $0.16 \pm 0.16$ & $2.34 \pm 0.47$ & $43.57 \pm \phantom{0}9.07$ \\
    \bottomrule
\end{tabular}

\end{table}

\clearpage
\subsection{Real-World Benchmark} \label{sec:rw}
On real-world test sets no ground truth is available, so we evaluate agreement with the NUTS posterior by the correlation of marginal posterior means ($r$), the ratio of posterior standard deviations ($\sigma$-ratio), a rank-based measure of posterior shape (rank-MAD), and the difference in predictive accuracy ($\Delta$LOO-NLL; \myappref{app:met}).

\mytabref{tab:real} summarizes agreement across all regimes for Gaussian models.
\mb achieves near-perfect alignment with NUTS in terms of posterior means ($r = 1.00$), posterior uncertainty ($\sigma$-ratio $1.00$), posterior shape ($\mathrm{rank\text{-}MAD} \leq 0.01$) and predictive accuracy ($\Delta\mathrm{LOO\text{-}NLL} \leq 0.01$).
ADVI matches NUTS in posterior means ($r \geq 0.97$) but is systematically overconfident ($\sigma$-ratio $0.75$--$0.91$) and disagrees more in shape ($\mathrm{rank\text{-}MAD}=0.03$--$0.05$). PF is closer in width ($\sigma$-ratio $0.85$--$0.95$), but its means drift as the sets grow ($r = 0.99$ to $0.90$).
The agreement carries over to the non-Gaussian models.
On real Bernoulli and Poisson data, \mb again reproduces NUTS' posterior means ($r = 1.00$),
posterior uncertainty ($\sigma$-ratio $0.97$--$1.00$),
posterior shape ($\mathrm{rank\text{-}MAD} \leq 0.01$)
and predictive accuracy ($\Delta\mathrm{LOO\text{-}NLL} = 0.00$).
ADVI remains $12$--$30\%$ overconfident on Bernoulli data and $32$--$41\%$ on count data, where its posterior shape departs most from NUTS ($\mathrm{rank\text{-}MAD}=0.08$--$0.11$). PF is $13$--$28\%$ overconfident on both.
\rev{
In an analysis of the worst parameter per dataset (\myappref{app:agree}), \mb agrees with NUTS on $98.9\%$ of the $5{,}328$ NUTS-converged datasets. Our safeguards catch $54$ of the remaining $60$ cases.}

\begin{table}[h]
  \caption{
     Posterior agreement with NUTS on the Gaussian real-world test sets (regimes as in \mysecref{sec:dat}), aggregated over all parameters.
     Entries are median $\pm$ MAD over datasets; $\Delta\mathrm{time}$ is the wall time minus that of NUTS in seconds per dataset, timed as in \mytabref{tab:oracle} (\myappref{app:rt}).
     Bernoulli and Poisson results are in \myappref{app:real}.
  }
  \label{tab:real}
  \centering
  \apptable
  \begin{tabular}{cc|ccccc}
    \toprule
    $\mathrm{regime}$ & $\mathrm{model}$ & $r$ & $\sigma\text{-ratio}$ & $\mathrm{rank\text{-}MAD}$ & $\Delta\mathrm{LOO\text{-}NLL}$ & $\Delta\mathrm{time}$ \\
    \midrule
      \texttt{small} & \texttt{MB} & $1.00 \pm 0.00$ & $1.00 \pm 0.01$ & $0.00 \pm 0.00$ & $0.00 \pm 0.00$ & $-76.29 \pm 34.78$ \\
         & \texttt{ADVI} & $1.00 \pm 0.00$ & $0.91 \pm 0.09$ & $0.03 \pm 0.02$ & $0.02 \pm 0.01$ & $-32.83 \pm 23.88$ \\
         & \texttt{PF} & $0.99 \pm 0.01$ & $0.85 \pm 0.13$ & $0.05 \pm 0.02$ & $0.01 \pm 0.01$ & $-42.53 \pm 22.53$ \\
    \midrule
      \texttt{medium} & \texttt{MB} & $1.00 \pm 0.00$ & $1.00 \pm 0.01$ & $0.00 \pm 0.00$ & $0.00 \pm 0.00$ & $-77.35 \pm 33.46$ \\
         & \texttt{ADVI} & $0.99 \pm 0.01$ & $0.83 \pm 0.10$ & $0.04 \pm 0.02$ & $0.02 \pm 0.01$ & $-38.33 \pm 23.74$ \\
         & \texttt{PF} & $0.97 \pm 0.02$ & $0.89 \pm 0.14$ & $0.05 \pm 0.02$ & $0.02 \pm 0.02$ & $-37.37 \pm 18.85$ \\
    \midrule
      \texttt{large} & \texttt{MB} & $1.00 \pm 0.00$ & $1.00 \pm 0.01$ & $0.00 \pm 0.00$ & $0.00 \pm 0.00$ & $-91.08 \pm 44.66$ \\
         & \texttt{ADVI} & $0.98 \pm 0.01$ & $0.78 \pm 0.09$ & $0.05 \pm 0.01$ & $0.01 \pm 0.01$ & $-50.27 \pm 36.26$ \\
         & \texttt{PF} & $0.94 \pm 0.03$ & $0.94 \pm 0.13$ & $0.05 \pm 0.02$ & $0.02 \pm 0.02$ & $-45.95 \pm 31.41$ \\
    \midrule
      \texttt{huge} & \texttt{MB} & $1.00 \pm 0.00$ & $1.00 \pm 0.01$ & $0.00 \pm 0.00$ & $0.00 \pm 0.00$ & $-83.57 \pm 33.02$ \\
         & \texttt{ADVI} & $0.97 \pm 0.02$ & $0.75 \pm 0.10$ & $0.05 \pm 0.02$ & $0.02 \pm 0.01$ & $-40.92 \pm 25.20$ \\
         & \texttt{PF} & $0.90 \pm 0.05$ & $0.95 \pm 0.13$ & $0.06 \pm 0.02$ & $0.02 \pm 0.02$ & $-30.06 \pm 17.69$ \\
    \bottomrule
\end{tabular}

\end{table}

\subsection{Synergy with MCMC} \label{sec:hyb_res}
The IMH head is what turns the raw flow into \mb (\mysecref{sec:hyb});
we isolate its contribution by ablating it on every benchmark (\myappref{app:abl}).
The raw flow already tracks NUTS in recovery and calibration, but its predictive gap to NUTS grows to $\Delta \mathrm{LOO{-}NLL}=0.06$--$0.82$ on the \textit{huge} sets (\mytabref{tab:abl_n}--\ref{tab:abl_p}).
The IMH head closes this gap to ${\leq}0.01$ at every size and in every family, leaving recovery and calibration mostly unchanged;
it adds $0.07$--$0.10$\,s per dataset on GPU and $0.2$--$1.8$\,s on CPU (\myappref{app:rt}).
IMH outperforms self-normalized importance sampling \citep[SNIS;][]{Tokdar.2010, Dax.2023, Barret.2026}, whose finite-sample bias grows with model size and leaves a residual gap of $0.01$--$0.13$ on the \textit{huge} benchmarks (\myappref{app:abl}).

The flow also accelerates MCMC: \texttt{MB+NUTS} initializes the chains and their mass matrix from \mb draws and samples with half the default tuning budget ($4$ chains; $500$ instead of $1{,}000$ tuning steps; \myappref{app:wn}).
Across all oracle benchmarks it matches cold-start NUTS on every accuracy metric (\mytabref{tab:abl_n}--\ref{tab:abl_p}), divergent transitions drop by more than $98\%$ (Gaussian $7{,}412 \to 55$, Bernoulli $4{,}385 \to 21$, Poisson $24{,}909 \to 426$), and sampling time per dataset falls by a factor of $3.5$--$13$ against \texttt{NUTS} (\mytabref{tab:warm}).

\clearpage
\subsection{Model Comparison} \label{sec:mc}
The importance weights calculated for IMH enable model comparison as a byproduct, since
their expectation under the flow is the marginal likelihood \citep{Kass.1995, Schmitt.2024a}:
$\mathbb{E}_{q}[w]
= \int \tilde p(\boldsymbol\vartheta \mid \mathbf D)\, \mathrm{d}\boldsymbol\vartheta
= \int p(\mathbf{D} \mid \boldsymbol\vartheta)\, p(\boldsymbol\vartheta)\, \mathrm{d}\boldsymbol\vartheta
= p(\mathbf{D}).$
Thus,
the Bayes factor between two candidate models can be estimated by the ratio of their mean importance weights, $\widehat{\mathrm{BF}}_{12} = \hat p(\mathbf{D} \mid \mathcal{M}_1)\,/\,\hat p(\mathbf{D} \mid \mathcal{M}_2)$.
These weights are less reliable where the flow under-covers the posterior, which is signaled by low IMH acceptance or low ESS (\mysecref{sec:hyb}).
\mb therefore estimates the evidence from one further importance-sampling pass whose proposal is a multivariate Student-$t$ fitted to the refined draws \citep{Geweke.1989}.
We validated this estimate on all oracle sets against bridge sampling on NUTS draws \citep{Gronau.2017}.
For the Bayes factor of a random slope against a random intercept only, the Jeffreys category ($\mathrm{BF} = 3, 10, 30, 100$; \citealp{Jeffreys.1961, Wagenmakers.2010}) matches the reference on $99$--$100\%$ of Gaussian datasets, $97$--$100\%$ of Bernoulli datasets, and $96$--$100\%$ of Poisson datasets.
Of the $20$ disagreements in total, $19$ are flagged by the Pareto-smoothed importance sampling diagnostic (PSIS-$\hat{k} > 0.7$; \myappref{app:safe}) or lie within $0.2$ nats of a category boundary (\myappref{app:ev}).

\subsection{What Prior Conditioning Buys} \label{sec:usecase}
Prior conditioning turns a sensitivity analysis into a batched forward pass.
We condition \mb on $132$, $60$ and $54$ priors
for one held-out real dataset per likelihood family, each with a correlated random slope (\texttt{sleepstudy}, \texttt{cbpp}, \texttt{Salamanders}; \myappref{app:prior_grid}).
Each grid takes $1.6$--$2.8$\,s with \mb ($0.3$--$0.8$\,s more for the evidence of every prior). NUTS needs $14$--$18$ core-hours per grid and fails to converge at $9$ of the $246$ priors.
Over the NUTS-converged cases, \mb tracks NUTS throughout ($r = 1.00$, $\sigma$-ratio $0.99$--$1.00$; \myfigref{fig:usecase}).
Its evidence estimate (\mysecref{sec:mc}) matches bridge sampling on NUTS within a median of $0.009$--$0.016$ nats per dataset (\myappref{app:prior_grid}).

Prior conditioning also guards the amortized model against prior shifts. A network trained at a fixed prior $P_0$ can be miscalibrated on problems under a different prior $P_1$. We emulate this by conditioning the raw flow on a fixed $P_0$ while data and NUTS stay under a varying $P_1$ on oracle test sets (\myappref{app:fixed_prior}).
A $P_0$ three times wider than $P_1$ makes the posterior $11$--$22\%$ too wide. Three times tighter makes the posterior $9$--$35\%$ too narrow. IMH cannot repair it, because the proposal is narrower than its target.
Posterior means and predictions are unaffected ($r \geq 0.98$): the damage is confined to uncertainty quantification, and a single dataset does not reveal it without a sampler or dedicated local calibration check \citep{Zhao.2021, Linhart.2023, Sailynoja.2025}.

\subsection{Robustness} \label{sec:rob}
Amortized inference is known to degrade silently when test data fall outside the training simulator \citep{Cannon.2022, Ward.2022, Schmitt.2024b}. %
We stress-test \mb against NUTS along five axes (\myappref{app:rob}):
misspecified likelihoods (outcomes regenerated from heavier-tailed or overdispersed families),
misspecified priors (scale, location or family deviate from the generative prior \rev{and are partly beyond the trained ranges}),
out-of-distribution predictors (infinite-variance marginals, tail-dependent copulas, longitudinal structure),
collinear designs (binned by the condition number $\kappa_2(\mathbf X)$ up to near-singular),
and data-poor groups (fewer observations than parameters).
On the first two axes, both methods target the same wrong posterior, so agreement with NUTS separates amortization error from modeling error. On the remaining three axes the fitted model stays correct and NUTS is an exact reference.

For each stress test, \mytabref{tab:rob} reports the condition that degrades the raw neural posterior the most, as measured by its predictive gap.
Posterior inference for full \mb nevertheless remains mostly indistinguishable from NUTS in every worst case ($r \geq 0.99$, $\sigma$-ratio $0.97$--$1.00$, $|\Delta\mathrm{LOO\text{-}NLL}| \leq 0.01$; \mytabref{tab:lik_mis_n}--\ref{tab:cond_p}).
Where the specified GLMM is wrong, \mb and NUTS degrade similarly against the generating parameters (e.g.\ global EACE $0.35$ vs.\ $0.34$ under target overdispersion, $0.27$ vs.\ $0.26$ under falsely certain priors; \mytabref{tab:lik_mis_p}--\ref{tab:prior_mis_n}).
The picture is the same for data poverty: in groups with fewer observations than parameters ($n_i < d + q$), \mb matches NUTS in local recovery and calibration (\mytabref{tab:poverty_n}--\ref{tab:poverty_p}).
\rev{Overall, \mb proved to be largely resilient against the stress we applied. The most notable disagreement with NUTS arose under extreme priors, far outside the training range (\myappref{app:prior_oos}). The safeguards reliably flagged such cases (\myappref{app:agree}).}

\begin{table}[h]
  \caption{
     Worst-case condition per stress test and likelihood family (\myappref{app:rob}).
     Entries are median $\pm$ MAD over NUTS-converged datasets; $\%\,\mathrm{conv.}$ is the NUTS convergence rate.
  }
  \label{tab:rob}
  \centering
  \apptable
  \begin{tabular}{lllc|ccc}
    \toprule
    $\mathrm{stress}$ & $\mathrm{family}$ & $\mathrm{worst\ condition}$ & $\%\,\mathrm{conv.}$ & $r$ & $\sigma\text{-ratio}$ & $\Delta\mathrm{LOO\text{-}NLL}$ \\
    \midrule
    \multirow{3}{*}{likelihood (\ref{app:lik_mis})}
       & Gaussian & Student-$t$, $\nu{=}3$ & 92 & $1.00 \pm 0.00$ & $1.00 \pm 0.01$ & $\phantom{-}0.00 \pm 0.00$ \\
       & Bernoulli & logit noise, SD $1$ & 95 & $1.00 \pm 0.00$ & $1.00 \pm 0.01$ & $\phantom{-}0.00 \pm 0.00$ \\
       & Poisson & Neg.\ Binomial, $\theta{=}1$ & 98 & $1.00 \pm 0.00$ & $1.00 \pm 0.01$ & $\phantom{-}0.00 \pm 0.00$ \\
    \midrule
    \multirow{3}{*}{prior (\ref{app:prior_mis})}
       & Gaussian & mean $+2$ prior SD & 90 & $1.00 \pm 0.00$ & $1.00 \pm 0.01$ & $\phantom{-}0.00 \pm 0.00$ \\
       & Bernoulli & scale ${\times}3$ & 98 & $1.00 \pm 0.00$ & $0.99 \pm 0.01$ & $\phantom{-}0.00 \pm 0.00$ \\
       & Poisson & scale ${\times}3$ & 95 & $1.00 \pm 0.00$ & $1.00 \pm 0.01$ & $\phantom{-}0.00 \pm 0.00$ \\
    \midrule
    \multirow{3}{*}{predictors (\ref{app:ood})}
       & Gaussian & Clayton, $\tau{=}0.9$ & 86 & $1.00 \pm 0.00$ & $1.00 \pm 0.01$ & $\phantom{-}0.00 \pm 0.00$ \\
       & Bernoulli & Cauchy $X$ & 95 & $1.00 \pm 0.00$ & $0.99 \pm 0.01$ & $\phantom{-}0.00 \pm 0.00$ \\
       & Poisson & Cauchy $X$ & 97 & $1.00 \pm 0.00$ & $1.00 \pm 0.01$ & $\phantom{-}0.00 \pm 0.00$ \\
    \midrule
    \multirow{3}{*}{collinearity (\ref{app:cond})}
       & Gaussian & $\kappa_2 \in [10, 10^6)$ & 87 & $1.00 \pm 0.00$ & $1.00 \pm 0.01$ & $\phantom{-}0.00 \pm 0.00$ \\
       & Bernoulli & $\kappa_2 \in [10, 10^6)$ & 97 & $1.00 \pm 0.00$ & $0.99 \pm 0.01$ & $\phantom{-}0.00 \pm 0.00$ \\
       & Poisson & $\kappa_2 \in [6, 10)$ & 98 & $1.00 \pm 0.00$ & $0.97 \pm 0.02$ & $\phantom{-}0.00 \pm 0.00$ \\
    \bottomrule
\end{tabular}

  \vspace{-4pt}
\end{table}

\section{Discussion} \label{sec:dis}
We introduced \mb, a pretrained prior-amortized in-context model for Bayesian GLMMs. %
On oracle datasets with known ground truth, on held-out real datasets, and under every stress test we applied, its posteriors closely match those of NUTS in parameter recovery, calibration, shape and predictive accuracy. Amortization also makes it two to three orders of magnitude faster.
The standard fast alternatives are overconfident in every family: ADVI, a known property of mean-field VI \citep{Turner.2011, Giordano.2018}, and Pathfinder, which additionally misplaces the variance components. Yet they are only $1.8$--$3.3\times$ and $2.1$--$4.9\times$ faster than NUTS (\myappref{app:rt}).
Our results remove the three obstacles named in \mysecref{sec:intro}:
(1) analyses over many datasets, model structures and priors become one batched forward pass (\mysecref{sec:usecase});
(2) wall time is governed by the network and IMH budget rather than by posterior geometry, so latency is bounded in advance;
(3) the model is tuning-free, and the IMH acceptance rate doubles as a diagnostic (\myappref{app:safe}).

\subsection{Limitations}
\textbf{Supported envelope.}
No simulation-based inference method can claim uniform accuracy, so our claims are restricted to the benchmarked model class: up to $16$ fixed effects, up to $5$ correlated random effects, one grouping factor, and Gaussian, Bernoulli or Poisson likelihoods.
This envelope covers a large majority of the reported GLMM analyses in our survey of $1{,}133$ publications (\myappref{app:sur}).
\rev{Furthermore, \mb generalizes well to dataset sizes outside the training range (\myappref{app:budget}).}
Beyond the tested range, both actual and potential underperformance are made detectable: a low IMH acceptance rate or ESS and a high PSIS-$\hat k$ flag the former; posteriors that drift from their analytical MAP anchors, and inputs outside the training ranges flag the latter (\myappref{app:safe}).

\textbf{Model class.}
Nested and crossed random effects occupy a meaningful share of GLMMs in practice but are not captured by a single set of group-level summaries. Extending \mb to such designs is future work.
Further likelihoods (negative-binomial, ordinal, zero-inflated) require new simulators, appropriate cheap analytical estimators and trained weights, not a new architecture.

\subsection{Outlook and Conclusion}
Two further directions stand out.
First, the forward-KL objective does not target proposal efficiency. Fine-tuning the flow on the $\chi^2$ divergence, whose Monte Carlo estimate is the variance of the importance weights \citep{Muller.2019}, could raise IMH acceptance.
Second, the evidence estimate of \mysecref{sec:mc} makes Bayes factors between candidate model structures or priors a single batched forward pass (\mysecref{sec:usecase}), opening a path to amortized Bayesian model comparison (\myappref{app:ev}).
\mb makes Bayesian GLMM inference a batched forward pass in three operating modes: (1) the raw flow in milliseconds on GPU; (2) default \mb with IMH refinement in ${\approx}0.1$\,s on GPU and $0.5$--$2.3$\,s on CPU; and (3) \texttt{MB+NUTS}, a warm start making NUTS up to $13\times$ faster and more stable.
Amortized inference and MCMC come out as collaborators rather than competitors.
\clearpage

\ifpreprint\else
\subsection*{AI Use Statement}
We used generative AI tools for the following tasks.
Experiments were designed by the authors and then discussed with, reviewed by, and refined with AI tools; parts of the experiment code were drafted by AI tools, including memory optimization and debugging.
The model source code was written by the authors and reviewed and refactored with the help of AI tools. Unit tests were extended by AI tools.
Results were partly discussed with AI tools.
Research ideas and model extensions originated with the authors and were refined in discussions with AI tools; the theoretical framework itself was developed by the authors with the help of the cited literature.
The survey of \myappref{app:sur} was carried out by an LLM pipeline; its models, the extraction protocol and the manual audit are described there.
No qualitative/thematic analysis beyond it used AI tools.
We did not use generative AI to generate datasets (the simulator is described in \mysecref{sec:dat} and \myappref{app:dat}), to formulate mathematical claims or proofs (all follow from cited sources, are textbook results or were derived by hand), for translation, for cleaning or reformatting datasets.
We have reviewed all AI-assisted work. Every piece of code written or altered by AI was inspected, and the survey extractions were checked for hallucinations by hand.
We take responsibility for the final content of this work.
\fi

\subsection*{Ethics Statement}
All datasets and software used in this work are open source and cited.
\mb is a tool for statistical analysis in the empirical sciences; we see no obvious opportunity for malicious use.
Its amortized posteriors are approximations whose reliability is flagged per dataset by the diagnostics of \mysecref{sec:hyb}.
We recommend that practitioners whose critical decisions depend on the results prototype with \mb.
If IMH acceptance or ESS is low or the neural proposal is flagged by the PSIS-$\hat k$ diagnostic, we recommend basing final conclusions on \texttt{MB+NUTS} (\myappref{app:safe}).

\subsection*{Reproducibility Statement}
The \ificlrfinal\else anonymized \fi source code linked in \mysecref{sec:intro} covers every part of this paper, from the simulator (\myappref{app:dat}) and model training (\myappref{app:arch}) to the evaluation protocol (\myappref{app:pro}), the experiments and the plots.
The software is released as a Python package with demo notebooks; network weights are downloaded automatically on first use.
The held-out test collection is described in \myappref{app:tes} and the survey data are released with the code.

\ificlrfinal
\subsection*{Acknowledgments}
We thank Niki Kilbertus for his early-stage tips on model architecture, Fabian Scheipl for his advice on input standardization, Jakob Macke, Tommy Rochussen, and Stefan Radev for their literature pointers, and Daniel Habermann for his valuable insights on data simulation and PyMC posterior sampling.
A.K. was supported by a doctoral scholarship of the German Academic Scholarship Foundation (Studienstiftung des deutschen Volkes) and by a doctoral position at Helmholtz Munich.
Compute was provided by Helmholtz Munich.
The authors declare no competing interests.
\fi

{\small %

}
\newpage

\appendix
\pdfbookmark[1]{Appendix}{appendix}

\renewcommand{\topfraction}{0.9}
\renewcommand{\bottomfraction}{0.85}
\renewcommand{\textfraction}{0.07}
\renewcommand{\floatpagefraction}{0.5}
\setcounter{topnumber}{3}
\setcounter{bottomnumber}{3}
\setcounter{totalnumber}{4}
\let\appsubsection\subsection
\renewcommand{\subsection}{\needspace{12\baselineskip}\appsubsection}

\makeatletter
\@addtoreset{figure}{section}
\@addtoreset{table}{section}
\makeatother
\renewcommand{\thefigure}{\thesection\arabic{figure}}
\renewcommand{\thetable}{\thesection\arabic{table}}

\makeatletter
\renewcommand{\toclevel@section}{2}
\renewcommand{\toclevel@subsection}{3}
\renewcommand{\toclevel@subsubsection}{4}
\makeatother

\section*{Appendix Contents}
\noindent
\startcontents[appendix]
{\small\setlength{\parskip}{0pt}\printcontents[appendix]{}{1}{\setcounter{tocdepth}{2}}}

\clearpage
\section{Evaluation Protocol} \label{app:pro}
This appendix describes the metrics (\myappref{app:met}), the test datasets (\myappref{app:tes}) and the reference methods (\myappref{app:tra}) used for the results.

\subsection{Evaluation Metrics} \label{app:met}

\paragraph{Parameter recovery.}
Pearson's $r$ measures linear agreement between posterior means
$\hat{\boldsymbol{\vartheta}}$ and ground-truth parameters
$\boldsymbol{\vartheta}^*$ across the test batch.
The root mean square error (RMSE) \rev{compares posterior means with the ground truth}.
It is normalized (NRMSE) by the batch standard deviation of ground-truth values \rev{(which simplifies comparisons across parameters)}.
\rev{For a test batch of size $B$,
\begin{equation*}
\mathrm{NRMSE}
= \frac{\mathrm{RMSE}(\hat{\boldsymbol{\vartheta}}, \boldsymbol{\vartheta}^*)}{\mathrm{SD}(\boldsymbol{\vartheta}^*)}
= \sqrt{\frac{\sum_{b=1}^{B} (\hat\vartheta_b - \vartheta^*_b)^2}
             {\sum_{b=1}^{B} (\vartheta^*_b - \bar{\vartheta}^*)^2}} \quad \text{with}
\quad \bar{\vartheta}^* = \frac{1}{B}\sum_{b=1}^{B} \vartheta^*_b .
\end{equation*}
}

\paragraph{Posterior calibration.}
Let $C(\alpha) = \frac{1}{P}\sum_{p=1}^{P} \mathbf{1}\!\left[\vartheta^*_p \in
\bigl[\hat{q}^{(p)}_{\alpha/2},\, \hat{q}^{(p)}_{1-\alpha/2}\bigr]\right]$
denote the empirical coverage at nominal level $1-\alpha$, where the sum runs
over all $P$ scalar parameters of a given type pooled across the test batch and
$\hat{q}^{(p)}_{\tau}$ is the $\tau$-quantile of the marginal posterior samples
for parameter $p$. This quantifies how often the $1-\alpha$ credible interval actually contains the true parameters in that class (e.g. how accurate the credible intervals are for fixed effects).
The signed expected coverage error
\begin{equation*}
\mathrm{ECE} = \frac{1}{|A|}\sum_{\alpha \in A} C(\alpha) - (1-\alpha)
\end{equation*}
diagnoses the direction of miscalibration: negative values indicate
overconfidence (undercoverage), positive values over-dispersion.
Its absolute counterpart
\begin{equation*}
\mathrm{EACE} = \frac{1}{|A|}\sum_{\alpha \in A} |C(\alpha) - (1-\alpha)|
\end{equation*}
measures the magnitude of miscalibration. Both are averaged over $A = \{0.05, 0.10, \ldots, 0.50\}$.

\paragraph{Simulation-based calibration (SBC).}
For each parameter $p$ in a dataset with known ground truth $\vartheta^*_p$, the
\emph{fractional rank} is the fraction of posterior samples strictly below the true value:
\begin{equation*}
u_p = \frac{1}{S}\sum_{s=1}^{S} \mathbf{1}\!\bigl[\vartheta^{(s)}_p < \vartheta^*_p\bigr].
\end{equation*}
Under a perfectly calibrated posterior, $u_p \sim \mathrm{Uniform}(0,1)$
\citep{Talts.2020}.
Fractional ranks are pooled over all (parameter, dataset) pairs within each parameter
type (fixed effects, random-effect scales, random-effect coefficients, etc.) and the
empirical CDF is compared with the diagonal. For visual clarity, we plot the empirical CDF against the $\Delta$ to the diagonal.
An S-shaped ECDF indicates over-dispersion (too-wide posteriors); an inverse-S indicates undercoverage.
The displayed bands are simultaneous $95\%$ ECDF bands following \citet{Sailynoja.2022}:
$n_\mathrm{sim}=2{,}000$ uniform samples of size $n_\mathrm{eff}$ are drawn, the
minimal coverage probability is computed for each simulated path, and the
$\alpha=0.05$ quantile of these probabilities determines the band width.
Simultaneous bands account for multiple comparisons across the entire ECDF.

\paragraph{Predictive fit.}
LOO-NLL is computed via PSIS leave-one-out cross-validation \citep{Vehtari.2015}:
\begin{equation*}
\mathrm{LOO\text{-}NLL} = -\frac{1}{n}\sum_{ij} \log \hat{p}(y_{ij} \mid \mathbf{D}_{-ij}),
\end{equation*}
where $\hat{p}(y_{ij} \mid \mathbf{D}_{-ij}) = \sum_s \tilde{w}_s\,
p(y_{ij} \mid \boldsymbol{\vartheta}^{(s)}, \boldsymbol{\alpha}_i^{(s)})$ is
the importance-weighted leave-one-out predictive density with PSIS-smoothed
weights $\tilde{w}_s \propto 1/p(y_{ij} \mid \boldsymbol{\vartheta}^{(s)},
\boldsymbol{\alpha}_i^{(s)})$.
The PSIS computation also yields per-observation Pareto shape parameters
$\hat{k}$; observations with $\hat{k} > 0.7$ indicate unreliable LOO estimates. This quantifies how likely any single left-out observation is under the GLMM, if the latter were fit without that observation.

\newpage
\paragraph{Posterior agreement.}
The following metrics quantify how closely a method's posterior approximates NUTS.
All three are computed against converged NUTS fits (\myappref{app:tra}).

\textit{Posterior mean correlation ($r$).}
For each dataset, all active posterior means
$\bar{\vartheta}_{\text{method}}$ and $\bar{\vartheta}_{\text{NUTS}}$
are concatenated into flat vectors across fixed effects, random-effect scales,
residual scale, and random-effect coefficients.
Pearson's $r$ is computed between these vectors; the reported statistic is
median $\pm$ MAD over datasets.

\textit{Posterior standard-deviation ratio ($\sigma$-ratio).}
For each dataset, the per-parameter std ratio
$\sigma_\mathrm{method}^{(p)} / \sigma_\mathrm{NUTS}^{(p)}$
is computed over all active parameters and the per-dataset median is taken.
The reported statistic is median $\pm$ MAD of these per-dataset medians over datasets;
values $> 1$ indicate a wider approximate posterior.

\textit{Rank-MAD.}
This metric applies the fractional-rank principle of SBC to the real-data setting: NUTS samples take the role of $\vartheta^*$.
For each active parameter and dataset, every NUTS sample is ranked within
the marginal \texttt{MB} distribution and expressed as a fraction $\in [0,1]$.
Under perfect agreement these fractions follow $\mathrm{Uniform}(0,1)$.
The metric is
\begin{equation*}
\mathrm{rank\text{-}MAD} = \frac{1}{|Q|}\sum_{q \in Q} \bigl|\hat{F}(q) - q\bigr|,
\end{equation*}
where $Q = \{0.10, 0.25, 0.50, 0.75, 0.90\}$ and $\hat{F}(q)$ is the empirical
$q$-quantile of the pooled rank fractions across all active (parameter, dataset) pairs.
A value of 0 indicates that the distributional shapes agree perfectly.

\begin{table}[p]
  \caption{Real-world test datasets: source package, likelihood, fixed-effect count $d$ (intercept included), groups $m$, observations $n$, range of observations per group, and group-size entropy ratio $H$ ($1$ = balanced).}
  \label{tab:test_datasets}
  \centering
  \apptable
  \setlength{\tabcolsep}{3pt}
  \begin{tabular}{lllrrrrr}
    \toprule
    Dataset & Package & Likelihood & $d$ & $m$ & $n$ & $n$/group & $H$ \\
    \midrule
    \texttt{Orange} & \texttt{datasets} & Gaussian & 2 & 5 & 35 & 7--7 & 1.000 \\
    \texttt{Machines} & \texttt{nlme} & Gaussian & 3 & 6 & 54 & 9--9 & 1.000 \\
    \texttt{Indometh} & \texttt{datasets} & Gaussian & 2 & 6 & 66 & 11--11 & 1.000 \\
    \texttt{Oats} & \texttt{nlme} & Gaussian & 4 & 6 & 72 & 12--12 & 1.000 \\
    \texttt{ergoStool} & \texttt{nlme} & Gaussian & 4 & 9 & 36 & 4--4 & 1.000 \\
    \texttt{Pastes} & \texttt{lme4} & Gaussian & 3 & 10 & 60 & 6--6 & 1.000 \\
    \texttt{Pixel} & \texttt{nlme} & Gaussian & 3 & 10 & 102 & 4--14 & 0.972 \\
    \texttt{Theoph} & \texttt{datasets} & Gaussian & 3 & 12 & 132 & 11--11 & 1.000 \\
    \texttt{sleepstudy} & \texttt{lme4} & Gaussian & 2 & 18 & 180 & 10--10 & 1.000 \\
    \texttt{Penicillin} & \texttt{lme4} & Gaussian & 6 & 24 & 144 & 6--6 & 1.000 \\
    \texttt{Oxboys} & \texttt{nlme} & Gaussian & 10 & 26 & 234 & 9--9 & 1.000 \\
    \texttt{Orthodont} & \texttt{nlme} & Gaussian & 3 & 27 & 108 & 4--4 & 1.000 \\
    \texttt{Exam} & \texttt{mlmRev} & Gaussian & 11 & 65 & 4059 & 2--198 & 0.974 \\
    \texttt{Gcsemv} & \texttt{mlmRev} & Gaussian & 3 & 73 & 1703 & 2--93 & 0.934 \\
    \texttt{Hsb82} & \texttt{mlmRev} & Gaussian & 5 & 160 & 7185 & 14--67 & 0.993 \\
    \texttt{MathAchieve} & \texttt{MEMSS} & Gaussian & 5 & 160 & 7185 & 14--67 & 0.993 \\
    \texttt{InstEval} & \texttt{lme4} & Gaussian & 12 & 2972 & 73421 & 1--92 & 0.978 \\
    \midrule
    \texttt{analcatdata\_boxing1} & \texttt{PMLB} & Bernoulli & 3 & 10 & 120 & 12--12 & 1.000 \\
    \texttt{analcatdata\_boxing2} & \texttt{PMLB} & Bernoulli & 3 & 11 & 132 & 12--12 & 1.000 \\
    \texttt{cbpp} & \texttt{lme4} & Bernoulli & 4 & 15 & 842 & 26--96 & 0.968 \\
    \texttt{cane} & \texttt{boot} & Bernoulli & 5 & 45 & 21266 & 232--826 & 0.988 \\
    \texttt{bacteria} & \texttt{MASS} & Bernoulli & 6 & 50 & 220 & 2--5 & 0.994 \\
    \texttt{respiratory} & \texttt{geepack} & Bernoulli & 7 & 56 & 444 & 4--8 & 0.999 \\
    \texttt{Contraception} & \texttt{mlmRev} & Bernoulli & 6 & 60 & 1934 & 2--118 & 0.948 \\
    \texttt{OME} & \texttt{MASS} & Bernoulli & 6 & 77 & 4196 & 19--134 & 0.970 \\
    \texttt{guImmun} & \texttt{mlmRev} & Bernoulli & 16 & 161 & 2159 & 1--55 & 0.961 \\
    \texttt{guPrenat} & \texttt{mlmRev} & Bernoulli & 22 & 161 & 2449 & 1--50 & 0.965 \\
    \texttt{respInf} & \texttt{gamlss.data} & Bernoulli & 9 & 275 & 1200 & 1--6 & 0.984 \\
    \texttt{VerbAgg} & \texttt{lme4} & Bernoulli & 7 & 316 & 7584 & 24--24 & 1.000 \\
    \texttt{ohio} & \texttt{geepack} & Bernoulli & 3 & 537 & 2148 & 4--4 & 1.000 \\
    \texttt{muscatine} & \texttt{geepack} & Bernoulli & 5 & 4856 & 9856 & 1--3 & 0.989 \\
    \midrule
    \texttt{Mmmec} & \texttt{mlmRev} & Poisson & 3 & 9 & 354 & 3--95 & 0.824 \\
    \texttt{Arabidopsis} & \texttt{lme4} & Poisson & 5 & 9 & 625 & 32--137 & 0.961 \\
    \texttt{analcatdata\_apnea1} & \texttt{PMLB} & Poisson & 3 & 19 & 475 & 25--25 & 1.000 \\
    \texttt{analcatdata\_apnea2} & \texttt{PMLB} & Poisson & 3 & 19 & 475 & 25--25 & 1.000 \\
    \texttt{Salamanders} & \texttt{glmmTMB} & Poisson & 12 & 23 & 644 & 28--28 & 1.000 \\
    \texttt{Owls} & \texttt{glmmTMB} & Poisson & 5 & 27 & 599 & 4--52 & 0.964 \\
    \texttt{epil} & \texttt{MASS} & Poisson & 5 & 59 & 236 & 4--4 & 1.000 \\
    \texttt{grouseticks} & \texttt{lme4} & Poisson & 4 & 118 & 403 & 1--10 & 0.969 \\
    \texttt{Chem97} & \texttt{mlmRev} & Poisson & 4 & 2410 & 31022 & 1--188 & 0.946 \\
    \bottomrule
  \end{tabular}
\end{table}

\subsection{Test Datasets} \label{app:tes}
The held-out collection comprises $40$ real hierarchical datasets (\mytabref{tab:test_datasets}): $36$ from standard R mixed-effects packages (\texttt{boot}, \texttt{datasets}, \texttt{gamlss.data}, \texttt{geepack}, \texttt{glmmTMB}, \texttt{lme4}, \texttt{MASS}, \texttt{MEMSS}, \texttt{mlmRev}, and \texttt{nlme}) and four (the \texttt{analcatdata} apnea and boxing sets) from PMLB \citep{Romano.2022}, all preprocessed as in \myappref{app:pre}.
In the table, $d$ counts the fixed effects including the intercept (after dummy-coding categoricals), $m$ the groups, $n$ the observations, and $H = -\sum_i p_i \log p_i / \log m$ with $p_i = n_i/n$ is the group-size entropy ratio, where $H = 1$ indicates perfectly balanced groups and $H \to 0$ extreme imbalance.
Aggregate binomial datasets are expanded to Bernoulli observations before these summaries are computed.
\texttt{InstEval}, \texttt{Chem97} and \texttt{muscatine} exceed the group budget of \mytabref{tab:sizes} and enter the test sets only through group subsampling.

\paragraph{Hierarchical bootstrap.}
Both test set types draw their designs from this collection with the hierarchical bootstrap of Algorithm~\ref{alg:subsampler}.
Oracle test sets then replace $\mathbf{y}$ with simulated outcomes; real-world test sets keep the original outcome and sample no regression parameters.
Their model specification follows \texttt{Bambi}'s weakly informative defaults \citep{Capretto.2022}, e.g.\ $\mathcal{N}(0, 2.5)$ fixed effects and half-Normal$(2.5)$ random-effect scales, with correlated random effects and an $\mathrm{LKJ}(1)$ prior on $\mathbf{R}$ whenever $q \ge 2$; all inference methods receive the identical specification.
Gaussian outcomes are rescaled to unit standard deviation, which leaves posterior shapes unchanged.

\begin{algorithm}[hbp]
\caption{Hierarchical bootstrap (\texttt{Subsampler})}
\label{alg:subsampler}
\begin{algorithmic}[1]
  \REQUIRE Source dataset with groups $\{G_j\}$, target counts $(d, m, n)$
  \ENSURE Subsampled $(\mathbf{X}, \mathbf{y}, \mathbf{ns})$ with original $\mathbf{y}$
  \STATE Randomly select $m$ eligible groups (groups with $\ge n_{\min}$ observations)
  \STATE Sample per-group observation budgets: seed every group at $n_{\min}$, then allocate the remainder by repeated $\mathrm{Dirichlet}(c\,\mathbf{1}_m)$--multinomial draws with concentration $c \sim \mathcal{U}(2, 20)$, clipped to each group's available count
  \STATE Randomly subsample $\mathrm{ns}_i$ observations from group $G_i$ without replacement
  \STATE Randomly select $d-1$ predictor columns; append intercept column
  \STATE Standardize predictors; return original $\mathbf{y}$ unchanged
\end{algorithmic}
\end{algorithm}

\subsection{Reference Sampler and Baselines} \label{app:tra}

\paragraph{PyMC parameterization.}
Since our model class explicitly includes correlated random effects (not supported by \texttt{Bambi}, \citealp{Capretto.2022}), we implement all baseline models directly in \texttt{PyMC} \citep{Abril-Pla.2023}. On the oracle and the real-world test sets alike, NUTS estimates $\mathbf{R}$ under its LKJ prior whenever $q \ge 2$.
Random effects use a non-centered parameterization throughout, which improves HMC geometry in weakly-identified settings (small $n_i$):

\textit{Independent random effects}: each dimension is parameterized separately as
\begin{equation*}
\boldsymbol{\alpha}_i^{(j)} = \zeta_i^{(j)} \cdot \sigma_{\alpha_j}, \quad \zeta_i^{(j)} \sim \mathcal{N}(0, 1), \quad \sigma_{\alpha_j} \sim \mathcal P (\tau_{\sigma_j}),
\end{equation*}
where $\mathcal P$ denotes the prior family drawn at simulation time (half-Normal, half-Student-$t$, or Exponential).

\textit{Correlated random effects}: the full covariance is $\mathbf{S} = \mathrm{diag}(\boldsymbol{\sigma}_\alpha)\,\mathbf{R}\,\mathrm{diag}(\boldsymbol{\sigma}_\alpha)$ with $\mathbf{R} \sim \mathrm{LKJ}(\eta)$.
The Cholesky factor $\mathbf{L}_S$ is obtained via \texttt{LKJCholeskyCov}, and random effects are drawn as
\begin{equation*}
\boldsymbol{\alpha}_i = \boldsymbol{\zeta}_i\, \mathbf{L}_S^\top, \quad \boldsymbol{\zeta}_i \sim \mathcal{N}_q(\mathbf{0}, \mathbf{I}).
\end{equation*}
We verify correctness by comparing the log-posterior against \texttt{Bambi} (uncorrelated case)
at 50 random unconstrained-space points across ten held-out datasets per size class; numerical differences are below $10^{-9}$ in both cases.
Data and prior hyperparameters are passed to \texttt{PyMC} in the same standardized form that \mb uses (\myappref{app:sta}), so both target the identical posterior.

\paragraph{NUTS.}
Every dataset is fitted with a fixed ladder of three NUTS budgets \citep{Hoffman.2014}, each with $4$ chains and \texttt{PyMC}'s step-size and diagonal mass-matrix adaptation.
Level 0 is the \texttt{PyMC} default: $1{,}000$ warm-up iterations and $1{,}000$ draws per chain, target acceptance $0.8$, maximum tree depth $10$.
Level 1 doubles the warm-up ($2{,}000$) and raises the target acceptance to $0.9$; level 2 uses $4{,}000$ warm-up iterations, $2{,}000$ draws per chain, target acceptance $0.99$ and tree depth $12$.
Level 2 ($8{,}000$ joint draws, later thinned to $4{,}000$ for equal comparison) is the reference posterior for the agreement metrics of \myappref{app:met}.
The \texttt{NUTS} row of the recovery, calibration and runtime tables is the posterior a practitioner obtains by escalating the budget until the diagnostics are clean: per dataset we compare against the cheapest level that passes the convergence criterion below. The reported wall time is the accumulated wall time until the chosen NUTS level.

\ifpreprint\else\clearpage\fi
\paragraph{Convergence criterion.}
A NUTS fit counts as converged over all model variables if the rank-normalized $\hat{R} \le 1.01$ and the bulk and tail effective sample sizes are $\ge 400$ \citep{Vehtari.2021}, and no transition diverged (levels 0 and 1) or at most $0.1\%$ of the post-warm-up transitions diverged (level 2).
On the oracle test sets ($512$ datasets each) this criterion retains $84$--$98\%$ of the datasets, while $41$--$67\%$ are kept at \texttt{PyMC}'s default budget and $66$--$88\%$ at level 1.
Restricting the oracle tables to these datasets changes no entry by more than $0.02$ ($r$, NRMSE) or $0.015$ (ECE, EACE) relative to scoring all $512$ datasets, so the filter removes NUTS failures without biasing the comparison.
\mytabref{tab:nuts_conv_n}--\ref{tab:nuts_conv_p} list the diagnostics of every level on every test set.
Level-0 and level-1 runs that fail it agree with the level-2 reference as closely as runs that pass: on all $23$ test sets, the median over datasets of the correlation of posterior means is $\ge 0.999$, of the $\sigma$-ratio $0.986$--$1.002$, and of the rank-MAD $\le 0.007$.
The ADVI and Pathfinder budgets, by contrast, reach $\sigma$-ratios of $0.49$--$2.1$.
Without a reference, however, a practitioner cannot tell such runs from wrong ones, which is why the escalating \texttt{NUTS} pays for the next level.

\paragraph{ADVI.}
We run mean-field ADVI \citep{Kucukelbir.2017} at \texttt{PyMC}'s defaults (\texttt{adagrad\_window}, base learning rate $10^{-3}$). We compare \texttt{PyMC}'s default budget ($10^4$ iterations) with an escalated one ($10^5$ iterations).
$4{,}000$ draws are taken from the approximation after $10^4$ iterations and after $10^5$ iterations. We report the latter for the \texttt{ADVI} row of the tables. Runtime and ELBO are logged for both.
Runs whose objective becomes non-finite are discarded as failed fits (Bernoulli data only).

\paragraph{Pathfinder.}
Pathfinder \citep{Zhang.2022} runs L-BFGS from jittered initial points, places a Gaussian at the ELBO-maximizing point of each optimization path, and pools the paths' draws by Pareto-smoothed importance resampling (\texttt{pymc-extras} defaults).
We run it with $4$ paths (its default) and with $20$ paths, which we report in the \texttt{PF} row. Again, we sample $4{,}000$ draws each, and record the Pareto $\hat{k}$ of the pooled importance weights (which exceeds $0.7$ on nearly every GLMM dataset).

\paragraph{Laplace approximation.}
We include Laplace approximation only as a failure check (\myappref{app:la}) for Pathfinder. We use \texttt{pymc-extras}' \texttt{fit\_laplace}: a BFGS search for the mode of the unconstrained parameterization (log scales, non-centered random effects, Cholesky-factor correlations), the dense Hessian at the mode by automatic differentiation, and $4{,}000$ draws from the resulting Gaussian are transformed back to the constrained space.
Where the Hessian is not positive definite the draws are not finite. Where the mode of a log-scale lies far in the tail the draws are astronomically large; a fit with any non-finite draw or any draw beyond $10^3$ standardized units is discarded as failed ($4$--$12\%$ of the oracle datasets).

We do not compare against the integrated nested Laplace approximation \citep[INLA;][]{Rue.2009, Lindgren.2015}: \texttt{R-INLA} cannot fit our model as specified (Normal fixed-effect priors only, a Wishart prior for correlated random effects). An experimental \texttt{PyMC} implementation exists (in \texttt{pymc-extras} 0.15.1), but it replaces INLA's deterministic integration over the hyperparameters by NUTS on the Laplace-marginalized posterior and returns no random-effect posteriors. Pathfinder takes INLA's place as the optimization-based alternative.

\paragraph{Compute and wall time.}
All reference methods run one dataset per job on the CPU nodes of a shared cluster (Intel Xeon Gold): NUTS on four cores (one per chain), ADVI, Pathfinder and LA on one core.
The recorded wall time covers model construction, \texttt{PyTensor} compilation and sampling.
\begin{table}[hbp]
  \caption{
     Convergence diagnostics of the cold-start \texttt{NUTS} fits at every budget level on every Gaussian test set.
     divg.: total divergent transitions; $\hat{R}_\mathrm{worst}$: largest $\hat{R}$; sat.: share of datasets whose tree-depth saturation rate exceeds $5\%$ of post-warm-up draws; conv.: share of datasets passing the convergence criterion; the last four columns give the share failing each of its checks ($\hat{R}$, bulk and tail ESS, divergences).
  }
  \label{tab:nuts_conv_n}
  \centering
  \apptablewide
  \begin{tabular}{llcrrrrrrrr}
    \toprule
    $\mathrm{regime}$ & $\mathrm{set}$ & $\mathrm{level}$ & $\mathrm{divg.}$ & $\hat{R}_\mathrm{worst}$ & $\%\,\mathrm{sat.}$ & $\%\,\mathrm{conv.}$ & $\%\,\hat{R}{>}1.01$ & $\%\,\mathrm{ESS}{<}400$ & $\%\,\mathrm{ESS}_\mathrm{t}{<}400$ & $\%\,\mathrm{divg.}$ \\
    \midrule
    \texttt{small} & oracle & $0$ & $1{,}769$ & $1.60$ & $0.4$ & $56$ & $15.0$ & $12.9$ & $9.6$ & $31.1$ \\
     &  & $1$ & $98$ & $1.11$ & $0.0$ & $77$ & $13.9$ & $12.5$ & $8.6$ & $6.1$ \\
     &  & $2$ & $0$ & $1.07$ & $0.0$ & $95$ & $4.3$ & $3.5$ & $2.1$ & $0.0$ \\
     & real & $0$ & $1{,}714$ & $1.05$ & $0.0$ & $53$ & $10.5$ & $6.6$ & $7.2$ & $39.8$ \\
     &  & $1$ & $306$ & $1.38$ & $0.0$ & $77$ & $6.6$ & $4.3$ & $2.7$ & $17.2$ \\
     &  & $2$ & $6$ & $1.07$ & $0.0$ & $99$ & $1.4$ & $0.4$ & $0.2$ & $0.0$ \\
    \texttt{medium} & oracle & $0$ & $1{,}924$ & $1.36$ & $0.0$ & $56$ & $21.5$ & $17.0$ & $11.1$ & $25.6$ \\
     &  & $1$ & $213$ & $1.13$ & $0.0$ & $76$ & $16.8$ & $17.6$ & $10.9$ & $5.3$ \\
     &  & $2$ & $0$ & $1.24$ & $0.0$ & $92$ & $6.8$ & $6.1$ & $3.5$ & $0.0$ \\
     & real & $0$ & $762$ & $1.54$ & $0.0$ & $58$ & $24.2$ & $18.8$ & $15.4$ & $24.4$ \\
     &  & $1$ & $40$ & $1.23$ & $0.0$ & $73$ & $19.5$ & $14.1$ & $11.3$ & $4.1$ \\
     &  & $2$ & $1$ & $1.03$ & $0.0$ & $95$ & $4.5$ & $2.0$ & $1.6$ & $0.0$ \\
    \texttt{large} & oracle & $0$ & $2{,}228$ & $1.60$ & $0.4$ & $60$ & $23.4$ & $19.9$ & $14.3$ & $20.7$ \\
     &  & $1$ & $90$ & $1.30$ & $0.0$ & $70$ & $23.4$ & $18.6$ & $12.5$ & $4.5$ \\
     &  & $2$ & $1$ & $1.06$ & $0.0$ & $91$ & $7.8$ & $6.8$ & $2.9$ & $0.0$ \\
     & real & $0$ & $2{,}229$ & $1.82$ & $0.6$ & $56$ & $31.2$ & $22.7$ & $22.9$ & $17.6$ \\
     &  & $1$ & $165$ & $1.54$ & $2.1$ & $68$ & $27.3$ & $22.7$ & $17.0$ & $3.7$ \\
     &  & $2$ & $0$ & $1.06$ & $0.8$ & $94$ & $5.7$ & $2.9$ & $1.8$ & $0.0$ \\
    \texttt{huge} & oracle & $0$ & $1{,}491$ & $1.68$ & $0.4$ & $56$ & $31.6$ & $26.2$ & $19.5$ & $14.6$ \\
     &  & $1$ & $139$ & $1.63$ & $0.2$ & $66$ & $27.7$ & $27.1$ & $16.6$ & $2.5$ \\
     &  & $2$ & $0$ & $1.09$ & $0.0$ & $84$ & $14.1$ & $11.9$ & $7.6$ & $0.0$ \\
     & real & $0$ & $1{,}045$ & $1.60$ & $0.4$ & $62$ & $31.1$ & $22.1$ & $21.1$ & $10.7$ \\
     &  & $1$ & $1{,}299$ & $1.61$ & $1.4$ & $67$ & $28.3$ & $21.3$ & $17.2$ & $2.7$ \\
     &  & $2$ & $2$ & $1.60$ & $0.6$ & $93$ & $6.4$ & $2.7$ & $2.3$ & $0.0$ \\
    \bottomrule
\end{tabular}

\end{table}

\begin{table}[hbp]
  \caption{Convergence diagnostics of the cold-start \texttt{NUTS} fits on every Bernoulli test set. Layout as in \mytabref{tab:nuts_conv_n}.}
  \label{tab:nuts_conv_b}
  \centering
  \apptablewide
  \begin{tabular}{llcrrrrrrrr}
    \toprule
    $\mathrm{regime}$ & $\mathrm{set}$ & $\mathrm{level}$ & $\mathrm{divg.}$ & $\hat{R}_\mathrm{worst}$ & $\%\,\mathrm{sat.}$ & $\%\,\mathrm{conv.}$ & $\%\,\hat{R}{>}1.01$ & $\%\,\mathrm{ESS}{<}400$ & $\%\,\mathrm{ESS}_\mathrm{t}{<}400$ & $\%\,\mathrm{divg.}$ \\
    \midrule
    \texttt{small} & oracle & $0$ & $943$ & $1.31$ & $0.0$ & $62$ & $6.6$ & $5.3$ & $6.1$ & $34.0$ \\
     &  & $1$ & $136$ & $1.10$ & $0.0$ & $88$ & $5.1$ & $3.3$ & $2.9$ & $6.6$ \\
     &  & $2$ & $0$ & $1.06$ & $0.0$ & $98$ & $1.2$ & $1.2$ & $1.2$ & $0.0$ \\
     & real & $0$ & $1{,}880$ & $1.25$ & $0.0$ & $66$ & $6.4$ & $4.1$ & $4.1$ & $31.1$ \\
     &  & $1$ & $237$ & $1.03$ & $0.0$ & $86$ & $4.3$ & $1.6$ & $2.1$ & $9.2$ \\
     &  & $2$ & $0$ & $1.02$ & $0.0$ & $99$ & $0.8$ & $0.2$ & $0.2$ & $0.0$ \\
    \texttt{medium} & oracle & $0$ & $1{,}435$ & $1.53$ & $0.0$ & $67$ & $13.1$ & $9.4$ & $11.3$ & $20.7$ \\
     &  & $1$ & $1{,}037$ & $1.53$ & $0.0$ & $83$ & $11.5$ & $8.6$ & $8.0$ & $3.7$ \\
     &  & $2$ & $3$ & $1.53$ & $0.2$ & $98$ & $1.6$ & $1.4$ & $0.8$ & $0.0$ \\
     & real & $0$ & $681$ & $1.17$ & $0.0$ & $62$ & $16.6$ & $13.1$ & $12.5$ & $25.0$ \\
     &  & $1$ & $98$ & $1.17$ & $0.0$ & $79$ & $13.3$ & $8.0$ & $7.6$ & $7.6$ \\
     &  & $2$ & $2$ & $1.19$ & $0.2$ & $97$ & $2.3$ & $0.8$ & $1.2$ & $0.0$ \\
    \texttt{large} & oracle & $0$ & $387$ & $1.28$ & $0.0$ & $66$ & $21.1$ & $16.0$ & $14.8$ & $13.1$ \\
     &  & $1$ & $19$ & $1.47$ & $0.0$ & $82$ & $14.5$ & $10.9$ & $9.6$ & $2.0$ \\
     &  & $2$ & $0$ & $1.03$ & $0.0$ & $96$ & $3.1$ & $2.0$ & $1.4$ & $0.0$ \\
     & real & $0$ & $1{,}045$ & $1.60$ & $0.2$ & $62$ & $26.4$ & $18.8$ & $19.9$ & $13.9$ \\
     &  & $1$ & $238$ & $2.01$ & $0.4$ & $75$ & $20.1$ & $13.5$ & $13.1$ & $4.3$ \\
     &  & $2$ & $1$ & $1.60$ & $0.4$ & $97$ & $2.9$ & $1.8$ & $1.6$ & $0.0$ \\
    \texttt{huge} & oracle & $0$ & $1{,}620$ & $1.55$ & $0.0$ & $64$ & $21.7$ & $17.4$ & $15.0$ & $14.3$ \\
     &  & $1$ & $1{,}072$ & $1.56$ & $0.0$ & $75$ & $19.9$ & $16.8$ & $14.5$ & $2.7$ \\
     &  & $2$ & $2$ & $1.54$ & $0.4$ & $94$ & $5.5$ & $3.3$ & $2.3$ & $0.0$ \\
     & real & $0$ & $243$ & $1.26$ & $0.0$ & $62$ & $30.9$ & $23.2$ & $21.3$ & $9.2$ \\
     &  & $1$ & $545$ & $2.13$ & $0.2$ & $71$ & $25.8$ & $19.7$ & $15.4$ & $2.1$ \\
     &  & $2$ & $1$ & $1.07$ & $0.2$ & $94$ & $4.7$ & $3.1$ & $3.1$ & $0.0$ \\
    \bottomrule
\end{tabular}

\end{table}

\begin{table}[hbp]
  \caption{Convergence diagnostics of the cold-start \texttt{NUTS} fits on every Poisson test set. Layout as in \mytabref{tab:nuts_conv_n}.}
  \label{tab:nuts_conv_p}
  \centering
  \apptablewide
  \begin{tabular}{llcrrrrrrrr}
    \toprule
    $\mathrm{regime}$ & $\mathrm{set}$ & $\mathrm{level}$ & $\mathrm{divg.}$ & $\hat{R}_\mathrm{worst}$ & $\%\,\mathrm{sat.}$ & $\%\,\mathrm{conv.}$ & $\%\,\hat{R}{>}1.01$ & $\%\,\mathrm{ESS}{<}400$ & $\%\,\mathrm{ESS}_\mathrm{t}{<}400$ & $\%\,\mathrm{divg.}$ \\
    \midrule
    \texttt{small} & oracle & $0$ & $4{,}025$ & $1.18$ & $0.0$ & $41$ & $10.0$ & $7.8$ & $7.0$ & $54.3$ \\
     &  & $1$ & $489$ & $1.06$ & $0.4$ & $77$ & $7.4$ & $5.1$ & $2.5$ & $16.0$ \\
     &  & $2$ & $2$ & $1.22$ & $0.0$ & $98$ & $2.0$ & $1.2$ & $1.0$ & $0.0$ \\
     & real & $0$ & $14{,}289$ & $1.56$ & $5.9$ & $23$ & $50.2$ & $38.9$ & $29.9$ & $69.3$ \\
     &  & $1$ & $2{,}564$ & $1.32$ & $7.2$ & $33$ & $44.9$ & $38.9$ & $25.6$ & $29.3$ \\
     &  & $2$ & $5$ & $1.04$ & $4.7$ & $76$ & $23.8$ & $15.2$ & $10.5$ & $0.0$ \\
    \texttt{medium} & oracle & $0$ & $7{,}666$ & $1.55$ & $0.8$ & $51$ & $17.2$ & $13.9$ & $12.1$ & $42.0$ \\
     &  & $1$ & $834$ & $1.07$ & $1.0$ & $76$ & $13.3$ & $10.5$ & $8.8$ & $13.3$ \\
     &  & $2$ & $1$ & $1.03$ & $0.6$ & $95$ & $4.5$ & $3.5$ & $1.8$ & $0.0$ \\
     & real & $0$ & $6{,}506$ & $1.65$ & $0.2$ & $30$ & $23.6$ & $15.2$ & $5.1$ & $64.6$ \\
     &  & $1$ & $1{,}003$ & $1.07$ & $0.0$ & $59$ & $19.3$ & $8.0$ & $2.3$ & $26.4$ \\
     &  & $2$ & $1$ & $1.01$ & $0.0$ & $98$ & $2.1$ & $0.2$ & $0.2$ & $0.0$ \\
    \texttt{large} & oracle & $0$ & $7{,}261$ & $1.50$ & $0.8$ & $55$ & $21.5$ & $17.4$ & $15.6$ & $31.4$ \\
     &  & $1$ & $675$ & $1.18$ & $1.2$ & $73$ & $16.8$ & $14.6$ & $10.5$ & $10.9$ \\
     &  & $2$ & $0$ & $1.08$ & $0.6$ & $92$ & $6.4$ & $6.4$ & $3.5$ & $0.0$ \\
     & real & $0$ & $5{,}155$ & $1.54$ & $0.0$ & $42$ & $20.7$ & $15.0$ & $7.2$ & $55.1$ \\
     &  & $1$ & $746$ & $1.32$ & $0.0$ & $66$ & $17.8$ & $9.0$ & $4.7$ & $22.9$ \\
     &  & $2$ & $0$ & $1.02$ & $0.2$ & $99$ & $1.0$ & $0.6$ & $0.4$ & $0.0$ \\
    \texttt{huge} & oracle & $0$ & $5{,}957$ & $1.21$ & $0.8$ & $59$ & $23.2$ & $19.7$ & $16.6$ & $24.4$ \\
     &  & $1$ & $1{,}505$ & $1.44$ & $0.6$ & $70$ & $23.2$ & $19.1$ & $13.9$ & $7.4$ \\
     &  & $2$ & $0$ & $1.08$ & $0.6$ & $91$ & $6.8$ & $6.8$ & $3.7$ & $0.0$ \\
    \bottomrule
\end{tabular}

\end{table}

\clearpage
\section{Amortized Inference meets MCMC} \label{app:hyb}
The amortized posterior can be combined with likelihood-based refinement in two ways:
as a proposal distribution for Independence Metropolis--Hastings (asymptotically exact at no MCMC tuning cost, \myappref{app:imh}), or as an initialization for NUTS (inheriting NUTS' asymptotic guarantees at a reduced tuning budget, \myappref{app:wn}).
Both use the exact unnormalized posterior, so their quality does not hinge on network generalization alone.
The first two subsections describe the procedure first and its empirical evaluation second.
An ablation compares them with the raw flow and with importance sampling (\myappref{app:abl}), and the safeguards that flag unreliable posteriors close this appendix (\myappref{app:safe}).

\subsection{Independence Metropolis--Hastings}\label{app:imh}
\paragraph{Procedure.}
The flow $q(\boldsymbol\vartheta) = p_{\boldsymbol\Pi_g}(\boldsymbol\vartheta \mid \mathbf{s})$ serves as a fixed proposal for an IMH chain \citep{Tierney.1994}.
The chain runs over the global parameters $\boldsymbol\vartheta$ only; the random effects are marginalized out of the log-weight
\begin{equation*}
    \log w(\boldsymbol\vartheta) = \log p(\mathbf{D} \mid \boldsymbol\vartheta) + \log p(\boldsymbol\vartheta) - \log q(\boldsymbol\vartheta).
\end{equation*}
All $S = C \times T$ proposals are drawn and weighted upfront.
Each of the $C$ chains starts at the first draw of its block, runs $T$ accept/reject steps and discards the first $T_\mathrm{burn}$.

\textbf{Gaussian outcomes.} The marginal likelihood $p(\mathbf{D} \mid \boldsymbol\vartheta)$ has a closed form.
For every retained state, the random effects are redrawn from their exact conditional $p(\boldsymbol\alpha_i \mid \boldsymbol\vartheta, \mathbf{D}_i)$ (Rao--Blackwellization).

\textbf{Bernoulli and Poisson outcomes.} The marginal likelihood has no closed form.
We replace it by the unbiased importance-sampling estimate \citep[IS$^2$;][]{Tran.2013}
\begin{equation*}
    \hat p(\mathbf{D} \mid \boldsymbol\vartheta) = \prod_{i=1}^m \frac{1}{K} \sum_{k=1}^K \frac{p(\mathbf{y}_i \mid \boldsymbol\alpha_{ik}, \boldsymbol\vartheta)\, p(\boldsymbol\alpha_{ik} \mid \boldsymbol\vartheta)}{g_i(\boldsymbol\alpha_{ik})},
    \qquad \boldsymbol\alpha_{ik} \sim g_i .
\end{equation*}
The inner proposal $g_i = (1-\epsilon)\,\mathcal{N}(\boldsymbol\alpha_i^*, \mathbf{H}_i^{-1}) + \epsilon\,\mathcal{N}(\mathbf{0}, \boldsymbol\Sigma_{\boldsymbol\vartheta})$ is a Laplace approximation of group $i$'s conditional (mode $\boldsymbol\alpha_i^*$ by damped Newton steps, Hessian $\mathbf{H}_i$), mixed with a small share of the random-effect prior for safety.
Because $\hat p$ is unbiased and each state keeps its own estimate, the chain is pseudo-marginal \citep{Andrieu.2009}.
Its $\boldsymbol\vartheta$-marginal is the exact posterior; the estimator noise only lowers the acceptance rate.
Each state also stores one inner draw per group, picked in proportion to its weight.
A rejection would repeat these draws, so every retained step refreshes them with one sampling--importance-resampling move \citep{Andrieu.2018}, which leaves $p(\boldsymbol\alpha_i \mid \boldsymbol\vartheta, \mathbf{D}_i)$ invariant.
The same estimate $\hat p$ gives the marginal likelihood of \myappref{app:ev} for the discrete families.
A plain Laplace approximation in place of $\hat p$ would be cheaper, but it targets a pseudo-posterior.
On the real Bernoulli test sets it leaves posteriors $2$--$3\%$ too narrow ($\sigma$-ratio $0.97$--$0.98$ against $0.99$--$1.00$ for the pseudo-marginal chain).
On the \textit{large} and \textit{huge} oracle sets it leaves a predictive gap of $0.004$--$0.008$ nats to NUTS, which the pseudo-marginal chain closes to ${\leq}0.002$.

\paragraph{Pool size, acceptance and ESS.}
The mean acceptance rate $\bar a$ measures how well the flow overlaps the posterior.
Rates below ${\approx}0.1$ indicate a large gap.
Where the flow lacks tail mass, the chains can only revisit the few pool members that cover the tails; the refined posterior can be too narrow even at moderate acceptance and a larger pool cannot reliably fix it.
\mb therefore also reports the effective sample size (ESS) of the $C$ chains. It uses the minimum rank-normalized split-chain bulk ESS over the global parameters \citep{Vehtari.2021}, and warns for ESS $< 150$ (\myappref{app:safe}).
We ran a pool size sweep in the hardest regime (Bernoulli, \textit{huge}; $S \in \{1, 2, 4\} \times 10^3$) using the pseudo-marginal chain.
The EACE for IMH decays proportional to $(\bar{a}S)^{-0.4}$; at $\bar{a}S \approx 760$ accepted draws it reaches EACE $=0.025$.
\mb thus reports a suggested pool size $S_\mathrm{sugg} = \lceil 700 / \bar{a} \rceil$ with its diagnostics, as a practical floor of about $700$ accepted draws, rounded up to a multiple of $500$ and clamped to $\min = 1{,}000$ and $\max = 16{,}000$ (\myappref{app:safe}).

\paragraph{Defaults.}
$C=4$ chains and $T_\mathrm{burn}=25$.
For the discrete families, $K=8$ inner draws and $\epsilon=0.01$; under the posterior, this keeps the standard deviation of $\log \hat p(\mathbf{D} \mid \boldsymbol\vartheta)$ at a median of $0.09$--$0.12$ nats.
The released implementation over-draws the pool to $S = C\,(\lceil n/C \rceil + T_\mathrm{burn})$, so that exactly the $n$ requested samples remain after burn-in (e.g.\ $n=1000$: $S=1{,}100$, $T=275$).
The benchmarks use a fixed pool of $S=4{,}000$ ($T=1{,}000$, $3{,}900$ samples), matching the $4 \times 1{,}000$ draws of NUTS.

\ifpreprint\else\clearpage\fi
\subsection{Warm-Started NUTS}\label{app:wn}
\paragraph{Procedure.}
Warm-started NUTS (\texttt{MB+NUTS}) targets the exact posterior, while \mb's proposal reduces the tuning budget needed to reach the typical set: $4$ chains with $500$ tuning steps and $500$ draws, against \texttt{PyMC}'s default of $1{,}000$ tuning steps and $1{,}000$ draws (\texttt{NUTS}$_0$), on the identical \texttt{PyMC} model (\myappref{app:tra}).
We use $S=4{,}000$ \mb samples for initialization. The $C$ chain starts are spaced evenly along the samples, and the initial diagonal mass matrix is the per-coordinate variance of $64$ draws.
Each start is back-transformed to \texttt{PyMC}'s non-centered parameterization (\myappref{app:tra}).

\needspace{6\baselineskip}
\paragraph{Results.}
\texttt{MB+NUTS} and cold-start NUTS are compared on the twelve oracle benchmarks ($512$ datasets each), both timed on four Xeon Gold CPU cores (\myappref{app:rt}); the reported \texttt{MB+NUTS} wall-clock time covers the complete procedure per dataset including \mb sampling.
Accuracy is at parity: \texttt{MB+NUTS} matches cold-start NUTS in recovery, calibration and predictive accuracy on every benchmark (\mytabref{tab:abl_n}--\ref{tab:abl_p}).
\mytabref{tab:warm} summarizes the sampler diagnostics.
The warm start stabilizes sampling: it removes more than $98\%$ of divergent transitions in each family.
It is also faster: against the escalating \texttt{NUTS} of the main text, the mean wall-clock time per dataset falls by $6.6$--$9.2\times$ (Gaussian), $7.1$--$13\times$ (Bernoulli) and $3.5$--$13\times$ (Poisson).
The escalation costs $1.7$--$3.2\times$ the default budget, so against \texttt{NUTS}$_0$ alone the gain is $1.1$--$7.5\times$, smallest on the \textit{huge} Poisson benchmark.

\begin{table}[hbp]
  \caption{
     Sampler diagnostics of warm-started (\texttt{MB+NUTS}: $4$ chains, $500$ tuning steps, $500$ draws) and cold-start NUTS on each oracle test set ($512$ datasets per regime): the default budget \texttt{NUTS}$_0$ ($4$ chains, $1{,}000$ tuning steps, $1{,}000$ draws) and the escalating \texttt{NUTS} of the main text (time only: below the top level, a run with any divergence fails the convergence criterion and is replaced by the next level, so its divergence count reflects this selection rather than the sampler; \myappref{app:tra}).
     Divergences are total divergent transitions over all datasets and chains; time is the mean wall-clock time in seconds per dataset over all $512$ datasets on four CPU cores,
   speed-ups are ratios of these means.
  }
  \label{tab:warm}
  \centering
  \apptable
  \begin{tabular}{ll|rr|rrr|rr}
    \toprule
     & & \multicolumn{2}{c|}{$\mathrm{divergences}$} & \multicolumn{3}{c|}{$\mathrm{time\ [s]}$} & \multicolumn{2}{c}{$\mathrm{speed\text{-}up}$} \\
    $\mathrm{family}$ & $\mathrm{regime}$ & \texttt{NUTS}$_0$ & \texttt{MB+NUTS} & \texttt{NUTS}$_0$ & \texttt{NUTS} & \texttt{MB+NUTS} & $\mathrm{vs.}$ \texttt{NUTS}$_0$ & $\mathrm{vs.}$ \texttt{NUTS} \\
    \midrule
    \multirow{4}{*}{Gaussian}
       & \texttt{small} & $1{,}769$ & $16$ & $53.9$ & $99.4$ & $12.5$ & $4.3\times$ & $8.0\times$ \\
       & \texttt{medium} & $1{,}924$ & $11$ & $64.3$ & $136.4$ & $14.9$ & $4.3\times$ & $9.2\times$ \\
       & \texttt{large} & $2{,}228$ & $26$ & $71.5$ & $172.2$ & $23.8$ & $3.0\times$ & $7.2\times$ \\
       & \texttt{huge} & $1{,}491$ & $2$ & $79.3$ & $221.5$ & $33.8$ & $2.3\times$ & $6.6\times$ \\
    \midrule
    \multirow{4}{*}{Bernoulli}
       & \texttt{small} & $943$ & $7$ & $53.8$ & $91.9$ & $7.2$ & $7.5\times$ & $12.8\times$ \\
       & \texttt{medium} & $1{,}435$ & $4$ & $67.4$ & $134.8$ & $17.3$ & $3.9\times$ & $7.8\times$ \\
       & \texttt{large} & $387$ & $1$ & $72.4$ & $196.6$ & $21.9$ & $3.3\times$ & $9.0\times$ \\
       & \texttt{huge} & $1{,}620$ & $9$ & $79.8$ & $258.7$ & $36.4$ & $2.2\times$ & $7.1\times$ \\
    \midrule
    \multirow{4}{*}{Poisson}
       & \texttt{small} & $4{,}025$ & $64$ & $54.9$ & $132.0$ & $10.1$ & $5.4\times$ & $13.1\times$ \\
       & \texttt{medium} & $7{,}666$ & $81$ & $67.0$ & $181.6$ & $18.1$ & $3.7\times$ & $10.0\times$ \\
       & \texttt{large} & $7{,}261$ & $129$ & $80.2$ & $256.4$ & $30.1$ & $2.7\times$ & $8.5\times$ \\
       & \texttt{huge} & $5{,}957$ & $152$ & $87.6$ & $264.7$ & $76.7$ & $1.1\times$ & $3.5\times$ \\
    \bottomrule
\end{tabular}

\end{table}

\clearpage
\subsection{Refinement Ablations} \label{app:abl}
We compare the full refinement ladder on the oracle benchmarks (512 datasets per size regime):
the raw amortized posterior (\MBz{}), i.e.\ \mb with its IMH head ablated;
self-normalized importance sampling \citep[SNIS;][]{Tokdar.2010} toward the same target (\MBz{}+IS, marginal weights for Gaussian, Laplace weights for the discrete families);
IMH (\MBz{}+IMH, the default pipeline \texttt{MB} of the main text);
warm-started NUTS (\texttt{MB+NUTS}, treated in \myappref{app:wn});
and default NUTS as the reference (\myappref{app:tra}).
SNIS is an established method for NPE refinement (\citealp{Dax.2023, Gebhard.2023, Barret.2026}\rev{; \citealp{Li.2026}}) and is a close relative of IMH \citep{Liu.1996}: IMH uses ratios of importance weights $w$ in an accept--reject step and returns unweighted draws, whereas SNIS reweights the draws by $w$ and carries a finite-sample bias that grows with the weight variance.
In both cases unreliable proposals are flagged by the Pareto-smoothed importance sampling shape diagnostic \citep[PSIS-$\hat k > 0.7$;][]{Vehtari.2015, Yao.2018} (\myappref{app:safe}).

\paragraph{Results.}
\mytabref{tab:abl_n}--\ref{tab:abl_p} support two observations.
First, the two weighted refinements act very similarly on recovery and calibration, and both shrink the raw posterior's predictive gap, with IMH ahead of IS on the largest regimes (Gaussian \textit{huge} LOO-NLL $2.398 \to 2.241$ vs.\ $2.379$, NUTS $2.245$; Poisson \textit{huge} $2.212 \to 1.397$ vs.\ $1.464$, NUTS $1.397$).
SNIS matches IMH up to the \textit{medium} regime, but its residual predictive gap to NUTS reaches $0.01$--$0.13$ on the \textit{huge} benchmarks, which is why IMH is the default head.
Second, \texttt{MB+NUTS} matches cold-start NUTS on every accuracy metric at half the default tuning budget. \mytabref{tab:abl_real} reports the corresponding ablation on the real-world test sets of \mysecref{sec:rw}.
The raw posterior agrees with NUTS in means and shape everywhere ($r \geq 0.99$, rank-MAD $\leq 0.04$) but shows family-specific width biases: over-dispersion that grows with size for Gaussian outcomes ($\sigma$-ratio up to $1.31$) and mild narrowing for Poisson (down to $0.90$). This goes along with a predictive gap that grows with size (up to $0.53$ for Gaussian and $1.64$ for Poisson outcomes).
The default IMH head removes the width biases ($\sigma$-ratio $0.97$--$1.00$) and closes the predictive gap (${\leq}0.01$) at every size and in every family.

\begin{table}[hbp]
  \caption{
     Posterior agreement of the raw flow (\MBz{}) and the default pipeline (\texttt{MB}) with \texttt{NUTS} on each real-world test set.
     Entries are median $\pm$ MAD over datasets; $r$ and rank-MAD are \MBz{} values (\texttt{MB} matches or improves them; \mytabref{tab:real} and \myappref{app:real}).
     No \textit{huge} regime exists for count outcomes (\mytabref{tab:real_p}).
  }
  \label{tab:abl_real}
  \centering
  \apptable
  \begin{tabular}{llcc|cc|cc}
    \toprule
     & & & & \multicolumn{2}{c|}{$\sigma\text{-ratio}$} & \multicolumn{2}{c}{$\Delta\mathrm{LOO\text{-}NLL}$} \\
    $\mathrm{family}$ & $\mathrm{regime}$ & $r$ & $\mathrm{rank\text{-}MAD}$ & \MBz{} & \texttt{MB} & \MBz{} & \texttt{MB} \\
    \midrule
    \multirow{4}{*}{Gaussian}
       & \texttt{small} & $1.00 \pm 0.00$ & $0.00 \pm 0.00$ & $1.00 \pm 0.02$ & $1.00 \pm 0.01$ & $0.00 \pm 0.00$ & $0.00 \pm 0.00$ \\
       & \texttt{medium} & $1.00 \pm 0.00$ & $0.01 \pm 0.01$ & $1.06 \pm 0.06$ & $1.00 \pm 0.01$ & $0.02 \pm 0.01$ & $0.00 \pm 0.00$ \\
       & \texttt{large} & $0.99 \pm 0.01$ & $0.02 \pm 0.01$ & $1.23 \pm 0.16$ & $1.00 \pm 0.01$ & $0.09 \pm 0.07$ & $0.00 \pm 0.00$ \\
       & \texttt{huge} & $0.99 \pm 0.01$ & $0.03 \pm 0.01$ & $1.31 \pm 0.16$ & $1.00 \pm 0.01$ & $0.53 \pm 0.40$ & $0.00 \pm 0.00$ \\
    \midrule
    \multirow{4}{*}{Bernoulli}
       & \texttt{small} & $1.00 \pm 0.00$ & $0.01 \pm 0.01$ & $1.02 \pm 0.03$ & $1.00 \pm 0.01$ & $0.00 \pm 0.00$ & $0.00 \pm 0.00$ \\
       & \texttt{medium} & $1.00 \pm 0.00$ & $0.01 \pm 0.00$ & $1.00 \pm 0.04$ & $1.00 \pm 0.01$ & $0.01 \pm 0.01$ & $0.00 \pm 0.00$ \\
       & \texttt{large} & $0.99 \pm 0.00$ & $0.01 \pm 0.00$ & $0.99 \pm 0.03$ & $1.00 \pm 0.01$ & $0.02 \pm 0.01$ & $0.00 \pm 0.00$ \\
       & \texttt{huge} & $0.99 \pm 0.01$ & $0.01 \pm 0.00$ & $0.99 \pm 0.04$ & $0.99 \pm 0.01$ & $0.05 \pm 0.02$ & $0.00 \pm 0.00$ \\
    \midrule
    \multirow{3}{*}{Poisson}
       & \texttt{small} & $1.00 \pm 0.00$ & $0.03 \pm 0.02$ & $1.02 \pm 0.13$ & $1.00 \pm 0.02$ & $0.21 \pm 0.21$ & $0.00 \pm 0.00$ \\
       & \texttt{medium} & $0.99 \pm 0.00$ & $0.03 \pm 0.01$ & $0.94 \pm 0.06$ & $0.99 \pm 0.01$ & $0.24 \pm 0.17$ & $0.00 \pm 0.00$ \\
       & \texttt{large} & $0.99 \pm 0.01$ & $0.03 \pm 0.01$ & $0.90 \pm 0.08$ & $0.97 \pm 0.02$ & $1.64 \pm 1.45$ & $0.00 \pm 0.00$ \\
    \bottomrule
\end{tabular}

\end{table}

\begin{table}[hbp]
  \caption{
     Refinement ablation on each Gaussian oracle test set ($512$ datasets per regime).
     Recovery ($r$, NRMSE) and calibration (ECE, EACE) entries are averaged over parameters as in \mytabref{tab:oracle}, but over all $512$ datasets rather than the NUTS-converged subset; LOO-NLL is the median over datasets.
     \MBz{}+IMH is the default pipeline \texttt{MB} of the main text; the NUTS budgets are those of \mytabref{tab:warm}.
  }
  \label{tab:abl_n}
  \centering
  \apptable
  \begin{tabular}{cl|cccc|c}
    \toprule
    $\mathrm{regime}$ & $\mathrm{model}$ & $r$ & $\mathrm{NRMSE}$ & $\mathrm{ECE}$ & $\mathrm{EACE}$ & $\mathrm{LOO\text{-}NLL}$ \\
    \midrule
    \texttt{small} & \MBz{} & $0.943$ & $0.314$ & $0.002$ & $0.020$ & $2.178$ \\
     & \MBz{}+IS & $0.947$ & $0.302$ & $0.002$ & $0.017$ & $2.177$ \\
     & \MBz{}+IMH & $0.947$ & $0.302$ & $0.001$ & $0.018$ & $2.178$ \\
     & \texttt{MB+NUTS} & $0.947$ & $0.303$ & $-0.000$ & $0.019$ & $2.176$ \\
     & \texttt{NUTS} & $0.947$ & $0.303$ & $0.001$ & $0.017$ & $2.177$ \\
    \midrule
    \texttt{medium} & \MBz{} & $0.951$ & $0.295$ & $0.008$ & $0.020$ & $2.284$ \\
     & \MBz{}+IS & $0.952$ & $0.290$ & $0.001$ & $0.021$ & $2.276$ \\
     & \MBz{}+IMH & $0.952$ & $0.290$ & $0.001$ & $0.021$ & $2.260$ \\
     & \texttt{MB+NUTS} & $0.952$ & $0.291$ & $0.001$ & $0.021$ & $2.260$ \\
     & \texttt{NUTS} & $0.952$ & $0.290$ & $0.001$ & $0.021$ & $2.259$ \\
    \midrule
    \texttt{large} & \MBz{} & $0.952$ & $0.295$ & $0.007$ & $0.017$ & $2.329$ \\
     & \MBz{}+IS & $0.952$ & $0.293$ & $0.001$ & $0.017$ & $2.301$ \\
     & \MBz{}+IMH & $0.955$ & $0.284$ & $-0.000$ & $0.017$ & $2.274$ \\
     & \texttt{MB+NUTS} & $0.955$ & $0.283$ & $-0.000$ & $0.019$ & $2.272$ \\
     & \texttt{NUTS} & $0.955$ & $0.284$ & $0.002$ & $0.018$ & $2.273$ \\
    \midrule
    \texttt{huge} & \MBz{} & $0.956$ & $0.284$ & $0.007$ & $0.016$ & $2.398$ \\
     & \MBz{}+IS & $0.957$ & $0.281$ & $-0.002$ & $0.015$ & $2.379$ \\
     & \MBz{}+IMH & $0.959$ & $0.275$ & $-0.003$ & $0.016$ & $2.241$ \\
     & \texttt{MB+NUTS} & $0.959$ & $0.274$ & $-0.001$ & $0.015$ & $2.240$ \\
     & \texttt{NUTS} & $0.959$ & $0.274$ & $0.000$ & $0.014$ & $2.245$ \\
    \bottomrule
\end{tabular}

\end{table}

\begin{table}[hbp]
  \caption{Refinement ablation on each Bernoulli oracle test set. Layout as in \mytabref{tab:abl_n}.}
  \label{tab:abl_b}
  \centering
  \apptable
  \begin{tabular}{cl|cccc|c}
    \toprule
    $\mathrm{regime}$ & $\mathrm{model}$ & $r$ & $\mathrm{NRMSE}$ & $\mathrm{ECE}$ & $\mathrm{EACE}$ & $\mathrm{LOO\text{-}NLL}$ \\
    \midrule
    \texttt{small} & \MBz{} & $0.897$ & $0.414$ & $-0.002$ & $0.024$ & $0.551$ \\
     & \MBz{}+IS & $0.899$ & $0.411$ & $-0.003$ & $0.021$ & $0.550$ \\
     & \MBz{}+IMH & $0.898$ & $0.413$ & $0.003$ & $0.020$ & $0.550$ \\
     & \texttt{MB+NUTS} & $0.897$ & $0.416$ & $0.001$ & $0.020$ & $0.550$ \\
     & \texttt{NUTS} & $0.898$ & $0.415$ & $0.002$ & $0.020$ & $0.551$ \\
    \midrule
    \texttt{medium} & \MBz{} & $0.881$ & $0.430$ & $-0.005$ & $0.018$ & $0.460$ \\
     & \MBz{}+IS & $0.884$ & $0.426$ & $-0.006$ & $0.019$ & $0.452$ \\
     & \MBz{}+IMH & $0.884$ & $0.426$ & $-0.004$ & $0.019$ & $0.449$ \\
     & \texttt{MB+NUTS} & $0.883$ & $0.427$ & $-0.006$ & $0.019$ & $0.449$ \\
     & \texttt{NUTS} & $0.883$ & $0.428$ & $-0.005$ & $0.019$ & $0.449$ \\
    \midrule
    \texttt{large} & \MBz{} & $0.883$ & $0.427$ & $0.007$ & $0.016$ & $0.460$ \\
     & \MBz{}+IS & $0.882$ & $0.426$ & $0.001$ & $0.015$ & $0.434$ \\
     & \MBz{}+IMH & $0.884$ & $0.424$ & $0.003$ & $0.015$ & $0.428$ \\
     & \texttt{MB+NUTS} & $0.883$ & $0.427$ & $0.004$ & $0.016$ & $0.428$ \\
     & \texttt{NUTS} & $0.882$ & $0.427$ & $0.006$ & $0.016$ & $0.427$ \\
    \midrule
    \texttt{huge} & \MBz{} & $0.850$ & $0.474$ & $0.003$ & $0.018$ & $0.471$ \\
     & \MBz{}+IS & $0.855$ & $0.468$ & $-0.003$ & $0.021$ & $0.423$ \\
     & \MBz{}+IMH & $0.859$ & $0.464$ & $-0.006$ & $0.020$ & $0.408$ \\
     & \texttt{MB+NUTS} & $0.860$ & $0.464$ & $0.003$ & $0.019$ & $0.409$ \\
     & \texttt{NUTS} & $0.860$ & $0.465$ & $0.004$ & $0.019$ & $0.409$ \\
    \bottomrule
\end{tabular}

\end{table}

\begin{table}[hbp]
  \caption{Refinement ablation on each Poisson oracle test set. Layout as in \mytabref{tab:abl_n}.}
  \label{tab:abl_p}
  \centering
  \apptable
  \begin{tabular}{cl|cccc|c}
    \toprule
    $\mathrm{regime}$ & $\mathrm{model}$ & $r$ & $\mathrm{NRMSE}$ & $\mathrm{ECE}$ & $\mathrm{EACE}$ & $\mathrm{LOO\text{-}NLL}$ \\
    \midrule
    \texttt{small} & \MBz{} & $0.931$ & $0.339$ & $-0.002$ & $0.032$ & $1.360$ \\
     & \MBz{}+IS & $0.932$ & $0.336$ & $0.001$ & $0.031$ & $1.347$ \\
      & \MBz{}+IMH & $0.932$ & $0.337$ & $0.002$ & $0.030$ & $1.341$ \\
     & \texttt{MB+NUTS} & $0.932$ & $0.338$ & $-0.001$ & $0.033$ & $1.339$ \\
     & \texttt{NUTS} & $0.932$ & $0.336$ & $0.002$ & $0.031$ & $1.341$ \\
    \midrule
    \texttt{medium} & \MBz{} & $0.927$ & $0.331$ & $-0.001$ & $0.015$ & $1.447$ \\
     & \MBz{}+IS & $0.926$ & $0.333$ & $-0.003$ & $0.018$ & $1.367$ \\
     & \MBz{}+IMH & $0.927$ & $0.332$ & $-0.003$ & $0.018$ & $1.357$ \\
     & \texttt{MB+NUTS} & $0.927$ & $0.331$ & $-0.003$ & $0.018$ & $1.357$ \\
     & \texttt{NUTS} & $0.928$ & $0.331$ & $-0.002$ & $0.018$ & $1.357$ \\
    \midrule
    \texttt{large} & \MBz{} & $0.941$ & $0.299$ & $-0.000$ & $0.022$ & $1.700$ \\
     & \MBz{}+IS & $0.943$ & $0.295$ & $-0.014$ & $0.021$ & $1.420$ \\
     & \MBz{}+IMH & $0.943$ & $0.294$ & $-0.014$ & $0.020$ & $1.395$ \\
     & \texttt{MB+NUTS} & $0.943$ & $0.295$ & $-0.010$ & $0.018$ & $1.394$ \\
     & \texttt{NUTS} & $0.943$ & $0.295$ & $-0.009$ & $0.018$ & $1.394$ \\
    \midrule
    \texttt{huge} & \MBz{} & $0.921$ & $0.336$ & $0.008$ & $0.024$ & $2.212$ \\
     & \MBz{}+IS & $0.922$ & $0.335$ & $-0.002$ & $0.021$ & $1.464$ \\
     & \MBz{}+IMH & $0.924$ & $0.330$ & $-0.014$ & $0.024$ & $1.397$ \\
     & \texttt{MB+NUTS} & $0.925$ & $0.328$ & $-0.001$ & $0.019$ & $1.397$ \\
     & \texttt{NUTS} & $0.925$ & $0.327$ & $0.001$ & $0.020$ & $1.397$ \\
    \bottomrule
\end{tabular}

\end{table}

\FloatBarrier %
\subsection{Deployment Safeguards} \label{app:safe}

\mb's inference pipeline ships five per-dataset safeguards that detect (and where possible repair) unreliable posteriors at deployment.

\paragraph{Support validation.}
Datasets whose dimensions fall outside the training ranges (\myappref{app:sizes}) receive an explicit warning.
\rev{The same goes for prior scales outside the trained hyperprior range (\myappref{app:prior}).}

\paragraph{Acceptance gate.}
The default IMH head reports its own reliability: the chain's mean acceptance rate $\bar a$ measures the overlap between the flow proposal and the exact posterior. The suggested pool size is derived from it (\myappref{app:imh}) and targets the budget at which the finite-pool calibration error approaches the level of the easy regimes.
A mean acceptance below $0.1$ flags a large flow--posterior gap; the refined posterior is returned together with a warning that recommends re-running at $S_\mathrm{sugg}$ or falling back to MCMC (e.g.\ \texttt{MB+NUTS}, \myappref{app:wn}).

\paragraph{ESS gate.}
The acceptance rate can stay moderate while the IMH chains revisit only a few pool members, for instance where the flow lacks tail mass; the refined posterior is then too narrow.
A minimum bulk ESS over the global parameters below $150$ (\myappref{app:imh}) returns the posterior with a warning that recommends falling back to MCMC (e.g.\ \texttt{MB+NUTS}).

\paragraph{PSIS-$\hat k$.}
The Pareto-smoothed shape diagnostic $\hat{k}$ \citep{Yao.2018} is an analogous gate (\myappref{app:ev}). Since IMH computes importance weights, their empirical distribution can be approximated with a Pareto distribution with the shape parameter $k$ (with low $k$ indicating relatively even spread): $\hat{k} \leq 0.7$ certifies the flow posterior as a usable proposal, $\hat{k} > 0.7$ returns the posterior with a warning.

\paragraph{MAP-consistency check.}
Posterior means are compared against the (flow-independent) analytical MAP estimates that already condition the summarizer (\myappref{app:suf}), via the per-dataset score $z = \max_j\, |\mathrm{mean}_j - \mathrm{MAP}_j| / \mathrm{SD}_j$.
A large $z$ means the flow posterior drifted away from its analytical anchor (a symptom of data outside the training distribution).
The warning threshold ($z^* = 5$) is calibrated on held-out simulated data ($99\%$ quantile).
In distribution it fires rarely ($\leq 0.7\%$ of test datasets). Under the design shifts of \myappref{app:ood} it fires up to ${\sim}7\times$ more often (validating the check).

Together with the refinements above, these checks make silent failure the exception.

\clearpage
\section{Benchmark Details} \label{app:res}
This appendix collects the remaining per-family results: the remaining oracle results (\myappref{app:oracle}), the separate random-effect correlation results (\myappref{app:corr}), the non-Gaussian real-world tables (\myappref{app:real}), \rev{the per-dataset agreement with NUTS (\myappref{app:agree})}, the runtime breakdown (\myappref{app:rt}) and the Laplace baseline (\myappref{app:la}).
Metrics and convergence criteria are defined in \myappref{app:pro}.

\subsection{Oracle Benchmark} \label{app:oracle}
The figures on the following pages complement \myfigref{fig:2}; columns and descriptors are reused throughout, and an additional top row shows the raw flow posterior (\MBz{}) without IMH refinement.
\mytabref{tab:oracle} in the main text summarizes the Gaussian regimes; \mytabref{tab:oracle_b} and \mytabref{tab:oracle_p} do so for Bernoulli and Poisson.

\begin{figure}[h]
	\centering
	\includegraphics[width=1.0\textwidth, page=1]{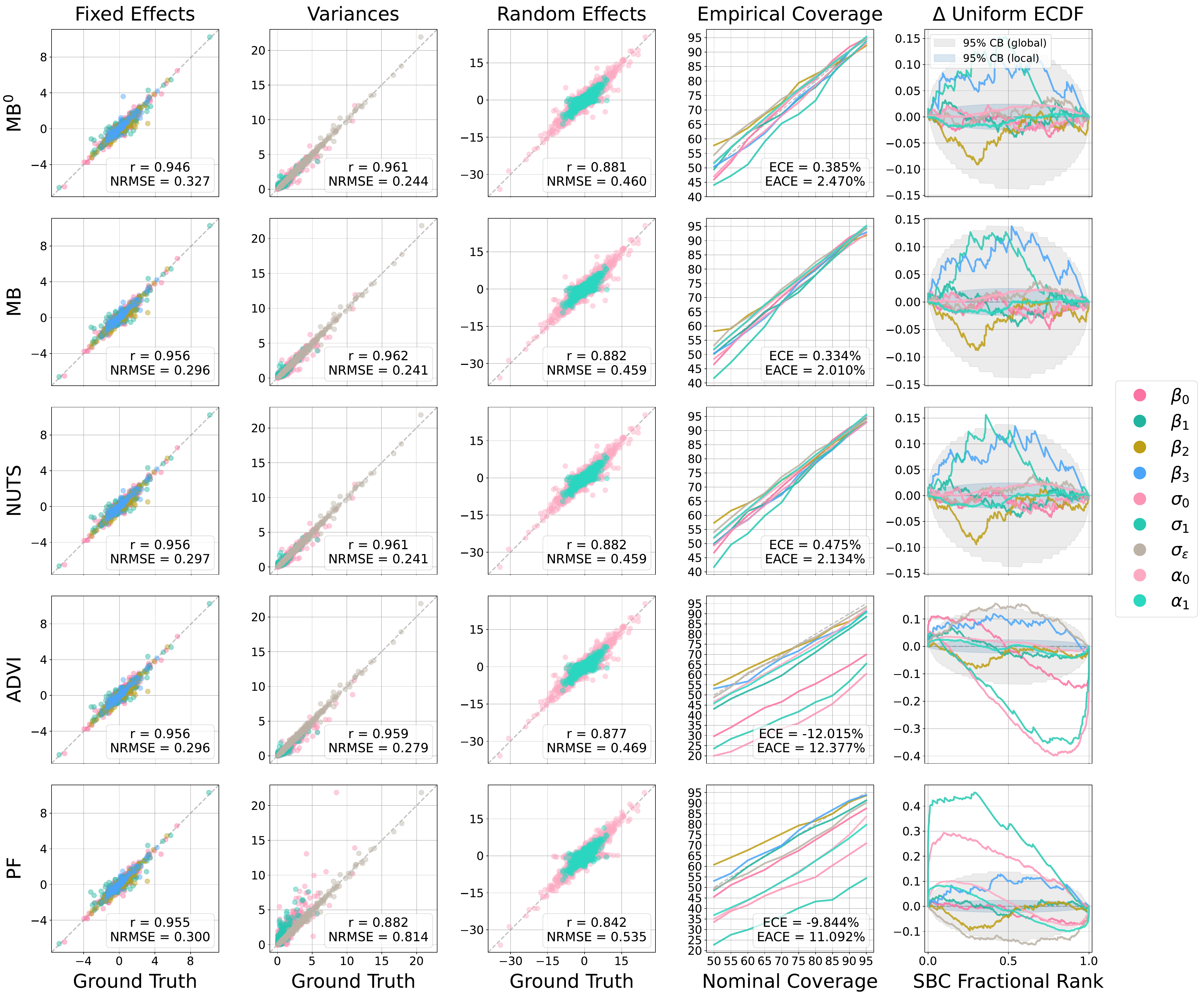}
   \caption{Performance on the \textit{small} Gaussian oracle benchmark; layout as in \myfigref{fig:2}.}
   \label{fig:normal-small}
\end{figure}
\begin{figure}[p]
	\centering
	\includegraphics[width=1.0\textwidth, page=1]{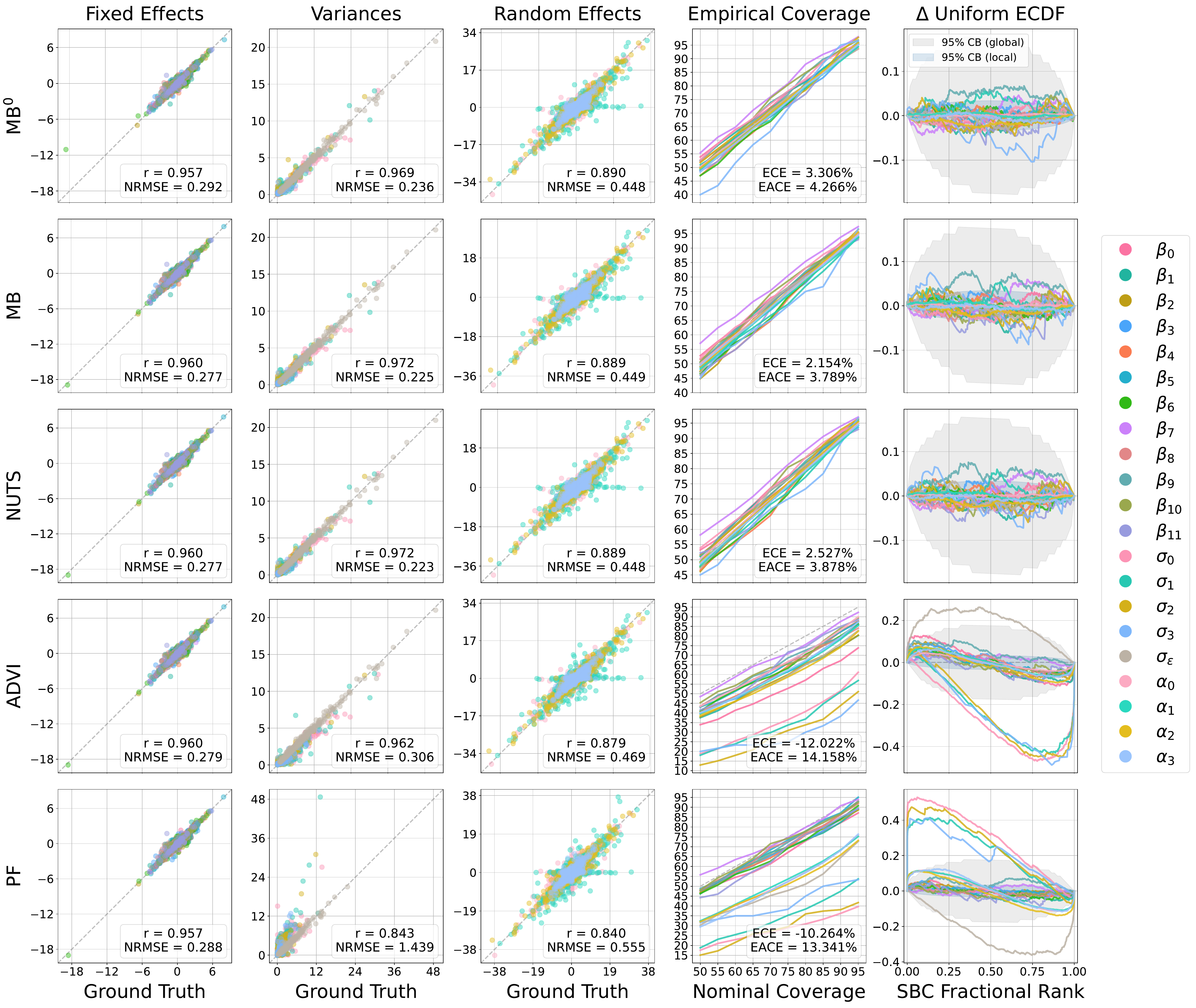}
   \caption{Performance on the \textit{large} Gaussian oracle benchmark; layout as in \myfigref{fig:2}.}
   \label{fig:normal-large}
\end{figure}
\begin{figure}[p]
	\centering
	\includegraphics[width=1.0\textwidth, page=1]{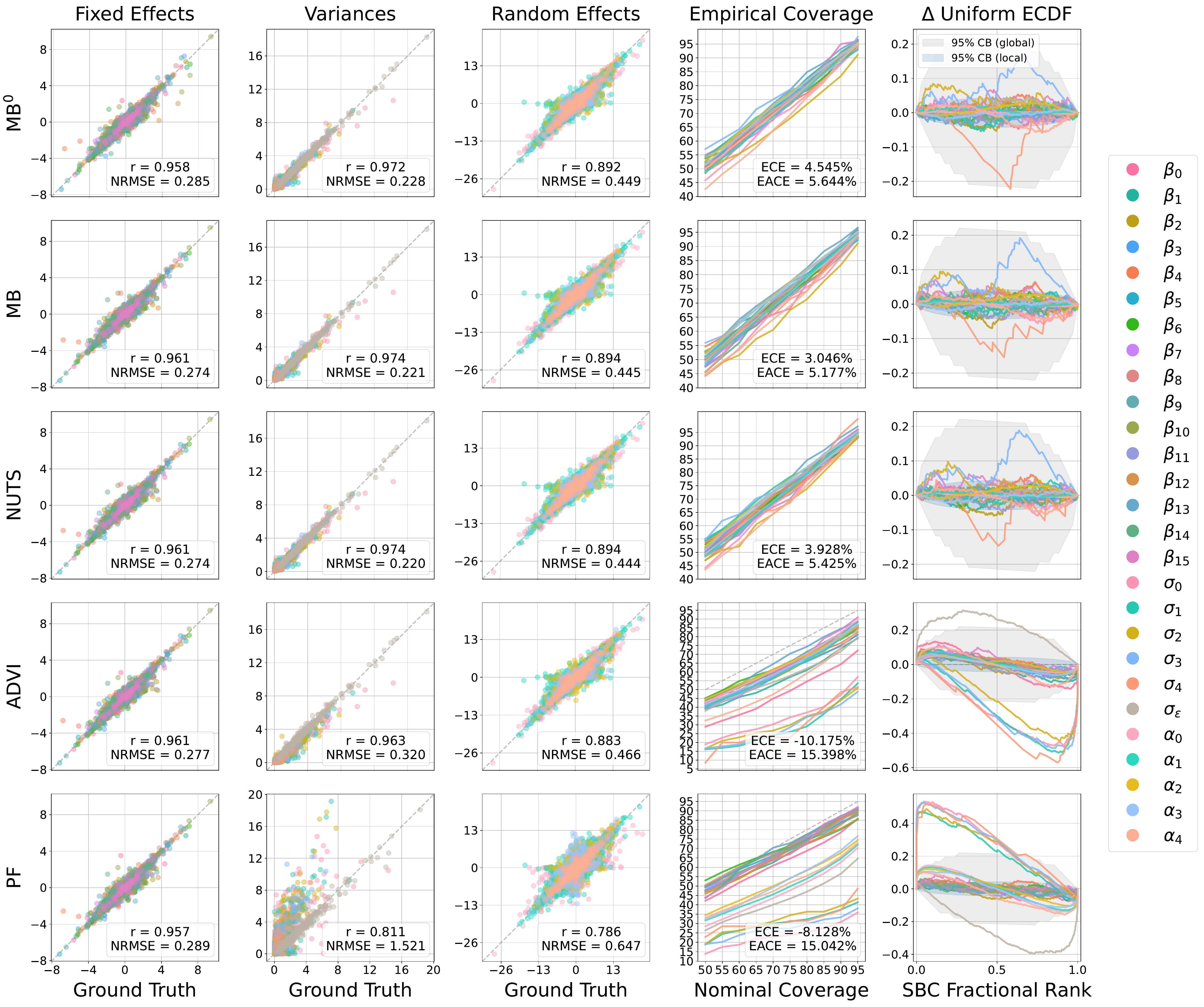}
   \caption{Performance on the \textit{huge} Gaussian oracle benchmark; layout as in \myfigref{fig:2}.}
   \label{fig:normal-huge}
\end{figure}
\clearpage

\begin{table}[p]
  \caption{
     Performance on each Bernoulli oracle test set averaged over all parameters.
     Layout as in \mytabref{tab:oracle}.
  }
  \label{tab:oracle_b}
  \centering
  \apptable
  \begin{tabular}{cc|cccccc}
    \toprule
    $\mathrm{regime}$ & $\mathrm{model}$ & $r$ & $\mathrm{NRMSE}$ & $\mathrm{ECE}$ & $\mathrm{EACE}$ & $\mathrm{LOO\text{-}NLL}$ & $\mathrm{time}$ \\
    \midrule
      \texttt{small} & \texttt{MB} & $0.89 \pm 0.09$ & $0.42 \pm 0.17$ & $\phantom{-}0.00 \pm 0.02$ & $0.02 \pm 0.01$ & $0.55 \pm 0.09$ & $\phantom{0}0.10 \pm \phantom{0}0.00$ \\
          & \texttt{NUTS} & $0.89 \pm 0.09$ & $0.43 \pm 0.17$ & $\phantom{-}0.00 \pm 0.02$ & $0.02 \pm 0.01$ & $0.55 \pm 0.09$ & $62.79 \pm 21.21$ \\
          & \texttt{ADVI} & $0.88 \pm 0.10$ & $0.45 \pm 0.18$ & $-0.12 \pm 0.08$ & $0.12 \pm 0.08$ & $0.57 \pm 0.09$ & $29.15 \pm \phantom{0}3.79$ \\
          & \texttt{PF} & $0.84 \pm 0.13$ & $0.57 \pm 0.28$ & $-0.09 \pm 0.10$ & $0.10 \pm 0.09$ & $0.56 \pm 0.09$ & $18.37 \pm \phantom{0}1.90$ \\
    \midrule
      \texttt{medium} & \texttt{MB} & $0.88 \pm 0.10$ & $0.43 \pm 0.18$ & $-0.01 \pm 0.02$ & $0.02 \pm 0.01$ & $0.45 \pm 0.06$ & $\phantom{0}0.11 \pm \phantom{0}0.01$ \\
          & \texttt{NUTS} & $0.88 \pm 0.10$ & $0.43 \pm 0.18$ & $-0.01 \pm 0.02$ & $0.02 \pm 0.01$ & $0.45 \pm 0.06$ & $78.14 \pm 28.35$ \\
          & \texttt{ADVI} & $0.88 \pm 0.11$ & $0.44 \pm 0.20$ & $-0.13 \pm 0.08$ & $0.13 \pm 0.08$ & $0.47 \pm 0.06$ & $37.28 \pm \phantom{0}8.43$ \\
          & \texttt{PF} & $0.78 \pm 0.24$ & $0.70 \pm 0.57$ & $-0.06 \pm 0.09$ & $0.08 \pm 0.07$ & $0.47 \pm 0.07$ & $26.90 \pm \phantom{0}6.12$ \\
    \midrule
      \texttt{large} & \texttt{MB} & $0.88 \pm 0.12$ & $0.43 \pm 0.18$ & $\phantom{-}0.00 \pm 0.02$ & $0.02 \pm 0.01$ & $0.43 \pm 0.05$ & $\phantom{0}0.11 \pm \phantom{0}0.01$ \\
          & \texttt{NUTS} & $0.88 \pm 0.12$ & $0.43 \pm 0.18$ & $\phantom{-}0.01 \pm 0.02$ & $0.02 \pm 0.01$ & $0.43 \pm 0.06$ & $71.75 \pm 21.17$ \\
          & \texttt{ADVI} & $0.87 \pm 0.13$ & $0.45 \pm 0.20$ & $-0.14 \pm 0.09$ & $0.14 \pm 0.09$ & $0.45 \pm 0.06$ & $42.81 \pm 10.64$ \\
          & \texttt{PF} & $0.80 \pm 0.20$ & $0.75 \pm 0.61$ & $-0.06 \pm 0.10$ & $0.08 \pm 0.08$ & $0.46 \pm 0.07$ & $35.49 \pm \phantom{0}9.65$ \\
    \midrule
      \texttt{huge} & \texttt{MB} & $0.86 \pm 0.12$ & $0.46 \pm 0.18$ & $\phantom{-}0.00 \pm 0.03$ & $0.02 \pm 0.02$ & $0.41 \pm 0.05$ & $\phantom{0}0.12 \pm \phantom{0}0.01$ \\
          & \texttt{NUTS} & $0.86 \pm 0.12$ & $0.46 \pm 0.18$ & $\phantom{-}0.00 \pm 0.03$ & $0.02 \pm 0.02$ & $0.41 \pm 0.05$ & $87.61 \pm 29.52$ \\
          & \texttt{ADVI} & $0.85 \pm 0.14$ & $0.49 \pm 0.22$ & $-0.14 \pm 0.08$ & $0.14 \pm 0.08$ & $0.43 \pm 0.06$ & $47.89 \pm 11.70$ \\
          & \texttt{PF} & $0.78 \pm 0.20$ & $0.85 \pm 0.76$ & $-0.08 \pm 0.10$ & $0.08 \pm 0.10$ & $0.44 \pm 0.07$ & $42.06 \pm 11.08$ \\
    \bottomrule
\end{tabular}

\end{table}
\begin{figure}[p]
	\centering
	\includegraphics[width=1.0\textwidth, page=1]{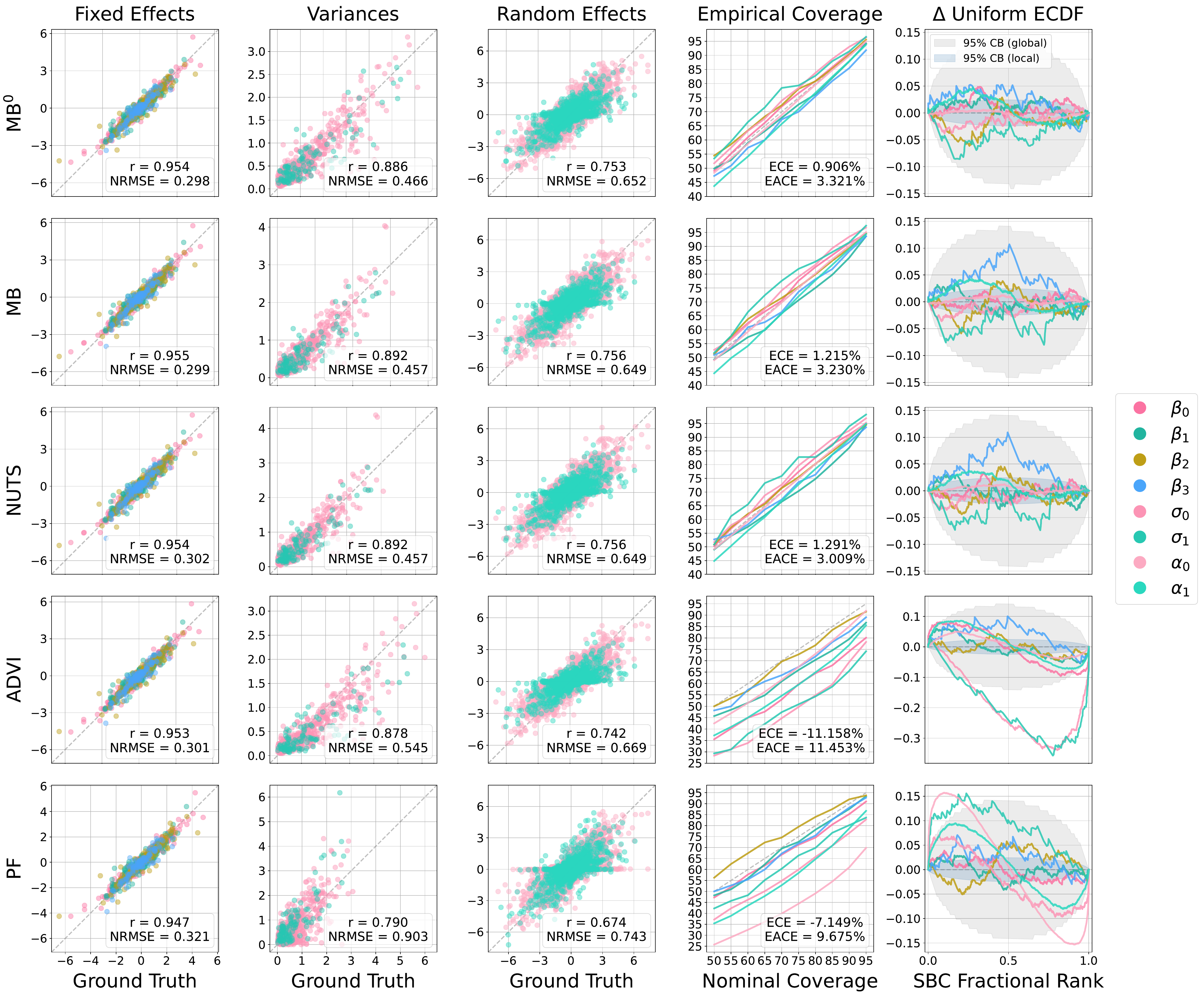}
   \caption{Performance on the \textit{small} Bernoulli oracle benchmark; layout as in \myfigref{fig:2}.}
   \label{fig:bernoulli-small}
\end{figure}
\begin{figure}[p]
	\centering
	\includegraphics[width=1.0\textwidth, page=1]{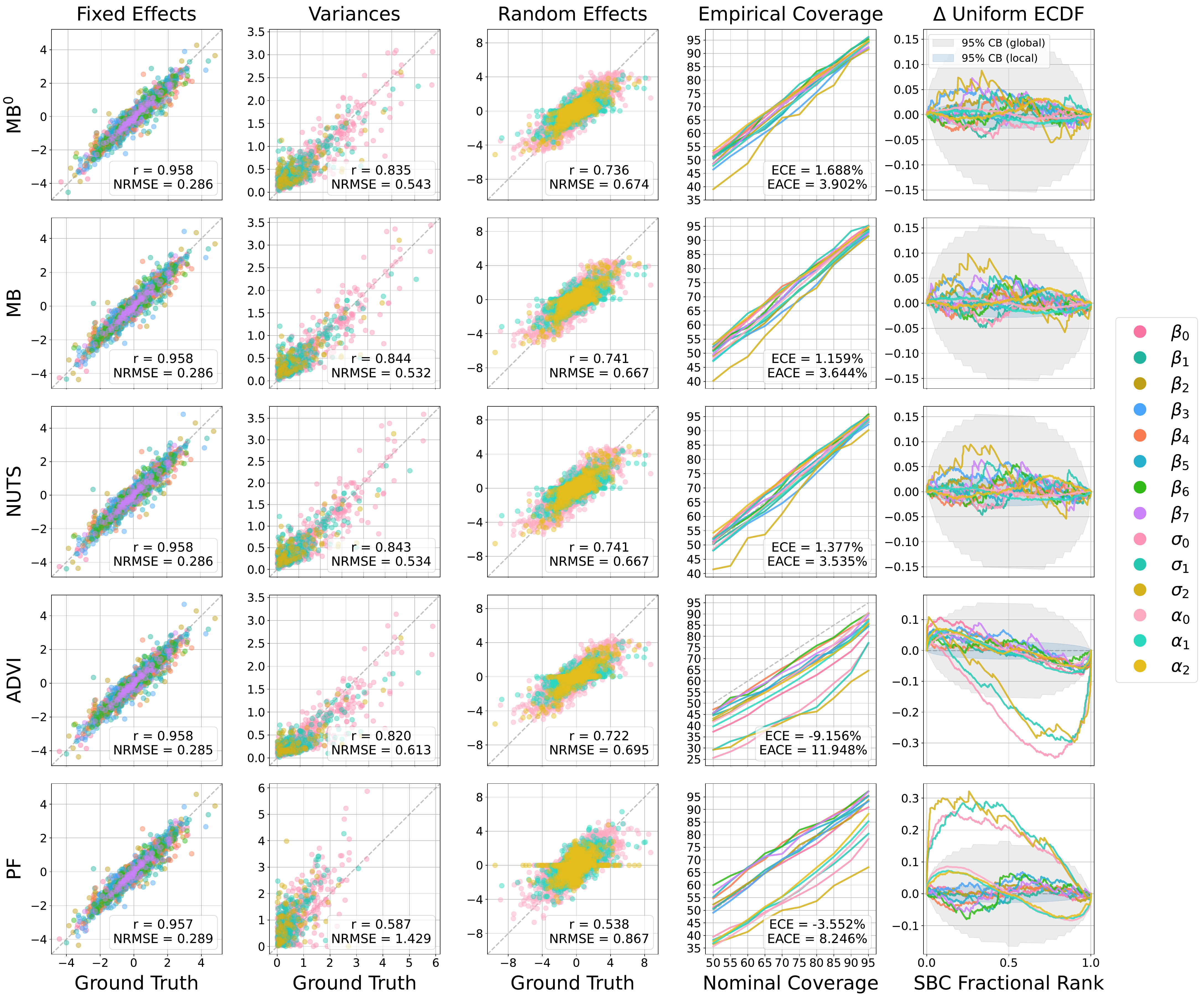}
   \caption{Performance on the \textit{medium} Bernoulli oracle benchmark; layout as in \myfigref{fig:2}.}
   \label{fig:bernoulli-medium}
\end{figure}
\begin{figure}[p]
	\centering
	\includegraphics[width=1.0\textwidth, page=1]{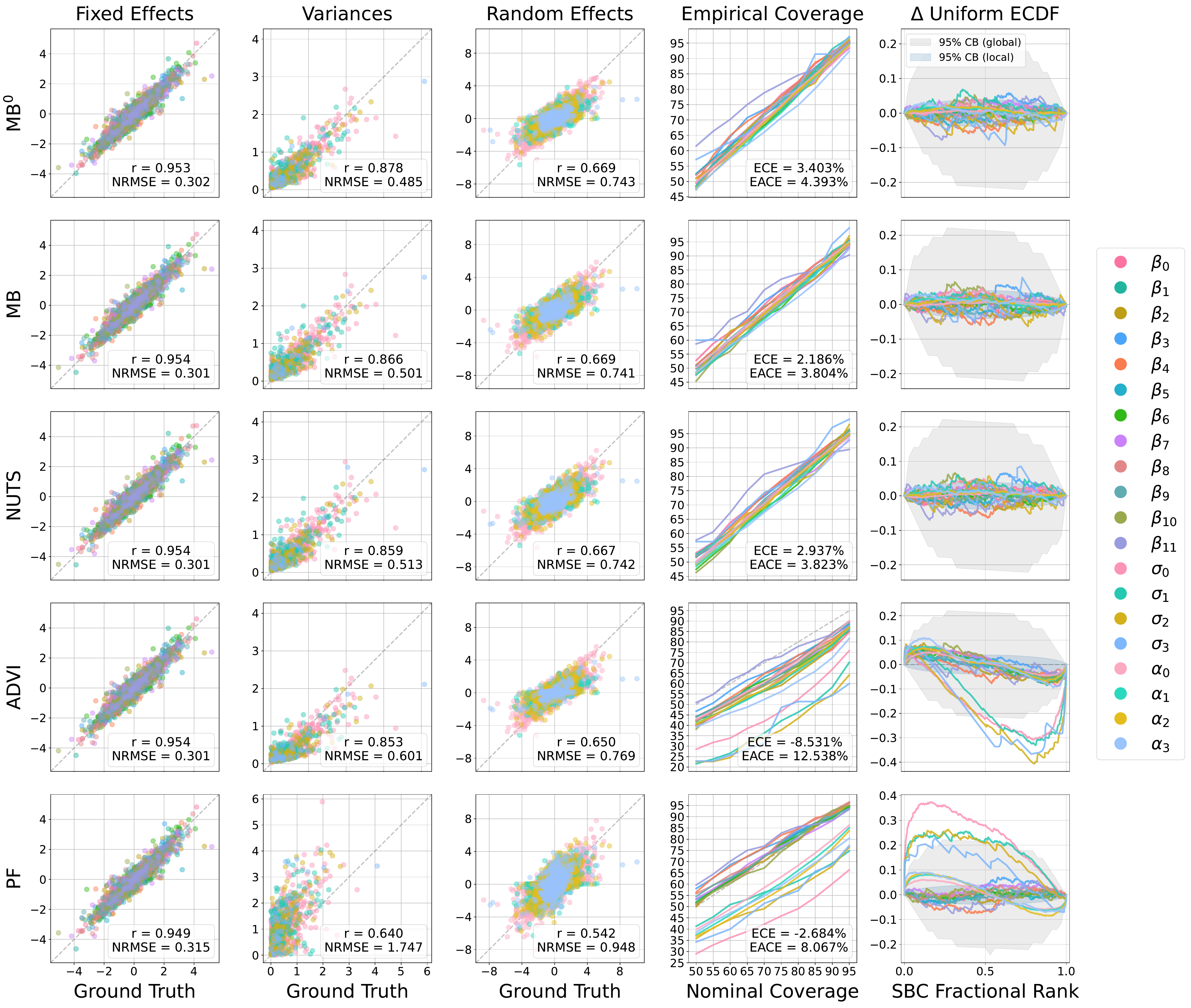}
   \caption{Performance on the \textit{large} Bernoulli oracle benchmark; layout as in \myfigref{fig:2}.}
   \label{fig:bernoulli-large}
\end{figure}
\begin{figure}[p]
	\centering
	\includegraphics[width=1.0\textwidth, page=1]{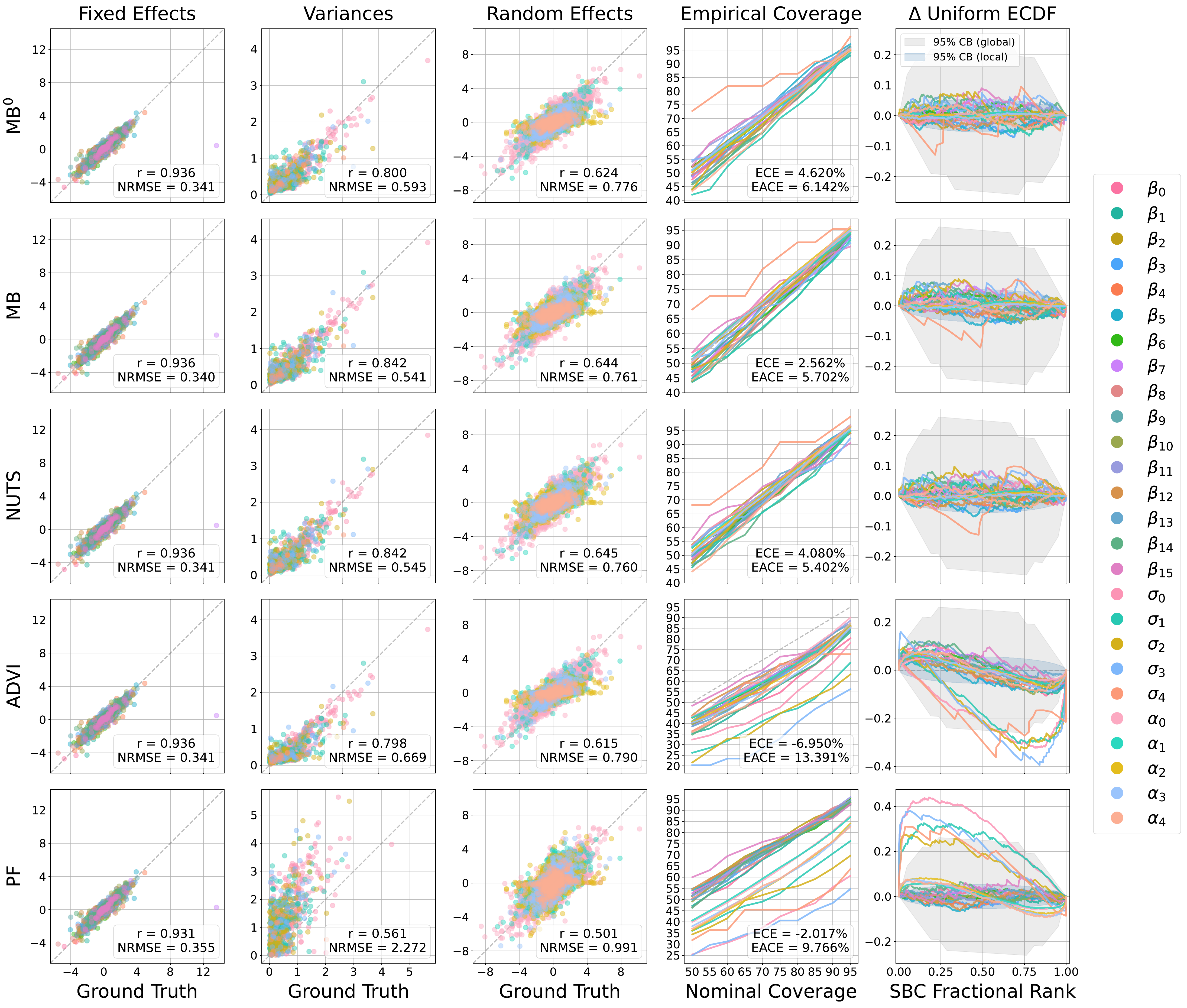}
   \caption{Performance on the \textit{huge} Bernoulli oracle benchmark; layout as in \myfigref{fig:2}.}
   \label{fig:bernoulli-huge}
\end{figure}
\clearpage

\begin{table}[p]
  \caption{
     Performance on each Poisson oracle test set averaged over all parameters.
     Layout as in \mytabref{tab:oracle}.
  }
  \label{tab:oracle_p}
  \centering
  \apptable
  \begin{tabular}{cc|cccccc}
    \toprule
    $\mathrm{regime}$ & $\mathrm{model}$ & $r$ & $\mathrm{NRMSE}$ & $\mathrm{ECE}$ & $\mathrm{EACE}$ & $\mathrm{LOO\text{-}NLL}$ & $\mathrm{time}$ \\
    \midrule
      \texttt{small} & \texttt{MB} & $0.92 \pm 0.07$ & $0.35 \pm 0.16$ & $\phantom{-}0.00 \pm 0.04$ & $0.03 \pm 0.02$ & $1.34 \pm 0.22$ & $\phantom{0}0.11 \pm \phantom{0}0.00$ \\
          & \texttt{NUTS} & $0.92 \pm 0.07$ & $0.35 \pm 0.16$ & $\phantom{-}0.00 \pm 0.04$ & $0.03 \pm 0.02$ & $1.34 \pm 0.22$ & $89.15 \pm 43.89$ \\
          & \texttt{ADVI} & $0.91 \pm 0.08$ & $0.39 \pm 0.18$ & $-0.19 \pm 0.11$ & $0.19 \pm 0.11$ & $1.39 \pm 0.25$ & $26.88 \pm \phantom{0}2.35$ \\
          & \texttt{PF} & $0.87 \pm 0.11$ & $0.53 \pm 0.34$ & $-0.10 \pm 0.11$ & $0.12 \pm 0.08$ & $1.36 \pm 0.23$ & $18.08 \pm \phantom{0}1.82$ \\
    \midrule
      \texttt{medium} & \texttt{MB} & $0.93 \pm 0.07$ & $0.33 \pm 0.16$ & $\phantom{-}0.00 \pm 0.02$ & $0.02 \pm 0.01$ & $1.36 \pm 0.18$ & $\phantom{0}0.11 \pm \phantom{0}0.01$ \\
          & \texttt{NUTS} & $0.93 \pm 0.07$ & $0.33 \pm 0.16$ & $\phantom{-}0.00 \pm 0.02$ & $0.02 \pm 0.01$ & $1.36 \pm 0.17$ & $84.47 \pm 36.42$ \\
          & \texttt{ADVI} & $0.92 \pm 0.08$ & $0.36 \pm 0.19$ & $-0.21 \pm 0.11$ & $0.21 \pm 0.11$ & $1.46 \pm 0.22$ & $33.10 \pm \phantom{0}5.40$ \\
          & \texttt{PF} & $0.88 \pm 0.12$ & $0.64 \pm 0.61$ & $-0.12 \pm 0.12$ & $0.12 \pm 0.11$ & $1.39 \pm 0.19$ & $26.88 \pm \phantom{0}5.90$ \\
    \midrule
      \texttt{large} & \texttt{MB} & $0.94 \pm 0.07$ & $0.30 \pm 0.16$ & $-0.01 \pm 0.02$ & $0.02 \pm 0.01$ & $1.39 \pm 0.16$ & $\phantom{0}0.12 \pm \phantom{0}0.01$ \\
          & \texttt{NUTS} & $0.94 \pm 0.07$ & $0.30 \pm 0.16$ & $-0.01 \pm 0.02$ & $0.02 \pm 0.01$ & $1.39 \pm 0.16$ & $85.68 \pm 32.25$ \\
          & \texttt{ADVI} & $0.93 \pm 0.08$ & $0.34 \pm 0.18$ & $-0.25 \pm 0.13$ & $0.25 \pm 0.13$ & $1.48 \pm 0.20$ & $37.25 \pm \phantom{0}7.41$ \\
          & \texttt{PF} & $0.89 \pm 0.12$ & $0.58 \pm 0.57$ & $-0.14 \pm 0.11$ & $0.14 \pm 0.11$ & $1.48 \pm 0.19$ & $33.86 \pm \phantom{0}8.01$ \\
    \midrule
      \texttt{huge} & \texttt{MB} & $0.92 \pm 0.09$ & $0.33 \pm 0.18$ & $-0.02 \pm 0.03$ & $0.03 \pm 0.02$ & $1.40 \pm 0.14$ & $\phantom{0}0.12 \pm \phantom{0}0.01$ \\
          & \texttt{NUTS} & $0.92 \pm 0.09$ & $0.33 \pm 0.18$ & $\phantom{-}0.00 \pm 0.02$ & $0.02 \pm 0.01$ & $1.40 \pm 0.13$ & $91.84 \pm 34.55$ \\
          & \texttt{ADVI} & $0.90 \pm 0.12$ & $0.38 \pm 0.22$ & $-0.25 \pm 0.10$ & $0.25 \pm 0.10$ & $1.51 \pm 0.18$ & $43.66 \pm 10.13$ \\
          & \texttt{PF} & $0.87 \pm 0.14$ & $0.70 \pm 0.76$ & $-0.15 \pm 0.11$ & $0.15 \pm 0.11$ & $1.49 \pm 0.18$ & $43.24 \pm 11.32$ \\
    \bottomrule
\end{tabular}

\end{table}
\begin{figure}[p]
	\centering
	\includegraphics[width=1.0\textwidth, page=1]{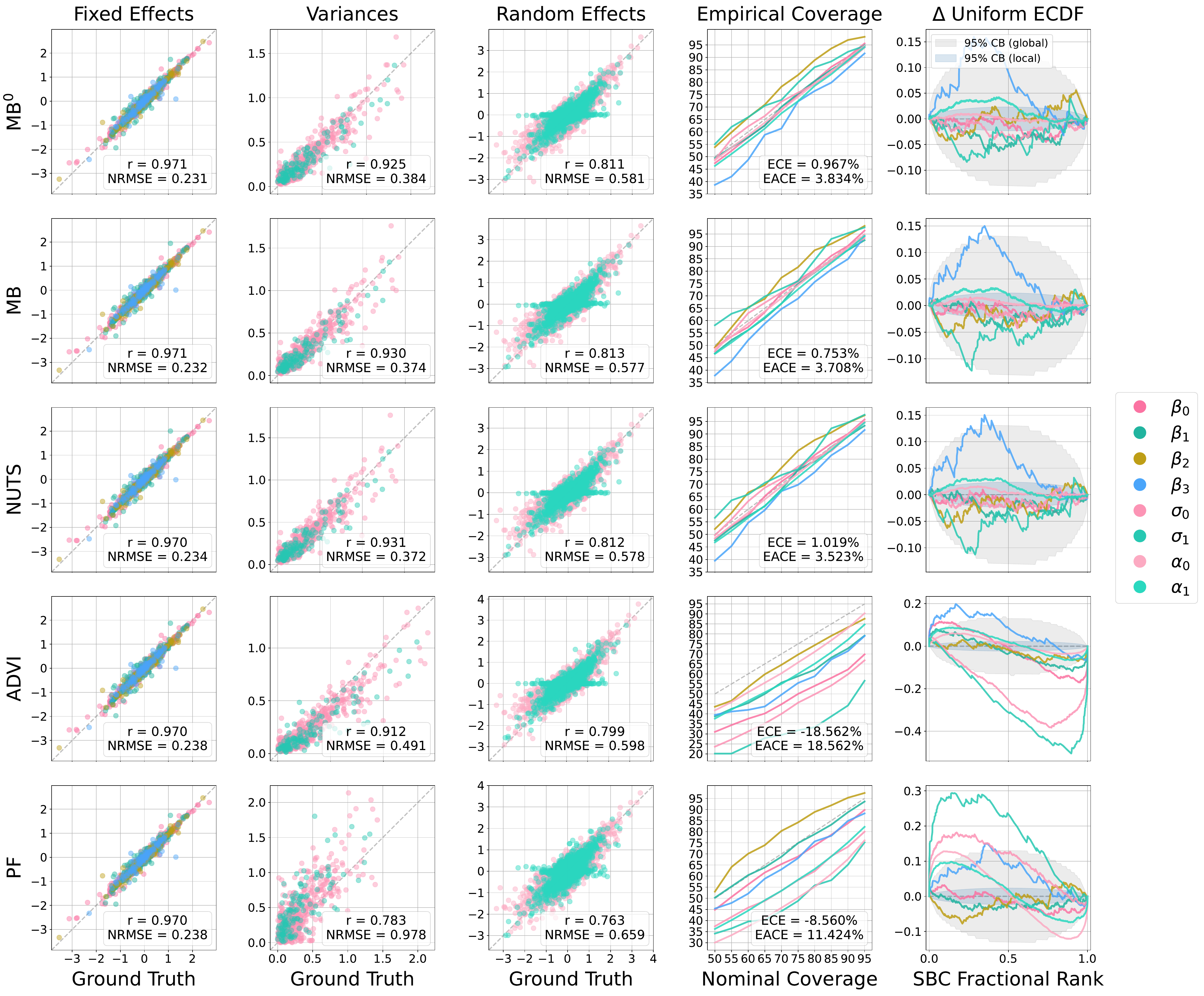}
   \caption{Performance on the \textit{small} Poisson oracle benchmark; layout as in \myfigref{fig:2}.}
   \label{fig:poisson-small}
\end{figure}
\begin{figure}[p]
	\centering
	\includegraphics[width=1.0\textwidth, page=1]{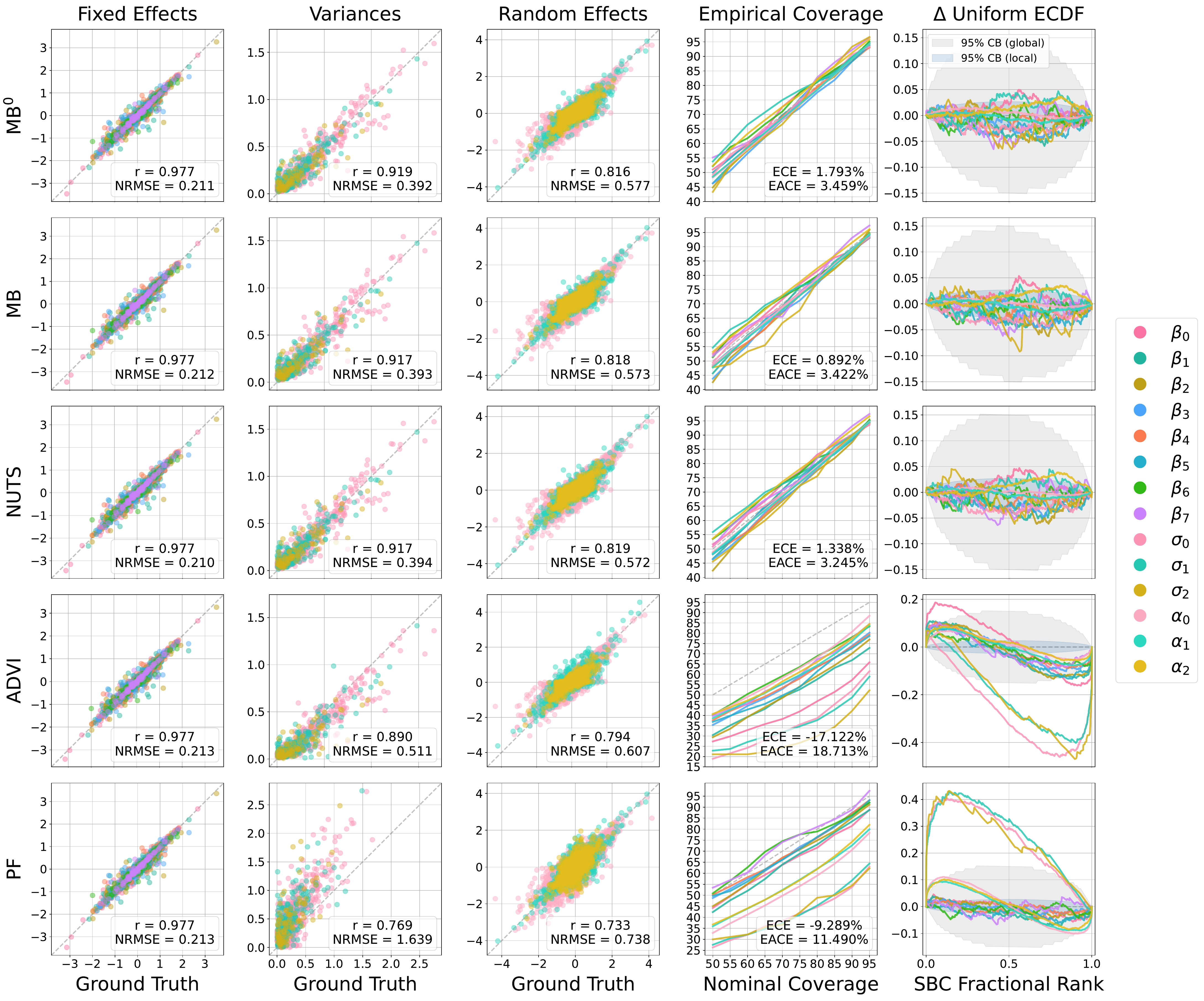}
   \caption{Performance on the \textit{medium} Poisson oracle benchmark; layout as in \myfigref{fig:2}.}
   \label{fig:poisson-medium}
\end{figure}
\begin{figure}[p]
	\centering
	\includegraphics[width=1.0\textwidth, page=1]{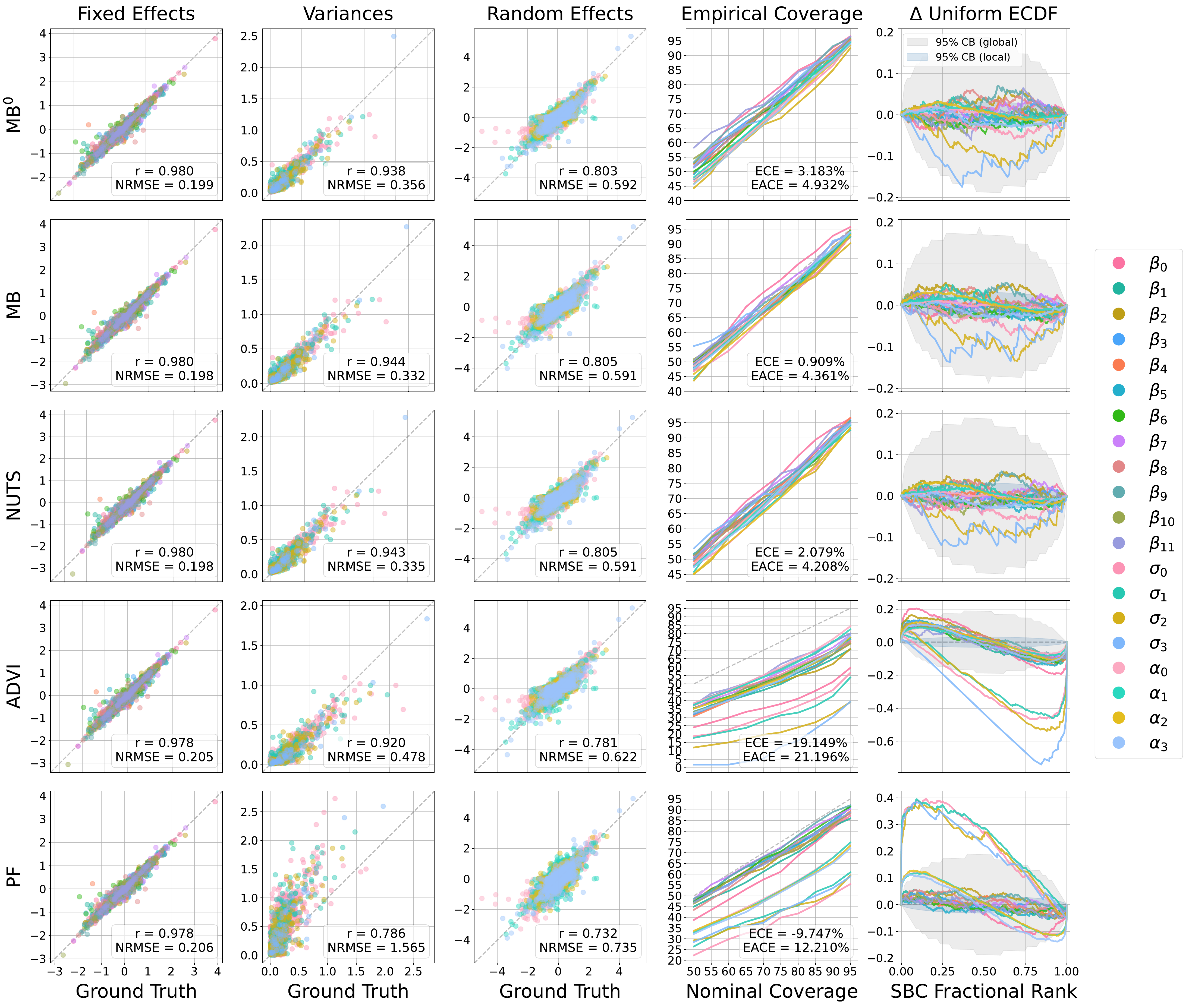}
   \caption{Performance on the \textit{large} Poisson oracle benchmark; layout as in \myfigref{fig:2}.}
   \label{fig:poisson-large}
\end{figure}
\begin{figure}[p]
	\centering
	\includegraphics[width=1.0\textwidth, page=1]{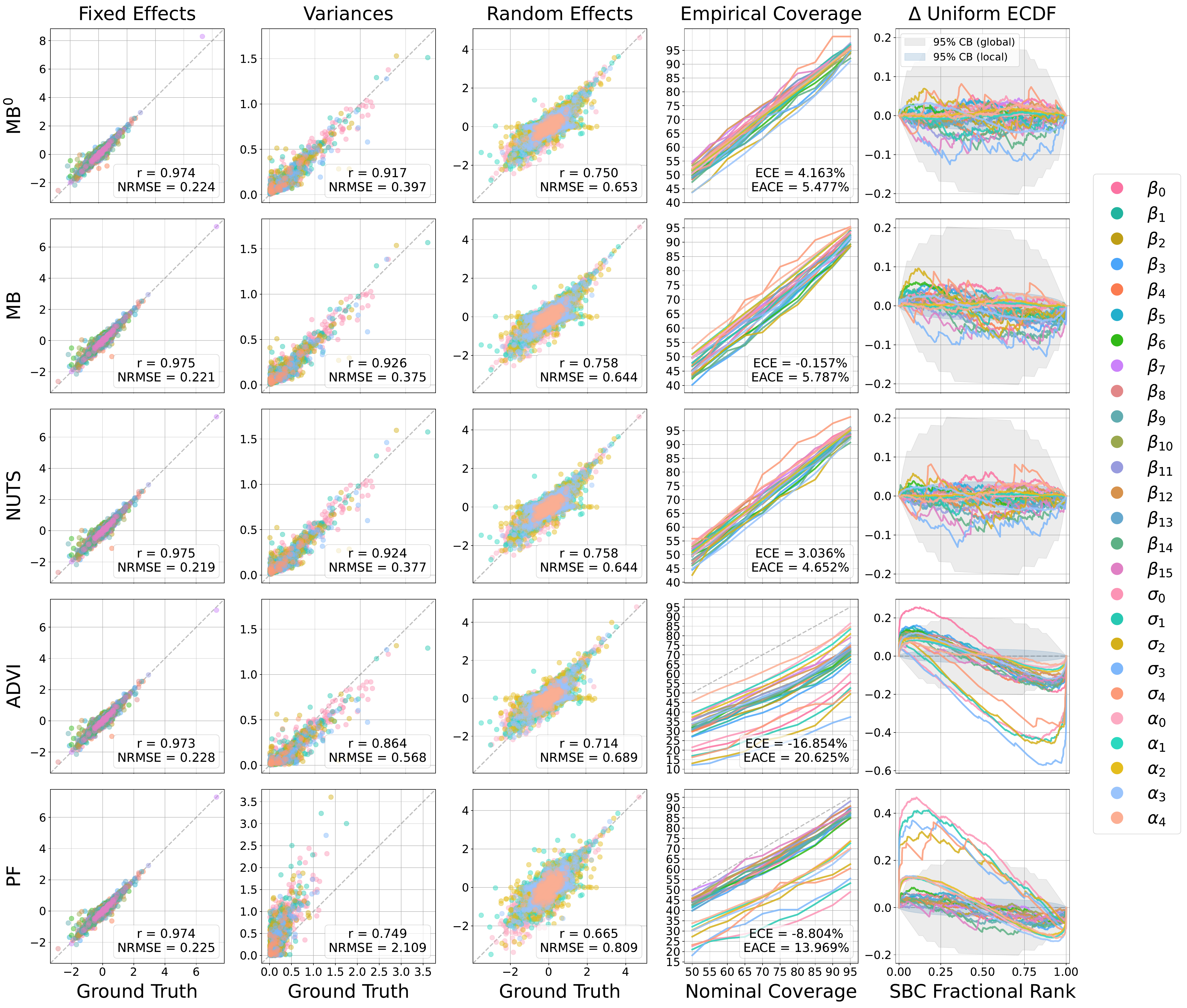}
   \caption{Performance on the \textit{huge} Poisson oracle benchmark; layout as in \myfigref{fig:2}.}
   \label{fig:poisson-huge}
\end{figure}
\clearpage

\subsection{Random-Effect Correlations} \label{app:corr}
\begin{figure}[h]
    \centering
    \includegraphics[width=1.0\textwidth]{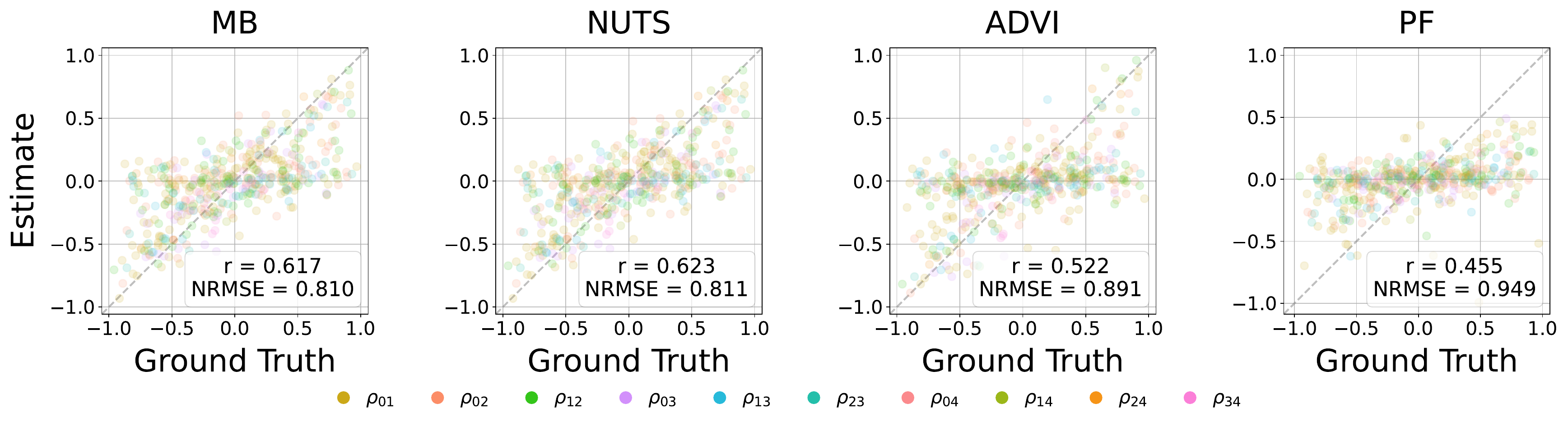}
    \caption{
       Recovery of the random-effect correlations on the \textit{huge} Gaussian oracle benchmark (correlated, NUTS-converged datasets): posterior mean against ground truth, one color per pair $\rho_{jk}$.
       All methods shrink their estimates toward zero; \mb and NUTS are nearly indistinguishable, ADVI loses the large $|\rho|$.
    }
    \label{fig:corr}
\end{figure}

\mytabref{tab:oracle}, \mytabref{tab:oracle_b} and \mytabref{tab:oracle_p} exclude the random-effect correlations ($\mathbf R$) from their averages. \mytabref{tab:corr} reports them separately for two reasons:
(1) The correlations are the least identified GLMM parameters (each is estimated from $m$ noisy pairs of group-level effects), and the up to $q(q-1)/2$ pairs would dominate the dimension-weighted averages of the oracle tables.
(2) NUTS notably struggles here: of the $110$--$200$ datasets per test set with correlated random effects (\myappref{app:prior}), NUTS converges on only $74$--$95\%$ (against $91$--$100\%$ with independent random effects) under the convergence criterion of \myappref{app:tra}.
We restrict the evaluation to the correlated datasets NUTS converged on ($101$--$176$ per regime), computing recovery and calibration per correlation pair and reporting mean $\pm$ SD over the pairs (a single pair in \textit{small}, no spread possible; up to ten in \textit{huge}).
\myfigref{fig:corr} shows the recovery for the \textit{huge} Gaussian benchmark.

All inference methods recover the correlations substantially worse than any other parameter class.
On the Gaussian sets, \mb and NUTS reach $r = 0.62$--$0.68$ and NRMSE $0.74$--$0.81$ (against $r \ge 0.94$ and NRMSE $\le 0.31$ for the parameter average in \mytabref{tab:oracle}).
The Poisson sets are somewhat lower ($r = 0.46$--$0.63$), and the Bernoulli sets degrade with size, down to $r = 0.32$ for NUTS and $0.34$ for \mb in the \textit{huge} regime, where the spread over pairs ($\pm 0.16$--$0.19$) is of the order of the mean.
Throughout, \mb tracks NUTS within $0.01$ in $r$ on the Gaussian sets and within $0.02$ on the discrete sets, sharing the common failure mode of \myfigref{fig:corr}: every method shrinks the posterior means toward zero relative to the ground truth.
The raw flow (\MBz{}) is the best-calibrated method for the correlations (EACE $\le 0.07$ throughout, ECE between $-0.06$ and $0.05$); the IMH head brings the correlations to the calibration of NUTS (ECE within $0.02$ of NUTS in every regime).
ADVI remains strongly overconfident (ECE $-0.16$ to $-0.25$ on the Gaussian and Poisson sets) and recovers the correlations worse than NUTS in every regime but one ($r$ lower by up to $0.19$).
PF is the weakest method on the correlations: $r = 0.45$--$0.54$ on the Gaussian and $0.19$--$0.52$ on the discrete sets, NRMSE $\ge 0.85$ throughout, and overconfident in every regime except the \textit{small} discrete ones (ECE $-0.07$ to $-0.22$).

\begin{table}[h]
  \caption{
     Recovery and calibration of the random-effect correlations $\rho_{jk}$ on the oracle test sets (correlated, NUTS-converged datasets; see text).
     Entries are mean $\pm$ SD over the $q(q-1)/2$ correlation pairs; metrics as in \mytabref{tab:oracle}; \MBz{} is the raw flow without the IMH head (\myappref{app:abl}).
  }
  \label{tab:corr}
  \centering
  \apptable\renewcommand{\arraystretch}{0.92}
  \begin{tabular}{ccc|cccc}
    \toprule
    $\mathrm{family}$ & $\mathrm{regime}$ & $\mathrm{model}$ & $r$ & $\mathrm{NRMSE}$ & $\mathrm{ECE}$ & $\mathrm{EACE}$ \\
    \midrule
      Gaussian & \texttt{small} & \MBz{} & $0.66 \pm 0.00$ & $0.75 \pm 0.00$ & $-0.01 \pm 0.00$ & $0.01 \pm 0.00$ \\
               &      & \texttt{MB} & $0.67 \pm 0.00$ & $0.75 \pm 0.00$ & $-0.01 \pm 0.00$ & $0.01 \pm 0.00$ \\
               &      & \texttt{NUTS} & $0.67 \pm 0.00$ & $0.75 \pm 0.00$ & $-0.01 \pm 0.00$ & $0.02 \pm 0.00$ \\
               &      & \texttt{ADVI} & $0.66 \pm 0.00$ & $0.76 \pm 0.00$ & $-0.18 \pm 0.00$ & $0.18 \pm 0.00$ \\
               &      & \texttt{PF} & $0.54 \pm 0.00$ & $0.85 \pm 0.00$ & $-0.12 \pm 0.00$ & $0.12 \pm 0.00$ \\
    \midrule
               & \texttt{medium} & \MBz{} & $0.66 \pm 0.06$ & $0.76 \pm 0.04$ & $\phantom{-}0.01 \pm 0.02$ & $0.03 \pm 0.01$ \\
               &      & \texttt{MB} & $0.68 \pm 0.07$ & $0.74 \pm 0.06$ & $-0.02 \pm 0.01$ & $0.03 \pm 0.01$ \\
               &      & \texttt{NUTS} & $0.68 \pm 0.07$ & $0.74 \pm 0.06$ & $-0.02 \pm 0.01$ & $0.03 \pm 0.02$ \\
               &      & \texttt{ADVI} & $0.59 \pm 0.09$ & $0.81 \pm 0.07$ & $-0.23 \pm 0.01$ & $0.23 \pm 0.01$ \\
               &      & \texttt{PF} & $0.49 \pm 0.02$ & $0.90 \pm 0.03$ & $-0.16 \pm 0.03$ & $0.16 \pm 0.03$ \\
    \midrule
               & \texttt{large} & \MBz{} & $0.61 \pm 0.11$ & $0.81 \pm 0.07$ & $-0.01 \pm 0.06$ & $0.05 \pm 0.03$ \\
               &      & \texttt{MB} & $0.63 \pm 0.08$ & $0.80 \pm 0.05$ & $-0.05 \pm 0.06$ & $0.06 \pm 0.05$ \\
               &      & \texttt{NUTS} & $0.62 \pm 0.08$ & $0.80 \pm 0.06$ & $-0.05 \pm 0.06$ & $0.06 \pm 0.05$ \\
               &      & \texttt{ADVI} & $0.50 \pm 0.07$ & $0.89 \pm 0.05$ & $-0.21 \pm 0.05$ & $0.21 \pm 0.05$ \\
               &      & \texttt{PF} & $0.45 \pm 0.15$ & $0.92 \pm 0.06$ & $-0.16 \pm 0.06$ & $0.16 \pm 0.06$ \\
    \midrule
               & \texttt{huge} & \MBz{} & $0.57 \pm 0.14$ & $0.85 \pm 0.16$ & $\phantom{-}0.02 \pm 0.09$ & $0.07 \pm 0.06$ \\
               &      & \texttt{MB} & $0.62 \pm 0.16$ & $0.81 \pm 0.17$ & $-0.02 \pm 0.11$ & $0.09 \pm 0.06$ \\
               &      & \texttt{NUTS} & $0.62 \pm 0.15$ & $0.81 \pm 0.16$ & $-0.01 \pm 0.11$ & $0.09 \pm 0.06$ \\
               &      & \texttt{ADVI} & $0.52 \pm 0.09$ & $0.89 \pm 0.12$ & $-0.16 \pm 0.12$ & $0.18 \pm 0.09$ \\
               &      & \texttt{PF} & $0.45 \pm 0.13$ & $0.95 \pm 0.11$ & $-0.07 \pm 0.12$ & $0.13 \pm 0.04$ \\
    \midrule
      Bernoulli & \texttt{small} & \MBz{} & $0.52 \pm 0.00$ & $0.86 \pm 0.00$ & $\phantom{-}0.05 \pm 0.00$ & $0.05 \pm 0.00$ \\
                &      & \texttt{MB} & $0.56 \pm 0.00$ & $0.83 \pm 0.00$ & $\phantom{-}0.05 \pm 0.00$ & $0.05 \pm 0.00$ \\
                &      & \texttt{NUTS} & $0.57 \pm 0.00$ & $0.82 \pm 0.00$ & $\phantom{-}0.05 \pm 0.00$ & $0.05 \pm 0.00$ \\
                &      & \texttt{ADVI} & $0.45 \pm 0.00$ & $0.90 \pm 0.00$ & $-0.08 \pm 0.00$ & $0.08 \pm 0.00$ \\
                &      & \texttt{PF} & $0.44 \pm 0.00$ & $0.90 \pm 0.00$ & $\phantom{-}0.01 \pm 0.00$ & $0.03 \pm 0.00$ \\
    \midrule
                & \texttt{medium} & \MBz{} & $0.35 \pm 0.18$ & $0.94 \pm 0.06$ & $-0.01 \pm 0.05$ & $0.04 \pm 0.02$ \\
                &      & \texttt{MB} & $0.38 \pm 0.14$ & $0.93 \pm 0.05$ & $-0.06 \pm 0.05$ & $0.06 \pm 0.05$ \\
                &      & \texttt{NUTS} & $0.38 \pm 0.15$ & $0.93 \pm 0.05$ & $-0.06 \pm 0.05$ & $0.06 \pm 0.05$ \\
                &      & \texttt{ADVI} & $0.26 \pm 0.22$ & $0.97 \pm 0.06$ & $-0.10 \pm 0.08$ & $0.10 \pm 0.08$ \\
                &      & \texttt{PF} & $0.25 \pm 0.09$ & $0.98 \pm 0.02$ & $-0.14 \pm 0.04$ & $0.14 \pm 0.04$ \\
    \midrule
                & \texttt{large} & \MBz{} & $0.31 \pm 0.18$ & $0.94 \pm 0.06$ & $\phantom{-}0.02 \pm 0.02$ & $0.03 \pm 0.01$ \\
                &      & \texttt{MB} & $0.38 \pm 0.13$ & $0.93 \pm 0.06$ & $-0.04 \pm 0.04$ & $0.05 \pm 0.03$ \\
                &      & \texttt{NUTS} & $0.39 \pm 0.14$ & $0.92 \pm 0.06$ & $-0.04 \pm 0.04$ & $0.05 \pm 0.03$ \\
                &      & \texttt{ADVI} & $0.34 \pm 0.20$ & $0.95 \pm 0.06$ & $-0.10 \pm 0.09$ & $0.10 \pm 0.08$ \\
                &      & \texttt{PF} & $0.21 \pm 0.22$ & $0.99 \pm 0.06$ & $-0.15 \pm 0.04$ & $0.15 \pm 0.04$ \\
    \midrule
                & \texttt{huge} & \MBz{} & $0.20 \pm 0.19$ & $0.98 \pm 0.05$ & $-0.02 \pm 0.05$ & $0.06 \pm 0.03$ \\
                &      & \texttt{MB} & $0.34 \pm 0.16$ & $0.95 \pm 0.06$ & $-0.09 \pm 0.05$ & $0.09 \pm 0.05$ \\
                &      & \texttt{NUTS} & $0.32 \pm 0.19$ & $0.95 \pm 0.07$ & $-0.08 \pm 0.05$ & $0.09 \pm 0.04$ \\
                &      & \texttt{ADVI} & $0.33 \pm 0.21$ & $1.00 \pm 0.15$ & $-0.11 \pm 0.09$ & $0.13 \pm 0.05$ \\
                &      & \texttt{PF} & $0.19 \pm 0.30$ & $0.99 \pm 0.08$ & $-0.12 \pm 0.05$ & $0.12 \pm 0.04$ \\
    \midrule
      Poisson & \texttt{small} & \MBz{} & $0.62 \pm 0.00$ & $0.79 \pm 0.00$ & $\phantom{-}0.05 \pm 0.00$ & $0.05 \pm 0.00$ \\
              &      & \texttt{MB} & $0.62 \pm 0.00$ & $0.79 \pm 0.00$ & $\phantom{-}0.03 \pm 0.00$ & $0.03 \pm 0.00$ \\
              &      & \texttt{NUTS} & $0.62 \pm 0.00$ & $0.79 \pm 0.00$ & $\phantom{-}0.03 \pm 0.00$ & $0.03 \pm 0.00$ \\
              &      & \texttt{ADVI} & $0.43 \pm 0.00$ & $0.93 \pm 0.00$ & $-0.24 \pm 0.00$ & $0.24 \pm 0.00$ \\
              &      & \texttt{PF} & $0.52 \pm 0.00$ & $0.85 \pm 0.00$ & $-0.02 \pm 0.00$ & $0.05 \pm 0.00$ \\
    \midrule
              & \texttt{medium} & \MBz{} & $0.45 \pm 0.11$ & $0.91 \pm 0.04$ & $-0.06 \pm 0.03$ & $0.07 \pm 0.03$ \\
              &      & \texttt{MB} & $0.46 \pm 0.07$ & $0.90 \pm 0.02$ & $-0.11 \pm 0.05$ & $0.11 \pm 0.05$ \\
              &      & \texttt{NUTS} & $0.47 \pm 0.07$ & $0.90 \pm 0.02$ & $-0.11 \pm 0.06$ & $0.11 \pm 0.06$ \\
              &      & \texttt{ADVI} & $0.42 \pm 0.08$ & $0.93 \pm 0.01$ & $-0.24 \pm 0.04$ & $0.24 \pm 0.04$ \\
              &      & \texttt{PF} & $0.38 \pm 0.09$ & $0.96 \pm 0.01$ & $-0.22 \pm 0.07$ & $0.22 \pm 0.07$ \\
    \midrule
              & \texttt{large} & \MBz{} & $0.62 \pm 0.08$ & $0.79 \pm 0.06$ & $\phantom{-}0.01 \pm 0.03$ & $0.03 \pm 0.02$ \\
              &      & \texttt{MB} & $0.63 \pm 0.10$ & $0.79 \pm 0.08$ & $-0.04 \pm 0.03$ & $0.04 \pm 0.02$ \\
              &      & \texttt{NUTS} & $0.62 \pm 0.11$ & $0.79 \pm 0.08$ & $-0.03 \pm 0.03$ & $0.04 \pm 0.02$ \\
              &      & \texttt{ADVI} & $0.46 \pm 0.17$ & $0.90 \pm 0.08$ & $-0.25 \pm 0.04$ & $0.25 \pm 0.04$ \\
              &      & \texttt{PF} & $0.37 \pm 0.24$ & $0.94 \pm 0.07$ & $-0.20 \pm 0.08$ & $0.20 \pm 0.08$ \\
    \midrule
              & \texttt{huge} & \MBz{} & $0.43 \pm 0.11$ & $0.92 \pm 0.07$ & $\phantom{-}0.00 \pm 0.06$ & $0.05 \pm 0.03$ \\
              &      & \texttt{MB} & $0.50 \pm 0.10$ & $0.88 \pm 0.06$ & $-0.07 \pm 0.08$ & $0.09 \pm 0.07$ \\
              &      & \texttt{NUTS} & $0.49 \pm 0.09$ & $0.88 \pm 0.07$ & $-0.06 \pm 0.07$ & $0.07 \pm 0.06$ \\
              &      & \texttt{ADVI} & $0.37 \pm 0.12$ & $0.95 \pm 0.05$ & $-0.18 \pm 0.07$ & $0.18 \pm 0.07$ \\
              &      & \texttt{PF} & $0.25 \pm 0.14$ & $0.99 \pm 0.06$ & $-0.11 \pm 0.06$ & $0.11 \pm 0.06$ \\
    \bottomrule
\end{tabular}

\end{table}

\clearpage

\subsection{Real-World Benchmark} \label{app:real}
\mytabref{tab:real_b} and \mytabref{tab:real_p} report the real-world agreement with NUTS for the Bernoulli and Poisson families; the Gaussian results are in \mytabref{tab:real} (\mysecref{sec:rw}).

\begin{table}[hbp]
  \caption{
     Posterior agreement on each Bernoulli real-world test set aggregated over all parameters.
     Layout as in \mytabref{tab:real}.
  }
  \label{tab:real_b}
  \centering
  \apptable
  \begin{tabular}{cc|ccccc}
    \toprule
    $\mathrm{regime}$ & $\mathrm{model}$ & $r$ & $\sigma\text{-ratio}$ & $\mathrm{rank\text{-}MAD}$ & $\Delta\mathrm{LOO\text{-}NLL}$ & $\Delta\mathrm{time}$ \\
    \midrule
      \texttt{small} & \texttt{MB} & $1.00 \pm 0.00$ & $1.00 \pm 0.01$ & $0.00 \pm 0.00$ & $0.00 \pm 0.00$ & $-53.26 \pm 17.13$ \\
         & \texttt{ADVI} & $0.98 \pm 0.02$ & $0.88 \pm 0.09$ & $0.04 \pm 0.01$ & $0.01 \pm 0.01$ & $-16.95 \pm \phantom{0}6.14$ \\
         & \texttt{PF} & $0.98 \pm 0.02$ & $0.80 \pm 0.14$ & $0.05 \pm 0.03$ & $0.01 \pm 0.01$ & $-24.56 \pm \phantom{0}8.33$ \\
    \midrule
      \texttt{medium} & \texttt{MB} & $1.00 \pm 0.00$ & $1.00 \pm 0.01$ & $0.00 \pm 0.00$ & $0.00 \pm 0.00$ & $-76.62 \pm 32.61$ \\
         & \texttt{ADVI} & $0.96 \pm 0.04$ & $0.74 \pm 0.11$ & $0.05 \pm 0.02$ & $0.02 \pm 0.01$ & $-21.57 \pm 12.56$ \\
         & \texttt{PF} & $0.96 \pm 0.04$ & $0.82 \pm 0.12$ & $0.05 \pm 0.02$ & $0.02 \pm 0.02$ & $-25.82 \pm 19.96$ \\
    \midrule
      \texttt{large} & \texttt{MB} & $1.00 \pm 0.00$ & $1.00 \pm 0.01$ & $0.00 \pm 0.00$ & $0.00 \pm 0.00$ & $-83.90 \pm 35.77$ \\
         & \texttt{ADVI} & $0.96 \pm 0.04$ & $0.74 \pm 0.11$ & $0.05 \pm 0.02$ & $0.02 \pm 0.01$ & $-27.19 \pm 16.43$ \\
         & \texttt{PF} & $0.94 \pm 0.06$ & $0.85 \pm 0.11$ & $0.05 \pm 0.02$ & $0.03 \pm 0.02$ & $-26.09 \pm 23.29$ \\
    \midrule
      \texttt{huge} & \texttt{MB} & $1.00 \pm 0.00$ & $0.99 \pm 0.01$ & $0.00 \pm 0.00$ & $0.00 \pm 0.00$ & $-84.80 \pm 33.46$ \\
         & \texttt{ADVI} & $0.95 \pm 0.04$ & $0.70 \pm 0.12$ & $0.06 \pm 0.02$ & $0.02 \pm 0.01$ & $-27.97 \pm 17.35$ \\
         & \texttt{PF} & $0.94 \pm 0.05$ & $0.87 \pm 0.10$ & $0.05 \pm 0.02$ & $0.03 \pm 0.02$ & $-24.00 \pm 20.68$ \\
    \bottomrule
\end{tabular}

\end{table}

\begin{table}[hbp]
  \caption{
     Posterior agreement on each Poisson real-world test set aggregated over all parameters.
     Layout as in \mytabref{tab:real}.
     \texttt{ADVI}'s posteriors are too narrow here ($\sigma$-ratio $0.59$--$0.68$) and depart most in shape (rank-MAD $0.08$--$0.11$); \texttt{PF} is closer in width ($\sigma$-ratio $0.72$--$0.80$).
     No \textit{huge} regime exists for count outcomes: no real hierarchical datasets with sufficiently many predictors were found.
  }
  \label{tab:real_p}
  \centering
  \apptable
  \begin{tabular}{cc|ccccc}
    \toprule
    $\mathrm{regime}$ & $\mathrm{model}$ & $r$ & $\sigma\text{-ratio}$ & $\mathrm{rank\text{-}MAD}$ & $\Delta\mathrm{LOO\text{-}NLL}$ & $\Delta\mathrm{time}$ \\
    \midrule
      \texttt{small} & \texttt{MB} & $1.00 \pm 0.00$ & $1.00 \pm 0.02$ & $0.01 \pm 0.00$ & $0.00 \pm 0.00$ & $-179.31 \pm 129.94$ \\
         & \texttt{ADVI} & $0.99 \pm 0.01$ & $0.59 \pm 0.27$ & $0.11 \pm 0.07$ & $0.11 \pm 0.09$ & $-148.96 \pm 127.57$ \\
         & \texttt{PF} & $0.99 \pm 0.01$ & $0.80 \pm 0.13$ & $0.07 \pm 0.02$ & $0.03 \pm 0.04$ & $-155.97 \pm 130.06$ \\
    \midrule
      \texttt{medium} & \texttt{MB} & $1.00 \pm 0.00$ & $0.99 \pm 0.01$ & $0.00 \pm 0.00$ & $0.00 \pm 0.00$ & $-168.49 \pm 103.08$ \\
         & \texttt{ADVI} & $0.99 \pm 0.00$ & $0.63 \pm 0.11$ & $0.09 \pm 0.02$ & $0.05 \pm 0.02$ & $-113.71 \pm \phantom{0}91.78$ \\
         & \texttt{PF} & $0.99 \pm 0.01$ & $0.76 \pm 0.06$ & $0.07 \pm 0.01$ & $0.01 \pm 0.01$ & $-\phantom{0}97.15 \pm \phantom{0}73.55$ \\
    \midrule
      \texttt{large} & \texttt{MB} & $1.00 \pm 0.00$ & $0.97 \pm 0.02$ & $0.01 \pm 0.00$ & $0.00 \pm 0.00$ & $-122.52 \pm \phantom{0}71.19$ \\
         & \texttt{ADVI} & $0.99 \pm 0.01$ & $0.68 \pm 0.06$ & $0.08 \pm 0.01$ & $0.05 \pm 0.02$ & $-\phantom{0}93.67 \pm \phantom{0}70.67$ \\
         & \texttt{PF} & $0.98 \pm 0.01$ & $0.72 \pm 0.05$ & $0.07 \pm 0.01$ & $0.01 \pm 0.01$ & $-\phantom{0}84.08 \pm \phantom{0}62.18$ \\
    \bottomrule
\end{tabular}

\end{table}

\subsection{Per-Dataset Agreement} \label{app:agree}
\rev{The benchmark tables summarize performance by medians over the test set. To get a sense of per-dataset performance, we compare \mb with NUTS on every dataset of the oracle and real test sets in this section.}
\rev{The comparison is on the worst case: each dataset is rated by the GLMM parameter where \mb and NUTS disagree the most. In this sense, the shares below are a conservative lower bound on agreement with NUTS.}

\paragraph{Setup.}
\begin{itemize}
  \item \rev{\textbf{Data.} The $23$ oracle and real test sets of \mysecref{sec:cb} and \mysecref{sec:rw} with the draws behind their tables, restricted to the $11{,}072$ datasets where NUTS converged (\myappref{app:tra}).}
  \item \rev{\textbf{Distance.} For every global parameter (fixed effects, random-effect SDs, residual SD), we measure how far \mb's posterior mean lies from NUTS' in NUTS posterior SDs (the \emph{shift}) and the ratio of the two posterior SDs (the $\sigma$-ratio).
  Per dataset we keep the worst case: the largest shift and the $\sigma$-ratio furthest from $1$.}
  \item \rev{\textbf{Criterion.} A dataset is off NUTS if its shift exceeds $1$ or its $\sigma$-ratio lies outside $[2/3, 3/2]$.}
  \item \rev{\textbf{Flags.} A dataset is flagged if its IMH acceptance rate lies below $0.1$ or its ESS lie below $150$ (\myappref{app:imh}).}
  \item \rev{\textbf{Control.} We apply the same rule to a NUTS run at \texttt{PyMC}'s default budget (level 0 of \myappref{app:tra}) against the reference run.}
\end{itemize}

\paragraph{Results.}
\begin{itemize}
  \item \rev{\textbf{Agreement.} \mb agrees with NUTS on $99.1\%$ of the datasets ($99.3\%$ oracle, $98.9\%$ real), with a median shift of $0.07$ NUTS SDs (\mytabref{tab:agree}).}
  \item \rev{\textbf{Control.} The default NUTS run is never off the reference, so the rule does not fire on a correct posterior.}
  \item \rev{\textbf{Flags.} The diagnostics flag $93$ of the $101$ datasets off NUTS. The other $8$ are too narrow by a factor of $1.5$--$2.3$.}
\end{itemize}

\begin{table}[hbp]
  \caption{
     \rev{Per-dataset agreement of \mb with NUTS over the NUTS-converged datasets of each family and test source.
     Shift and $\sigma$-ratio show the median and $[5\%, 95\%]$ percentiles over datasets. Comparison is over the GLMM parameter per dataset, where \mb and NUTS disagree the most.
     The last row compares the NUTS reference with NUTS at \texttt{PyMC}'s default budget.}
  }
  \label{tab:agree}
  \centering
  \apptable
  \begin{tabular}{ll|rrrcc}
    \toprule
    $\mathrm{family}$ & $\mathrm{test\ sets}$ & $\mathrm{datasets}$ & $\mathrm{off\ NUTS}$ & $\mathrm{unflagged}$ & $\mathrm{shift}$ & $\sigma\text{-ratio}$ \\
    \midrule
    Gaussian & oracle & $1{,}854$ & $0$ & $0$ & $0.07\ [0.02, 0.14]$ & $0.97\ [0.91, 1.08]$ \\
             & real   & $1{,}946$ & $14$ & $2$ & $0.08\ [0.03, 0.26]$ & $0.96\ [0.84, 1.08]$ \\
    \midrule
    Bernoulli & oracle & $1{,}978$ & $10$ & $2$ & $0.07\ [0.02, 0.17]$ & $0.97\ [0.86, 1.08]$ \\
              & real   & $1{,}986$ & $3$ & $0$ & $0.08\ [0.02, 0.18]$ & $0.96\ [0.86, 1.08]$ \\
    \midrule
    Poisson & oracle & $1{,}912$ & $31$ & $0$ & $0.07\ [0.02, 0.17]$ & $0.97\ [0.83, 1.08]$ \\
            & real   & $1{,}396$ & $43$ & $4$ & $0.09\ [0.03, 0.37]$ & $0.93\ [0.73, 1.07]$ \\
    \midrule
    all & & $11{,}072$ & $101$ & $8$ & $0.07\ [0.02, 0.21]$ & $0.96\ [0.84, 1.08]$ \\
    \midrule
    \multicolumn{2}{l|}{\texttt{NUTS} default} & $6{,}416$ & $0$ & -- & $0.05\ [0.02, 0.09]$ & $0.98\ [0.94, 1.06]$ \\
    \bottomrule
\end{tabular}

\end{table}

\subsection{Runtime Results} \label{app:rt}

\paragraph{Setup.}
Runtimes were measured on the twelve oracle benchmarks (three likelihood families $\times$ four size regimes, $512$ datasets each).
\mb draws $4{,}000$ posterior samples per dataset, the draw count every reference method is evaluated at; the IMH head runs $4$ chains of $1{,}000$ steps on these draws.
NUTS, ADVI, Pathfinder and LA run via \texttt{PyMC} on the CPU nodes described in \myappref{app:tra} (NUTS on four cores, the others on one); their wall times are those recorded at fit time, failed fits excluded.
\texttt{NUTS} is the escalating competitor of \myappref{app:tra}, so its time is the cumulative cost of every budget level tried; \texttt{NUTS}$_0$ is \texttt{PyMC}'s default budget alone.
\mb is timed on one NVIDIA H100 GPU and on four Xeon Gold CPU cores.
Every dataset is processed separately (batch of one) after a warm-up pass, as the released API runs one when the model is loaded. We use device synchronization around the timed region.
Two savings available in deployment, batching several datasets per pass and caching the analytical statistics, are not used.
Speed-ups are paired per dataset: the reference method's recorded wall time on a dataset divided by \mb's time on the same dataset, summarized by the median and the 5th--95th percentile of these ratios over all $6{,}144$ datasets (\mytabref{tab:rt_ratio}).

\paragraph{What is timed.}
\texttt{MB} is the default pipeline: one forward pass through the summary networks and the global posterior network $\boldsymbol\Pi_g$, which draws $4{,}000$ global samples, followed by the IMH head, which reimputes the random effects (from their analytical conditional for Gaussian outcomes, otherwise from the inner draws of the pseudo-marginal chain; \myappref{app:imh}).
The IMH head is listed separately for comparison.
\MBz{} is the raw flow with refinement switched off; for the discrete families it draws the random effects from the local posterior network $\boldsymbol\Pi_l$, which the default pipeline never evaluates (Gaussian \MBz{} uses the closed-form conditional instead, \myappref{app:arch}).

\begin{figure}[hbp]
    \centering
    \begin{subfigure}{0.72\textwidth}
        \centering
        \includegraphics[width=\textwidth]{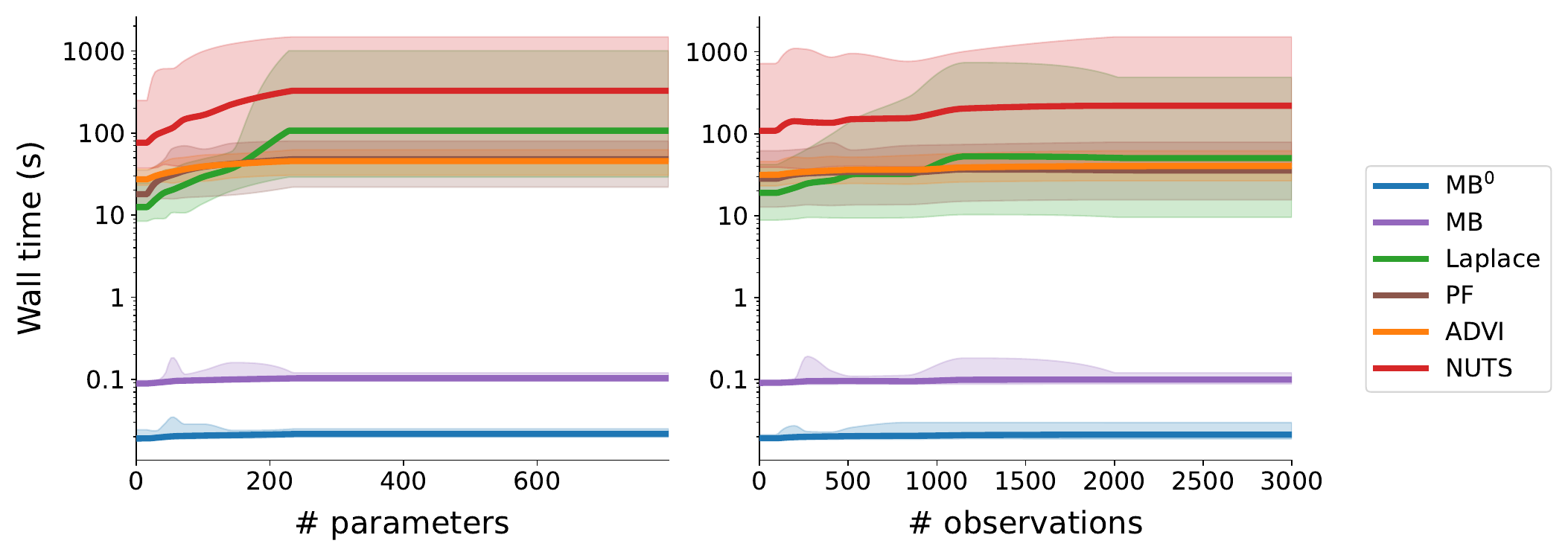}
        \caption{Gaussian}
        \label{fig:rt_n}
    \end{subfigure}
    \par\medskip
    \begin{subfigure}{0.72\textwidth}
        \centering
        \includegraphics[width=\textwidth]{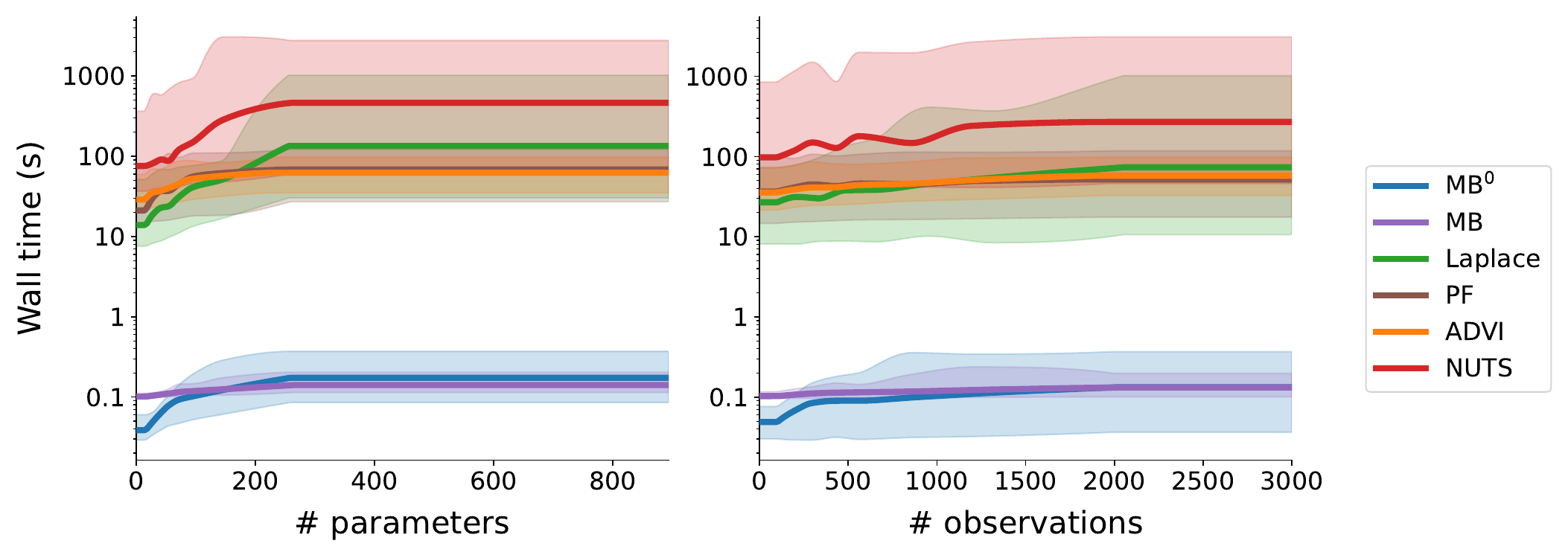}
        \caption{Bernoulli}
        \label{fig:rt_b}
    \end{subfigure}
    \par\medskip
    \begin{subfigure}{0.72\textwidth}
        \centering
        \includegraphics[width=\textwidth]{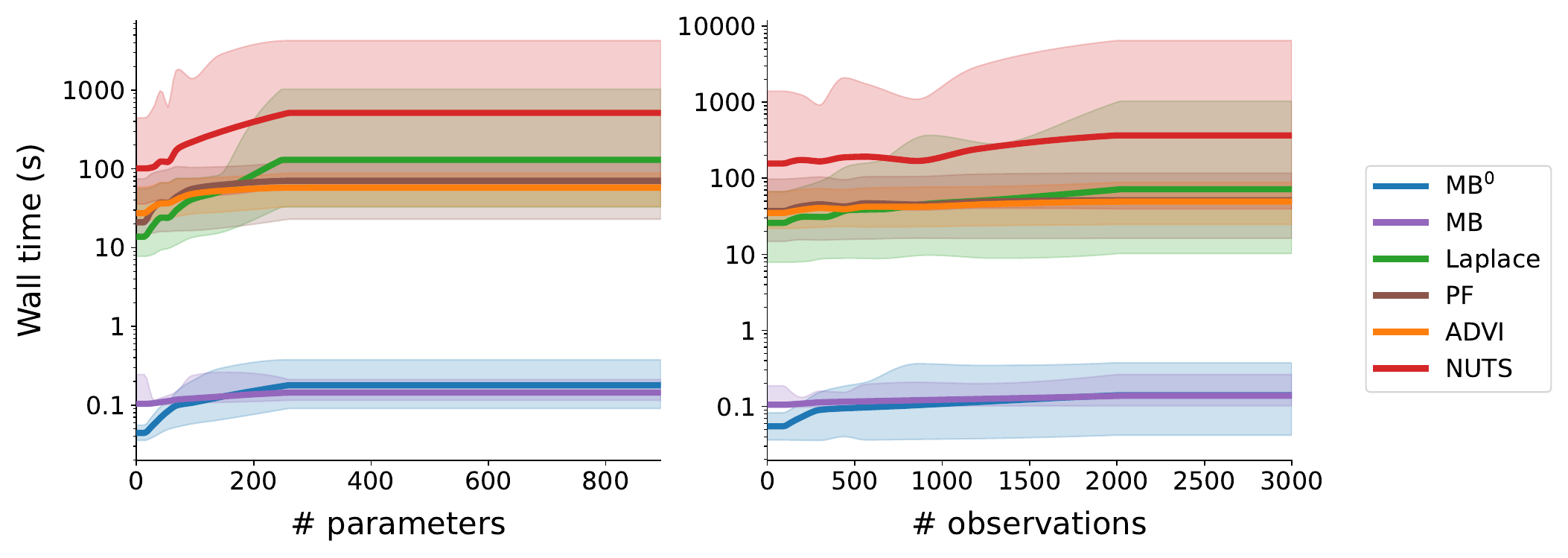}
        \caption{Poisson}
        \label{fig:rt_p}
    \end{subfigure}
    \caption{
       Wall-clock time per dataset vs.\ model complexity ($n_\mathrm{params} = d + q + mq$, left) and total observations ($n$, right), pooled over the four size regimes of each likelihood family.
       Lines show binned means; bands span the 0.5th--99.5th percentile.
       \mb appears as the raw flow (\MBz{}) and the default pipeline (\texttt{MB}), one dataset per forward pass on the GPU; \texttt{NUTS} (escalating composite), \texttt{ADVI}, \texttt{PF} and \texttt{LA} as recorded at fit time on the CPU (\texttt{NUTS} on four cores, the others on one).
       Note the log ordinate: \texttt{MB} is flat in both panels because its cost tracks the architecture, whereas the sampler baselines track the data; \MBz{} of the discrete families inherits a mild dependence on the group count from the local posterior network.
    }
    \label{fig:rt}
\end{figure}

\paragraph{Results.}
\mytabref{tab:rt_full} and \myfigref{fig:rt} give the breakdown. Three take-aways follow:
\begin{enumerate}
   \item \textbf{The IMH head permits a shortcut around the local posterior network}. \mb returns a full posterior in $0.09$--$0.12$\,s per dataset on the GPU and $0.54$--$2.3$\,s on four CPU cores.
      The $\boldsymbol\Pi_g$ (global posterior) pass costs $0.02$\,s (GPU) resp. $0.2$--$0.5$\,s (CPU) in every family. The IMH head adds $0.07$--$0.10$\,s resp. $0.2$--$1.8$\,s.
The head is also what makes the global pass sufficient: by marginalizing the random effects out, it avoids $\boldsymbol\Pi_l$, whose evaluation per group and draw makes \MBz{} $4$--$8\times$ slower than \texttt{MB} on CPU for Bernoulli and Poisson ($4.3$--$10.6$\,s).
   \item \textbf{Wall-clock time scales more favorably for \mb}.
The flow pass grows by at most $13\%$ (GPU) and about $50\%$ (CPU) from \textit{small} to \textit{huge}; only the head's per-group work on CPU grows significantly with the group count ($2$--$3\times$). The 95th percentile of \texttt{MB} stays within $1.5\times$ its median on GPU, where the slowest of all $6{,}144$ datasets took $0.32$\,s. The 95th percentile is $0.8$--$6.2$\,s on CPU.
\texttt{NUTS}' 95th percentile is $2.9$--$14\times$ its median, because the escalation adds the full cost of every failed level; its slowest dataset took $7{,}556$\,s. \texttt{NUTS}$_0$ alone has a 95th percentile of $1.5$--$2.0\times$ its median.
\item \textbf{\mb works best on GPU and remains useful on CPU}. Per dataset, \texttt{MB} is $410$--$5{,}500\times$ faster than \texttt{NUTS} on the GPU and $21$--$540\times$ faster on four Xeon Gold CPU cores (5th--95th percentile; medians $800\times$ and $83\times$, \mytabref{tab:rt_ratio}).
The interval is long on the right because the escalation makes the slowest NUTS fits very slow (up to $56{,}000\times$), and short on the left: no dataset falls below $175\times$ on the GPU or $3.5\times$ on the CPU.
Against LA, \texttt{MB} is $110$--$740\times$ faster on the GPU and $6$--$69\times$ on the CPU, and against ADVI $240$--$560\times$ and $12$--$68\times$.
The cheaper references save little time: at the median of each benchmark, ADVI is $1.8$--$3.3\times$ and \texttt{PF} $2.1$--$4.9\times$ faster than \texttt{NUTS}, and \texttt{NUTS}$_0$, the default budget that converges on $41$--$67\%$ of the datasets, $1.2$--$2.0\times$.
Counting only converged \texttt{NUTS} runs (\myappref{app:tra}) widens the gap, since the slowest runs converge least often: $94$--$292$\,s per converged NUTS posterior, $900$--$2{,}700\times$ (GPU) and $120$--$310\times$ (CPU) the cost of \texttt{MB}.
\end{enumerate}

\begin{table}[h]
  \caption{
     Median wall-clock time in seconds per dataset ($512$ datasets per cell, one dataset per forward pass), for \mb on one H100 GPU (left) and on four Xeon Gold CPU cores (right).
     \texttt{MB} is the default pipeline; its IMH head is listed separately. \MBz{} is the raw flow.
     \texttt{LA}, \texttt{PF}, \texttt{ADVI}, \texttt{NUTS}$_0$ and \texttt{NUTS} run on the CPU only (\texttt{NUTS} on four cores, the others on one); \texttt{NUTS} is the escalating composite of \myappref{app:tra}, its time cumulative over the levels tried, \texttt{NUTS}$_0$ the default budget alone.
  }
  \label{tab:rt_full}
  \centering
  \apptable
  \begin{tabular}{l|cccc|cccc}
    \toprule
     & \multicolumn{4}{c|}{H100 GPU} & \multicolumn{4}{c}{4 CPU cores} \\
    $\mathrm{method}$ & \textit{small} & \textit{medium} & \textit{large} & \textit{huge} & \textit{small} & \textit{medium} & \textit{large} & \textit{huge} \\
    \midrule
    \multicolumn{9}{l}{Gaussian} \\
    \texttt{MB}                & $0.090$ & $0.093$ & $0.095$ & $0.098$ & $0.539$ & $0.605$ & $0.704$ & $0.857$ \\
    \quad IMH only & $0.074$ & $0.076$ & $0.078$ & $0.081$ & $0.196$ & $0.218$ & $0.280$ & $0.361$ \\
    \MBz{}                     & $0.019$ & $0.020$ & $0.020$ & $0.021$ & $0.371$ & $0.458$ & $0.512$ & $0.629$ \\
    \texttt{LA}                & -- & -- & -- & -- & $12.6$ & $20.5$ & $26.3$ & $33.3$ \\
    \texttt{PF}                & -- & -- & -- & -- & $18.2$ & $29.1$ & $37.2$ & $44.9$ \\
    \texttt{ADVI}              & -- & -- & -- & -- & $27.7$ & $33.8$ & $37.5$ & $41.5$ \\
    \texttt{NUTS}$_0$          & -- & -- & -- & -- & $45.3$ & $55.1$ & $64.4$ & $73.9$ \\
    \texttt{NUTS}              & -- & -- & -- & -- & $80.6$ & $85.2$ & $87.9$ & $94.4$ \\
    \midrule
    \multicolumn{9}{l}{Bernoulli} \\
    \texttt{MB}                & $0.104$ & $0.108$ & $0.114$ & $0.118$ & $0.781$ & $1.03$ & $1.34$ & $2.21$ \\
    \quad IMH only & $0.086$ & $0.090$ & $0.096$ & $0.099$ & $0.552$ & $0.749$ & $1.01$ & $1.79$ \\
    \MBz{}                     & $0.048$ & $0.068$ & $0.086$ & $0.102$ & $4.26$ & $6.78$ & $10.0$ & $10.3$ \\
    \texttt{LA}                & -- & -- & -- & -- & $12.9$ & $17.4$ & $27.2$ & $42.0$ \\
    \texttt{PF}                & -- & -- & -- & -- & $18.5$ & $27.1$ & $36.4$ & $43.9$ \\
    \texttt{ADVI}              & -- & -- & -- & -- & $29.2$ & $37.5$ & $43.2$ & $49.7$ \\
    \texttt{NUTS}$_0$          & -- & -- & -- & -- & $46.5$ & $59.0$ & $64.7$ & $75.0$ \\
    \texttt{NUTS}              & -- & -- & -- & -- & $72.8$ & $78.6$ & $76.8$ & $91.2$ \\
    \midrule
    \multicolumn{9}{l}{Poisson} \\
    \texttt{MB}                & $0.105$ & $0.109$ & $0.116$ & $0.120$ & $1.15$ & $1.55$ & $2.01$ & $2.26$ \\
    \quad IMH only & $0.087$ & $0.091$ & $0.098$ & $0.102$ & $0.870$ & $1.22$ & $1.64$ & $1.82$ \\
    \MBz{}                     & $0.054$ & $0.073$ & $0.093$ & $0.106$ & $4.39$ & $7.47$ & $9.24$ & $10.6$ \\
    \texttt{LA}                & -- & -- & -- & -- & $12.9$ & $17.9$ & $27.9$ & $41.9$ \\
    \texttt{PF}                & -- & -- & -- & -- & $18.2$ & $27.5$ & $36.2$ & $45.6$ \\
    \texttt{ADVI}              & -- & -- & -- & -- & $27.0$ & $33.7$ & $38.6$ & $45.8$ \\
    \texttt{NUTS}$_0$          & -- & -- & -- & -- & $45.3$ & $56.4$ & $70.0$ & $78.6$ \\
    \texttt{NUTS}              & -- & -- & -- & -- & $89.6$ & $91.3$ & $92.7$ & $97.7$ \\
    \bottomrule
  \end{tabular}
\end{table}

\begin{table}[htbp]
   \caption{
     Per-dataset speed-up of \texttt{MB} over each reference method, pooled over the twelve oracle benchmarks ($6{,}144$ datasets; $5{,}679$--$6{,}144$ with a completed reference fit).
     Each entry is a quantile of the ratio of the reference method's wall time to \texttt{MB}'s on the same dataset, with \texttt{MB} on the H100 GPU or on four Xeon Gold CPU cores.
   }
   \label{tab:rt_ratio}
   \centering
   \apptable
   \begin{tabular}{l|ccccc|ccccc}
     \toprule
      & \multicolumn{5}{c|}{\texttt{MB} on H100 GPU} & \multicolumn{5}{c}{\texttt{MB} on 4 CPU cores} \\
     $\mathrm{vs.}$ & min & $5\%$ & median & $95\%$ & max & min & $5\%$ & median & $95\%$ & max \\
     \midrule
     \texttt{NUTS}     & $175$ & $410$ & $800$ & $5{,}500$ & $56{,}000$ & $3.5$ & $21$ & $83$ & $540$ & $4{,}200$ \\
     \texttt{NUTS}$_0$ & $170$ & $390$ & $560$ & $1{,}000$ & $4{,}000$  & $3.5$ & $19$ & $60$ & $140$ & $310$ \\
     \texttt{ADVI}     & $100$ & $240$ & $340$ & $560$     & $880$      & $2.2$ & $12$ & $37$ & $68$  & $140$ \\
     \texttt{PF}       & $32$  & $160$ & $300$ & $740$     & $1{,}100$  & $1.3$ & $9$  & $32$ & $80$  & $180$ \\
     \texttt{LA}       & $33$  & $110$ & $190$ & $740$     & $9{,}400$  & $1.6$ & $6$  & $23$ & $69$  & $780$ \\
     \bottomrule
   \end{tabular}
\end{table}

\subsection{Laplace Baseline} \label{app:la}

Laplace approximation (LA) is the classic fast approximation for Bayesian inference, and Pathfinder (PF) builds on the same idea: both run quasi-Newton optimization and fit a Gaussian from its curvature. LA does this at the mode and PF at the ELBO-maximizing point of each optimization path. PF additionally uses importance resampling across paths (\myappref{app:tra}).
Notably, GLMM posteriors are far from Gaussian in their variance components; their marginals are skewed and concentrated near zero, making LA ill-suited for this model class.
We report LA only as a failure check: it shows what the Gaussian assumption costs, and how much of that cost PF removes.
\mytabref{tab:la} breaks recovery down by parameter class on the oracle benchmarks.

LA recovers the fixed effects (NRMSE within $0.04$ of NUTS) but fails on everything that depends on the variance components.
Their NRMSE is $3.1$, $7.3$ and $5.0$ against $0.28$, $0.52$ and $0.37$ for NUTS (Gaussian, Bernoulli, Poisson), with $r = 0.28$--$0.68$.
The random effects inherit the error through their conditional (NRMSE $0.81$--$1.93$ against $0.45$--$0.71$), and the predictive accuracy is the worst of all methods in every family (LOO-NLL $0.4$--$1.7$ above NUTS).
On the \textit{medium} Bernoulli benchmark, for instance, its variance-component posteriors are $3.9\times$ as wide as the NUTS reference and their means are uncorrelated with it ($r \approx 0$).
The mode search itself is fragile: BFGS reports precision loss on more than $90\%$ of the datasets, and $4$--$12\%$ of the fits fail outright (\myappref{app:tra}).
PF removes most but not all of it; its variance components are still off (NRMSE $1.6$ in every family against $0.36$--$0.60$ for ADVI; $r = 0.64$--$0.81$), but its random effects (NRMSE $0.58$--$0.90$) and predictive accuracy (LOO-NLL within $0.06$ of NUTS) are much closer to the reference.

\begin{table}[hbp]
  \caption{
    Recovery by parameter class on the oracle benchmarks, averaged over the four size regimes (datasets where the level-2 NUTS reference converged, as in \mytabref{tab:oracle}).
    Entries are NRMSE ($r$ in parentheses) for fixed effects ($\boldsymbol\beta$), variance components ($\boldsymbol\sigma$) and random effects ($\boldsymbol\alpha$), median per-dataset LOO-NLL, and median wall-clock time in seconds per dataset (\texttt{MB} on one GPU, \texttt{NUTS} on four CPU cores, \texttt{ADVI}, \texttt{PF} and \texttt{LA} on one; \mytabref{tab:rt_full}).
    \texttt{MB} is the default pipeline; \texttt{NUTS} the escalating competitor of \myappref{app:tra}. LOO-NLL and time are the regime averages of \mytabref{tab:oracle}, \mytabref{tab:oracle_b} and \mytabref{tab:oracle_p}.
    Bold marks NRMSE at least $1.5\times$ the NUTS value of the same cell and LOO-NLL at least $0.1$ above it.
  }
  \label{tab:la}
  \centering
  \apptable
  \begin{tabular}{ll|ccc|cc}
    \toprule
    $\mathrm{family}$ & $\mathrm{model}$ & $\boldsymbol\beta$: NRMSE ($r$) & $\boldsymbol\sigma$: NRMSE ($r$) & $\boldsymbol\alpha$: NRMSE ($r$) & $\mathrm{LOO\text{-}NLL}$ & $\mathrm{time\ [s]}$ \\
    \midrule
      \multirow{5}{*}{Gaussian}
        & \texttt{MB} & $0.28$ ($0.96$) & $0.28$ ($0.96$) & $0.45$ ($0.89$) & $2.27$ & $\phantom{0}0.1$ \\
        & \texttt{NUTS} & $0.28$ ($0.96$) & $0.28$ ($0.96$) & $0.45$ ($0.89$) & $2.27$ & $83.5$ \\
        & \texttt{ADVI} & $0.28$ ($0.96$) & $0.36$ ($0.95$) & $0.47$ ($0.88$) & $2.33$ & $34.5$ \\
        & \texttt{PF} & $0.29$ ($0.96$) & $\mathbf{1.55}$ ($0.81$) & $0.58$ ($0.82$) & $2.30$ & $31.5$ \\
        & \texttt{LA} & $0.30$ ($0.95$) & $\mathbf{3.14}$ ($0.68$) & $\mathbf{0.81}$ ($0.74$) & $\mathbf{3.92}$ & $22.1$ \\
    \midrule
      \multirow{5}{*}{Bernoulli}
        & \texttt{MB} & $0.31$ ($0.95$) & $0.51$ ($0.86$) & $0.71$ ($0.70$) & $0.46$ & $\phantom{0}0.1$ \\
        & \texttt{NUTS} & $0.31$ ($0.95$) & $0.52$ ($0.86$) & $0.71$ ($0.70$) & $0.46$ & $75.1$ \\
        & \texttt{ADVI} & $0.31$ ($0.95$) & $0.60$ ($0.84$) & $0.73$ ($0.68$) & $0.48$ & $39.3$ \\
        & \texttt{PF} & $0.32$ ($0.94$) & $\mathbf{1.64}$ ($0.64$) & $0.90$ ($0.56$) & $0.48$ & $30.7$ \\
        & \texttt{LA} & $0.35$ ($0.94$) & $\mathbf{7.27}$ ($0.28$) & $\mathbf{1.93}$ ($0.43$) & $\mathbf{0.85}$ & $21.8$ \\
    \midrule
      \multirow{5}{*}{Poisson}
        & \texttt{MB} & $0.22$ ($0.97$) & $0.37$ ($0.93$) & $0.60$ ($0.80$) & $1.37$ & $\phantom{0}0.1$ \\
        & \texttt{NUTS} & $0.21$ ($0.98$) & $0.37$ ($0.93$) & $0.59$ ($0.80$) & $1.37$ & $87.8$ \\
        & \texttt{ADVI} & $0.22$ ($0.97$) & $0.51$ ($0.90$) & $0.63$ ($0.77$) & $1.46$ & $35.2$ \\
        & \texttt{PF} & $0.22$ ($0.97$) & $\mathbf{1.58}$ ($0.77$) & $0.73$ ($0.72$) & $1.43$ & $30.5$ \\
        & \texttt{LA} & $0.23$ ($0.97$) & $\mathbf{5.00}$ ($0.48$) & $\mathbf{1.37}$ ($0.57$) & $\mathbf{2.66}$ & $20.9$ \\
    \bottomrule
\end{tabular}

\end{table}

\clearpage
\section{Model Comparison and Prior Conditioning} \label{app:mc}
This appendix holds the analyses behind \mysecref{sec:mc} and \mysecref{sec:usecase}: the marginal likelihood from importance weights (\myappref{app:ev}), a prior sensitivity analysis on real data (\myappref{app:prior_grid}) and an emulated fixed-prior baseline (\myappref{app:fixed_prior}).
It closes with a simulation-based power analysis (\myappref{app:power}).

\subsection{Marginal Likelihood from Importance Weights} \label{app:ev}
Every marginal importance weight $w_s = p(\mathbf{D} \mid \boldsymbol\vartheta_s)\,p(\boldsymbol\vartheta_s) / q(\boldsymbol\vartheta_s)$ divides the unnormalized posterior by the flow density, so $\mathbb{E}_q[w] = p(\mathbf{D})$ (\mysecref{sec:mc}) and $\hat p(\mathbf{D}) = S^{-1} \sum_s w_s$ is an unbiased estimate of the marginal likelihood \citep{Kass.1995}.
The identity rests on three requirements:
\begin{enumerate}
   \item the proposal $q$ must be normalized, which the flow guarantees by construction;
   \item the numerator must retain every normalizing constant of likelihood and prior:
      the analytical marginalization of the random effects provides this for Gaussian outcomes, and the unbiased IS$^2$ estimate of \myappref{app:imh} for Bernoulli and Poisson outcomes;
   \item $q$ must cover the support of the posterior, which the mass-covering forward-KL objective favors but does not guarantee.
\end{enumerate}

Requirement 3 fails where the flow under-covers the posterior.
\mb estimates the evidence from one further importance-sampling pass based on a multivariate Student-$t_5$ fitted to \mb draws \citep{Geweke.1989}.
Importance weights use the closed-form marginal likelihood for Gaussian outcomes and its pseudo-marginal estimate for Bernoulli and Poisson outcomes.
The estimate is flagged when the PSIS-$\hat k$ of this pass or of the IMH weights (\myappref{app:safe}) exceeds $0.7$.

\paragraph{Setup.}
\begin{itemize}
   \item \textbf{Data.} The oracle test sets of all three families (\myappref{app:tes}), restricted to the datasets where NUTS converged (\myappref{app:tra}; $432$--$503$ of $512$ per regime).
   \item \textbf{Reference.} Iterative bridge sampling \citep{Meng.1996, Gronau.2017} on the NUTS draws. Its warped-Gaussian proposal is fitted on one half of the draws and evaluated on the other; swapping the halves changes the reference by a median of $0.006$--$0.013$ nats. For Bernoulli and Poisson outcomes, the reference integrates out the random effects by adaptive Gauss--Hermite quadrature (AGQ; $20$, $12$, $9$, $7$ and $5$ nodes per dimension for $q = 1, \dots, 5$). %
   \item \textbf{Bayes factors.} Every dataset with a random slope is also fit with a random intercept only, by \mb and by NUTS. We compare their $\ln \mathrm{BF}$ (slope vs.\ no slope) by Jeffreys category on the $1{,}783$ datasets where both NUTS fits converged ($110$--$252$ per regime).
   \item \textbf{Baseline.} The IMH weights themselves (\myappref{app:imh}), from $4{,}000$ flow draws.
\end{itemize}

\paragraph{Results.}
\mytabref{tab:ev} compares both estimators with the reference. Four take-aways:
\begin{enumerate}
   \item \textbf{The evidence is accurate.} The median error of the Student-$t$ estimate is $0.008$--$0.019$ nats on Gaussian data and $0.010$--$0.025$ nats on Bernoulli and Poisson data. %
   \item \textbf{Larger models benefit from the extra pass.} The IMH weights are as accurate as the Student-$t$ pass on \textit{small} and \textit{medium} ($0.005$--$0.015$ nats). Their error grows with model size, and so does their flagged share ($9$--$16\%$ on \textit{large}, $27$--$42\%$ on \textit{huge}). On \textit{huge}, the Student-$t$ pass cuts the median error from $0.054$--$0.074$ to $0.019$--$0.025$ nats. The $90\%$ quantile is cut from $0.31$--$0.45$ to $0.06$--$0.09$ nats.
   \item \textbf{Bayes factors match the reference.} The Jeffreys category agrees on $99.3\%$, $98.7\%$ and $98.2\%$ of Gaussian, Bernoulli and Poisson datasets. %
      Easy cases do not drive this result: decisive Bayes factors ($\mathrm{BF} > 100$ either way) make up $42\%$ of the references, and agreement on the others is $98\%$. Of the $20$ disagreements in total, $19$ are flagged or have a reference $\ln\mathrm{BF}$ within $0.2$ nats of a category boundary.
   \item \textbf{The flag errs on the side of caution.} It catches all $10$ estimates that are off by more than one nat. %
   It fires on $10$--$38\%$ of the \textit{large} and \textit{huge} datasets. %
      Most of these estimates are still fine: $86\%$ are within $0.1$ nats of the reference.
   \item \rev{\textbf{\mb makes this workflow practical.} Per dataset, \mb needs a median of $1.0$\,s on one GPU for both fits and both evidence passes, against $226$\,s of NUTS sampling on four CPU cores. In total, the pass took $36$ minutes for MB and
$655$
core hours for NUTS.}
\end{enumerate}

\begin{table}[hbp]
  \caption{
     Evidence vs.\ bridge sampling on \texttt{NUTS} draws, oracle datasets.
     IMH: IMH weights only; $t$: additional Student-$t$ pass.
     $|\Delta|$: absolute error of $\log p(\mathbf{D})$.
     BF cat.\ agree: share of the Bayes-factor comparisons whose $\ln \mathrm{BF}$ falls into the reference's Jeffreys category.
  }
  \label{tab:ev}
  \centering
  \apptablewide
  \begin{tabular}{ll l cc c cc}
    \toprule
    $\mathrm{family}$ & $\mathrm{regime}$ & $\mathrm{source}$ & $\mathrm{med}\,|\Delta|$ & $q_{90}\,|\Delta|$ & $\mathrm{frac}\,\hat k>0.7$ & $\mathrm{BF\ cat.\ agree}$ & $\mathrm{med}\,|\Delta\ln \mathrm{BF}|$ \\
    \midrule
    Gaussian & \texttt{small} & IMH & $0.005$ & $0.020$ & $0.01$ & $1.00$ & $0.015$ \\
     &  & $t$ & $0.008$ & $0.021$ & $0.01$ & $1.00$ & $0.014$ \\
    \midrule
     & \texttt{medium} & IMH & $0.014$ & $0.039$ & $0.02$ & $0.97$ & $0.028$ \\
     &  & $t$ & $0.012$ & $0.033$ & $0.02$ & $0.99$ & $0.016$ \\
    \midrule
     & \texttt{large} & IMH & $0.027$ & $0.130$ & $0.12$ & $0.98$ & $0.047$ \\
     &  & $t$ & $0.013$ & $0.043$ & $0.10$ & $1.00$ & $0.024$ \\
    \midrule
     & \texttt{huge} & IMH & $0.057$ & $0.421$ & $0.30$ & $0.96$ & $0.089$ \\
     &  & $t$ & $0.019$ & $0.062$ & $0.28$ & $0.99$ & $0.030$ \\
    \midrule\midrule
    Bernoulli & \texttt{small} & IMH & $0.009$ & $0.030$ & $0.02$ & $1.00$ & $0.023$ \\
     &  & $t$ & $0.010$ & $0.031$ & $0.03$ & $1.00$ & $0.020$ \\
    \midrule
     & \texttt{medium} & IMH & $0.015$ & $0.049$ & $0.04$ & $0.98$ & $0.032$ \\
     &  & $t$ & $0.017$ & $0.047$ & $0.04$ & $0.99$ & $0.026$ \\
    \midrule
     & \texttt{large} & IMH & $0.028$ & $0.117$ & $0.09$ & $0.98$ & $0.052$ \\
     &  & $t$ & $0.020$ & $0.073$ & $0.11$ & $0.97$ & $0.033$ \\
    \midrule
     & \texttt{huge} & IMH & $0.054$ & $0.314$ & $0.27$ & $0.94$ & $0.090$ \\
     &  & $t$ & $0.025$ & $0.088$ & $0.27$ & $0.98$ & $0.033$ \\
    \midrule\midrule
    Poisson & \texttt{small} & IMH & $0.007$ & $0.025$ & $0.01$ & $0.98$ & $0.026$ \\
     &  & $t$ & $0.010$ & $0.027$ & $0.02$ & $0.99$ & $0.020$ \\
    \midrule
     & \texttt{medium} & IMH & $0.013$ & $0.045$ & $0.02$ & $0.95$ & $0.052$ \\
     &  & $t$ & $0.015$ & $0.040$ & $0.02$ & $1.00$ & $0.022$ \\
    \midrule
     & \texttt{large} & IMH & $0.030$ & $0.161$ & $0.16$ & $0.92$ & $0.158$ \\
     &  & $t$ & $0.017$ & $0.054$ & $0.13$ & $0.97$ & $0.041$ \\
    \midrule
     & \texttt{huge} & IMH & $0.074$ & $0.446$ & $0.42$ & $0.89$ & $0.192$ \\
     &  & $t$ & $0.024$ & $0.094$ & $0.38$ & $0.96$ & $0.060$ \\
    \bottomrule
\end{tabular}

\end{table}

\clearpage
\subsection{Zero-Shot Prior Sensitivity Analysis} \label{app:prior_grid}
Prior conditioning promises a prior sensitivity analysis in a single batched forward pass (\mysecref{sec:usecase}).
We test this promise on real data against NUTS, which is refit at every prior.

\paragraph{Setup.}
\begin{itemize}
   \item \textbf{Data.} One held-out dataset per likelihood family (\mytabref{tab:prior_grid_setup}; \citealp{Bates.2015, Brooks.2017}), each with a correlated random intercept and slope: \texttt{Reaction \textasciitilde{} Days + (1 + Days | Subject)}, \texttt{incidence \textasciitilde{} period + (1 + period4 | herd)} and \texttt{count \textasciitilde{} mined + wtemp + (1 + wtemp | site)}. The \textit{key effects} are \texttt{Days}, \texttt{period4} and \texttt{mined}.
  \item \textbf{Prior grid.} Twelve log-spaced Normal slope-prior scales $\tau_\beta \in [0.05, 5]$ crossed with half-Normal, half-Student-$t_5$ and Exponential priors on the random-effect SDs at scales $\tau_\sigma \in \{0.25, 1, 2.5, 5\}$; all other priors at the Bambi defaults. Priors above a family's training hyper-prior range (\myappref{app:prior}) are dropped, leaving $132$, $60$ and $54$ priors.
  \item \textbf{Methods.} \mb ($4{,}000$ draws per prior, one batched call per SD-prior family) and the raw flow \MBz{}. NUTS refits every grid point with the budget ladder of \myappref{app:tra}. Level 2 is the reference. The reported NUTS cost is that of the escalating run, which stops at the cheapest level that passes the convergence criterion.
  \item \textbf{Disagreement flag.} The posterior median of the key effect shifts by more than $0.25$ SDs of the NUTS posterior (about three Monte Carlo standard errors of the difference of two medians) or the $90\%$-interval width ratio leaves $[0.8, 1.25]$. Only priors at which NUTS converged (\myappref{app:tra}) are compared ($237$ of $246$).
\end{itemize}

\begin{table}[hbp]
  \caption{
    Prior-sensitivity datasets, grid sizes and reference cost. conv.: priors at which \texttt{NUTS} converged (\myappref{app:tra}). \texttt{NUTS}: median seconds per prior of the escalating run (four chains on four Xeon CPU cores) and core-hours for the full grid.
  }
  \label{tab:prior_grid_setup}
  \centering
  \apptablewide
  \begin{tabular}{lcc|cc|cc}
    \toprule
    & & & \multicolumn{2}{c|}{priors} & \multicolumn{2}{c}{\texttt{NUTS}} \\
    dataset & $m$ & $n$ & grid & conv. & per prior & full grid \\
    & & & & & [s] & [core-h] \\
    \midrule
    \texttt{sleepstudy} (Gaussian)     & 18 & 180 & 132 & 127 & \phantom{0}86 & 18.3 \\
    \texttt{cbpp} (Bernoulli)          & 15 & 842 & \phantom{0}60 & \phantom{0}59 & \phantom{0}98 & 14.2 \\
    \texttt{Salamanders} (Poisson)     & 23 & 644 & \phantom{0}54 & \phantom{0}51 & 192 & 15.6 \\
    \bottomrule
\end{tabular}

\end{table}

\begin{figure}[htbp]
  \centering
  \includegraphics[width=\linewidth]{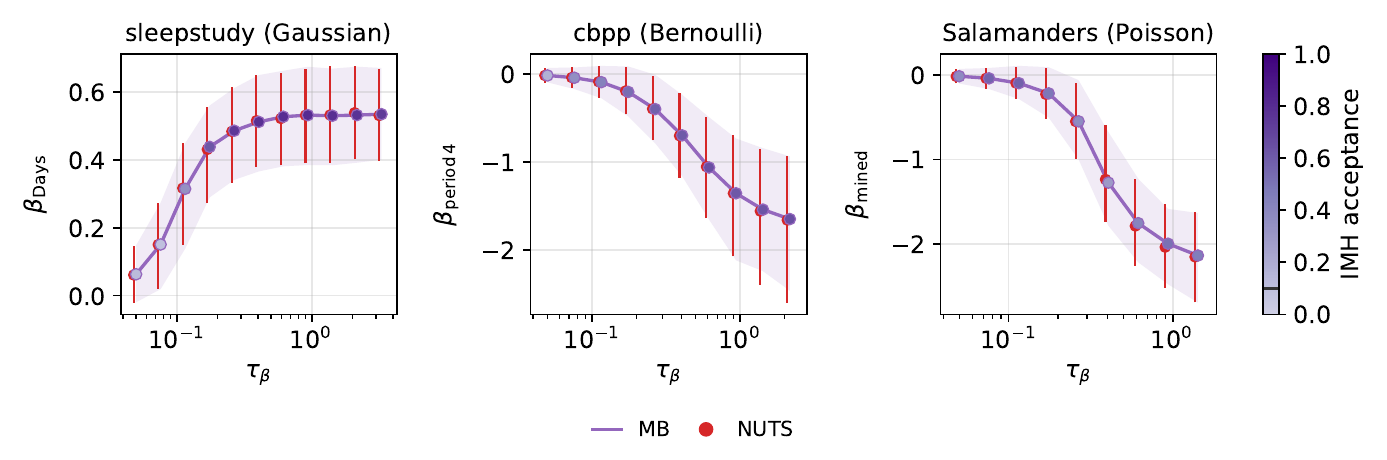}
  \caption{
    Prior sensitivity in one pass: posterior median and $90\%$ interval of the key effect against the slope-prior scale $\tau_\beta$, with the random-effect SD prior fixed at half-Normal($2.5$) (\texttt{sleepstudy}) or half-Normal($1$).
    \mb as lines and bands, markers shaded by IMH acceptance (the colorbar line marks the warning threshold $0.1$); NUTS as red markers.
    The Bayes-factor curves are in \myfigref{fig:prior_grid_bf}.
  }
  \label{fig:usecase}
\end{figure}

\begin{figure}[htbp]
  \centering
  \includegraphics[width=\linewidth]{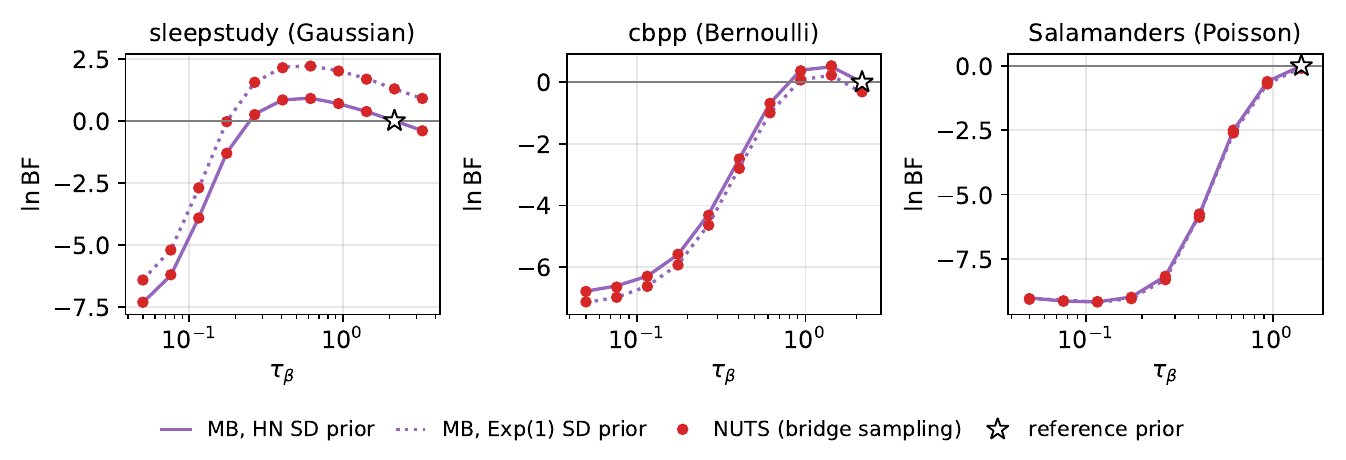}
  \caption{
    $\ln \mathrm{BF}$ of each slope prior against the dataset's Bambi-like reference prior (star), under half-Normal (HN; scale $2.5$ on \texttt{sleepstudy}, $1$ otherwise) and Exponential($1$) SD priors: Student-$t$ importance sampling on the \mb draws (lines) and bridge sampling on the NUTS draws (markers).
  }
  \label{fig:prior_grid_bf}
\end{figure}

\paragraph{Results.}
\mytabref{tab:prior_grid} gives the agreement over the $237$ converged priors, \myfigref{fig:usecase} shows prior sensitivity for the key effect.
Three take-aways:
\begin{enumerate}
  \item \textbf{The grid is interactive with \mb.} Each grid takes $1.6$--$2.8$\,s on one H100 GPU, IMH included, and its evidence $0.3$--$0.8$\,s more, against $14$--$18$ core-hours for NUTS (\mytabref{tab:prior_grid_setup}). %
  \item \textbf{\mb reproduces NUTS across the whole grid} (\mytabref{tab:prior_grid}). The flag fires at none of the $237$ priors for \mb, in each of three seeded runs, against $84$ for the raw flow. %
  \item \textbf{This includes the priors where the flow is hardest to correct.} At the tightest slope priors of \texttt{sleepstudy} and \texttt{cbpp} ($\tau_\beta \leq 0.076$), IMH acceptance drops as low as $0.02$ (pale markers in \myfigref{fig:usecase}), yet the key effect stays within $0.10$ NUTS SDs of the reference and its interval width within $4\%$ of NUTS'.
  At $\tau_\beta = 0.41$ on \texttt{Salamanders}, where the key effect's posterior is skewed toward zero and tied to the random-intercept SD, \mb's $90\%$ interval is $2$--$16\%$ narrower than NUTS'.
\end{enumerate}
\paragraph{Evidence.}
The IMH weights are least reliable where the flow under-covers the posterior: at the tightest slope prior of \texttt{sleepstudy} (IMH acceptance $0.03$) their evidence errs by up to $1.3$ nats, flagged by $\hat k > 1$; on \texttt{Salamanders} the $90\%$ quantile of the error is $0.25$ nats.
The Student-$t$ pass of \myappref{app:ev} removes these errors.
Over the $237$ NUTS-converged priors, its evidence matches bridge sampling on the NUTS draws within a median of $0.009$--$0.016$ nats per dataset (at most $0.03$, $0.05$ and $0.08$; bridge noise floor $0.008$--$0.011$).
Its own PSIS-$\hat k$ stays below $0.7$ at every prior; the IMH weights exceed $0.7$ at $16$ priors, which the default flag therefore marks.
The resulting Bayes-factor curves (\myfigref{fig:prior_grid_bf}) favor moderate slope priors on \texttt{sleepstudy} ($\tau_\beta \approx 0.4$--$0.6$; $\ln \mathrm{BF} = +0.9$ against the reference under its half-Normal SD prior) and \texttt{cbpp} ($\tau_\beta \approx 0.9$--$1.4$; $\ln \mathrm{BF} \leq +0.5$), and the widest slope prior in the grid on \texttt{Salamanders} (the reference, $\tau_\beta = 1.42$); all three reject $\tau_\beta = 0.05$ ($\ln \mathrm{BF} \leq -6$).
For \mb this means that a full prior sensitivity analysis, including the evidence, takes a few seconds and agrees with NUTS at every prior.

\begin{table}[hbp]
  \caption{
    Agreement with \texttt{NUTS} over all grid priors at which \texttt{NUTS} converged, median $\pm$ MAD: $r$, $\sigma$-ratio, rank-MAD and $\Delta\mathrm{LOO\text{-}NLL}$ over all parameters, the median shift of the key effect in NUTS posterior SDs, the number of flagged priors, and the IMH acceptance of \texttt{MB}.
  }
  \label{tab:prior_grid}
  \centering
  \apptablewide
  \begin{tabular}{ll|ccccc|c|c}
    \toprule
    dataset & model & $r$ & $\sigma$-ratio & rank-MAD & $\Delta\mathrm{LOO\text{-}NLL}$ & $|\Delta \mathrm{med}| / \mathrm{sd}$ & flagged & IMH acc. \\
    \midrule
    \multirow{2}{*}{\texttt{sleepstudy}}
      & \MBz{}                 & $1.00 \pm 0.00$ & $1.03 \pm 0.01$ & $0.01 \pm 0.01$ & $\phantom{-}0.00 \pm 0.00$ & $0.06 \pm 0.05$ & $\phantom{0}33/127$ & -- \\
      & \texttt{MB}            & $1.00 \pm 0.00$ & $1.00 \pm 0.01$ & $0.00 \pm 0.00$ & $\phantom{-}0.00 \pm 0.00$ & $0.03 \pm 0.02$ & $\phantom{00}0/127$ & $0.74 \pm 0.07$ \\
    \midrule
    \multirow{2}{*}{\texttt{cbpp}}
      & \MBz{}                 & $0.99 \pm 0.00$ & $0.93 \pm 0.01$ & $0.02 \pm 0.00$ & $\phantom{-}0.00 \pm 0.00$ & $0.12 \pm 0.08$ & $14/59$ & -- \\
      & \texttt{MB}            & $1.00 \pm 0.00$ & $0.99 \pm 0.01$ & $0.00 \pm 0.00$ & $\phantom{-}0.00 \pm 0.00$ & $0.02 \pm 0.01$ & $\phantom{0}0/59$ & $0.61 \pm 0.05$ \\
    \midrule
    \multirow{2}{*}{\texttt{Salamanders}}
      & \MBz{}                 & $1.00 \pm 0.00$ & $1.02 \pm 0.07$ & $0.02 \pm 0.02$ & $\phantom{-}0.01 \pm 0.00$ & $0.12 \pm 0.09$ & $37/51$ & -- \\
      & \texttt{MB}            & $1.00 \pm 0.00$ & $1.00 \pm 0.01$ & $0.00 \pm 0.00$ & $\phantom{-}0.00 \pm 0.00$ & $0.03 \pm 0.03$ & $\phantom{0}0/51$ & $0.47 \pm 0.06$ \\
    \bottomrule
\end{tabular}

\end{table}

\clearpage

\clearpage
\subsection{Simulation-Based Power Analysis} \label{app:power}
\rev{
Before any data collection, a power analysis (PA) can be used to guide researchers in determining how large their study must be to likely detect an effect of a given size \citep{Cohen.1988}.
Frequentist PA is based on null-hypothesis significance tests, Bayesian PA is based on the marginal posterior for the effect of interest \citep{Wang.2002}.
For GLMMs, both approaches to PA are simulation-based: they simulate many datasets (typically $1{,}000$ per condition), fit a model to each, and count how often the effect of interest is detected \citep{Kumle.2021}.
Bayesian PA is considerably slower than its frequentist counterpart \citep{Vasishth.2023} and would strongly profit from amortization. In this sense, \mb can not only save considerable time during statistical analysis, but also during study planning and data collection.}

\paragraph{Setup.}
\begin{itemize}
   \item \rev{\textbf{Design.} We simulate a cluster-randomized controlled trial with $m \in \{10, 20, 40, 60, 80\}$ schools, half of which receive an educational program. $20$ students per school are tested and can either pass or not. We set the control pass rate to $50\%$, the odds ratio (OR) of the program to $1.8$, and the SD of the school log-odds to $0.5$. The GLMM formula is $$\texttt{passed $\sim$ program + (1 | school)}$$ We simulate $1{,}000$ trials per $m$ at OR $1.8$ and separately at OR $1.0$ (to check type I error).}
   \item \rev{\textbf{Priors.} We use a weakly informative prior with intercept $\sim \mathcal N(0, 1.5)$, \texttt{program} $\sim \mathcal N(0, 1)$ and random intercept SD $\sim \mathcal{HN}(1)$. Alternatively, we use a skeptical prior with \texttt{program} $\sim \mathcal N(0, 0.36)$, which puts only $5\%$ prior mass on OR $\geq 1.8$ \citep{Spiegelhalter.1994}.}
  \item \rev{\textbf{Decision rule.} A trial detects the effect if the $95\%$ credible interval of \texttt{program} excludes $0$.}
  \item \rev{\textbf{Methods.} \mb draws $4{,}000$ samples per posterior. It fits all $20{,}000$ posteriors in ten batched calls on one H100 GPU. NUTS refits the first $100$ trials of every cell under both priors ($2{,}000$ fits in total, under the usual budget ladder).}
\end{itemize}
\rev{In every cell, \mb's detection rate should lie within two standard errors of NUTS', and its decisions should agree with NUTS about as often as two NUTS runs agree with each other. Its type I error should stay within $[0.025, 0.075]$ under the weakly informative prior \citep{Bradley.1978}. It should match that of NUTS under the skeptical prior (which is conservative by design).}

\begin{figure}[htbp]
  \centering
  \includegraphics[width=\linewidth]{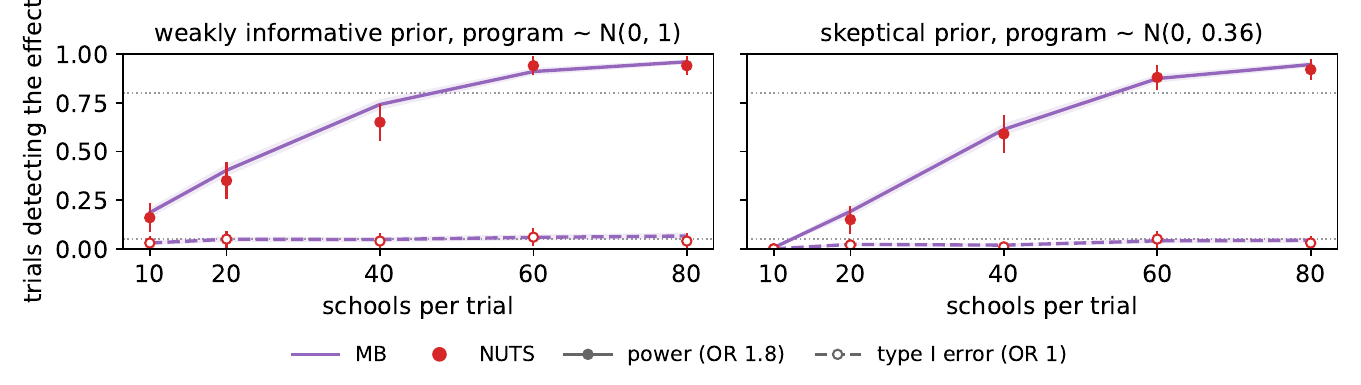}
  \caption{
    \rev{Detection rate against the number of schools under the weakly informative (left) and the skeptical prior (right). Error bands span two Monte Carlo SEs. \mb has ten times more trials, so its SEs are at most $0.016$.}
  }
  \label{fig:power}
\end{figure}

\paragraph{Results.}
\rev{\myfigref{fig:power} shows the power curves (detection rate over number of groups).}
\begin{enumerate}
  \item \rev{\textbf{\mb reaches the same conclusions as NUTS.} On the $2{,}000$
posteriors
fitted by both, \mb and NUTS agree on $98.7\%$ of the decisions. Two NUTS runs agree with each other in $99.0\%$ of the cases. Power estimates differ between the two by at most $0.04$ in any cell, and their posteriors match closely ($\sigma$-ratio $0.99$--$1.00$, rank-MAD ${\leq}\,0.003$). Our hypothesis holds in $19$ of $20$ cells. In the remaining cell (OR $1.8$, $m = 20$, skeptical prior), \mb detects $4$ trials that NUTS misses; all of them are borderline cases where the credible interval borders on $0$. With $20$ cells, the probability of at least one such failure is $0.61$, even if \mb equaled NUTS.}
  \item \rev{\textbf{The analysis takes minutes.} \mb fit all $20{,}000$ posteriors in $9.2$ minutes. NUTS took $111$ core-hours for the $2{,}000$ refits (which extrapolates to about $1{,}110$ core-hours for all $20{,}000$ posteriors).}
  \item \rev{\textbf{The skeptical prior costs power with few schools.} Under both priors, $80\%$ power requires between $40$ and $60$ schools. With $10$ schools, the skeptical prior has a power of only $0.01$ (against $0.19$ for the weakly informative prior). In exchange, its type I error stays below $0.05$ at every $m$ (weakly informative: $0.03$--$0.07$).}
\end{enumerate}

\begin{table}[hbp]
  \caption{
    Emulated fixed-prior baseline on the oracle sets (all four size regimes; NUTS-converged datasets whose $P_0$ lies within the training hyper-prior, $n$ per row).
    \MBz{}($P_0$) is the raw flow conditioned on the reference prior $P_0$ while data and \texttt{NUTS} stay under the analyst's prior $P_1$; $\tau_1/\tau_0$ is the realized median scale ratio.
    Medians over datasets; $\sigma\text{-ratio}_{Q1}$ restricted to the lowest $m/q$ quartile; $\mathrm{EACE}_g$ against the generating parameters (\texttt{NUTS}: ${\leq}0.007$).
    +IMH($P_1$): the same flow refined as IMH proposal towards $P_1$.
    IMH acceptance of \MBz{}($P_0$) as proposal towards $P_1$ and towards $P_0$ (emulation check).
  }
  \label{tab:fixed_prior}
  \centering
  \apptablewide\setlength{\tabcolsep}{3pt}
  \begin{tabular}{ll r r ccc cc cc}
    \toprule
     & & & & \multicolumn{3}{c}{\MBz{}($P_0$) vs.\ \texttt{NUTS}($P_1$)} & \multicolumn{2}{c}{+IMH($P_1$)} & \multicolumn{2}{c}{IMH acceptance} \\
    \cmidrule(lr){5-7}\cmidrule(lr){8-9}\cmidrule(lr){10-11}
    family & $P_0$ & $n$ & $\tau_1/\tau_0$ & $\sigma$-ratio & $\sigma$-ratio$_{Q1}$ & $\mathrm{EACE}_g$ & $\sigma$-ratio & $\mathrm{EACE}_g$ & $\to P_1$ & $\to P_0$ \\
    \midrule
    Gaussian & $P_0 = P_1$ (\texttt{MB}) & 371 & 1 & $1.012$ & $1.020$ & $0.004$ & $0.995$ & $0.027$ & $0.50$ & --- \\
     & $\tau_0 = 3\tau_1$ & 64 & 1/3 & $1.113$ & $1.242$ & $0.035$ & $0.994$ & $0.030$ & $0.41$ & $0.63$ \\
     & $\tau_0 = \tau_1/3$ & 365 & 3 & $0.915$ & $0.874$ & $0.225$ & $0.951$ & $0.189$ & $0.04$ & $0.22$ \\
     & $\nu_0 = \nu_1 + 2\tau_1$ & 371 & 1 & $1.031$ & $1.046$ & $0.089$ & $0.992$ & $0.091$ & $0.08$ & $0.33$ \\
     & hyper-prior mode & 371 & 1.4 & $1.011$ & $1.022$ & $0.088$ & $0.994$ & $0.100$ & $0.09$ & $0.38$ \\
    \midrule
    Bernoulli & $P_0 = P_1$ (\texttt{MB}) & 397 & 1 & $0.999$ & $0.989$ & $0.004$ & $0.997$ & $0.013$ & $0.52$ & --- \\
     & $\tau_0 = 3\tau_1$ & 31 & 1/3 & $1.219$ & $1.295$ & $0.063$ & $1.002$ & $0.012$ & $0.30$ & $0.63$ \\
     & $\tau_0 = \tau_1/3$ & 394 & 3 & $0.646$ & $0.610$ & $0.321$ & $0.843$ & $0.200$ & $0.04$ & $0.43$ \\
     & $\nu_0 = \nu_1 + 2\tau_1$ & 397 & 1 & $1.000$ & $1.013$ & $0.063$ & $0.992$ & $0.048$ & $0.14$ & $0.41$ \\
     & hyper-prior mode & 397 & 1 & $0.997$ & $1.008$ & $0.021$ & $0.995$ & $0.029$ & $0.22$ & $0.51$ \\
    \midrule
    Poisson & $P_0 = P_1$ (\texttt{MB}) & 382 & 1 & $1.009$ & $1.003$ & $0.010$ & $1.000$ & $0.015$ & $0.47$ & --- \\
     & $\tau_0 = 3\tau_1$ & 45 & 1/3 & $1.171$ & $1.200$ & $0.025$ & $1.002$ & $0.048$ & $0.16$ & $0.21$ \\
     & $\tau_0 = \tau_1/3$ & 375 & 3 & $0.889$ & $0.829$ & $0.169$ & $0.961$ & $0.110$ & $0.07$ & $0.34$ \\
     & $\nu_0 = \nu_1 + 2\tau_1$ & 382 & 1 & $1.018$ & $1.018$ & $0.036$ & $0.996$ & $0.044$ & $0.17$ & $0.34$ \\
     & hyper-prior mode & 382 & 1/1.2 & $1.027$ & $1.025$ & $0.003$ & $0.998$ & $0.029$ & $0.23$ & $0.35$ \\
    \bottomrule
\end{tabular}

\end{table}

\clearpage
\subsection{Emulated Fixed-Prior Baseline} \label{app:fixed_prior}
The amortized alternatives closest to \mb in model class fix the prior at training time \citep{Habermann.2025, Arruda.2024}; \rev{prior-conditioned estimators exist for non-hierarchical models \citep{Elsemuller.2024, Whittle.2026, Chang.2025, Yang.2026} or at fixed data \citep{Battaglia.2024}}.
We quantify what this costs an analyst whose prior differs from the training prior. This isolates what prior conditioning buys without training a reference network with a different architecture, capacity and convergence.

\paragraph{Setup.}
\begin{itemize}
  \item \textbf{Data.} The oracle sets of all four size regimes ($128$ datasets per family and regime, NUTS-converged subset, \myappref{app:tra}).
  \item \textbf{Reference priors $P_0$.} Derived from each dataset's $P_1$ (all scales three times wider or tighter; fixed-effect locations shifted by $2\tau_1$; prior families rotated as in \myappref{app:prior_mis}) or fixed for all datasets at the mode of the training hyper-prior (\myappref{app:prior}), the prior a fixed-prior network would sensibly be trained at. The LKJ concentration stays at $P_1$. A dataset enters a condition only if every scale of $P_0$ lies within the training hyper-prior range, so the network never extrapolates; this keeps most datasets for tighter and few for wider $P_0$ ($n$ in \mytabref{tab:fixed_prior}).
  \item \textbf{Arms.} \MBz{}($P_0$); the same flow as IMH proposal towards $P_1$ (\myappref{app:imh}); \mb at $P_1$ as reference. Emulation check: the flow as IMH proposal towards $P_0$, whose acceptance relative to \mb measures how well the network fits at $P_0$.
  \item \textbf{Metrics}, fixed before the larger size regimes were run: $\sigma$-ratio against \texttt{NUTS}($P_1$), overall and in the lowest quartile of groups per random effect ($m/q$); EACE against the generating parameters; IMH acceptance.
\end{itemize}

\paragraph{Results.}
\mytabref{tab:fixed_prior}. Three take-aways:
\begin{enumerate}
  \item \textbf{A fixed prior distorts uncertainty exactly where data are scarce.} A $P_0$ three times wider than $P_1$ makes the posterior $11$--$22\%$ too wide overall, $20$--$30\%$ in the lowest $m/q$ quartile and $4$--$10\%$ in the highest. A $P_0$ three times tighter makes it $9$--$35\%$ too narrow and raises global EACE to $0.17$--$0.32$, against ${\leq}0.007$ for NUTS; for binary outcomes, whose variance components stay prior-dominated, this overconfidence grows with the size regime ($\sigma$-ratio $0.74$ at small to $0.53$ at huge). Posterior means and predictions are insensitive throughout ($r \geq 0.98$, $\Delta\mathrm{LOO\text{-}NLL} \leq 0.06$), and so is rotating the prior families at equal scale ($\sigma$-ratio $1.00$--$1.02$): the damage is confined to uncertainty quantification.
  \item \textbf{IMH repairs a vague fixed prior, not a tight one.} As proposal towards $P_1$, the wider-prior flow recovers NUTS at $0.3$--$0.8$ times the acceptance of \mb. The tighter-prior flow cannot supply the tails of its target: acceptance falls to $0.04$--$0.07$ and the posterior stays $4$--$16\%$ too narrow (global EACE $0.11$--$0.20$). For both, the acceptance gap to \mb widens as posteriors sharpen, from $1.1$--$4$ times at small to $2$--$12$ times at huge.
  \item \textbf{Even a well-chosen fixed prior costs proposal quality.} The hyper-prior mode and a location shift of two prior SDs leave the raw posterior nearly intact ($\sigma$-ratio $1.00$--$1.03$, global EACE $0.003$--$0.09$) but cut acceptance towards $P_1$ to $0.08$--$0.23$, against $0.47$--$0.52$ for \mb.
\end{enumerate}
For \mb this means that prior conditioning buys correct uncertainty quantification without a sampler on the datasets where an analyst cannot check it (few groups, tight priors), and the proposal quality on which the exactness path rests: a fixed-prior network used as IMH proposal pays in acceptance (up to thirteen-fold here) and cannot be repaired once its prior is tighter than the analyst's.

\clearpage
\section{Robustness and Stress Tests} \label{app:rob}
This appendix collects all experiments that deliberately place the test distribution outside the training simulator and probes regimes where amortized inference is expected to fail:
misspecified likelihoods (\myappref{app:lik_mis}), misspecified priors (\myappref{app:prior_mis}),
\rev{out-of-support priors (\myappref{app:prior_oos}),}
out-of-distribution predictors (\myappref{app:ood}),
ill-conditioned designs (\myappref{app:cond}),
data-poor groups (\myappref{app:poverty})
\rev{and oversized datasets (\myappref{app:budget})}
.
The safeguards that flag unreliable posteriors are described in \myappref{app:safe}.

\paragraph{Common protocol.}
NUTS is refit under the same perturbation in every suite.
Under likelihood and prior misspecification both methods therefore target the same wrong posterior, so their agreement separates amortization error from modeling error; under out-of-distribution predictors, ill-conditioned designs and data poverty the fitted GLMM stays correctly specified and NUTS is an exact reference.
Agreement metrics (\myappref{app:met}) are computed pairwise on the NUTS-converged subset (\myappref{app:tra}). Where the generating parameters are known, quality against them is reported for global ($g$) and local ($l$) parameters separately.
Each suite reports \mb next to the ablated raw flow \MBz{} (\myappref{app:abl}), whose predictive gap to NUTS measures how hard a condition hits the amortized proposal.

\paragraph{Common outcome.}
The same picture emerges in every suite
\rev{except under extreme priors (\myappref{app:prior_oos})}%
, so we state it once and report only the suite-specific numbers below.
\mb stays faithful to NUTS under every perturbation: $r \geq 0.99$, posterior widths within a few percent of NUTS, rank-MAD $\leq 0.008$ and $|\Delta\mathrm{LOO\text{-}NLL}| \leq 0.003$ in every condition except a two-dataset Poisson bin ($0.022$; \mytabref{tab:cond_p}). NUTS converges on $75$--$100\%$ of the datasets per condition.
Where the fitted GLMM is wrong, \mb and NUTS degrade in lockstep against the generating parameters.
The raw flow \MBz{} agrees with NUTS in means and largely in marginal shape, but its predictive gap grows with the severity of the stress, and its posterior widths drift where the stress is strongest.
The IMH head closes the predictive gap and removes the width bias in every suite, at the price of a mild finite-pool overconfidence (${\leq}0.03$ EACE over NUTS).

\subsection{Likelihood Misspecification} \label{app:lik_mis}

\paragraph{Setup.}
We use the oracle test sets ($32$ datasets per size regime, $128$ per condition and family) and keep the predictors and generating parameters. We regenerate the outcomes $\mathbf y$ under contaminated likelihoods at three severities per family:
\begin{itemize}
   \item Gaussian $\to$ Student-$t$ ($\nu \in \{10, 5, 3\}$);
   \item Bernoulli $\to$ latent logit noise ($\sigma_c \in \{0.5, 1, 2\}$), the binary analogue of overdispersion;
   \item Poisson $\to$ Negative Binomial (size $\theta \in \{10, 3, 1\}$; smaller $\theta$ = more overdispersion).
\end{itemize}
The raw flow, \mb and NUTS fit the original GLMM, and their posteriors are compared.

\paragraph{Results.}
\mytabref{tab:lik_mis_n}--\ref{tab:lik_mis_p}.
The raw flow's predictive gap grows with severity for Poisson outcomes (median $\Delta$LOO-NLL $0.05 \to 0.48$ at $\theta{=}1$).
Calibration against the generating parameters degrades at the same pace for \mb and NUTS (e.g.\ global EACE $0.24$ vs.\ $0.24$ at $\sigma_c{=}2$; $0.35$ vs.\ $0.34$ at $\theta{=}1$): when the GLMM is wrong, \mb is as wrong as NUTS, but not worse.

\begin{table}[hbp]
  \caption{
    Likelihood misspecification on the Gaussian oracle sets (outcomes regenerated from Student-$t$ with $\nu$ degrees of freedom; $128$ datasets per condition).
    Left: agreement with \texttt{NUTS} fitting the same misspecified model, median $\pm$ MAD over the conv.\ of $128$ datasets that pass the convergence criterion (\myappref{app:tra}); $r$ and rank-MAD as medians only (their MAD is $\leq 0.005$ and $\leq 0.01$ throughout).
    Right: quality against the generating parameters, recovery (NRMSE) and calibration (EACE) for global ($g$) and local ($l$) parameters.
  }
  \label{tab:lik_mis_n}
  \centering
  \apptablewide\setlength{\tabcolsep}{1.3pt}
  \begin{tabular}{ll r c c c c | c c c c}
    \toprule
     & & & \multicolumn{4}{c|}{agreement with \texttt{NUTS}} & \multicolumn{4}{c}{quality vs.\ generating parameters} \\
    $\mathrm{condition}$ & $\mathrm{model}$ & $\mathrm{conv.}$ & $r$ & $\sigma\text{-ratio}$ & $\mathrm{rank\text{-}MAD}$ & $\Delta\mathrm{LOO\text{-}NLL}$ & $\mathrm{NRMSE}_g$ & $\mathrm{EACE}_g$ & $\mathrm{NRMSE}_l$ & $\mathrm{EACE}_l$ \\
    \midrule
    baseline & \MBz{} & 115 & $1.000$ & $1.013 \pm 0.013$ & $0.005$ & $\phantom{-}0.005 \pm 0.004$ & $0.199$ & $0.004$ & $0.345$ & $0.006$ \\
     & \texttt{MB} &  & $1.000$ & $0.999 \pm 0.005$ & $0.002$ & $0.000 \pm 0.000$ & $0.196$ & $0.011$ & $0.346$ & $0.002$ \\
     & \texttt{NUTS} &  & --- & --- & --- & --- & $0.195$ & $0.007$ & $0.346$ & $0.002$ \\
    \midrule
    $t(\nu{=}10)$ & \MBz{} & 117 & $1.000$ & $1.014 \pm 0.018$ & $0.006$ & $\phantom{-}0.004 \pm 0.004$ & $0.255$ & $0.047$ & $0.371$ & $0.011$ \\
     & \texttt{MB} &  & $1.000$ & $1.000 \pm 0.006$ & $0.002$ & $0.000 \pm 0.000$ & $0.251$ & $0.052$ & $0.370$ & $0.014$ \\
     & \texttt{NUTS} &  & --- & --- & --- & --- & $0.251$ & $0.050$ & $0.371$ & $0.014$ \\
    \midrule
    $t(\nu{=}5)$ & \MBz{} & 118 & $1.000$ & $1.009 \pm 0.018$ & $0.006$ & $\phantom{-}0.004 \pm 0.004$ & $0.378$ & $0.062$ & $0.409$ & $0.008$ \\
     & \texttt{MB} &  & $1.000$ & $0.999 \pm 0.006$ & $0.002$ & $0.000 \pm 0.000$ & $0.375$ & $0.067$ & $0.408$ & $0.008$ \\
     & \texttt{NUTS} &  & --- & --- & --- & --- & $0.374$ & $0.064$ & $0.408$ & $0.007$ \\
    \midrule
    $t(\nu{=}3)$ & \MBz{} & 118 & $1.000$ & $1.008 \pm 0.017$ & $0.006$ & $\phantom{-}0.005 \pm 0.005$ & $0.610$ & $0.074$ & $0.457$ & $0.003$ \\
     & \texttt{MB} &  & $1.000$ & $0.999 \pm 0.005$ & $0.001$ & $0.000 \pm 0.001$ & $0.598$ & $0.075$ & $0.459$ & $0.005$ \\
     & \texttt{NUTS} &  & --- & --- & --- & --- & $0.597$ & $0.074$ & $0.459$ & $0.005$ \\
    \bottomrule
\end{tabular}

\end{table}

\begin{table}[hbp]
  \caption{Likelihood misspecification on the Bernoulli oracle sets (unmodeled latent logit noise of scale $\sigma_c$). Layout as in \mytabref{tab:lik_mis_n}.}
  \label{tab:lik_mis_b}
  \centering
  \apptablewide\setlength{\tabcolsep}{1.3pt}
  \begin{tabular}{ll r c c c c | c c c c}
    \toprule
     & & & \multicolumn{4}{c|}{agreement with \texttt{NUTS}} & \multicolumn{4}{c}{quality vs.\ generating parameters} \\
    $\mathrm{condition}$ & $\mathrm{model}$ & $\mathrm{conv.}$ & $r$ & $\sigma\text{-ratio}$ & $\mathrm{rank\text{-}MAD}$ & $\Delta\mathrm{LOO\text{-}NLL}$ & $\mathrm{NRMSE}_g$ & $\mathrm{EACE}_g$ & $\mathrm{NRMSE}_l$ & $\mathrm{EACE}_l$ \\
    \midrule
    baseline & \MBz{} & 124 & $0.998$ & $1.001 \pm 0.030$ & $0.007$ & $0.011 \pm 0.010$ & $0.398$ & $0.007$ & $0.732$ & $0.002$ \\
     & \texttt{MB} &  & $1.000$ & $0.998 \pm 0.009$ & $0.002$ & $0.000 \pm 0.000$ & $0.379$ & $0.005$ & $0.712$ & $0.009$ \\
     & \texttt{NUTS} &  & --- & --- & --- & --- & $0.381$ & $0.006$ & $0.712$ & $0.011$ \\
    \midrule
    $\sigma_c{=}0.5$ & \MBz{} & 126 & $0.998$ & $0.997 \pm 0.028$ & $0.007$ & $0.009 \pm 0.009$ & $0.418$ & $0.010$ & $0.753$ & $0.014$ \\
     & \texttt{MB} &  & $1.000$ & $0.997 \pm 0.009$ & $0.002$ & $0.000 \pm 0.000$ & $0.407$ & $0.019$ & $0.737$ & $0.010$ \\
     & \texttt{NUTS} &  & --- & --- & --- & --- & $0.406$ & $0.013$ & $0.735$ & $0.008$ \\
    \midrule
    $\sigma_c{=}1$ & \MBz{} & 122 & $0.998$ & $0.998 \pm 0.026$ & $0.007$ & $0.011 \pm 0.011$ & $0.422$ & $0.056$ & $0.762$ & $0.027$ \\
     & \texttt{MB} &  & $1.000$ & $0.996 \pm 0.008$ & $0.002$ & $0.000 \pm 0.000$ & $0.417$ & $0.068$ & $0.758$ & $0.024$ \\
     & \texttt{NUTS} &  & --- & --- & --- & --- & $0.416$ & $0.062$ & $0.759$ & $0.022$ \\
    \midrule
    $\sigma_c{=}2$ & \MBz{} & 124 & $0.998$ & $1.007 \pm 0.034$ & $0.007$ & $0.009 \pm 0.009$ & $0.551$ & $0.225$ & $0.801$ & $0.082$ \\
     & \texttt{MB} &  & $1.000$ & $0.999 \pm 0.008$ & $0.002$ & $0.000 \pm 0.000$ & $0.544$ & $0.244$ & $0.793$ & $0.088$ \\
     & \texttt{NUTS} &  & --- & --- & --- & --- & $0.542$ & $0.237$ & $0.790$ & $0.085$ \\
    \bottomrule
\end{tabular}

\end{table}

\begin{table}[hbp]
  \caption{Likelihood misspecification on the Poisson oracle sets (outcomes regenerated from a Negative Binomial with size $\theta$; smaller $\theta$ = more overdispersion). Layout as in \mytabref{tab:lik_mis_n}.}
  \label{tab:lik_mis_p}
  \centering
  \apptablewide\setlength{\tabcolsep}{1.3pt}
  \begin{tabular}{ll r c c c c | c c c c}
    \toprule
     & & & \multicolumn{4}{c|}{agreement with \texttt{NUTS}} & \multicolumn{4}{c}{quality vs.\ generating parameters} \\
    $\mathrm{condition}$ & $\mathrm{model}$ & $\mathrm{conv.}$ & $r$ & $\sigma\text{-ratio}$ & $\mathrm{rank\text{-}MAD}$ & $\Delta\mathrm{LOO\text{-}NLL}$ & $\mathrm{NRMSE}_g$ & $\mathrm{EACE}_g$ & $\mathrm{NRMSE}_l$ & $\mathrm{EACE}_l$ \\
    \midrule
    baseline & \MBz{} & 117 & $0.999$ & $1.005 \pm 0.031$ & $0.008$ & $\phantom{-}0.054 \pm 0.054$ & $0.251$ & $0.007$ & $0.543$ & $0.008$ \\
     & \texttt{MB} &  & $1.000$ & $0.999 \pm 0.007$ & $0.002$ & $\phantom{-}0.000 \pm 0.001$ & $0.258$ & $0.024$ & $0.537$ & $0.007$ \\
     & \texttt{NUTS} &  & --- & --- & --- & --- & $0.260$ & $0.013$ & $0.537$ & $0.007$ \\
    \midrule
    $\theta{=}10$ & \MBz{} & 122 & $0.999$ & $1.008 \pm 0.023$ & $0.008$ & $\phantom{-}0.095 \pm 0.095$ & $0.330$ & $0.087$ & $0.618$ & $0.007$ \\
     & \texttt{MB} &  & $1.000$ & $0.998 \pm 0.006$ & $0.002$ & $0.000 \pm 0.001$ & $0.327$ & $0.111$ & $0.618$ & $0.009$ \\
     & \texttt{NUTS} &  & --- & --- & --- & --- & $0.329$ & $0.095$ & $0.619$ & $0.008$ \\
    \midrule
    $\theta{=}3$ & \MBz{} & 123 & $0.998$ & $1.004 \pm 0.022$ & $0.010$ & $\phantom{-}0.252 \pm 0.251$ & $0.444$ & $0.180$ & $0.773$ & $0.035$ \\
     & \texttt{MB} &  & $1.000$ & $0.997 \pm 0.006$ & $0.002$ & $-0.001 \pm 0.002$ & $0.450$ & $0.208$ & $0.791$ & $0.042$ \\
     & \texttt{NUTS} &  & --- & --- & --- & --- & $0.456$ & $0.194$ & $0.797$ & $0.041$ \\
    \midrule
    $\theta{=}1$ & \MBz{} & 126 & $0.996$ & $1.007 \pm 0.023$ & $0.015$ & $\phantom{-}0.476 \pm 0.472$ & $0.796$ & $0.321$ & $1.168$ & $0.118$ \\
     & \texttt{MB} &  & $1.000$ & $0.998 \pm 0.005$ & $0.002$ & $-0.003 \pm 0.004$ & $0.826$ & $0.352$ & $1.221$ & $0.132$ \\
     & \texttt{NUTS} &  & --- & --- & --- & --- & $0.835$ & $0.344$ & $1.229$ & $0.131$ \\
    \bottomrule
\end{tabular}

\end{table}

\clearpage
\subsection{Prior Misspecification} \label{app:prior_mis}

\paragraph{Setup.}
The prior hyperparameters stored with generated test sets are rewritten while the observations and the true parameters remain untouched: the fitted prior systematically deviates from the one that generated the data.
Five conditions cover wrong scale ($\tau{\times}c$, all scale hyperparameters multiplied by $c \in \{1/3, 3\}$), wrong location ($\nu_\beta{+}k\tau_\beta$, fixed-effect prior means shifted by $k$ prior SDs), and wrong family (family$+1$, all prior-family indices rotated, e.g.\ Normal $\to$ Student-$t$); each uses $128$ datasets ($32$ per size regime) per likelihood family.
\rev{Scale ${\times}3$ already leaves the trained range (\myappref{app:prior}) for $96\%$ of Bernoulli and $94\%$ of Poisson datasets.}

\paragraph{Results.}
\mytabref{tab:prior_mis_n}--\ref{tab:prior_mis_p}.
Prior misspecification opens no gap between the amortized and the exact posterior ($\sigma$-ratio within $1\%$ of one, $|\Delta$LOO-NLL$| \leq 0.001$); NUTS converges on $88$--$98\%$ of the datasets.
Quality against the generating parameters degrades in step with NUTS, and asymmetrically. An over-tight prior ($\tau{\times}1/3$) is the most damaging for all models (global EACE rises to $0.21$--$0.32$ for \mb and $0.20$--$0.32$ for NUTS, from ${\leq}0.03$ and ${\leq}0.01$ under the correct prior). An over-wide prior ($\tau{\times}3$) is largely benign.
The raw flow's predictive gap stays at most $0.03$ for Gaussian and Bernoulli outcomes and reaches $0.34$ for Poisson.

\begin{table}[hbp]
  \caption{
    Prior misspecification on the Gaussian oracle sets (prior scale $\tau{\times}c$, fixed-effect prior location $\nu_\beta{+}k\tau_\beta$, or prior family rotated; \texttt{NUTS} refit under the same perturbed prior; $128$ datasets per condition, pooled over the four size regimes).
    Layout as in \mytabref{tab:lik_mis_n}.
  }
  \label{tab:prior_mis_n}
  \centering
  \apptablewide\setlength{\tabcolsep}{1.3pt}
  \begin{tabular}{ll r c c c c | c c c c}
    \toprule
     & & & \multicolumn{4}{c|}{agreement with \texttt{NUTS}} & \multicolumn{4}{c}{quality vs.\ generating parameters} \\
    $\mathrm{condition}$ & $\mathrm{model}$ & $\mathrm{conv.}$ & $r$ & $\sigma\text{-ratio}$ & $\mathrm{rank\text{-}MAD}$ & $\Delta\mathrm{LOO\text{-}NLL}$ & $\mathrm{NRMSE}_g$ & $\mathrm{EACE}_g$ & $\mathrm{NRMSE}_l$ & $\mathrm{EACE}_l$ \\
    \midrule
    correct & \MBz{} & 115 & $1.000$ & $1.013 \pm 0.021$ & $0.007$ & $0.005 \pm 0.005$ & $0.199$ & $0.003$ & $0.346$ & $0.004$ \\
     & \texttt{MB} &  & $1.000$ & $0.999 \pm 0.005$ & $0.002$ & $0.000 \pm 0.000$ & $0.196$ & $0.010$ & $0.346$ & $0.002$ \\
     & \texttt{NUTS} &  & --- & --- & --- & --- & $0.195$ & $0.007$ & $0.346$ & $0.002$ \\
    \midrule
    $\tau{\times}1/3$ & \MBz{} & 120 & $0.999$ & $1.014 \pm 0.030$ & $0.010$ & $0.011 \pm 0.011$ & $0.283$ & $0.238$ & $0.372$ & $0.088$ \\
     & \texttt{MB} &  & $1.000$ & $0.998 \pm 0.006$ & $0.002$ & $0.000 \pm 0.000$ & $0.262$ & $0.266$ & $0.362$ & $0.091$ \\
     & \texttt{NUTS} &  & --- & --- & --- & --- & $0.260$ & $0.258$ & $0.361$ & $0.086$ \\
    \midrule
    $\tau{\times}3$ & \MBz{} & 115 & $1.000$ & $1.020 \pm 0.023$ & $0.007$ & $0.009 \pm 0.009$ & $0.226$ & $0.023$ & $0.357$ & $0.037$ \\
     & \texttt{MB} &  & $1.000$ & $0.998 \pm 0.005$ & $0.002$ & $0.000 \pm 0.000$ & $0.219$ & $0.012$ & $0.356$ & $0.031$ \\
     & \texttt{NUTS} &  & --- & --- & --- & --- & $0.219$ & $0.014$ & $0.356$ & $0.032$ \\
    \midrule
    $\nu_\beta{+}1\tau_\beta$ & \MBz{} & 118 & $1.000$ & $1.013 \pm 0.018$ & $0.007$ & $0.006 \pm 0.006$ & $0.206$ & $0.031$ & $0.348$ & $0.002$ \\
     & \texttt{MB} &  & $1.000$ & $1.000 \pm 0.006$ & $0.002$ & $0.000 \pm 0.000$ & $0.204$ & $0.036$ & $0.347$ & $0.003$ \\
     & \texttt{NUTS} &  & --- & --- & --- & --- & $0.203$ & $0.034$ & $0.347$ & $0.003$ \\
    \midrule
    $\nu_\beta{+}2\tau_\beta$ & \MBz{} & 115 & $0.999$ & $1.016 \pm 0.020$ & $0.008$ & $0.011 \pm 0.010$ & $0.256$ & $0.090$ & $0.379$ & $0.006$ \\
     & \texttt{MB} &  & $1.000$ & $0.998 \pm 0.005$ & $0.002$ & $0.000 \pm 0.000$ & $0.238$ & $0.102$ & $0.378$ & $0.008$ \\
     & \texttt{NUTS} &  & --- & --- & --- & --- & $0.237$ & $0.097$ & $0.378$ & $0.006$ \\
    \midrule
    family$+1$ & \MBz{} & 116 & $1.000$ & $1.014 \pm 0.019$ & $0.007$ & $0.006 \pm 0.005$ & $0.201$ & $0.004$ & $0.350$ & $0.006$ \\
     & \texttt{MB} &  & $1.000$ & $0.999 \pm 0.005$ & $0.002$ & $0.000 \pm 0.000$ & $0.199$ & $0.009$ & $0.350$ & $0.003$ \\
     & \texttt{NUTS} &  & --- & --- & --- & --- & $0.199$ & $0.008$ & $0.349$ & $0.003$ \\
    \bottomrule
\end{tabular}

\end{table}

\begin{table}[hbp]
  \caption{Prior misspecification on the Bernoulli oracle sets. Layout as in \mytabref{tab:prior_mis_n}.}
  \label{tab:prior_mis_b}
  \centering
  \apptablewide\setlength{\tabcolsep}{1.3pt}
  \begin{tabular}{ll r c c c c | c c c c}
    \toprule
     & & & \multicolumn{4}{c|}{agreement with \texttt{NUTS}} & \multicolumn{4}{c}{quality vs.\ generating parameters} \\
    $\mathrm{condition}$ & $\mathrm{model}$ & $\mathrm{conv.}$ & $r$ & $\sigma\text{-ratio}$ & $\mathrm{rank\text{-}MAD}$ & $\Delta\mathrm{LOO\text{-}NLL}$ & $\mathrm{NRMSE}_g$ & $\mathrm{EACE}_g$ & $\mathrm{NRMSE}_l$ & $\mathrm{EACE}_l$ \\
    \midrule
    correct & \MBz{} & 124 & $0.998$ & $1.000 \pm 0.029$ & $0.007$ & $0.012 \pm 0.010$ & $0.398$ & $0.008$ & $0.732$ & $0.002$ \\
     & \texttt{MB} &  & $1.000$ & $0.999 \pm 0.009$ & $0.002$ & $0.000 \pm 0.000$ & $0.381$ & $0.003$ & $0.712$ & $0.009$ \\
     & \texttt{NUTS} &  & --- & --- & --- & --- & $0.381$ & $0.006$ & $0.712$ & $0.011$ \\
    \midrule
    $\tau{\times}1/3$ & \MBz{} & 125 & $0.998$ & $1.001 \pm 0.078$ & $0.013$ & $0.006 \pm 0.005$ & $0.610$ & $0.316$ & $0.810$ & $0.209$ \\
     & \texttt{MB} &  & $1.000$ & $1.000 \pm 0.010$ & $0.002$ & $0.000 \pm 0.000$ & $0.581$ & $0.324$ & $0.788$ & $0.198$ \\
     & \texttt{NUTS} &  & --- & --- & --- & --- & $0.579$ & $0.317$ & $0.786$ & $0.196$ \\
    \midrule
    $\tau{\times}3$ & \MBz{} & 125 & $0.996$ & $0.946 \pm 0.061$ & $0.014$ & $0.030 \pm 0.028$ & $0.520$ & $0.060$ & $0.769$ & $0.044$ \\
     & \texttt{MB} &  & $1.000$ & $0.990 \pm 0.013$ & $0.003$ & $0.000 \pm 0.000$ & $0.616$ & $0.023$ & $0.779$ & $0.066$ \\
     & \texttt{NUTS} &  & --- & --- & --- & --- & $0.740$ & $0.023$ & $0.823$ & $0.074$ \\
    \midrule
    $\nu_\beta{+}1\tau_\beta$ & \MBz{} & 121 & $0.998$ & $0.987 \pm 0.040$ & $0.009$ & $0.013 \pm 0.012$ & $0.414$ & $0.012$ & $0.745$ & $0.007$ \\
     & \texttt{MB} &  & $1.000$ & $0.996 \pm 0.008$ & $0.002$ & $0.000 \pm 0.000$ & $0.398$ & $0.030$ & $0.728$ & $0.007$ \\
     & \texttt{NUTS} &  & --- & --- & --- & --- & $0.399$ & $0.025$ & $0.727$ & $0.010$ \\
    \midrule
    $\nu_\beta{+}2\tau_\beta$ & \MBz{} & 112 & $0.998$ & $0.969 \pm 0.044$ & $0.011$ & $0.012 \pm 0.010$ & $0.489$ & $0.083$ & $0.745$ & $0.006$ \\
     & \texttt{MB} &  & $1.000$ & $0.996 \pm 0.006$ & $0.002$ & $0.000 \pm 0.000$ & $0.497$ & $0.103$ & $0.737$ & $0.011$ \\
     & \texttt{NUTS} &  & --- & --- & --- & --- & $0.510$ & $0.099$ & $0.742$ & $0.013$ \\
    \midrule
    family$+1$ & \MBz{} & 123 & $0.998$ & $0.991 \pm 0.043$ & $0.009$ & $0.013 \pm 0.012$ & $0.401$ & $0.019$ & $0.732$ & $0.004$ \\
     & \texttt{MB} &  & $1.000$ & $0.996 \pm 0.009$ & $0.002$ & $0.000 \pm 0.000$ & $0.385$ & $0.004$ & $0.714$ & $0.008$ \\
     & \texttt{NUTS} &  & --- & --- & --- & --- & $0.389$ & $0.010$ & $0.714$ & $0.012$ \\
    \bottomrule
\end{tabular}

\end{table}

\begin{table}[hbp]
  \caption{Prior misspecification on the Poisson oracle sets. Layout as in \mytabref{tab:prior_mis_n}.}
  \label{tab:prior_mis_p}
  \centering
  \apptablewide\setlength{\tabcolsep}{1.3pt}
  \begin{tabular}{ll r c c c c | c c c c}
    \toprule
     & & & \multicolumn{4}{c|}{agreement with \texttt{NUTS}} & \multicolumn{4}{c}{quality vs.\ generating parameters} \\
    $\mathrm{condition}$ & $\mathrm{model}$ & $\mathrm{conv.}$ & $r$ & $\sigma\text{-ratio}$ & $\mathrm{rank\text{-}MAD}$ & $\Delta\mathrm{LOO\text{-}NLL}$ & $\mathrm{NRMSE}_g$ & $\mathrm{EACE}_g$ & $\mathrm{NRMSE}_l$ & $\mathrm{EACE}_l$ \\
    \midrule
    correct & \MBz{} & 117 & $0.999$ & $1.006 \pm 0.033$ & $0.010$ & $0.054 \pm 0.054$ & $0.254$ & $0.008$ & $0.545$ & $0.009$ \\
     & \texttt{MB} &  & $1.000$ & $0.999 \pm 0.007$ & $0.002$ & $0.000 \pm 0.001$ & $0.257$ & $0.026$ & $0.538$ & $0.008$ \\
     & \texttt{NUTS} &  & --- & --- & --- & --- & $0.260$ & $0.013$ & $0.537$ & $0.007$ \\
    \midrule
    $\tau{\times}1/3$ & \MBz{} & 121 & $0.999$ & $1.041 \pm 0.046$ & $0.012$ & $0.036 \pm 0.036$ & $0.371$ & $0.168$ & $0.580$ & $0.090$ \\
     & \texttt{MB} &  & $1.000$ & $0.997 \pm 0.006$ & $0.002$ & $0.000 \pm 0.000$ & $0.372$ & $0.214$ & $0.568$ & $0.102$ \\
     & \texttt{NUTS} &  & --- & --- & --- & --- & $0.371$ & $0.202$ & $0.568$ & $0.100$ \\
    \midrule
    $\tau{\times}3$ & \MBz{} & 121 & $0.997$ & $1.021 \pm 0.063$ & $0.012$ & $0.340 \pm 0.339$ & $0.344$ & $0.012$ & $0.572$ & $0.028$ \\
     & \texttt{MB} &  & $1.000$ & $0.998 \pm 0.010$ & $0.003$ & $0.000 \pm 0.001$ & $0.348$ & $0.043$ & $0.556$ & $0.025$ \\
     & \texttt{NUTS} &  & --- & --- & --- & --- & $0.357$ & $0.014$ & $0.556$ & $0.027$ \\
    \midrule
    $\nu_\beta{+}1\tau_\beta$ & \MBz{} & 121 & $0.998$ & $1.018 \pm 0.043$ & $0.010$ & $0.076 \pm 0.075$ & $0.282$ & $0.020$ & $0.569$ & $0.010$ \\
     & \texttt{MB} &  & $1.000$ & $0.997 \pm 0.008$ & $0.002$ & $0.000 \pm 0.001$ & $0.277$ & $0.045$ & $0.558$ & $0.012$ \\
     & \texttt{NUTS} &  & --- & --- & --- & --- & $0.278$ & $0.031$ & $0.557$ & $0.013$ \\
    \midrule
    $\nu_\beta{+}2\tau_\beta$ & \MBz{} & 123 & $0.998$ & $1.017 \pm 0.041$ & $0.011$ & $0.131 \pm 0.130$ & $0.310$ & $0.054$ & $0.561$ & $0.008$ \\
     & \texttt{MB} &  & $1.000$ & $0.997 \pm 0.008$ & $0.002$ & $0.000 \pm 0.000$ & $0.303$ & $0.083$ & $0.549$ & $0.010$ \\
     & \texttt{NUTS} &  & --- & --- & --- & --- & $0.303$ & $0.066$ & $0.550$ & $0.009$ \\
    \midrule
    family$+1$ & \MBz{} & 118 & $0.998$ & $1.016 \pm 0.042$ & $0.011$ & $0.121 \pm 0.120$ & $0.262$ & $0.007$ & $0.532$ & $0.003$ \\
     & \texttt{MB} &  & $1.000$ & $0.998 \pm 0.008$ & $0.003$ & $0.000 \pm 0.001$ & $0.259$ & $0.024$ & $0.521$ & $0.004$ \\
     & \texttt{NUTS} &  & --- & --- & --- & --- & $0.261$ & $0.012$ & $0.521$ & $0.003$ \\
    \bottomrule
\end{tabular}

\end{table}

\clearpage
\subsection{Out-of-Support Priors} \label{app:prior_oos}
\rev{The five conditions above were not designed to leave the trained ranges. To test behavior on out-of-sample priors, we deliberately moved the prior beyond the training range.}
\paragraph{Setup.}
\rev{We re-ran the experiment in \myappref{app:prior_mis} on the medium regime with four additional conditions:}
\begin{itemize}
  \item \rev{$\nu_\beta{+}5\tau_\beta$: fixed-effect prior means shifted by $5$ prior SDs,}
  \item \rev{$\tau{=}1.1{\times}\max$ and $\tau{=}2{\times}\max$: every scale hyperparameter was set at $1.1$ resp. $2$ times the upper end of its training range,}
  \item \rev{$\tau{=}\min/2$: every scale hyperparameter was set at half the lower end of its training range.}
\end{itemize}
\rev{\mytabref{tab:prior_oos} counts the datasets where \mb is off NUTS on its worst parameter (see \myappref{app:agree}).}

\paragraph{Results.}
\rev{\mytabref{tab:prior_oos} shows the comparison.}
\begin{itemize}
  \item \rev{\textbf{Extreme prior means} ($\nu_\beta{+}5\tau_\beta$). The raw flow leaves NUTS on $25$--$100\%$ of the datasets, \mb on up to $46\%$.}
  \item \rev{\textbf{Borderline prior scales} ($\tau{=}1.1{\times}\max$). The raw flow leaves NUTS on $48$--$92\%$ of the datasets, \mb on one dataset.}
  \item \rev{\textbf{Extreme prior scales} ($\tau{=}2{\times}\max$). The raw flow leaves NUTS on $95$--$100\%$ of the datasets, \mb on at most $8\%$.}
  \item \rev{\textbf{Highly critical priors} ($\tau{=}\min/2$). The raw flow leaves NUTS on every dataset, \mb on up to $68\%$.}
  \item \rev{\textbf{Flags.} The safeguards flag every dataset where \mb leaves NUTS.}
\end{itemize}
\rev{Beyond the trained ranges, \mb repairs most of the stress to the raw flow. However, under extreme prior means and highly critical priors \mb also leaves NUTS. The safeguards raise a flag in every such case, so \mb can degrade out of support but it never degrades silently.}

\begin{table}[hbp]
  \caption{
    \rev{Per-dataset agreement of \MBz{} and \mb with NUTS under out-of-support priors, over the NUTS-converged datasets of the medium regime.
    Each dataset is rated by its worst parameter (\myappref{app:agree}).
    Silent failure counts the datasets where \mb is off NUTS without a flag.}
  }
  \label{tab:prior_oos}
  \centering
  \apptable
  \begin{tabular}{ll|rrrrr}
    \toprule
     & & & \multicolumn{2}{c}{$\mathrm{off\ NUTS}$} & & \\
    \cmidrule(lr){4-5}
    $\mathrm{family}$ & $\mathrm{prior}$ & $\mathrm{datasets}$ & \MBz{} & \texttt{MB} & $\text{silent failure}$ & $\mathrm{flagged}$ \\
    \midrule
    Gaussian  & correct & $117$ & $17$ & $0$ & $0$ & $14\%$ \\
              & $\nu_\beta{+}5\tau_\beta$ & $119$ & $119$ & $55$ & $0$ & $100\%$ \\
              & $\tau{=}1.1{\times}\max$ & $117$ & $108$ & $0$ & $0$ & $66\%$ \\
              & $\tau{=}2{\times}\max$ & $118$ & $118$ & $9$ & $0$ & $100\%$ \\
              & $\tau{=}\min/2$ & $28$ & $28$ & $5$ & $0$ & $100\%$ \\
    \midrule
    Bernoulli & correct & $121$ & $8$ & $0$ & $0$ & $7\%$ \\
              & $\nu_\beta{+}5\tau_\beta$ & $70$ & $53$ & $5$ & $0$ & $64\%$ \\
              & $\tau{=}1.1{\times}\max$ & $121$ & $64$ & $1$ & $0$ & $28\%$ \\
              & $\tau{=}2{\times}\max$ & $121$ & $118$ & $4$ & $0$ & $77\%$ \\
              & $\tau{=}\min/2$ & $31$ & $31$ & $18$ & $0$ & $100\%$ \\
    \midrule
    Poisson   & correct & $121$ & $18$ & $0$ & $0$ & $13\%$ \\
              & $\nu_\beta{+}5\tau_\beta$ & $118$ & $29$ & $0$ & $0$ & $25\%$ \\
              & $\tau{=}1.1{\times}\max$ & $120$ & $57$ & $0$ & $0$ & $31\%$ \\
              & $\tau{=}2{\times}\max$ & $122$ & $116$ & $1$ & $0$ & $66\%$ \\
              & $\tau{=}\min/2$ & $31$ & $31$ & $21$ & $0$ & $100\%$ \\
    \bottomrule
\end{tabular}

\end{table}

\clearpage
\subsection{Out-of-Distribution Predictors} \label{app:ood}

\paragraph{Setup.}
The training design space is broad but structurally constrained: all simulated and real training designs have finite variance, asymptotically independent tails, and no temporal structure within groups.
We probe each constraint directly.
Taking the same held-out test sets as in \myappref{app:lik_mis} and keeping the priors, likelihood and parameters the same, we re-sample the predictors from distributions strictly outside the training support and regenerate the outcomes:
\begin{itemize}
   \item heavy-tailed marginals ($X \sim t(\nu{=}2)$ and $X \sim \mathrm{Cauchy}$, i.e.\ infinite variance);
   \item Clayton-copula dependence (Kendall $\tau \in \{0.5, 0.9\}$, inducing tail dependence);
   \item and longitudinal growth-curve designs (time as an additional predictor with a random slope).
\end{itemize}
The fitted GLMM therefore remains correctly specified, and any gap isolates the amortization error.

\paragraph{Results.}
\mytabref{tab:ood_n}--\ref{tab:ood_p}.
The raw flow's predictive performance decreases on the extremes (median $\Delta$LOO-NLL $0.24$ for Bernoulli and $1.29$ for Poisson under Cauchy predictors), which the IMH head repairs.
Even strictly outside the training support there is barely a gap between \mb and NUTS (posterior widths within $1\%$, $|\Delta$LOO-NLL$| \leq 0.001$).

\begin{table}[hbp]
  \caption{
    Out-of-distribution predictors on the Gaussian oracle sets: predictors re-sampled from $t(\nu{=}2)$ or Cauchy marginals, Clayton copulas with Kendall $\tau \in \{0.5, 0.9\}$, or longitudinal designs, outcomes regenerated under the original likelihood ($128$ datasets per condition).
    Layout as in \mytabref{tab:lik_mis_n}.
  }
  \label{tab:ood_n}
  \centering
  \apptablewide\setlength{\tabcolsep}{1.3pt}
  \begin{tabular}{ll r c c c c | c c c c}
    \toprule
     & & & \multicolumn{4}{c|}{agreement with \texttt{NUTS}} & \multicolumn{4}{c}{quality vs.\ generating parameters} \\
    $\mathrm{condition}$ & $\mathrm{model}$ & $\mathrm{conv.}$ & $r$ & $\sigma\text{-ratio}$ & $\mathrm{rank\text{-}MAD}$ & $\Delta\mathrm{LOO\text{-}NLL}$ & $\mathrm{NRMSE}_g$ & $\mathrm{EACE}_g$ & $\mathrm{NRMSE}_l$ & $\mathrm{EACE}_l$ \\
    \midrule
    baseline & \MBz{} & 115 & $1.000$ & $1.013 \pm 0.013$ & $0.005$ & $\phantom{-}0.005 \pm 0.004$ & $0.199$ & $0.004$ & $0.345$ & $0.006$ \\
     & \texttt{MB} &  & $1.000$ & $0.999 \pm 0.005$ & $0.002$ & $0.000 \pm 0.000$ & $0.196$ & $0.011$ & $0.346$ & $0.002$ \\
     & \texttt{NUTS} &  & --- & --- & --- & --- & $0.195$ & $0.007$ & $0.346$ & $0.002$ \\
    \midrule
    $t(\nu{=}2)$ & \MBz{} & 118 & $1.000$ & $1.006 \pm 0.012$ & $0.005$ & $\phantom{-}0.004 \pm 0.004$ & $0.166$ & $0.006$ & $0.370$ & $0.012$ \\
     & \texttt{MB} &  & $1.000$ & $1.000 \pm 0.005$ & $0.002$ & $0.000 \pm 0.000$ & $0.164$ & $0.010$ & $0.370$ & $0.009$ \\
     & \texttt{NUTS} &  & --- & --- & --- & --- & $0.163$ & $0.010$ & $0.369$ & $0.009$ \\
    \midrule
    Cauchy & \MBz{} & 118 & $1.000$ & $1.013 \pm 0.022$ & $0.007$ & $\phantom{-}0.022 \pm 0.021$ & $0.240$ & $0.006$ & $0.577$ & $0.010$ \\
     & \texttt{MB} &  & $1.000$ & $0.998 \pm 0.006$ & $0.002$ & $0.000 \pm 0.001$ & $0.209$ & $0.009$ & $0.574$ & $0.003$ \\
     & \texttt{NUTS} &  & --- & --- & --- & --- & $0.209$ & $0.007$ & $0.573$ & $0.002$ \\
    \midrule
    Clayton $0.5$ & \MBz{} & 114 & $1.000$ & $1.006 \pm 0.017$ & $0.006$ & $\phantom{-}0.003 \pm 0.003$ & $0.193$ & $0.036$ & $0.336$ & $0.003$ \\
     & \texttt{MB} &  & $1.000$ & $0.999 \pm 0.005$ & $0.002$ & $\phantom{-}0.000 \pm 0.000$ & $0.182$ & $0.019$ & $0.333$ & $0.004$ \\
     & \texttt{NUTS} &  & --- & --- & --- & --- & $0.182$ & $0.019$ & $0.333$ & $0.004$ \\
    \midrule
    Clayton $0.9$ & \MBz{} & 110 & $0.999$ & $1.023 \pm 0.023$ & $0.009$ & $\phantom{-}0.022 \pm 0.021$ & $0.280$ & $0.052$ & $0.317$ & $0.009$ \\
     & \texttt{MB} &  & $1.000$ & $1.000 \pm 0.007$ & $0.002$ & $0.000 \pm 0.000$ & $0.263$ & $0.011$ & $0.314$ & $0.003$ \\
     & \texttt{NUTS} &  & --- & --- & --- & --- & $0.263$ & $0.011$ & $0.314$ & $0.003$ \\
    \midrule
    longitudinal & \MBz{} & 114 & $1.000$ & $1.008 \pm 0.016$ & $0.005$ & $\phantom{-}0.003 \pm 0.002$ & $0.160$ & $0.013$ & $0.390$ & $0.005$ \\
     & \texttt{MB} &  & $1.000$ & $1.000 \pm 0.005$ & $0.002$ & $0.000 \pm 0.000$ & $0.155$ & $0.005$ & $0.390$ & $0.005$ \\
     & \texttt{NUTS} &  & --- & --- & --- & --- & $0.155$ & $0.005$ & $0.390$ & $0.005$ \\
    \bottomrule
\end{tabular}

\end{table}

\begin{table}[hbp]
  \caption{Out-of-distribution predictors on the Bernoulli oracle sets. Layout as in \mytabref{tab:ood_n}.}
  \label{tab:ood_b}
  \centering
  \apptablewide\setlength{\tabcolsep}{1.3pt}
  \begin{tabular}{ll r c c c c | c c c c}
    \toprule
     & & & \multicolumn{4}{c|}{agreement with \texttt{NUTS}} & \multicolumn{4}{c}{quality vs.\ generating parameters} \\
    $\mathrm{condition}$ & $\mathrm{model}$ & $\mathrm{conv.}$ & $r$ & $\sigma\text{-ratio}$ & $\mathrm{rank\text{-}MAD}$ & $\Delta\mathrm{LOO\text{-}NLL}$ & $\mathrm{NRMSE}_g$ & $\mathrm{EACE}_g$ & $\mathrm{NRMSE}_l$ & $\mathrm{EACE}_l$ \\
    \midrule
    baseline & \MBz{} & 124 & $0.998$ & $1.001 \pm 0.030$ & $0.007$ & $0.011 \pm 0.010$ & $0.398$ & $0.007$ & $0.732$ & $0.002$ \\
     & \texttt{MB} &  & $1.000$ & $0.998 \pm 0.009$ & $0.002$ & $0.000 \pm 0.000$ & $0.379$ & $0.005$ & $0.712$ & $0.009$ \\
     & \texttt{NUTS} &  & --- & --- & --- & --- & $0.381$ & $0.006$ & $0.712$ & $0.011$ \\
    \midrule
    $t(\nu{=}2)$ & \MBz{} & 123 & $0.998$ & $1.000 \pm 0.028$ & $0.007$ & $0.017 \pm 0.016$ & $0.337$ & $0.003$ & $0.719$ & $0.011$ \\
     & \texttt{MB} &  & $1.000$ & $0.999 \pm 0.008$ & $0.002$ & $0.000 \pm 0.000$ & $0.323$ & $0.008$ & $0.714$ & $0.002$ \\
     & \texttt{NUTS} &  & --- & --- & --- & --- & $0.325$ & $0.005$ & $0.713$ & $0.001$ \\
    \midrule
    Cauchy & \MBz{} & 121 & $0.993$ & $1.002 \pm 0.033$ & $0.011$ & $0.239 \pm 0.148$ & $0.488$ & $0.041$ & $0.817$ & $0.017$ \\
     & \texttt{MB} &  & $1.000$ & $0.993 \pm 0.012$ & $0.003$ & $0.000 \pm 0.000$ & $0.450$ & $0.012$ & $0.805$ & $0.005$ \\
     & \texttt{NUTS} &  & --- & --- & --- & --- & $0.443$ & $0.018$ & $0.804$ & $0.003$ \\
    \midrule
    Clayton $0.5$ & \MBz{} & 125 & $0.998$ & $0.990 \pm 0.032$ & $0.008$ & $0.007 \pm 0.006$ & $0.335$ & $0.018$ & $0.705$ & $0.018$ \\
     & \texttt{MB} &  & $1.000$ & $0.995 \pm 0.009$ & $0.002$ & $0.000 \pm 0.000$ & $0.316$ & $0.005$ & $0.688$ & $0.003$ \\
     & \texttt{NUTS} &  & --- & --- & --- & --- & $0.318$ & $0.004$ & $0.687$ & $0.004$ \\
    \midrule
    Clayton $0.9$ & \MBz{} & 120 & $0.997$ & $1.005 \pm 0.052$ & $0.011$ & $0.013 \pm 0.011$ & $0.521$ & $0.039$ & $0.759$ & $0.015$ \\
     & \texttt{MB} &  & $1.000$ & $0.997 \pm 0.009$ & $0.002$ & $0.000 \pm 0.000$ & $0.507$ & $0.005$ & $0.741$ & $0.004$ \\
     & \texttt{NUTS} &  & --- & --- & --- & --- & $0.506$ & $0.005$ & $0.741$ & $0.004$ \\
    \midrule
    longitudinal & \MBz{} & 125 & $0.999$ & $0.990 \pm 0.030$ & $0.007$ & $0.005 \pm 0.004$ & $0.312$ & $0.025$ & $0.686$ & $0.020$ \\
     & \texttt{MB} &  & $1.000$ & $0.997 \pm 0.007$ & $0.002$ & $0.000 \pm 0.000$ & $0.294$ & $0.022$ & $0.672$ & $0.004$ \\
     & \texttt{NUTS} &  & --- & --- & --- & --- & $0.295$ & $0.017$ & $0.672$ & $0.003$ \\
    \bottomrule
\end{tabular}

\end{table}

\begin{table}[hbp]
  \caption{Out-of-distribution predictors on the Poisson oracle sets. Layout as in \mytabref{tab:ood_n}.}
  \label{tab:ood_p}
  \centering
  \apptablewide\setlength{\tabcolsep}{1.3pt}
  \begin{tabular}{ll r c c c c | c c c c}
    \toprule
     & & & \multicolumn{4}{c|}{agreement with \texttt{NUTS}} & \multicolumn{4}{c}{quality vs.\ generating parameters} \\
    $\mathrm{condition}$ & $\mathrm{model}$ & $\mathrm{conv.}$ & $r$ & $\sigma\text{-ratio}$ & $\mathrm{rank\text{-}MAD}$ & $\Delta\mathrm{LOO\text{-}NLL}$ & $\mathrm{NRMSE}_g$ & $\mathrm{EACE}_g$ & $\mathrm{NRMSE}_l$ & $\mathrm{EACE}_l$ \\
    \midrule
    baseline & \MBz{} & 117 & $0.999$ & $1.005 \pm 0.031$ & $0.008$ & $0.054 \pm 0.054$ & $0.251$ & $0.007$ & $0.543$ & $0.008$ \\
     & \texttt{MB} &  & $1.000$ & $0.999 \pm 0.007$ & $0.002$ & $0.000 \pm 0.001$ & $0.258$ & $0.024$ & $0.537$ & $0.007$ \\
     & \texttt{NUTS} &  & --- & --- & --- & --- & $0.260$ & $0.013$ & $0.537$ & $0.007$ \\
    \midrule
    $t(\nu{=}2)$ & \MBz{} & 112 & $0.999$ & $0.994 \pm 0.022$ & $0.009$ & $0.215 \pm 0.215$ & $0.232$ & $0.010$ & $0.575$ & $0.004$ \\
     & \texttt{MB} &  & $1.000$ & $0.998 \pm 0.007$ & $0.002$ & $0.000 \pm 0.001$ & $0.225$ & $0.012$ & $0.572$ & $0.015$ \\
     & \texttt{NUTS} &  & --- & --- & --- & --- & $0.226$ & $0.008$ & $0.571$ & $0.015$ \\
    \midrule
    Cauchy & \MBz{} & 118 & $0.998$ & $1.011 \pm 0.029$ & $0.008$ & $1.293 \pm 1.291$ & $0.311$ & $0.018$ & $0.750$ & $0.003$ \\
     & \texttt{MB} &  & $1.000$ & $1.000 \pm 0.008$ & $0.002$ & $0.000 \pm 0.001$ & $0.306$ & $0.005$ & $0.743$ & $0.004$ \\
     & \texttt{NUTS} &  & --- & --- & --- & --- & $0.306$ & $0.005$ & $0.743$ & $0.002$ \\
    \midrule
    Clayton $0.5$ & \MBz{} & 113 & $0.999$ & $1.009 \pm 0.032$ & $0.009$ & $0.040 \pm 0.040$ & $0.249$ & $0.025$ & $0.536$ & $0.012$ \\
     & \texttt{MB} &  & $1.000$ & $0.999 \pm 0.006$ & $0.002$ & $0.000 \pm 0.000$ & $0.243$ & $0.007$ & $0.529$ & $0.013$ \\
     & \texttt{NUTS} &  & --- & --- & --- & --- & $0.245$ & $0.008$ & $0.529$ & $0.012$ \\
    \midrule
    Clayton $0.9$ & \MBz{} & 115 & $0.999$ & $1.011 \pm 0.034$ & $0.011$ & $0.261 \pm 0.260$ & $0.372$ & $0.062$ & $0.596$ & $0.022$ \\
     & \texttt{MB} &  & $1.000$ & $0.998 \pm 0.006$ & $0.002$ & $0.000 \pm 0.001$ & $0.359$ & $0.025$ & $0.587$ & $0.022$ \\
     & \texttt{NUTS} &  & --- & --- & --- & --- & $0.358$ & $0.015$ & $0.587$ & $0.018$ \\
    \midrule
    longitudinal & \MBz{} & 115 & $0.999$ & $1.009 \pm 0.026$ & $0.008$ & $0.052 \pm 0.051$ & $0.188$ & $0.021$ & $0.516$ & $0.011$ \\
     & \texttt{MB} &  & $1.000$ & $1.000 \pm 0.006$ & $0.002$ & $0.000 \pm 0.000$ & $0.186$ & $0.005$ & $0.512$ & $0.008$ \\
     & \texttt{NUTS} &  & --- & --- & --- & --- & $0.185$ & $0.005$ & $0.511$ & $0.007$ \\
    \bottomrule
\end{tabular}

\end{table}

\clearpage
\subsection{Ill-Conditioned Designs} \label{app:cond}

\paragraph{Setup.}
Real design matrices are often collinear, so we quantify agreement with NUTS as a function of the condition number $\kappa_2(\mathbf{X}) = \sigma_{\max} / \sigma_{\min}$ of each real dataset's fixed-effects design. Here, $\sigma_*$ denotes singular values of the design's SVD, and a large ratio indicates high collinearity.
Unlike the preceding suites this is a post-hoc analysis: simulated designs are near-orthogonal ($\kappa_2 \approx 1$).
In the real-world benchmarks of each family, collinearity arises naturally and $\kappa_2$ spans $1$ to ${\sim}10^{15}$ with a near-singular tail. Results are binned by $\kappa_2$.

\paragraph{Results.}
\mytabref{tab:cond_n}--\ref{tab:cond_p}.
Collinearity is the axis that stresses the amortized proposal most.
The raw flow remains faithful in posterior means ($r \geq 0.99$ at all $\kappa_2$) but degrades in width and predictive fit: for Gaussian outcomes it widens ($\sigma$-ratio up to $1.43$) and its predictive gap grows to $0.79$ with $\kappa_2$; for the discrete families it narrows instead ($\sigma$-ratio down to $0.96$ for Bernoulli, $0.85$ for Poisson). Predictive gaps stay small for Bernoulli (${\leq}0.06$) but grow large for Poisson ($2.51$ in the $[6, 10)$ bin).
NUTS itself converges less often on collinear Gaussian data ($99\% \to 87\%$ before the near-singular tail).
\mb holds NUTS-level agreement across the whole range, including the near-singular tail at $\kappa_2 \approx 10^{15}$, with one visible residue: Poisson posteriors narrow mildly at moderate to high $\kappa_2$ ($\sigma$-ratio down to $0.97$), whereas Bernoulli widths stay at NUTS level ($\sigma$-ratio $\geq 0.99$).

A bin-free analysis confirms this pattern on the Gaussian benchmarks: Spearman rank correlations over all converged datasets show the raw flow degrading monotonically with $\kappa_2$ ($r_s(\kappa_2, \Delta\mathrm{LOO}) = +0.81$, $r_s(\kappa_2, \sigma\text{-ratio}) = +0.65$), while for \mb both rank correlations nearly vanish ($r_s = +0.02$ and $+0.11$) over differences too small to matter ($\sigma$-ratio $1.00$ and $|\Delta\mathrm{LOO}| < 0.001$ in every bin).

\begin{table}[hbp]
  \caption{
    Posterior agreement vs.\ design condition number $\kappa_2(\mathbf{X})$ on the Gaussian real-world benchmarks.
    Entries are median $\pm$ MAD over the NUTS-converged datasets of each bin; $\%\,\mathrm{conv.}$ is the NUTS convergence rate per bin (\myappref{app:tra}) and $\tilde\kappa$ the median condition number within the bin.
  }
  \label{tab:cond_n}
  \centering
  \apptablewide
  \begin{tabular}{llrrrr|cccc}
    \toprule
    $\kappa_2(X)$ & $\mathrm{model}$ & $n$ & $n_{\mathrm{conv}}$ & $\%\,\mathrm{conv.}$ & $\tilde\kappa$ & $r$ & $\sigma\text{-ratio}$ & $\mathrm{rank\text{-}MAD}$ & $\Delta\mathrm{LOO\text{-}NLL}$ \\
    \midrule
    $[1, 3)$ & \MBz{} & 376 & 374 & 99 & $1.00$ & $1.000 \pm 0.000$ & $1.000 \pm 0.018$ & $0.005 \pm 0.003$ & $\phantom{-}0.001 \pm 0.001$ \\
     & \texttt{MB} &  &  &  &  & $1.000 \pm 0.000$ & $0.997 \pm 0.007$ & $0.002 \pm 0.001$ & $\phantom{-}0.000 \pm 0.000$ \\
    \midrule
    $[3, 6)$ & \MBz{} & 555 & 529 & 95 & $4.44$ & $0.999 \pm 0.001$ & $1.044 \pm 0.050$ & $0.011 \pm 0.007$ & $\phantom{-}0.010 \pm 0.008$ \\
     & \texttt{MB} &  &  &  &  & $1.000 \pm 0.000$ & $0.999 \pm 0.006$ & $0.002 \pm 0.001$ & $\phantom{-}0.000 \pm 0.000$ \\
    \midrule
    $[6, 10)$ & \MBz{} & 919 & 864 & 94 & $7.95$ & $0.990 \pm 0.008$ & $1.219 \pm 0.147$ & $0.024 \pm 0.010$ & $\phantom{-}0.158 \pm 0.132$ \\
     & \texttt{MB} &  &  &  &  & $1.000 \pm 0.000$ & $1.000 \pm 0.007$ & $0.002 \pm 0.001$ & $\phantom{-}0.000 \pm 0.000$ \\
    \midrule
    $[10, 10^{6})$ & \MBz{} & 127 & 110 & 87 & $11.01$ & $0.988 \pm 0.010$ & $1.431 \pm 0.223$ & $0.031 \pm 0.014$ & $\phantom{-}0.786 \pm 0.731$ \\
     & \texttt{MB} &  &  &  &  & $1.000 \pm 0.000$ & $1.003 \pm 0.011$ & $0.003 \pm 0.002$ & $\phantom{-}0.000 \pm 0.001$ \\
    \midrule
    $[10^{6}, \infty)$ & \MBz{} & 71 & 69 & 97 & $3.5\times10^{15}$ & $0.987 \pm 0.009$ & $1.422 \pm 0.251$ & $0.030 \pm 0.014$ & $\phantom{-}0.220 \pm 0.164$ \\
     & \texttt{MB} &  &  &  &  & $0.999 \pm 0.001$ & $1.001 \pm 0.007$ & $0.002 \pm 0.001$ & $\phantom{-}0.000 \pm 0.000$ \\
    \bottomrule
\end{tabular}

\end{table}

\begin{table}[hbp]
  \caption{Posterior agreement vs.\ $\kappa_2(\mathbf{X})$ on the Bernoulli real-world benchmarks. Layout as in \mytabref{tab:cond_n}.}
  \label{tab:cond_b}
  \centering
  \apptablewide
  \begin{tabular}{llrrrr|cccc}
    \toprule
    $\kappa_2(X)$ & $\mathrm{model}$ & $n$ & $n_{\mathrm{conv}}$ & $\%\,\mathrm{conv.}$ & $\tilde\kappa$ & $r$ & $\sigma\text{-ratio}$ & $\mathrm{rank\text{-}MAD}$ & $\Delta\mathrm{LOO\text{-}NLL}$ \\
    \midrule
    $[1, 3)$ & \MBz{} & 354 & 351 & 99 & $1.05$ & $0.999 \pm 0.001$ & $1.027 \pm 0.025$ & $0.010 \pm 0.005$ & $\phantom{-}0.002 \pm 0.002$ \\
     & \texttt{MB} &  &  &  &  & $1.000 \pm 0.000$ & $0.998 \pm 0.007$ & $0.002 \pm 0.001$ & $\phantom{-}0.000 \pm 0.000$ \\
    \midrule
    $[3, 6)$ & \MBz{} & 637 & 620 & 97 & $4.65$ & $0.996 \pm 0.003$ & $1.006 \pm 0.032$ & $0.009 \pm 0.004$ & $\phantom{-}0.007 \pm 0.005$ \\
     & \texttt{MB} &  &  &  &  & $1.000 \pm 0.000$ & $0.998 \pm 0.006$ & $0.002 \pm 0.001$ & $\phantom{-}0.000 \pm 0.000$ \\
    \midrule
    $[6, 10)$ & \MBz{} & 961 & 921 & 96 & $7.68$ & $0.993 \pm 0.006$ & $0.992 \pm 0.035$ & $0.008 \pm 0.004$ & $\phantom{-}0.029 \pm 0.018$ \\
     & \texttt{MB} &  &  &  &  & $0.999 \pm 0.000$ & $0.996 \pm 0.007$ & $0.002 \pm 0.001$ & $\phantom{-}0.000 \pm 0.000$ \\
    \midrule
    $[10, 10^{6})$ & \MBz{} & 67 & 65 & 97 & $10.50$ & $0.993 \pm 0.005$ & $0.964 \pm 0.030$ & $0.010 \pm 0.004$ & $\phantom{-}0.064 \pm 0.029$ \\
     & \texttt{MB} &  &  &  &  & $0.999 \pm 0.000$ & $0.990 \pm 0.009$ & $0.002 \pm 0.001$ & $-0.001 \pm 0.001$ \\
    \midrule
    $[10^{6}, \infty)$ & \MBz{} & 29 & 29 & 100 & $3.9\times10^{15}$ & $0.998 \pm 0.002$ & $0.957 \pm 0.039$ & $0.009 \pm 0.004$ & $\phantom{-}0.018 \pm 0.009$ \\
     & \texttt{MB} &  &  &  &  & $0.999 \pm 0.000$ & $0.999 \pm 0.010$ & $0.002 \pm 0.001$ & $-0.001 \pm 0.001$ \\
    \bottomrule
\end{tabular}

\end{table}

\begin{table}[h]
  \caption{
    Posterior agreement vs.\ $\kappa_2(\mathbf{X})$ on the Poisson real-world benchmarks. Layout as in \mytabref{tab:cond_n}.
    The near-singular bin is absent: only two Poisson datasets exceed $\kappa_2 = 10$.
  }
  \label{tab:cond_p}
  \centering
  \apptablewide
  \begin{tabular}{llrrrr|cccc}
    \toprule
    $\kappa_2(X)$ & $\mathrm{model}$ & $n$ & $n_{\mathrm{conv}}$ & $\%\,\mathrm{conv.}$ & $\tilde\kappa$ & $r$ & $\sigma\text{-ratio}$ & $\mathrm{rank\text{-}MAD}$ & $\Delta\mathrm{LOO\text{-}NLL}$ \\
    \midrule
    $[1, 3)$ & \MBz{} & 479 & 361 & 75 & $1.00$ & $0.995 \pm 0.003$ & $0.996 \pm 0.111$ & $0.035 \pm 0.018$ & $\phantom{-}0.114 \pm 0.110$ \\
     & \texttt{MB} &  &  &  &  & $1.000 \pm 0.000$ & $1.001 \pm 0.015$ & $0.005 \pm 0.003$ & $\phantom{-}0.000 \pm 0.002$ \\
    \midrule
    $[3, 6)$ & \MBz{} & 688 & 672 & 98 & $4.28$ & $0.990 \pm 0.005$ & $0.936 \pm 0.070$ & $0.031 \pm 0.012$ & $\phantom{-}0.319 \pm 0.249$ \\
     & \texttt{MB} &  &  &  &  & $1.000 \pm 0.000$ & $0.983 \pm 0.014$ & $0.005 \pm 0.002$ & $-0.001 \pm 0.001$ \\
    \midrule
    $[6, 10)$ & \MBz{} & 367 & 361 & 98 & $8.21$ & $0.987 \pm 0.007$ & $0.895 \pm 0.073$ & $0.031 \pm 0.013$ & $\phantom{-}2.510 \pm 2.262$ \\
     & \texttt{MB} &  &  &  &  & $1.000 \pm 0.000$ & $0.968 \pm 0.016$ & $0.006 \pm 0.002$ & $-0.002 \pm 0.002$ \\
    \midrule
    $[10, 10^{6})$ & \MBz{} & 2 & 2 & 100 & $11.43$ & $0.997 \pm 0.002$ & $0.852 \pm 0.008$ & $0.033 \pm 0.000$ & $\phantom{-}284.704 \pm 280.484$ \\
     & \texttt{MB} &  &  &  &  & $1.000 \pm 0.000$ & $0.978 \pm 0.007$ & $0.008 \pm 0.005$ & $-0.022 \pm 0.029$ \\
    \bottomrule
\end{tabular}

\end{table}

\subsection{Data-Poor Groups} \label{app:poverty}

\paragraph{Setup.}
Hierarchical models are routinely fit to groups with fewer observations than parameters, so we quantify posterior quality as a function of the per-group information ratio $\rho = n_i / (d + q)$. For GLMMs, $\rho < 1$ is the local $n < p$ regime.
The oracle test sets cover this regime densely: $37$--$47\%$ of all groups fall below $\rho = 1$ across the three likelihood families (minima $0.26$--$0.33$).
We complement $\rho$ with the most adversarial global count, $\gamma = n / p_\mathrm{all}$ with $p_\mathrm{all} = d + mq + q(q+1)/2 + 1$, i.e.\ every random effect counted as a free parameter.
Even so, $\gamma$ almost never drops below $1$ (minimum $1.1$), as identifying the $q \times q$ random-effect covariance requires $m \gtrsim q(q+1)/2$ groups.
Metrics aggregate over the parameter entries pooled within each bin, size regimes pooled and NUTS restricted to its converged subset, so entries are single values rather than median $\pm$ MAD.

\paragraph{Results.}
\mytabref{tab:poverty_n}--\ref{tab:poverty_p} score the local (random-effect) posteriors by $\rho$ bin and the global posteriors by $\gamma$ bin against the ground truth.
\mb matches NUTS in local recovery and calibration in the data-poorest bins ($\rho < 1$) of every family and is never meaningfully worse elsewhere (local NRMSE within $0.01$ of NUTS where it trails).
For the global parameters, 
recovery and calibration match NUTS throughout; on Poisson data the head is also what makes the predictive fit usable in the smallest bin (mean LOO-NLL $21.6$ raw $\to 1.46$, NUTS $1.45$).
Exactly where amortization is often expected to fail, in groups carrying less than one observation per parameter, \mb delivers honest uncertainty and NUTS-level recovery.

\begin{table}[hbp]
  \caption{
    Posterior quality in the data-poor regime on the Gaussian oracle benchmarks, size regimes pooled.
    Top: local (random-effect) posteriors by per-group $\rho = n_i/(d+q)$; \#grp and \#conv count all groups and those in NUTS-converged datasets.
    Bottom: global parameters by per-dataset $\gamma = n/p_\mathrm{all}$ (every random effect counted as a free parameter); \#ds and \#conv count datasets.
    Entries pool the parameter entries within each bin; $c_{90}$ is the coverage of the $90\%$ credible interval (nominally $0.900$) and LOO the mean per-dataset LOO-NLL.
  }
  \label{tab:poverty_n}
  \centering
  \apptablewide\setlength{\tabcolsep}{1.2pt}
  \begin{tabular}{lrr|ccc|ccc|ccc}
    \toprule
     & & & \multicolumn{3}{c|}{\MBz{}} & \multicolumn{3}{c|}{\texttt{MB}} & \multicolumn{3}{c}{\texttt{NUTS}} \\
    $\rho=n_i/(d+q)$ & $\#\,\mathrm{grp}$ & $\#\,\mathrm{conv}$ & $\mathrm{EACE}$ & $\mathrm{NRMSE}$ & $c_{90}$ & $\mathrm{EACE}$ & $\mathrm{NRMSE}$ & $c_{90}$ & $\mathrm{EACE}$ & $\mathrm{NRMSE}$ & $c_{90}$ \\
    \midrule
    $[0,0.5)$ & 17029 & 14770 & $0.002$ & $0.476$ & $0.899$ & $0.004$ & $0.473$ & $0.898$ & $0.005$ & $0.497$ & $0.897$ \\
    $[0.5,1)$ & 23262 & 20042 & $0.001$ & $0.418$ & $0.902$ & $0.002$ & $0.417$ & $0.898$ & $0.004$ & $0.444$ & $0.896$ \\
    $[1,2)$ & 24180 & 20612 & $0.002$ & $0.348$ & $0.902$ & $0.001$ & $0.347$ & $0.899$ & $0.001$ & $0.389$ & $0.899$ \\
    $[2,5)$ & 16514 & 13997 & $0.015$ & $0.362$ & $0.909$ & $0.012$ & $0.360$ & $0.906$ & $0.011$ & $0.411$ & $0.905$ \\
    $[5,\infty)$ & 7901 & 7113 & $0.007$ & $0.271$ & $0.905$ & $0.007$ & $0.270$ & $0.904$ & $0.002$ & $0.277$ & $0.900$ \\
    \bottomrule
\end{tabular}
\par\medskip
\begin{tabular}{lrr|cccc|cccc|cccc}
    \toprule
     & & & \multicolumn{4}{c|}{\MBz{}} & \multicolumn{4}{c|}{\texttt{MB}} & \multicolumn{4}{c}{\texttt{NUTS}} \\
    $\gamma=n/p$ & $\#\,\mathrm{ds}$ & $\#\,\mathrm{conv}$ & $\mathrm{EACE}$ & $\mathrm{NRMSE}$ & $c_{90}$ & $\mathrm{LOO}$ & $\mathrm{EACE}$ & $\mathrm{NRMSE}$ & $c_{90}$ & $\mathrm{LOO}$ & $\mathrm{EACE}$ & $\mathrm{NRMSE}$ & $c_{90}$ & $\mathrm{LOO}$ \\
    \midrule
    $[0,2)$ & 197 & 175 & $0.011$ & $0.264$ & $0.911$ & $2.474$ & $0.004$ & $0.245$ & $0.903$ & $2.373$ & $0.006$ & $0.254$ & $0.905$ & $2.393$ \\
    $[2,4)$ & 447 & 401 & $0.003$ & $0.219$ & $0.901$ & $2.495$ & $0.006$ & $0.218$ & $0.891$ & $2.334$ & $0.004$ & $0.226$ & $0.897$ & $2.348$ \\
    $[4,8)$ & 558 & 518 & $0.006$ & $0.186$ & $0.905$ & $2.306$ & $0.001$ & $0.182$ & $0.898$ & $2.216$ & $0.001$ & $0.187$ & $0.900$ & $2.224$ \\
    $[8,\infty)$ & 846 & 760 & $0.008$ & $0.167$ & $0.907$ & $2.272$ & $0.003$ & $0.159$ & $0.899$ & $2.194$ & $0.003$ & $0.167$ & $0.900$ & $2.245$ \\
    \bottomrule
\end{tabular}

\end{table}

\begin{table}[hbp]
  \caption{Posterior quality in the data-poor regime on the Bernoulli oracle benchmarks. Layout as in \mytabref{tab:poverty_n}.}
  \label{tab:poverty_b}
  \centering
  \apptablewide\setlength{\tabcolsep}{1.2pt}
  \begin{tabular}{lrr|ccc|ccc|ccc}
    \toprule
     & & & \multicolumn{3}{c|}{\MBz{}} & \multicolumn{3}{c|}{\texttt{MB}} & \multicolumn{3}{c}{\texttt{NUTS}} \\
    $\rho=n_i/(d+q)$ & $\#\,\mathrm{grp}$ & $\#\,\mathrm{conv}$ & $\mathrm{EACE}$ & $\mathrm{NRMSE}$ & $c_{90}$ & $\mathrm{EACE}$ & $\mathrm{NRMSE}$ & $c_{90}$ & $\mathrm{EACE}$ & $\mathrm{NRMSE}$ & $c_{90}$ \\
    \midrule
    $[0,0.5)$ & 17854 & 16919 & $0.008$ & $0.826$ & $0.892$ & $0.002$ & $0.822$ & $0.896$ & $0.003$ & $0.825$ & $0.900$ \\
    $[0.5,1)$ & 23797 & 22439 & $0.004$ & $0.735$ & $0.902$ & $0.005$ & $0.730$ & $0.903$ & $0.007$ & $0.731$ & $0.904$ \\
    $[1,2)$ & 21924 & 20762 & $0.009$ & $0.644$ & $0.898$ & $0.007$ & $0.638$ & $0.897$ & $0.007$ & $0.634$ & $0.899$ \\
    $[2,5)$ & 16630 & 15953 & $0.004$ & $0.555$ & $0.898$ & $0.003$ & $0.551$ & $0.900$ & $0.004$ & $0.562$ & $0.899$ \\
    $[5,\infty)$ & 8695 & 8461 & $0.014$ & $0.484$ & $0.895$ & $0.006$ & $0.477$ & $0.900$ & $0.006$ & $0.481$ & $0.900$ \\
    \bottomrule
\end{tabular}
\par\medskip
\begin{tabular}{lrr|cccc|cccc|cccc}
    \toprule
     & & & \multicolumn{4}{c|}{\MBz{}} & \multicolumn{4}{c|}{\texttt{MB}} & \multicolumn{4}{c}{\texttt{NUTS}} \\
    $\gamma=n/p$ & $\#\,\mathrm{ds}$ & $\#\,\mathrm{conv}$ & $\mathrm{EACE}$ & $\mathrm{NRMSE}$ & $c_{90}$ & $\mathrm{LOO}$ & $\mathrm{EACE}$ & $\mathrm{NRMSE}$ & $c_{90}$ & $\mathrm{LOO}$ & $\mathrm{EACE}$ & $\mathrm{NRMSE}$ & $c_{90}$ & $\mathrm{LOO}$ \\
    \midrule
    $[0,2)$ & 184 & 175 & $0.004$ & $0.562$ & $0.904$ & $0.582$ & $0.006$ & $0.559$ & $0.899$ & $0.463$ & $0.005$ & $0.564$ & $0.909$ & $0.465$ \\
    $[2,4)$ & 406 & 387 & $0.005$ & $0.453$ & $0.902$ & $0.508$ & $0.007$ & $0.444$ & $0.895$ & $0.450$ & $0.002$ & $0.448$ & $0.899$ & $0.450$ \\
    $[4,8)$ & 550 & 531 & $0.006$ & $0.404$ & $0.907$ & $0.477$ & $0.003$ & $0.403$ & $0.894$ & $0.449$ & $0.004$ & $0.410$ & $0.896$ & $0.451$ \\
    $[8,\infty)$ & 908 & 885 & $0.007$ & $0.308$ & $0.906$ & $0.476$ & $0.003$ & $0.303$ & $0.898$ & $0.459$ & $0.006$ & $0.302$ & $0.902$ & $0.460$ \\
    \bottomrule
\end{tabular}

\end{table}

\begin{table}[hbp]
  \caption{Posterior quality in the data-poor regime on the Poisson oracle benchmarks. Layout as in \mytabref{tab:poverty_n}.}
  \label{tab:poverty_p}
  \centering
  \apptablewide\setlength{\tabcolsep}{1.2pt}
  \begin{tabular}{lrr|ccc|ccc|ccc}
    \toprule
     & & & \multicolumn{3}{c|}{\MBz{}} & \multicolumn{3}{c|}{\texttt{MB}} & \multicolumn{3}{c}{\texttt{NUTS}} \\
    $\rho=n_i/(d+q)$ & $\#\,\mathrm{grp}$ & $\#\,\mathrm{conv}$ & $\mathrm{EACE}$ & $\mathrm{NRMSE}$ & $c_{90}$ & $\mathrm{EACE}$ & $\mathrm{NRMSE}$ & $c_{90}$ & $\mathrm{EACE}$ & $\mathrm{NRMSE}$ & $c_{90}$ \\
    \midrule
    $[0,0.5)$ & 6919 & 6172 & $0.022$ & $0.678$ & $0.890$ & $0.018$ & $0.673$ & $0.889$ & $0.014$ & $0.673$ & $0.896$ \\
    $[0.5,1)$ & 15641 & 14262 & $0.005$ & $0.623$ & $0.895$ & $0.004$ & $0.619$ & $0.892$ & $0.008$ & $0.621$ & $0.889$ \\
    $[1,2)$ & 15988 & 14465 & $0.012$ & $0.516$ & $0.895$ & $0.014$ & $0.514$ & $0.891$ & $0.016$ & $0.530$ & $0.890$ \\
    $[2,5)$ & 14207 & 13004 & $0.025$ & $0.473$ & $0.882$ & $0.020$ & $0.469$ & $0.887$ & $0.021$ & $0.516$ & $0.886$ \\
    $[5,\infty)$ & 8471 & 8159 & $0.015$ & $0.422$ & $0.895$ & $0.011$ & $0.420$ & $0.897$ & $0.012$ & $0.432$ & $0.896$ \\
    \bottomrule
\end{tabular}
\par\medskip
\begin{tabular}{lrr|cccc|cccc|cccc}
    \toprule
     & & & \multicolumn{4}{c|}{\MBz{}} & \multicolumn{4}{c|}{\texttt{MB}} & \multicolumn{4}{c}{\texttt{NUTS}} \\
    $\gamma=n/p$ & $\#\,\mathrm{ds}$ & $\#\,\mathrm{conv}$ & $\mathrm{EACE}$ & $\mathrm{NRMSE}$ & $c_{90}$ & $\mathrm{LOO}$ & $\mathrm{EACE}$ & $\mathrm{NRMSE}$ & $c_{90}$ & $\mathrm{LOO}$ & $\mathrm{EACE}$ & $\mathrm{NRMSE}$ & $c_{90}$ & $\mathrm{LOO}$ \\
    \midrule
    $[0,2)$ & 94 & 86 & $0.010$ & $0.390$ & $0.898$ & $21.621$ & $0.010$ & $0.383$ & $0.887$ & $1.464$ & $0.013$ & $0.382$ & $0.892$ & $1.448$ \\
    $[2,4)$ & 273 & 256 & $0.005$ & $0.309$ & $0.899$ & $7.102$ & $0.008$ & $0.310$ & $0.894$ & $1.448$ & $0.004$ & $0.317$ & $0.903$ & $1.443$ \\
    $[4,8)$ & 420 & 397 & $0.004$ & $0.252$ & $0.896$ & $9.504$ & $0.008$ & $0.249$ & $0.893$ & $1.369$ & $0.006$ & $0.252$ & $0.893$ & $1.369$ \\
    $[8,\infty)$ & 743 & 715 & $0.003$ & $0.250$ & $0.902$ & $4.419$ & $0.005$ & $0.244$ & $0.895$ & $1.348$ & $0.003$ & $0.252$ & $0.899$ & $1.341$ \\
    \bottomrule
\end{tabular}

\end{table}

\clearpage
\subsection{Oversized Datasets} \label{app:budget}
\rev{\mb was trained on datasets with at most $200$ groups and $3{,}000$ observations (\myappref{app:sizes}), and the released code flags datasets beyond these limits.
To check whether \mb generalizes outside them, we tested it directly on datasets with up to three times its training limits.}

\paragraph{Setup.}
\begin{itemize}
   \item \rev{\textbf{Design.} We vary the number of groups $m$ and the number of observations $N$ over ten cells: three cells within the training limits ($m \le 200$, $N \le 3{,}000$), three beyond them in groups (up to $m = 600$), two beyond them in observations (up to $N = 6{,}000$), and two beyond them in both.}
  \item \rev{\textbf{Simulated data.} We simulate $16$ datasets per family with $d = 4$ and $q = 1$, each once with $600$ groups. Each cell takes the first $m$ groups and their first $N/m$ observations. GLMM parameters
are kept constant across cells.}
  \item \rev{\textbf{Real data.} We subsample three datasets in the same way as above ($16$ times per source): \textit{Stroop reaction times} (Gaussian, $685$ participants; \citealp{Singmann.2025}), \textit{Yelp reviews} (Bernoulli, $888$ restaurants; \citealp{Roback.2021}) and \textit{Chem97 exam scores} (Poisson, $2{,}410$ schools; \citealp{Bates.2020}).}
  \item \rev{\textbf{Priors.} Each dataset and subsample gets one prior drawn from the training hyperprior, shared across its cells.}
  \item \rev{\textbf{Methods.} We pass the inference problems to the raw flow, \mb, and NUTS in the usual way, and rate each dataset by its worst parameter (see \myappref{app:agree}).}
\end{itemize}
\rev{Beyond the training limits, \mb should leave NUTS no more often than within them, and the safeguards should flag every dataset where it does.}

\paragraph{Results.}
\rev{\mytabref{tab:budget} shows where the raw flow and \mb disagree with NUTS.}
\begin{enumerate}
   \item \rev{\textbf{\mb agrees with NUTS beyond the training limits.} Of the $648$ datasets beyond the limits, \mb is off NUTS on none. Within the limits it is off on one of $279$: a Stroop subsample with an IMH acceptance of $0.005$ (flagged by the safeguards).}
  \item \rev{\textbf{The raw flow drifts as the datasets grow.} Within the limits, the raw flow is off NUTS on $0$--$31\%$ of the datasets per cell. Beyond them it is off on up to $47\%$ when only $m$ grows, up to $69\%$ when only $N$ grows and up to $100\%$ when both grow.}
  \item \rev{\textbf{The safeguards grow more cautious.} The median IMH acceptance falls to $0.14$--$0.60$ at $600 \times 9{,}000$. There the safeguards flag up to $47\%$ of the simulated and up to $100\%$ of the real datasets, although \mb agrees with NUTS on each of them. Within the limits they flag $0$--$12\%$.}
\end{enumerate}
\rev{\mb generalizes to datasets with up to three times as many groups and observations as seen in training. However, larger datasets lower the IMH acceptance. In this experiment, the safeguards never miss a disagreement with NUTS.}

\begin{table}[hbp]
  \caption{
     \rev{Per-dataset agreement of \MBz{} and \texttt{MB} with NUTS on oversized datasets. The comparison is limited to the cases where NUTS converged. The first three cells of each family lie within the training limits.
    Each dataset is rated by its worst parameter (\myappref{app:agree}).
    \texttt{silent} counts the cases where \mb is off NUTS without being flagged by the safeguards.}
  }
  \label{tab:budget}
  \centering
  \apptable
  \begin{tabular}{lrr@{\hspace{10pt}}rrrrr@{\hspace{10pt}}rrrrr}
    \toprule
     & & & \multicolumn{5}{c}{simulated} & \multicolumn{5}{c}{real} \\
    \cmidrule(lr){4-8} \cmidrule(lr){9-13}
     & & & & \multicolumn{2}{c}{$\mathrm{off\ NUTS}$} & & & & \multicolumn{2}{c}{$\mathrm{off\ NUTS}$} & & \\
    \cmidrule(lr){5-6} \cmidrule(lr){10-11}
    $\mathrm{family}$ & $m$ & $N$ & $\mathrm{datasets}$ & \MBz{} & \texttt{MB} & \texttt{silent} & $\mathrm{flagged}$ & $\mathrm{datasets}$ & \MBz{} & \texttt{MB} & \texttt{silent} & $\mathrm{flagged}$ \\
    \midrule
    Gaussian  & $100$ & $3{,}000$ & $12$ & $1$ & $0$ & $0$ & $0\%$ & $16$ & $5$ & $1$ & $0$ & $12\%$ \\
              & $200$ & $1{,}500$ & $15$ & $0$ & $0$ & $0$ & $0\%$ & $16$ & $2$ & $0$ & $0$ & $6\%$ \\
              & $200$ & $3{,}000$ & $15$ & $0$ & $0$ & $0$ & $0\%$ & $16$ & $4$ & $0$ & $0$ & $6\%$ \\
              & $250$ & $3{,}000$ & $14$ & $0$ & $0$ & $0$ & $0\%$ & $16$ & $3$ & $0$ & $0$ & $6\%$ \\
              & $400$ & $3{,}000$ & $15$ & $1$ & $0$ & $0$ & $0\%$ & $16$ & $3$ & $0$ & $0$ & $6\%$ \\
              & $600$ & $3{,}000$ & $14$ & $1$ & $0$ & $0$ & $0\%$ & $16$ & $3$ & $0$ & $0$ & $44\%$ \\
              & $200$ & $3{,}500$ & $13$ & $0$ & $0$ & $0$ & $0\%$ & $16$ & $4$ & $0$ & $0$ & $6\%$ \\
              & $200$ & $6{,}000$ & $12$ & $1$ & $0$ & $0$ & $0\%$ & $16$ & $4$ & $0$ & $0$ & $6\%$ \\
              & $400$ & $6{,}000$ & $14$ & $0$ & $0$ & $0$ & $0\%$ & $16$ & $3$ & $0$ & $0$ & $12\%$ \\
              & $600$ & $9{,}000$ & $13$ & $1$ & $0$ & $0$ & $0\%$ & $16$ & $2$ & $0$ & $0$ & $31\%$ \\
    \midrule
    Bernoulli & $100$ & $3{,}000$ & $16$ & $2$ & $0$ & $0$ & $0\%$ & $16$ & $0$ & $0$ & $0$ & $6\%$ \\
              & $200$ & $1{,}500$ & $16$ & $2$ & $0$ & $0$ & $6\%$ & $16$ & $0$ & $0$ & $0$ & $0\%$ \\
              & $200$ & $3{,}000$ & $16$ & $3$ & $0$ & $0$ & $6\%$ & $16$ & $0$ & $0$ & $0$ & $0\%$ \\
              & $250$ & $3{,}000$ & $16$ & $3$ & $0$ & $0$ & $19\%$ & $16$ & $0$ & $0$ & $0$ & $0\%$ \\
              & $400$ & $3{,}000$ & $16$ & $6$ & $0$ & $0$ & $25\%$ & $16$ & $0$ & $0$ & $0$ & $0\%$ \\
              & $600$ & $3{,}000$ & $15$ & $7$ & $0$ & $0$ & $33\%$ & $16$ & $1$ & $0$ & $0$ & $0\%$ \\
              & $200$ & $3{,}500$ & $16$ & $2$ & $0$ & $0$ & $0\%$ & $16$ & $0$ & $0$ & $0$ & $0\%$ \\
              & $200$ & $6{,}000$ & $16$ & $4$ & $0$ & $0$ & $19\%$ & $16$ & $4$ & $0$ & $0$ & $12\%$ \\
              & $400$ & $6{,}000$ & $16$ & $7$ & $0$ & $0$ & $44\%$ & $16$ & $6$ & $0$ & $0$ & $25\%$ \\
              & $600$ & $9{,}000$ & $15$ & $10$ & $0$ & $0$ & $47\%$ & $16$ & $16$ & $0$ & $0$ & $75\%$ \\
    \midrule
    Poisson   & $100$ & $3{,}000$ & $15$ & $1$ & $0$ & $0$ & $0\%$ & $15$ & $1$ & $0$ & $0$ & $0\%$ \\
              & $200$ & $1{,}500$ & $16$ & $0$ & $0$ & $0$ & $0\%$ & $16$ & $1$ & $0$ & $0$ & $6\%$ \\
              & $200$ & $3{,}000$ & $15$ & $0$ & $0$ & $0$ & $0\%$ & $16$ & $4$ & $0$ & $0$ & $6\%$ \\
              & $250$ & $3{,}000$ & $15$ & $0$ & $0$ & $0$ & $0\%$ & $16$ & $5$ & $0$ & $0$ & $19\%$ \\
              & $400$ & $3{,}000$ & $16$ & $0$ & $0$ & $0$ & $6\%$ & $16$ & $3$ & $0$ & $0$ & $0\%$ \\
              & $600$ & $3{,}000$ & $16$ & $0$ & $0$ & $0$ & $6\%$ & $16$ & $5$ & $0$ & $0$ & $0\%$ \\
              & $200$ & $3{,}500$ & $15$ & $0$ & $0$ & $0$ & $0\%$ & $16$ & $7$ & $0$ & $0$ & $25\%$ \\
              & $200$ & $6{,}000$ & $15$ & $1$ & $0$ & $0$ & $13\%$ & $16$ & $11$ & $0$ & $0$ & $38\%$ \\
              & $400$ & $6{,}000$ & $15$ & $1$ & $0$ & $0$ & $20\%$ & $16$ & $14$ & $0$ & $0$ & $31\%$ \\
              & $600$ & $9{,}000$ & $15$ & $3$ & $0$ & $0$ & $27\%$ & $16$ & $16$ & $0$ & $0$ & $100\%$ \\
    \bottomrule
\end{tabular}

\end{table}

\clearpage
\section{Architecture and Training Details} \label{app:arch}

\subsection{Data Representation and Summary Networks} \label{app:emb}\label{app:set}
Observations within each group are concatenated to $\mathbf{D}_i = [\mathbf{y}_i, \mathbf{X}_i, \mathbf{Z}_i]$ (intercept columns excluded) and processed as an unordered set; group structure is a separate tensor dimension, preserving permutation invariance over groups and over observations within groups.
Inputs shorter than the model's maximum dimensions are zero-padded, and boolean masks $\boldsymbol{\mu}$ propagate through every layer to exclude padded entries from attention, summaries, flow densities, and losses.
To spread the learning signal evenly across predictor positions, a shared random permutation is applied per dataset to the columns of $\mathbf{X}$, $\mathbf{Z}$, $\boldsymbol{\beta}$, $\boldsymbol{\alpha}_{1:m}$, and $\mathbf{S}$.

$\boldsymbol\Sigma_l$ and $\boldsymbol\Sigma_g$ are Set Transformer encoders \citep{Lee.2019} with identical layout: a linear projection to $d_\mathrm{model}=128$ (GELU), three attention blocks (4 heads, feed-forward width $512$), and a learned pool token whose final state is projected to the summary, $h_l = h_g = 32$.
The global encoder replaces its first block by an ISAB block with $32$ inducing points, which runs on the group summaries before the pool token is inserted; the local encoder uses three MAB blocks.
Permutation invariance is enforced by omitting positional encodings and causal masking.
Both posterior networks receive analytical features (\myappref{app:suf})
in addition to the learned summary and the prior embedding.

\subsection{Normalizing Flows} \label{app:nf}

Both $\boldsymbol\Pi_g$ and $\boldsymbol\Pi_l$ are conditional coupling flows \citep{Dinh.2017, Durkan.2019, Papamakarios.2021}.
Let $\boldsymbol{v}$ denote the vector being modeled and $\mathbf{c}$ the full conditioning vector: for the global flow, $\boldsymbol{v}$ is the unconstrained representation of $\boldsymbol{\vartheta}$ (see below) and $\mathbf{c}$ is the augmented global summary $\mathbf{s}$. 
For the local flow, $\boldsymbol{v} = \boldsymbol{\alpha}_i$ and $\mathbf{c} = (\boldsymbol{\vartheta}, \mathbf{s}_i)$ plus the per-group features of \myappref{app:suf}.
A normalizing flow is a learned invertible map $T(\cdot\,;\mathbf{c}) = T_B \circ \cdots \circ T_1$ from parameter space to a base space. Intermediate states are $\boldsymbol{v}^{(0)} = \boldsymbol{v}$ and $\boldsymbol{v}^{(b)} = T_b(\boldsymbol{v}^{(b-1)};\mathbf{c})$. The base space is a diagonal Normal density $p_0(\cdot \mid \mathbf{c})$ whose per-coordinate location and scale are predicted from $\mathbf{c}$:
\begin{equation*}
\log p_{\boldsymbol\Pi}(\boldsymbol{v} \mid \mathbf{c}) = \log p_0\!\left(\boldsymbol{v}^{(B)} \mid \mathbf{c}\right) + \sum_{b=1}^{B} \log \left|\det J_{T_b}\!\left(\boldsymbol{v}^{(b-1)}\right)\right|.
\end{equation*}
Density evaluation is one forward pass; sampling draws from $p_0$ and applies the blocks in reverse.
The global flow uses $B=4$ blocks, the local flow $B=3$.

\paragraph{Block structure.}
Each block is \textit{ActNorm} $\to$ \textit{Permute} $\to$ \textit{DualCoupling}.
\textit{ActNorm} \citep{Kingma.2018} standardizes each coordinate with a learned shift and scale, $v_j \mapsto (v_j - t_j)/a_j$, initialized from the first training batch; its log-determinant is $-\sum_j \log a_j$.
\textit{Permute} applies a fixed random permutation (log-determinant zero), so that every coordinate eventually conditions on every other.
\textit{DualCoupling} splits $\boldsymbol{v} = (\boldsymbol{v}_1, \boldsymbol{v}_2)$ at the midpoint and applies two coupling steps, first transforming $\boldsymbol{v}_2$ given $\boldsymbol{v}_1$ and then $\boldsymbol{v}_1$ given the updated $\boldsymbol{v}_2$, so that every coordinate is modified in every block.
A single coupling step maps
\begin{equation*}
\boldsymbol{v}_2 \mapsto f\!\left(\boldsymbol{v}_2;\; g_\psi(\boldsymbol{v}_1, \mathbf{c})\right), \qquad \boldsymbol{v}_1 \mapsto \boldsymbol{v}_1,
\end{equation*}
where $f$ acts elementwise and the conditioner $g_\psi$ is a depth-3 residual network (ReLU, width $256$, GLU gating by the context, no biases, zero-initialized output layer, so each step starts as the identity).
The Jacobian is block-triangular, $\log|\det J| = \sum_j \log|\partial f_j / \partial v_{2,j}|$.

\paragraph{Rational-quadratic splines.}
Each $f_j$ is a monotone rational-quadratic spline \citep{Durkan.2019} with $K=8$ bins on a domain $[l, r]$, whose knots and derivatives are predicted by $g_\psi$.
Inside bin $k$, with input knots $c_k$ and widths $w_k = c_{k+1} - c_k$, output knots $c'_k$ and heights $h_k = c'_{k+1} - c'_k$, secant slopes $s_k = h_k / w_k$, and knot derivatives $\delta_k$ (interior ones made positive by a softplus),
\begin{equation}
  f_j(v) = c'_k + h_k \cdot \frac{s_k\,\xi^2 + \delta_k\,\xi(1-\xi)}{s_k + (\delta_{k+1} + \delta_k - 2s_k)\,\xi(1-\xi)}, \qquad \xi = \frac{v-c_k}{w_k} \in [0,1),
  \label{eq:rqs}
\end{equation}
which is invertible by solving a quadratic in $\xi$.
Outside $[l, r]$ the map is affine with slope and offset predicted per coordinate; the output domain $[c'_1, c'_{K+1}]$ is the affine image of $[l, r]$ and the boundary derivatives equal the tail slope, so the map is $C^1$ (\emph{adaptive tails}).
$g_\psi$ also predicts $l$ and $r-l$ per coordinate, so the bins concentrate where each posterior has its mass (\emph{adaptive domain}).

\paragraph{Masking.}
Padded coordinates (mask $\boldsymbol{\mu}$) are skipped by \textit{ActNorm}, suppressed in the inputs to $g_\psi$, passed through the couplings as identity with zero log-determinant, and excluded from the base log-density, so one trained flow handles every $(d, q)$ within its size class.

\paragraph{Unconstrained parameterization.} \label{app:unc}
The global flow models $\boldsymbol{\vartheta}$ through a bijection $\boldsymbol{\vartheta} \mapsto \boldsymbol{v} \in \mathbb{R}^{\dim(\boldsymbol\vartheta)}$.
Scales ($\boldsymbol{\sigma}_\alpha$, $\sigma_\varepsilon$) are mapped by the inverse softplus, $\sigma \mapsto \log(e^{\sigma} - 1)$, which is close to the identity away from zero and stretches only the region near it.
The correlation matrix $\mathbf{R}$ ($q \ge 2$) is represented by its $q(q-1)/2$ canonical partial correlations \citep{Lewandowski.2009}: writing $\mathbf{R} = \mathbf{L}\mathbf{L}^\top$ with lower-triangular $\mathbf{L}$, each strictly lower entry is normalized by the norm its row has left, $\tilde\rho_{kj} = L_{kj} / \sqrt{1 - \sum_{j' < j} L_{kj'}^2} \in (-1, 1)$ for $1 \le j < k \le q$, and mapped to $\gamma_{kj} = \operatorname{artanh}(\tilde\rho_{kj}) \in \mathbb{R}$.
The inverse rebuilds $\mathbf{L}$ row by row from $\tilde\rho_{kj} = \tanh(\gamma_{kj})$ with unit-norm rows, so every $\boldsymbol\gamma \in \mathbb{R}^{q(q-1)/2}$ yields a valid correlation matrix and $\boldsymbol\gamma = \mathbf{0}$ gives the identity.
Samples are mapped back to the constrained scale before any evaluation, and so is the density: with $\mathbf{r}$ the lower triangle of $\mathbf{R}$, the reported global density is
\begin{gather*}
\log q(\boldsymbol\vartheta) = \log p_{\boldsymbol\Pi_g}(\boldsymbol{v} \mid \mathbf{s}) - \sum_{\sigma} \log \mathrm{softplus}'(v_\sigma) - \log\left|\det \frac{\partial \mathbf{r}}{\partial \boldsymbol\gamma}\right|, \\
\log\left|\det \frac{\partial \mathbf{r}}{\partial \boldsymbol\gamma}\right| = \tfrac{1}{2}\sum_{k > j} (q - j + 1)\log\bigl(1 - \tanh^2 \gamma_{kj}\bigr),
\end{gather*}
where the sum over $\sigma$ runs over the scale coordinates.
This $q(\boldsymbol\vartheta)$ lives in the same coordinates as prior and likelihood and is the density that enters the IMH acceptance ratio (\myappref{app:imh}) and the evidence weights (\myappref{app:ev}).

\subsection{Analytical Context Features} \label{app:suf}
Before any data enter the networks, we compute classical point estimates of every model parameter: $\hat{\boldsymbol\beta}$, $\hat{\mathbf S}$, $\hat\phi$, and per-group BLUPs $\hat{\boldsymbol\alpha}_i$ \citep{Laird.1982, Searle.1992}. These are fed to the normalizing flows as conditioning inputs.
All estimates are closed-form expressions or derived from short fixed-length iterations (vectorized over datasets and groups).
They depend only on the data and its stored prior, so they are computed once per dataset and cached.

\paragraph{Gaussian outcomes.}
Variance components have (near) closed-form estimators \citep{Henderson.1953, Searle.1992}, and we use a four-stage pipeline.
\begin{enumerate}
    \item \emph{Residual variance.} Outcomes and predictors are projected onto the within-group complement of $\mathbf{Z}_i$ (group-demeaning when $q=1$), such that the random effects vanish; pooled OLS on the projected data yields $\hat \phi = \hat\sigma_\varepsilon$.
    \item \emph{Random-effect covariance.} Per-group OLS estimates of the random effects are combined into a method-of-moments estimate $\hat{\mathbf{S}}$ that subtracts the share of their spread \citep{Henderson.1953, Cochran.1954}.
    \item \emph{Fixed effects and BLUPs.} Given $(\hat\sigma_\varepsilon, \hat{\mathbf{S}})$, generalized least squares yields $\hat{\boldsymbol{\beta}}$, and each group's random effect is the shrinkage compromise between its own data and the population \citep{Robinson.1991},
    \begin{equation*}
    \hat{\boldsymbol{\alpha}}_i = \bigl(\hat\sigma_\varepsilon^2\hat{\mathbf{S}}^{-1} + \mathbf{Z}_i^\top\mathbf{Z}_i\bigr)^{-1}\mathbf{Z}_i^\top\bigl(\mathbf{y}_i - \mathbf{X}_i\hat{\boldsymbol{\beta}}\bigr).
    \end{equation*}
    This pulls sparsely observed groups toward zero and densely observed ones toward their own least-squares fit.
    \item \emph{EM polish.} A few EM iterations \citep{Dempster.1977} refine $\hat{\mathbf{S}}$ and $\hat\sigma_\varepsilon$, since the moment estimate truncates at zero exactly where between-group variation is weakest.
\end{enumerate}

\paragraph{Bernoulli and Poisson outcomes.}
A pooled GLM fit \citep{McCullagh.1989} that ignores the group structure gives initial fixed effects and a Pearson dispersion statistic $\hat\phi$. The latter is kept as a global feature that signals excess variability that the random effects must absorb.
Penalized quasi-likelihood \citep[PQL;][]{Breslow.1993} then rewrites the model locally as a weighted LMM, and the Gaussian pipeline supplies a first estimate $\hat{\mathbf{S}}$.
An alternating Newton--Laplace--GLS loop refines all estimates jointly: damped Newton steps move each group to its posterior mode $\hat{\boldsymbol{\alpha}}_i$, a Laplace approximation step re-estimates the covariance from the modes and their curvature, $\hat{\mathbf{S}} = \frac1m \sum_i (\hat{\boldsymbol{\alpha}}_i\hat{\boldsymbol{\alpha}}_i^\top + \mathbf{H}_i^{-1})$, and a GLS step updates $\hat{\boldsymbol{\beta}}$.
Each pass is one EM step on the Laplace-approximated marginal likelihood, targeting the same fixed point as \texttt{lme4}'s outer optimizer \citep{Bates.2015}. The loop is stopped by a batch-wide convergence test (and has at most 6 iterations).

\paragraph{Prior calibration (empirical-Bayes MAP).}
The estimates so far ignore the prior, but the flows approximate a Bayesian posterior: with few groups or weakly identified parameters the posterior sits between likelihood and prior, and an unpenalized anchor would point the flows to the wrong place exactly where prior guidance matters most.
A final stage therefore moves the estimates to the mode of the Laplace-approximated marginal posterior under the dataset's stored prior.
For Gaussian outcomes this is a cheap low-dimensional optimization over the variance components (which is why this family's features are the fastest to compute, \myappref{app:rt}).
For Bernoulli and Poisson, $(\hat{\boldsymbol{\beta}}, \log\hat{\boldsymbol{\sigma}}_\alpha)$ are optimized jointly for 24 Adam steps, re-solving the group modes with a few damped Newton steps inside every iteration; a continuation schedule releases a cap on $\hat{\boldsymbol{\sigma}}_\alpha$ gradually so the scales cannot inflate before $\hat{\boldsymbol{\beta}}$ settles. The result is accepted only where it improves the Laplace objective.

\paragraph{Context features.}
The global flow receives $\hat{\boldsymbol{\beta}}$, $\hat{\boldsymbol{\sigma}}_\alpha$, $\hat\phi$, and the lower triangle of the estimated correlation matrix $\hat{\mathbf R}$.
The local flow receives BLUPs $\hat{\boldsymbol{\alpha}}_i$, the BLUP's standard deviation, and the shrinkage factor $\hat\lambda_{ik} = 1 - \mathrm{Var}(\hat\alpha_{ik})/\hat\sigma_{\alpha,k}^2$ for random effect $k$. The latter measures how firmly the group's own data pin down the random effect ($\hat\lambda \to 1$ group-dominated, $\hat\lambda \to 0$ population-dominated).
For Gaussian outcomes, BLUPs are computed in closed form, for discrete outcomes they are the Laplace modes from the pipeline above.
All analytical features are conditioning inputs the flows are free to overrule, not the final inference.

\subsection{Optimization and Hyperparameters}

All models are trained with Schedule-Free AdamW \citep{Defazio.2024} ($\beta_1 = 0.9$, $\beta_2 = 0.999$, weight decay $0$, learning rate $3\times10^{-4}$, gradient clipping at norm $1.0$, batch size $32$).
The global and local losses are summed with equal weight; local losses are averaged over active groups within each dataset.
Each model is trained for $8{,}000$--$16{,}000$ epochs of $4{,}096$ datasets precomputed by the simulator. This totals approximately $33$--$66\mathrm{M}$ unique training datasets and $1$--$2\mathrm{M}$ gradient steps, on a single NVIDIA H100 80\,GB GPU (4 CPU cores, 64\,GB RAM) in 3--5 days.

\paragraph{Efficient mini-batch construction.}
Because datasets vary in group count $m$ and maximum per-group observation count $n_{i,\max}$, naive random batching wastes memory on zero-padding.
We use a \emph{sortish} batch sampler: within consecutive windows of $50$ mini-batches, indices are sorted by $m$ (primary key) and $n_{i,\max}$ (secondary key). The resulting batches are then globally reshuffled.
Sorting by $m$ first matters because $m$ spans a wider relative range than $n_{i,\max}$ and dominates cross-axis padding waste.

\clearpage
\section{Simulator and Datasets} \label{app:dat}
This appendix details the simulation procedure sketched in \mysecref{sec:dat}: the literature survey that grounds the simulation ranges (\myappref{app:sur}), the size classes and their sampling distributions (\myappref{app:sizes}), the prior hyperparameters (\myappref{app:prior}), the two predictor sources (\myappref{app:scm}, \myappref{app:emu}), the preprocessing of real datasets (\myappref{app:pre}) and the analytical standardization that keeps parameters and hyperparameters coherent under $z$-scoring (\myappref{app:sta}).

\subsection{A Survey of Published GLMM Analyses} \label{app:sur}

To quantify how much of published GLMM practice \mb's model class covers, we surveyed $1{,}133$ randomly sampled papers from the cognitive and behavioral sciences\footnote{\textit{J.~Exp.~Psychol.} (General~\& LMC; 121), \textit{Cognition} (67), \textit{Cogn.~Sci.} (54), \textit{Psychol.~Sci.} (47), \textit{Behav.~Res.~Methods} (36), \textit{Psychon.~Bull.~\& Rev.} (31), \textit{Cogn.~Psychol.} (30), \textit{Judgm.~Decis.~Mak.} (21), \textit{J.~Mem.~Lang.} (20), \textit{Nat.~Hum.~Behav.} (19), and 27 other venues (54).}.
An LLM pipeline (Claude Haiku 4.5 for screening, Claude Sonnet 5 for extraction) identified $576$ papers ($51\%$) reporting at least one GLMM analysis and extracted the model equation for $500$ of them.
Each extraction stores the verbatim equation together with its text or code reference for auditability, and a manual check of $20$ randomly drawn extractions against their source papers found no extraction errors; the full dataset is released alongside our code (\texttt{model\_equations.csv}).

\begin{figure}[hbp]
    \centering
    \includegraphics[width=1.0\textwidth]{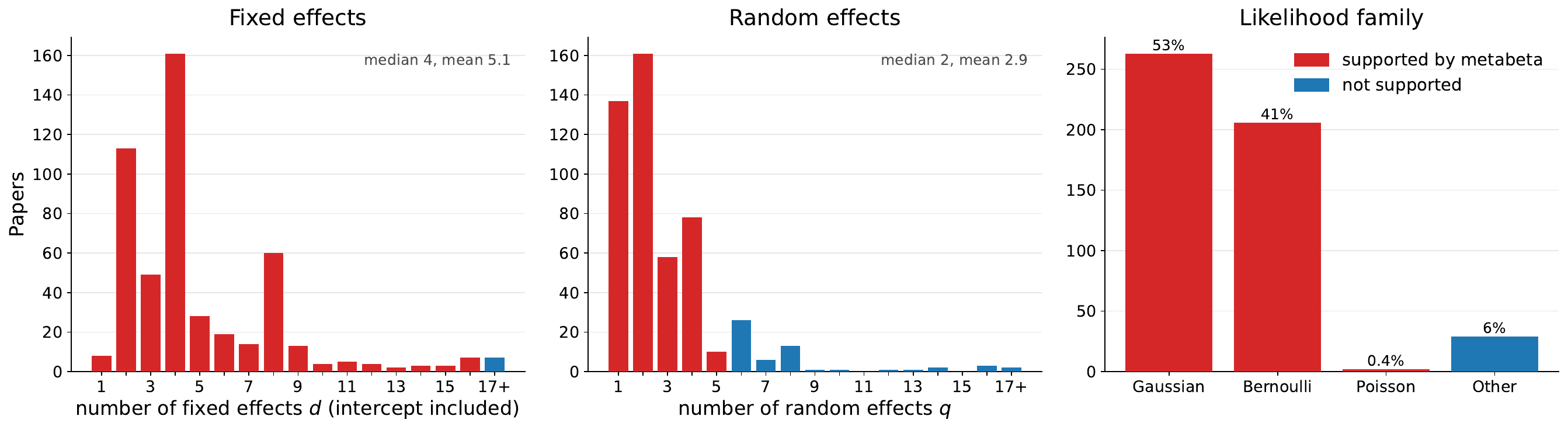}
    \caption{
       Survey of published GLMM analyses: distributions of the fixed-effect count $d$ (intercept included, as throughout the paper), the random-effect count $q$, and the likelihood family over the $500$ extracted models.
       Red marks analyses within \mb's supported ranges ($d \leq 16$, $q \leq 5$, Gaussian/Bernoulli/Poisson), blue those outside.
    }
    \label{fig:survey}
\end{figure}

The extracted models contain on average $d = 5.14$ fixed effects including the intercept ($95\%$ CI $[4.74, 5.55]$; median $4$) and $q = 2.91$ random effects ($[2.69, 3.13]$; median $2$); Gaussian ($53\%$), Bernoulli ($41\%$), and Poisson ($0.4\%$) likelihoods together account for $94\%$ of all models (\myfigref{fig:survey}). The remaining $6\%$ are predominantly ordinal (cumulative-link) models for rating scales, alongside a few heavy-tailed response-time (ex-Gaussian, log-normal) and bounded-proportion (beta) likelihoods.
Taken together, $83\%$ of the surveyed analyses fall within \mb's supported model class ($d \leq 16$, $q \leq 5$, Gaussian/Bernoulli/Poisson).

\rev{Of the surveyed papers containing any GLMM, $14\%$ report a Bayesian analysis (e.g. with \texttt{brms}, \texttt{Stan}, \texttt{rstanarm}, \texttt{MCMCglmm}, \texttt{JAGS/BUGS} or \texttt{PyMC}).
According to \texttt{OpenAlex} \citep{Priem.2022}, software citations show a similar share: $12\%$ of the cited software for GLMMs was Bayesian in 2025; seven years before that, it used to be $9\%$. Citations of \texttt{brms} \citep{Burkner.2017} alone grew tenfold over the same period (from $187$ to $1{,}941$ per year).}

\rev{While adoption of the Bayesian workflow is on the rise, many authors report that the cost of MCMC constrained their analyses:}
\begin{itemize}
   \item \rev{A simulation study of random slopes was capped to only $40$ replications per condition, because the simulation took several months \citep{Oberauer.2022}.}
   \item \rev{A Bayesian power analysis was capped to only $100$ simulated datasets per sample size, because computation would otherwise not have been tractable \citep{Vasishth.2023}. Its authors report that it ran $120$ times slower than its frequentist counterpart.}
   \item \rev{Two analyses of large observational datasets had to be constrained to subsamples for computational reasons: one kept at most $1{,}000$ bird-watching checklists per county \citep{Schrimpf.2021}, the other kept about $10\%$ of $3.4$ million records \citep{Arslan.2017}.}
   \item \rev{A pre-registered study checked only a single alternative prior in its prior sensitivity analysis to keep compute time manageable \citep{Wang.2021}. For the same reason, it aggregated its data over participants to drop random effects.}
\end{itemize}
\rev{Overall, GLMMs are a widely used tool for scientific inference on grouped data. While adoption of Bayesian methods in applied research is on the rise, it is (literally) slowed down by computationally expensive MCMC.}

\subsection{Size Classes} \label{app:sizes}
Every dataset first draws its size class (\mytabref{tab:sizes}) and then its structural dimensions: the fixed-effect count $d$, the random-effect count $q$, the group count $m$, and the group sizes $n_{1:m}$.
The classes differ only in the band for $d$ and the cap $q_{\max}$; all share one data budget of $m \le 200$ groups, $n_i \le 150$ observations per group, and $N_{\max} = 3{,}000$ observations in total.

\begin{table}[hbp]
\centering
\caption{%
  Size-class parameter ranges.
  $d$ is the fixed-effect dimension band (intercept included) and $q_{\max}$ the cap on the random-effect dimension, drawn log-uniformly from $[1, \min(q_{\max}, d)]$.
  All regimes share the same data budget: $m \in [5, 200]$ groups, $n_i \in [5, 150]$ observations per group, and a total-observation cap of $N_{\max} = 3{,}000$.
}
\label{tab:sizes}
\apptable
\begin{tabular}{lcc}
\toprule
$\mathrm{regime}$ & $d$ & $q_{\max}$ \\
\midrule
\textit{small}  & $1$--$4$   & $2$ \\
\textit{medium} & $5$--$8$   & $3$ \\
\textit{large}  & $9$--$12$  & $4$ \\
\textit{huge}   & $13$--$16$ & $5$ \\
\bottomrule
\end{tabular}
\end{table}

\paragraph{Identifiability.}
Estimating $d$ fixed effects requires $m > d$ groups, and estimating a $q \times q$ random-effect covariance requires $m > q(q+1)/2$.
The group count is therefore floored at
\begin{equation*}
m_\mathrm{low} = \max\!\bigl(m_{\min},\; \max(d,\; \tfrac{q(q+1)}{2}) + 4\bigr),
\end{equation*}
which leaves at least four between-group degrees of freedom above the tighter constraint.
Emulated designs (\myappref{app:emu}) can deliver fewer groups than requested, so a post-hoc clamp trims $q$ to the largest value identifiable from the \emph{realized} $m$, $q^* = \max\bigl(1,\, \lfloor(\sqrt{1+8(m-4)}-1)/2\rfloor\bigr)$.
Within groups, the minimum size is $\max(n_{\min},\, q+2)$, so every group keeps at least two residual degrees of freedom after fitting its own random effects.
The floor rises with $d$ and $q$: in the realized test sets, the median group count grows from $21$ in the \emph{small} class to $42$ in \emph{huge}.
The cap $N_{\max}$ only bounds the largest datasets; it binds for well under $1\%$ of oracle test datasets, whose median total is $406$--$512$ observations.

\paragraph{Sampling distributions.}
\begin{itemize}
  \item \emph{Fixed effects $d$}: uniform integer over the class band, giving equal coverage across complexity levels.
  \item \emph{Random effects $q$}: log-uniform over $[1, \min(q_{\max}, d)]$.
  This concentrates coverage on the small random-effect dimensions that dominate applied practice (\myappref{app:sur}: median $2$; $71\%$ of surveyed models have $q \leq 3$ and $89\%$ have $q \leq 5$), and the cap at $d$ keeps the random-effect design a sub-design of the fixed-effect one.
      \rev{The random-effect design $\mathbf{Z}$ is the first $q$ columns of $\mathbf{X}$. This is WLOG, because the released code orders the columns of $\mathbf{X}$ to match a supplied model specification (e.g. by a \texttt{brms}-styled formula).}
  \item \emph{Groups $m$}: a two-component mixture.
  With probability $0.8$, a mode-preserving Beta (defined in \myappref{app:prior}) on $[m_\mathrm{low}, m_{\max}]$ with concentration $\kappa = 8$ and mode $m_\mathrm{low} + \min\bigl(0.08\,(m_{\max}-m_\mathrm{low}),\, 20\bigr)$, floored to an integer; this places most mass near the identifiability floor, in line with the test collection ($61\%$ of its datasets have $m \leq 20$).
  With probability $0.2$, $m$ is uniform over $[\max(m_\mathrm{low}, 60),\, m_{\max}]$, which keeps many-group designs represented.
  \item \emph{Group sizes $n_i$ for synthetic designs}: a mean group size $\bar{n}$ is drawn on a log scale over $[n_{\min},\, \min(n_{i\max},\, N_{\max}/m)]$, at position $0.15 + 0.65\,\tilde{m}$ of the log-span plus Gaussian noise with scale $0.25$ of the span, where $\tilde{m} \in [0,1]$ is the log-scaled position of $m$ in its range.
  With probability $0.65$ every group receives $\bar{n}$ observations; otherwise the budget $m\bar{n}$ is split by $\mathrm{Dirichlet}(c\,\mathbf{1}_m)$ proportions with $c \sim \mathrm{LogUniform}(8, 40)$.
  \item \emph{Group sizes $n_i$ for emulated designs} (the source of all oracle test sets): only $m$ and the total budget are requested; the hierarchical bootstrap of \myappref{app:tes} allocates group sizes, clipped to what each source group actually contains.
  Group sizes are bounded by real data rather than by a parametric family, which reproduces the heavy right tail and imbalance of real hierarchical data.%
\end{itemize}

\subsection{Prior Hyperparameters} \label{app:prior}
Each dataset draws its prior families independently for the fixed-effect coefficients (Normal $0.80$, Student-$t$ $0.20$), the random-effect scales (half-Normal $0.60$, half-Student-$t$ $0.30$, Exponential $0.10$). Under the Gaussian likelihood, the residual scale is added (half-Normal $0.50$, half-Student-$t$ $0.40$, Exponential $0.10$). All Student-$t$ families have $5$ degrees of freedom.
The hyperparameters are then drawn as follows.

\paragraph{Locations.}
Locations are sampled from a spike-and-slab distribution:
each component of $\boldsymbol{\nu}_\beta \in \mathbb{R}^d$ is zero with probability $0.90$ and otherwise drawn from $\mathrm{Laplace}(0, s)$, with $s = 0.7$ for the Bernoulli likelihood and $s = 0.5$ otherwise.
The network thus sees mostly zero-centered priors with occasional strongly shifted ones.

\paragraph{Scales.}
Scale hyperparameters follow a \emph{mode-preserving Beta}: a Beta distribution rescaled to $[a, b]$ with mode $\omega$ and concentration $\kappa$.
Writing $t = (\omega - a)/(b - a)$ for the mode's relative position,
\begin{equation*}
\tau \sim a + (b - a)\cdot\mathrm{Beta}\bigl(1 + (\kappa-1)\,t,\; 1 + (\kappa-1)(1-t)\bigr),
\end{equation*}
so $\kappa = 1$ gives the uniform distribution on $[a, b]$ and larger $\kappa$ concentrates mass around $\omega$ without displacing it.
Ranges are tighter for Bernoulli and Poisson likelihoods to prevent extreme linear predictors:
\begin{itemize}
  \item $\boldsymbol{\tau}_\beta$: $[0.01, 4.0]$, mode $0.5$, $\kappa = 8$ (Gaussian); $[0.01, 3.0]$, mode $0.8$, $\kappa = 6$ (Bernoulli); $[0.01, 1.5]$, mode $0.5$, $\kappa = 6$ (Poisson).
  \item $\boldsymbol{\tau}_\sigma$: $[0.01, 5.0]$, mode $1.0$ (Gaussian); $[0.01, 2.0]$, mode $0.7$ (Bernoulli); $[0.01, 1.0]$, mode $0.3$ (Poisson); $\kappa = 5$ throughout.
  \item $\tau_\varepsilon$: $[0.05, 8.0]$, mode $2.0$, $\kappa = 5$ (Gaussian likelihood only).
\end{itemize}
Sampled scales $\sigma_\varepsilon$ and $\sigma_{\alpha_j}$ are floored at $10^{-3}$ to keep the likelihood well defined.
The floor is not mirrored in the \texttt{PyMC} priors of the baselines, since raising it would remove high-$R^2$ datasets from the benchmark and change the training distribution seen by \mb.

\paragraph{Correlations.}
Random effects are correlated in only part of the draws: never for $q = 1$, with probability $0.8$ for $q = 2$, and with probability $0.5$ for $q \geq 3$.
When they are, the correlation matrix is drawn from $\mathrm{LKJ}(\eta)$ with $\eta \sim \mathcal{U}(1, 2)$; otherwise it is the identity.

\paragraph{Outcome calibration.}
Prior draws that would produce degenerate outcomes are corrected before $\mathbf{y}$ is sampled.
Under the \emph{Gaussian} likelihood, an explanatory-power ceiling $R^2_{\max} \sim \mathcal{N}(0.68 + 0.015\,(d-1) + 0.010\,(q-1),\, 0.10^2)$, clipped to $[0.45, 0.95]$, is drawn per dataset; whenever $\mathbb{V}(\boldsymbol\eta) / (\mathbb{V}(\boldsymbol\eta) + \sigma_\varepsilon^2)$ exceeds it, $\sigma_\varepsilon$ is inflated to meet the ceiling.
This bounds the signal-to-noise ratio of the Gaussian benchmarks, and it loosens with design size because more predictors legitimately explain more variance.
Under the \emph{Bernoulli} and \emph{Poisson} likelihoods, the parameters are instead redrawn (up to $20$ times) while too large a share of the linear predictor is extreme: $|\eta| > 10$ for more than $5\%$ of observations, or $\eta > 10$ for more than $1\%$, respectively.
Any remaining excess is removed by scaling all regression parameters so that $\mathrm{sd}(\boldsymbol\eta)$ meets a per-dataset cap drawn from $\mathcal{U}(2, 4)$ (Bernoulli) or $\mathcal{U}(1, 2)$ (Poisson); Poisson predictors are additionally clipped to $\pm 5$ after standardization.

\subsection{SCM Generator} \label{app:scm}
The SCM Generator samples predictor matrices from structural causal models whose graph, weights, activations and marginal transforms are re-drawn on every call, so each dataset carries a distinct dependency structure.
Per dataset it picks one of two variants, a layered MLP-SCM (probability $0.6$) or a sparse DAG-SCM ($0.4$), and one of three equiprobable presets that jointly set the activation pool and the root-cause distribution, trading smoothness against variability.

\paragraph{Root causes.}
In the MLP-SCM, each exogenous column is assigned independently to one of nine marginal families (Gaussian, uniform, multinomial, Zipf, gamma, log-normal, beta, Student-$t$, Gaussian mixture), so right-skewed, heavy-tailed, bounded and multimodal marginals coexist within a dataset; the smoothest preset uses uniform causes only.
The DAG-SCM has no separate cause stage: its root nodes are standard normal, and all structure arises during propagation.

\paragraph{Propagation.}
The MLP-SCM passes the root causes through $\ell$ blocks of \texttt{Linear} $\to$ \texttt{NoiseLayer} $\to$ \texttt{Activation}.
Block-diagonal weight matrices cluster features into correlated groups, and a noiseless pilot pass scales the additive noise at each depth to a fixed fraction of the signal's interquartile range.
Three optional stages add structure the weights alone do not produce: shared latent noise over random feature subsets, a low-rank factor injection, and per-feature monotone maps that diversify marginal shapes while leaving the rank structure, and hence the dependency graph, intact.
The DAG-SCM instead samples a scale-free (Barab\'asi--Albert) or sparse random (Erd\H{o}s--R\'enyi) graph over the observed features plus a set of latent nodes, and evaluates $x_i = \phi(\mathbf{W}\mathbf{x}_{\mathrm{pa}(i)}) + \varepsilon_i$ in topological order, reading observed features off leaf nodes wherever the graph supplies enough of them.
It shares the noise calibration but none of the optional stages, and contributes sparse, explicitly causal dependencies alongside the dense ones of the layered variant.

\paragraph{Activation pool.}
Each layer draws from 25 fixed nonlinearities, both standard (ReLU, Tanh, SiLU) and irregular (Abs, Sine, Mod, Ceil, Sign), and from random Fourier feature approximations of Gaussian process draws under squared-exponential, Matérn and fractional kernels.
\texttt{RandomChoice} layers apply different activations to different feature subsets within one layer.
The preset decides how much of the pool is available, from a small, smooth, GP-dominated selection to the full set.

\paragraph{Post-hoc transforms.}
Finally, a random subset of continuous outputs is replaced by discrete or bounded variants (\mytabref{tab:posthoc}).
Each transform first mixes several SCM features with Dirichlet weights and then applies its mapping, either per column or as a block that emits several dummy-coded or ordinal columns at once.
The ordinal block thresholds one shared latent direction, so its columns are strongly rank-correlated, as multiple ordinal predictors typically are in real data.
The result is mixed-type feature matrices with nonlinear inter-predictor dependencies that parametric families cannot reproduce.

\begin{table}[hbp]
\centering
\apptable
\caption{Post-hoc feature transforms of the SCM Generator and the predictor type each one emits.}
\label{tab:posthoc}
\begin{tabular}{ll}
\toprule
Transform & Output \\
\midrule
\texttt{Threshold} & Binary (thresholded at zero) \\
\texttt{MultiThreshold} & Ordinal integer (Gaussian-sampled thresholds) \\
\texttt{QuantileBins} & Discrete bins (data-driven quantile cut points) \\
\texttt{Clamp} & Continuous with random quantile floor/ceiling \\
\texttt{CensoredFloor} & Left-censored continuous (detection-limit effect) \\
\texttt{Categorical} / \texttt{CategoricalBlock} & Dummy-coded factor(s), mutually exclusive within group \\
\texttt{OrdinalBlock} & Correlated ordinal integers from a shared latent direction \\
\texttt{Poisson} / \texttt{NegativeBinomial} & Integer counts \\
\bottomrule
\end{tabular}
\end{table}

\subsection{Emulator} \label{app:emu}
The Emulator draws predictors from a bank of $596$ real datasets from PMLB \citep{Romano.2022}, SRM \citep{Lichtenberg.2017} and UCI \citep{Dua.2019}, each preprocessed by the pipeline of \myappref{app:pre} and stored as a compressed array; the test collection of \myappref{app:tes} is excluded from the bank.
Existing hierarchical grouping is preserved where present, otherwise groups are assigned at random.
For a requested predictor count $d$, a source dataset with $d_{\mathrm{src}}$ predictors is selected with probability proportional to $\min(d_{\mathrm{src}},\, d) / d$: sources with $d_{\mathrm{src}} \ge d$ receive full weight and fill all $d-1$ predictor slots with real columns, while smaller sources receive proportionally less weight and have the shortfall padded with SCM-generated columns.
This soft weighting avoids a hard eligibility cutoff, which would shrink the source pool sharply at values of $d$ that few datasets reach and thereby make source diversity uneven across $d$.

\subsection{Preprocessing Pipeline}
\label{app:pre}

\paragraph{Design philosophy.}

The preprocessing pipeline is a \emph{canonicalization map}, not a statistical cleaning step: it projects raw inputs onto a common representation that is numerically stable for the Set Transformer, consistent between simulation and deployment, and interpretable at the level of regression coefficients.
All transforms are monotone (log1p, clipping) or scale-equivariant (z-standardization), so the qualitative structure of the likelihood --- in particular the monotonicity of the predictor--outcome mapping --- is preserved; mild biases (e.g.\ a small slope bias from winsorization) are accepted in exchange for reduced representational heterogeneity.
The pipeline is \emph{stateful}: all statistics used to transform data (column medians, winsorization bounds, encoder vocabularies) are fitted once and replayed identically on new data via \texttt{fit}/\texttt{transform} methods analogous to \texttt{sklearn} estimators, eliminating covariate shift between training and deployment.

\paragraph{Pipeline steps.}

\mytabref{tab:preprocessing} lists every step in order with its motivation; the first six define the structure of the dataset and contain the only row or column removals, the remaining seven are pure transformations.
Three steps carry decisions that the table cannot convey in a line.

\begin{table}[hbp]
\centering
\caption{Preprocessing steps in order of application, with the decision taken and its primary motivation. Steps 1--6 fix the structure of a dataset and contain the only row or column removals; steps 7--13 are pure transformations.}
\label{tab:preprocessing}
\apptablewide
\begin{tabular}{rlll}
\toprule
& Step & Decision & Primary motivation \\
\midrule
1 & Sentinel replacement & $-999 \to \texttt{NaN}$ & avoid contaminating distribution estimates \\
2 & Heavy-column removal & drop if $> 25\%$ missing & imputation unreliable \citep{Sterne.2009} \\
3 & Outcome coercion & detect binary / count / continuous & selects the likelihood family \\
4 & Grouping detection & explicit, heuristic, or high-cardinality & high-cardinality fixed effects are misspecified \\
5 & Rare-category rows & drop pools with prevalence $< 5\%$ & cannot be assigned to a reference group \\
6 & Imputation & group-aware median/mode & train/deploy consistency \\
\midrule
7 & Winsorization & clip to $[\mu \pm 4\sigma]$ & input regularization for the summarizer \\
8 & Near-constant removal & drop if $> 95\%$ identical & collinear with intercept \\
9 & Correlation filter & drop one of each $|r| > 0.95$ pair & well-conditioned posterior geometry \\
10 & Categorical lumping & pool rare levels as \texttt{other} & avoid wide sparse dummies \\
11 & Numeric transforms & binary: pass; skewed counts: log1p; z-score & keep interpretation, remove structural skew \\
12 & Categorical encoding & modal reference level & matches \texttt{contr.treatment} \\
13 & Outcome standardization & continuous: z-score (\myappref{app:sta}); else raw & common scale for loss and flow \\
\bottomrule
\end{tabular}
\end{table}

\paragraph{Outcome type detection.}
The outcome is coerced to a numeric array (two-class strings are mapped to $\{0,1\}$; rows with non-finite \texttt{y} are removed) and its type determines the likelihood family:
binary outcomes stay integer indicators, counts (non-negative integers with zeros) stay raw integers for the Poisson likelihood, continuous outcomes are z-standardized in the final step, and multiclass strings are treated as pseudo-continuous under the Gaussian likelihood.
Datasets without a designated target are kept as design-matrix sources for the Emulator only.

\paragraph{Grouping variable detection.}
The grouping variable is (i) the explicitly specified column if given; otherwise (ii) the top-scoring integer/categorical column under a heuristic rewarding intermediate cardinality (5--200 levels), average group size near 20, low imbalance, and clustering-typical names (\texttt{subject}, \texttt{school}, \texttt{site}, \ldots); otherwise (iii) a categorical predictor with more than \texttt{max\_categories} levels and ${\geq}2$ observations per level, since high-cardinality categoricals are misspecified as fixed effects (near-rank-deficient design) and incompatible with the network's input contract.
The dataset is sorted by the resulting integer group codes; crossed or nested structures are approximated by the single dominant factor.
Categorical predictors with more than \texttt{max\_categories} levels keep their $K$ most frequent levels; the rest are pooled into \texttt{other} if their combined prevalence is at least 5\% and are otherwise removed (matching \texttt{forcats::fct\_lump\_n} in R).

\paragraph{Numeric transforms.}
Numeric predictors are winsorized to $[\mu \pm 4\sigma]$ (training-data statistics): extreme values push the summarizer outside its training distribution, and the conservative threshold limits the impact on skewed predictors.
Binary predictors pass through unchanged, since standardization would destroy the reference-level interpretation.
Count-like predictors with $|\mathrm{skewness}| > 1$ are log($x{+}1$)-transformed then z-standardized (their structural $\mathbb{N}_0$ support produces systematic right-skew); low-skew counts and continuous predictors are z-standardized only, because their skew has no consistent direction and direction-specific transforms would heterogenize the representation.
Categoricals are encoded as treatment contrasts with the training-modal category as reference level (matching R's \texttt{contr.treatment}); unseen categories at deployment map to the zero vector, i.e.\ the reference effect.

\clearpage
\subsection{Standardization} \label{app:sta}\label{app:glmm}
All continuous predictors are $z$-standardized over groups and observations, as is standard for GLMMs, and for training stability so are continuous outcomes $y$. To keep the dependence structure intact, parameters and hyperparameters are standardized analytically during training and un-standardized after sampling:
\begin{gather*}
\beta_k^* = \beta_k \frac{\sigma_{x_k}}{\sigma_y}, \qquad
\alpha_{ik}^* = \alpha_{ik} \frac{\sigma_{z_k}}{\sigma_y} \sim \mathcal{N}\left(0, \sigma_k^{*2}\right), \\
\sigma_k^{*2} = \sigma_k^2 \frac{\sigma_{z_k}^2}{\sigma_y^2},\quad \sigma_\varepsilon^{*2} = \frac{\sigma_{\varepsilon}^2}{\sigma_y^2},
\end{gather*}
where $\sigma_{x_k}$ and $\sigma_y$ are the standard deviations of the $k$th predictor and the outcome, and $\beta^*_k$ is the standardized $k$th slope. The intercepts require special care:
\begin{gather*}
\beta_0^* = \frac{\beta_0 + \sum_{k=1}^d \mu_{x_k} \beta_k - \mu_y}{\sigma_y}, \\
\alpha_{i0}^* = \frac{\alpha_{i0} + \sum_{k=1}^q \mu_{z_k} \alpha_{ik}}{\sigma_y} = \frac{\sum_{k=0}^q \mu_{z_k} \alpha_{ik}}{\sigma_y} \sim \mathcal{N}\left(0, \sigma_{0}^{*2} \right),
\end{gather*}
where $\mu_{x_k}$ is the mean of the $k$th predictor. Due to the sum term in the latter,
\begin{equation*}
\sigma_{0}^{*2} = \frac{1}{\sigma_y^2}\left[\mathbb{V}\left(\alpha_{i0}\right) + \mathbb{V}\left(\sum_{k=1}^q \mu_{z_k} \alpha_{ik}\right) + 2 \cdot \mathrm{Cov}\left(\alpha_{i0}, \sum_{k=1}^q \mu_{z_k} \alpha_{ik}\right)\right],
\end{equation*}
which is equivalent to summing up the covariance matrix of the random vector $\boldsymbol \mu_z \odot \boldsymbol \alpha_i$ and dividing by $\sigma_y^2$.

\textit{Proof}:
\begingroup\allowdisplaybreaks %
\begin{align*}
    y^*_{ij} &= \frac{y_{ij} - \mu_y}{\sigma_y} \\
    &= \frac{1}{\sigma_y} \left(\beta_0 + \sum_{k=1}^d x_{ijk} \beta_k + \alpha_{i0} + \sum_{k=1}^q z_{ijk} \alpha_{ik} + \varepsilon_{ij} - \mu_y \right) \\
    &\stackrel{!}{=} \beta^*_0 + \sum_{k=1}^d x^*_{ijk} \beta^*_k + \alpha^*_{i0} + \sum_{k=1}^q z^*_{ijk} \alpha^*_{ik} + \varepsilon^*_{ij} \\
    &= \beta^*_0 + \sum_{k=1}^d \left(\frac{x_{ijk} - \mu_{x_k}}{\sigma_{x_k}}\right) \beta^*_k + \alpha^*_{i0} + \sum_{k=1}^q \left(\frac{z_{ijk} - \mu_{z_k}}{\sigma_{z_k}}\right) \alpha^*_{ik} + \varepsilon^*_{ij} \\
    &= \beta^*_0 + \sum_{k=1}^d \left(\frac{x_{ijk} - \mu_{x_k}}{\sigma_{x_k}}\right) \left(\beta_k \frac{\sigma_{x_k}}{\sigma_y}\right) + \alpha^*_{i0} + \sum_{k=1}^q \left(\frac{z_{ijk} - \mu_{z_k}}{\sigma_{z_k}}\right) \left(\alpha_{ik} \frac{\sigma_{z_k}}{\sigma_y}\right) + \frac{\varepsilon_{ij}}{\sigma_y} \\
    &= \beta^*_0 - \sum_{k=1}^d \frac{\mu_{x_k}\beta_k }{\sigma_y}  + \sum_{k=1}^d \frac{x_{ijk}\beta_k }{\sigma_y} + \alpha^*_{i0} - \sum_{k=1}^q \frac{\mu_{z_k}\alpha_{ik}}{\sigma_y} + \sum_{k=1}^q \frac{z_{ijk}\alpha_{ik}}{\sigma_y} + \frac{\varepsilon_{ij}}{\sigma_y} \\
    &= \frac{\beta_0 - \mu_y}{\sigma_y} + \sum_{k=1}^d \frac{x_{ijk}\beta_k }{\sigma_y} + \frac{\alpha_{i0}}{\sigma_y} + \sum_{k=1}^q \frac{z_{ijk}\alpha_{ik}}{\sigma_y} + \frac{\varepsilon_{ij}}{\sigma_y} \qquad \square
\end{align*}
\endgroup
The distributions of the standardized random effects and noise follow from scaling normal variables, and the random-intercept variance from the variance of a sum \citep{Wasserman.2010}.


\begin{thebibliography}{124}
\providecommand{\natexlab}[1]{#1}
\providecommand{\url}[1]{\texttt{#1}}
\expandafter\ifx\csname urlstyle\endcsname\relax
  \providecommand{\doi}[1]{doi: #1}\else
  \providecommand{\doi}{doi: \begingroup \urlstyle{rm}\Url}\fi

\bibitem[{Abril-Pla} et~al.(2023){Abril-Pla}, Andreani, Carroll, Dong,
  Fonnesbeck, Kochurov, Kumar, Lao, Luhmann, Martin, Osthege, Vieira, Wiecki,
  and Zinkov]{Abril-Pla.2023}
O.~{Abril-Pla}, V.~Andreani, C.~Carroll, L.~Dong, C.~J. Fonnesbeck,
  M.~Kochurov, R.~Kumar, J.~Lao, C.~C. Luhmann, O.~A. Martin, M.~Osthege,
  R.~Vieira, T.~Wiecki, and R.~Zinkov.
\newblock {{PyMC}}: A modern, and comprehensive probabilistic programming
  framework in {{Python}}.
\newblock \emph{PeerJ Computer Science}, 9:\penalty0 e1516, Sept. 2023.
\newblock ISSN 2376-5992.
\newblock \doi{10.7717/peerj-cs.1516}.

\bibitem[Andrieu and Roberts(2009)]{Andrieu.2009}
C.~Andrieu and G.~O. Roberts.
\newblock The {{Pseudo-Marginal Approach}} for {{Efficient Monte Carlo
  Computations}}.
\newblock \emph{The Annals of Statistics}, 37\penalty0 (2):\penalty0 697--725,
  2009.
\newblock \doi{10.1214/07-AOS574}.

\bibitem[Andrieu et~al.(2018)Andrieu, Lee, and Vihola]{Andrieu.2018}
C.~Andrieu, A.~Lee, and M.~Vihola.
\newblock Uniform {{Ergodicity}} of the {{Iterated Conditional SMC}} and
  {{Geometric Ergodicity}} of {{Particle Gibbs Samplers}}.
\newblock \emph{Bernoulli}, 24\penalty0 (2):\penalty0 842--872, 2018.
\newblock \doi{10.3150/15-BEJ785}.

\bibitem[Arruda et~al.(2024)Arruda, Sch{\"a}lte, Peiter, Teplytska, Jaehde, and
  Hasenauer]{Arruda.2024}
J.~Arruda, Y.~Sch{\"a}lte, C.~Peiter, O.~Teplytska, U.~Jaehde, and
  J.~Hasenauer.
\newblock An amortized approach to non-linear mixed-effects modeling based on
  neural posterior estimation.
\newblock In \emph{Proceedings of the 41st {{International Conference}} on
  {{Machine Learning}}}, pages 1865--1901. PMLR, July 2024.

\bibitem[Arslan et~al.(2017)Arslan, Willf{\"u}hr, Frans, Verweij, B{\"u}rkner,
  Myrskyl{\"a}, Voland, Almqvist, Zietsch, and Penke]{Arslan.2017}
R.~C. Arslan, K.~P. Willf{\"u}hr, E.~M. Frans, K.~J.~H. Verweij, P.-C.
  B{\"u}rkner, M.~Myrskyl{\"a}, E.~Voland, C.~Almqvist, B.~P. Zietsch, and
  L.~Penke.
\newblock Older fathers' children have lower evolutionary fitness across four
  centuries and in four populations.
\newblock \emph{Proceedings of the Royal Society B: Biological Sciences},
  284\penalty0 (1862):\penalty0 20171562, Sept. 2017.
\newblock ISSN 0962-8452.
\newblock \doi{10.1098/rspb.2017.1562}.

\bibitem[Barret and Dupourqu{\'e}(2026)]{Barret.2026}
D.~Barret and S.~Dupourqu{\'e}.
\newblock Simulation-based inference with neural posterior estimation applied
  to {{X}}-ray spectral fitting. {{III}}. {{Deriving}} exact posteriors with
  dimension reduction and importance sampling.
\newblock \emph{Astronomy \& Astrophysics}, 708:\penalty0 A280, Apr. 2026.
\newblock \doi{10.1051/0004-6361/202557639}.

\bibitem[Bates and Bolker(2020)]{Bates.2020}
D.~Bates and M.~M. a.~B. Bolker.
\newblock {{mlmRev}}: {{Examples}} from {{Multilevel Modelling Software
  Review}}, Apr. 2020.

\bibitem[Bates et~al.(2015)Bates, M{\"a}chler, Bolker, and Walker]{Bates.2015}
D.~Bates, M.~M{\"a}chler, B.~Bolker, and S.~Walker.
\newblock Fitting {{Linear Mixed-Effects Models Using}} {\textbf{lme4}}.
\newblock \emph{Journal of Statistical Software}, 67\penalty0 (1), 2015.
\newblock ISSN 1548-7660.
\newblock \doi{10.18637/jss.v067.i01}.

\bibitem[Battaglia et~al.(2024)Battaglia, Carmona, Haines, Anderson~Loake,
  Benskin, and Nicholls]{Battaglia.2024}
L.~Battaglia, C.~U. Carmona, R.~A. Haines, M.~Anderson~Loake, M.~Benskin, and
  G.~K. Nicholls.
\newblock Amortising over {{Hyperparameters}} in {{Generalised Bayesian
  Inference}}.
\newblock \emph{arXiv}, Dec. 2024.
\newblock \doi{10.48550/arXiv.2412.16419}.

\bibitem[Betancourt and Girolami(2013)]{Betancourt.2013}
M.~J. Betancourt and M.~Girolami.
\newblock Hamiltonian {{Monte Carlo}} for {{Hierarchical Models}}.
\newblock \emph{arXiv}, Dec. 2013.
\newblock \doi{10.48550/arXiv.1312.0906}.

\bibitem[Bradley(1978)]{Bradley.1978}
J.~V. Bradley.
\newblock Robustness?
\newblock \emph{British Journal of Mathematical and Statistical Psychology},
  31\penalty0 (2):\penalty0 144--152, Nov. 1978.
\newblock ISSN 0007-1102.
\newblock \doi{10.1111/j.2044-8317.1978.tb00581.x}.

\bibitem[Breslow and Clayton(1993)]{Breslow.1993}
N.~E. Breslow and D.~G. Clayton.
\newblock Approximate {{Inference}} in {{Generalized Linear Mixed Models}}.
\newblock \emph{Journal of the American Statistical Association}, 88\penalty0
  (421):\penalty0 9--25, 1993.
\newblock ISSN 0162-1459.
\newblock \doi{10.1080/01621459.1993.10594284}.

\bibitem[Brooks et~al.(2017)Brooks, Kristensen, {van Benthem}, Magnusson, Berg,
  Nielsen, Skaug, M{\"a}chler, and Bolker]{Brooks.2017}
M.~E. Brooks, K.~Kristensen, K.~J. {van Benthem}, A.~Magnusson, C.~W. Berg,
  A.~Nielsen, H.~J. Skaug, M.~M{\"a}chler, and B.~M. Bolker.
\newblock {{glmmTMB Balances Speed}} and {{Flexibility Among Packages}} for
  {{Zero-Inflated Generalized Linear Mixed Modeling}}.
\newblock \emph{The R Journal}, 9\penalty0 (2):\penalty0 378--400, 2017.
\newblock \doi{10.32614/RJ-2017-066}.

\bibitem[B{\"u}rkner(2017)]{Burkner.2017}
P.-C. B{\"u}rkner.
\newblock Brms: {{An R Package}} for {{Bayesian Multilevel Models Using Stan}}.
\newblock \emph{Journal of Statistical Software}, 80\penalty0 (1):\penalty0
  1--28, 2017.
\newblock ISSN 1548-7660.
\newblock \doi{10.18637/jss.v080.i01}.

\bibitem[B{\"u}rkner(2018)]{Burkner.2018}
P.-C. B{\"u}rkner.
\newblock Advanced {{Bayesian Multilevel Modeling}} with the {{R Package}}
  brms.
\newblock \emph{The R Journal}, 10\penalty0 (1):\penalty0 395, 2018.
\newblock ISSN 2073-4859.
\newblock \doi{10.32614/RJ-2018-017}.

\bibitem[Cannon et~al.(2022)Cannon, Ward, and Schmon]{Cannon.2022}
P.~Cannon, D.~Ward, and S.~M. Schmon.
\newblock Investigating the {{Impact}} of {{Model Misspecification}} in
  {{Neural Simulation}}-based {{Inference}}.
\newblock \emph{arXiv}, Sept. 2022.
\newblock \doi{10.48550/arXiv.2209.01845}.

\bibitem[Capretto et~al.(2022)Capretto, Piho, Kumar, Westfall, Yarkoni, and
  Martin]{Capretto.2022}
T.~Capretto, C.~Piho, R.~Kumar, J.~Westfall, T.~Yarkoni, and O.~A. Martin.
\newblock Bambi: {{A Simple Interface}} for {{Fitting Bayesian Linear Models}}
  in {{Python}}.
\newblock \emph{Journal of Statistical Software}, 103:\penalty0 1--29, Aug.
  2022.
\newblock ISSN 1548-7660.
\newblock \doi{10.18637/jss.v103.i15}.

\bibitem[Chang et~al.(2025)Chang, Loka, Huang, Remes, Kaski, and
  Acerbi]{Chang.2025}
P.~E. Chang, N.~R. B.~S. Loka, D.~Huang, U.~Remes, S.~Kaski, and L.~Acerbi.
\newblock Amortized {{Probabilistic Conditioning}} for {{Optimization}},
  {{Simulation}} and {{Inference}}.
\newblock In \emph{Proceedings of the 28th {{International Conference}} on
  {{Artificial Intelligence}} and {{Statistics}} ({{AISTATS}})}, volume 258 of
  \emph{Proceedings of {{Machine Learning Research}}}, 2025.

\bibitem[Cochran(1954)]{Cochran.1954}
W.~G. Cochran.
\newblock The combination of estimates from different experiments.
\newblock \emph{Biometrics. Journal of the International Biometric Society},
  10\penalty0 (1):\penalty0 101--129, 1954.
\newblock ISSN 0006-341X.
\newblock \doi{10.2307/3001666}.

\bibitem[Cohen(1988)]{Cohen.1988}
J.~Cohen.
\newblock \emph{Statistical {{Power Analysis}} for the {{Behavioral
  Sciences}}}.
\newblock Lawrence Erlbaum Associates, Hillsdale, NJ, 2 edition, 1988.
\newblock ISBN 978-0-8058-0283-2.

\bibitem[Cranmer et~al.(2020)Cranmer, Brehmer, and Louppe]{Cranmer.2020}
K.~Cranmer, J.~Brehmer, and G.~Louppe.
\newblock The frontier of simulation-based inference.
\newblock \emph{Proceedings of the National Academy of Sciences}, 117\penalty0
  (48):\penalty0 30055--30062, Dec. 2020.
\newblock ISSN 0027-8424, 1091-6490.
\newblock \doi{10.1073/pnas.1912789117}.

\bibitem[Dax et~al.(2023)Dax, Green, Gair, P{\"u}rrer, Wildberger, Macke,
  Buonanno, and Sch{\"o}lkopf]{Dax.2023}
M.~Dax, S.~R. Green, J.~Gair, M.~P{\"u}rrer, J.~Wildberger, J.~H. Macke,
  A.~Buonanno, and B.~Sch{\"o}lkopf.
\newblock Neural {{Importance Sampling}} for {{Rapid}} and {{Reliable
  Gravitational-Wave Inference}}.
\newblock \emph{Physical Review Letters}, 130\penalty0 (17):\penalty0 171403,
  Apr. 2023.
\newblock ISSN 0031-9007, 1079-7114.
\newblock \doi{10.1103/PhysRevLett.130.171403}.

\bibitem[Defazio et~al.(2024)Defazio, Yang, Mehta, Mishchenko, Khaled, and
  Cutkosky]{Defazio.2024}
A.~Defazio, X.~Yang, H.~Mehta, K.~Mishchenko, A.~Khaled, and A.~Cutkosky.
\newblock The {{Road Less Scheduled}}.
\newblock \emph{Advances in Neural Information Processing Systems},
  37:\penalty0 9974--10007, Dec. 2024.

\bibitem[Dempster et~al.(1977)Dempster, Laird, and Rubin]{Dempster.1977}
A.~P. Dempster, N.~M. Laird, and D.~B. Rubin.
\newblock Maximum {{Likelihood}} from {{Incomplete Data}} via the {{EM
  Algorithm}}.
\newblock \emph{Journal of the Royal Statistical Society. Series B
  (Methodological)}, 39\penalty0 (1):\penalty0 1--38, 1977.
\newblock ISSN 0035-9246.

\bibitem[Dinh et~al.(2017)Dinh, {Sohl-Dickstein}, and Bengio]{Dinh.2017}
L.~Dinh, J.~{Sohl-Dickstein}, and S.~Bengio.
\newblock Density estimation using {{Real NVP}}.
\newblock In \emph{International {{Conference}} on {{Learning
  Representations}}}, Feb. 2017.

\bibitem[Dua and Graff(2019)]{Dua.2019}
D.~Dua and C.~Graff.
\newblock {{UCI}} machine learning repository, 2019.

\bibitem[Durkan et~al.(2019)Durkan, Bekasov, Murray, and
  Papamakarios]{Durkan.2019}
C.~Durkan, A.~Bekasov, I.~Murray, and G.~Papamakarios.
\newblock Neural {{Spline Flows}}.
\newblock In \emph{Advances in {{Neural Information Processing Systems}}},
  volume~32. Curran Associates, Inc., 2019.

\bibitem[Elsem{\"u}ller et~al.(2024)Elsem{\"u}ller, Olischl{\"a}ger, Schmitt,
  B{\"u}rkner, Koethe, and Radev]{Elsemuller.2024}
L.~Elsem{\"u}ller, H.~Olischl{\"a}ger, M.~Schmitt, P.-C. B{\"u}rkner,
  U.~Koethe, and S.~T. Radev.
\newblock Sensitivity-{{Aware Amortized Bayesian Inference}}.
\newblock \emph{Transactions on Machine Learning Research}, May 2024.
\newblock ISSN 2835-8856.

\bibitem[{Figueroa-Z{\'u}{\~n}iga} et~al.(2013){Figueroa-Z{\'u}{\~n}iga},
  {Arellano-Valle}, and Ferrari]{Figueroa-Zuniga.2013}
J.~I. {Figueroa-Z{\'u}{\~n}iga}, R.~B. {Arellano-Valle}, and S.~L.~P. Ferrari.
\newblock Mixed beta regression: {{A Bayesian}} perspective.
\newblock \emph{Computational Statistics \& Data Analysis}, 61:\penalty0
  137--147, May 2013.
\newblock ISSN 0167-9473.
\newblock \doi{10.1016/j.csda.2012.12.002}.

\bibitem[Gabri{\'e} et~al.(2022)Gabri{\'e}, Rotskoff, and
  {Vanden-Eijnden}]{Gabrie.2022}
M.~Gabri{\'e}, G.~M. Rotskoff, and E.~{Vanden-Eijnden}.
\newblock Adaptive {{Monte Carlo}} augmented with normalizing flows.
\newblock \emph{Proceedings of the National Academy of Sciences}, 119\penalty0
  (10):\penalty0 e2109420119, Mar. 2022.
\newblock \doi{10.1073/pnas.2109420119}.

\bibitem[Gebhard et~al.(2023)Gebhard, Wildberger, Dax, Angerhausen, Quanz, and
  Sch{\"o}lkopf]{Gebhard.2023}
T.~D. Gebhard, J.~Wildberger, M.~Dax, D.~Angerhausen, S.~P. Quanz, and
  B.~Sch{\"o}lkopf.
\newblock Inferring {{Atmospheric Properties}} of {{Exoplanets}} with {{Flow
  Matching}} and {{Neural Importance Sampling}}.
\newblock \emph{arXiv}, Dec. 2023.
\newblock \doi{10.48550/arXiv.2312.08295}.

\bibitem[Gelfand and Smith(1990)]{Gelfand.1990}
A.~E. Gelfand and A.~F.~M. Smith.
\newblock Sampling-{{Based Approaches}} to {{Calculating Marginal Densities}}.
\newblock \emph{Journal of the American Statistical Association}, 85\penalty0
  (410):\penalty0 398--409, 1990.
\newblock ISSN 0162-1459.
\newblock \doi{10.2307/2289776}.

\bibitem[Gelman and Hill(2007)]{Gelman.2007}
A.~Gelman and J.~Hill.
\newblock \emph{Data Analysis Using Regression and Multilevel/Hierarchical
  Models}.
\newblock Analytical Methods for Social Research. Cambridge Univ. Press,
  Cambridge, 23rd printing edition, 2007.
\newblock ISBN 978-0-521-68689-1 978-0-521-86706-1.

\bibitem[Gelman et~al.(2013)Gelman, Carlin, Stern, Dunson, Vehtari, and
  Rubin]{Gelman.2013}
A.~Gelman, J.~B. Carlin, H.~S. Stern, D.~B. Dunson, A.~Vehtari, and D.~B.
  Rubin.
\newblock \emph{Bayesian Data Analysis}.
\newblock Texts in Statistical Science Series. CRC Press, Taylor \& Francis
  Group, Boca Raton London New York, third edition edition, 2013.
\newblock ISBN 978-1-4398-4095-5.

\bibitem[Gelman et~al.(2026)Gelman, Vehtari, McElreath, Simpson, Margossian,
  Yao, Kennedy, Gabry, B{\"u}rkner, Modr{\'a}k, and Leos~Barajas]{Gelman.2026}
A.~Gelman, A.~Vehtari, R.~McElreath, D.~Simpson, C.~C. Margossian, Y.~Yao,
  L.~Kennedy, J.~Gabry, P.-C. B{\"u}rkner, M.~Modr{\'a}k, and V.~Leos~Barajas.
\newblock \emph{Bayesian Workflow}.
\newblock CRC Press, Boca Raton, FL ; Abingdon, Oxon, first edition edition,
  2026.
\newblock ISBN 978-0-367-49018-8 978-0-367-49014-0.

\bibitem[Geweke(1989)]{Geweke.1989}
J.~Geweke.
\newblock Bayesian {{Inference}} in {{Econometric Models Using Monte Carlo
  Integration}}.
\newblock \emph{Econometrica}, 57\penalty0 (6):\penalty0 1317--1339, Nov. 1989.
\newblock ISSN 0012-9682.
\newblock \doi{10.2307/1913710}.

\bibitem[Giordano et~al.(2018)Giordano, Broderick, and Jordan]{Giordano.2018}
R.~Giordano, T.~Broderick, and M.~I. Jordan.
\newblock Covariances, {{Robustness}}, and {{Variational Bayes}}.
\newblock \emph{Journal of Machine Learning Research}, 19\penalty0
  (51):\penalty0 1--49, 2018.
\newblock ISSN 1533-7928.

\bibitem[Gloeckler et~al.(2024)Gloeckler, Deistler, Weilbach, Wood, and
  Macke]{Gloeckler.2024}
M.~Gloeckler, M.~Deistler, C.~D. Weilbach, F.~Wood, and J.~H. Macke.
\newblock All-in-one simulation-based inference.
\newblock In \emph{Proceedings of the 41st {{International Conference}} on
  {{Machine Learning}}}, pages 15735--15766. PMLR, July 2024.

\bibitem[Gordon(2019)]{Gordon.2019}
K.~R. Gordon.
\newblock How {{Mixed-Effects Modeling Can Advance Our Understanding}} of
  {{Learning}} and {{Memory}} and {{Improve Clinical}} and {{Educational
  Practice}}.
\newblock \emph{Journal of Speech, Language, and Hearing Research : JSLHR},
  62\penalty0 (3):\penalty0 507--524, Mar. 2019.
\newblock ISSN 1092-4388.
\newblock \doi{10.1044/2018_JSLHR-L-ASTM-18-0240}.

\bibitem[Greenberg et~al.(2019)Greenberg, Nonnenmacher, and
  Macke]{Greenberg.2019}
D.~Greenberg, M.~Nonnenmacher, and J.~Macke.
\newblock Automatic {{Posterior Transformation}} for {{Likelihood-Free
  Inference}}.
\newblock In \emph{Proceedings of the 36th {{International Conference}} on
  {{Machine Learning}}}, pages 2404--2414. PMLR, May 2019.

\bibitem[Grenioux et~al.(2023)Grenioux, Oliviero~Durmus, Moulines, and
  Gabri{\'e}]{Grenioux.2023}
L.~Grenioux, A.~Oliviero~Durmus, {\'E}.~Moulines, and M.~Gabri{\'e}.
\newblock On {{Sampling}} with {{Approximate Transport Maps}}.
\newblock In \emph{Proceedings of the 40th {{International Conference}} on
  {{Machine Learning}}}, volume 202, pages 11698--11733. PMLR, July 2023.

\bibitem[Gronau et~al.(2017)Gronau, Sarafoglou, Matzke, Ly, Boehm, Marsman,
  Leslie, Forster, Wagenmakers, and Steingroever]{Gronau.2017}
Q.~F. Gronau, A.~Sarafoglou, D.~Matzke, A.~Ly, U.~Boehm, M.~Marsman, D.~S.
  Leslie, J.~J. Forster, E.-J. Wagenmakers, and H.~Steingroever.
\newblock A tutorial on bridge sampling.
\newblock \emph{Journal of Mathematical Psychology}, 81:\penalty0 80--97, 2017.
\newblock ISSN 0022-2496.
\newblock \doi{10.1016/j.jmp.2017.09.005}.

\bibitem[Habermann et~al.(2025)Habermann, Schmitt, K{\"u}hmichel, Bulling,
  Radev, and B{\"u}rkner]{Habermann.2025}
D.~Habermann, M.~Schmitt, L.~K{\"u}hmichel, A.~Bulling, S.~T. Radev, and P.-C.
  B{\"u}rkner.
\newblock Amortized {{Bayesian Multilevel Models}}.
\newblock \emph{Bayesian Analysis}, -1\penalty0 (-1):\penalty0 1--30, Jan.
  2025.
\newblock ISSN 1936-0975, 1931-6690.
\newblock \doi{10.1214/25-BA1570}.

\bibitem[Harrison et~al.(2018)Harrison, Donaldson, {Correa-Cano}, Evans,
  Fisher, Goodwin, Robinson, Hodgson, and Inger]{Harrison.2018}
X.~A. Harrison, L.~Donaldson, M.~E. {Correa-Cano}, J.~Evans, D.~N. Fisher,
  C.~E.~D. Goodwin, B.~S. Robinson, D.~J. Hodgson, and R.~Inger.
\newblock A brief introduction to mixed effects modelling and multi-model
  inference in ecology.
\newblock \emph{PeerJ}, 6:\penalty0 e4794, May 2018.
\newblock ISSN 2167-8359.
\newblock \doi{10.7717/peerj.4794}.

\bibitem[Hastings(1970)]{Hastings.1970}
W.~K. Hastings.
\newblock Monte {{Carlo Sampling Methods Using Markov Chains}} and {{Their
  Applications}}.
\newblock \emph{Biometrika}, 57\penalty0 (1):\penalty0 97--109, 1970.
\newblock ISSN 0006-3444.
\newblock \doi{10.2307/2334940}.

\bibitem[He et~al.(2016)He, Zhang, Ren, and Sun]{He.2016}
K.~He, X.~Zhang, S.~Ren, and J.~Sun.
\newblock Deep {{Residual Learning}} for {{Image Recognition}}.
\newblock In \emph{2016 {{IEEE Conference}} on {{Computer Vision}} and
  {{Pattern Recognition}} ({{CVPR}})}, pages 770--778, June 2016.
\newblock \doi{10.1109/CVPR.2016.90}.

\bibitem[Henderson(1953)]{Henderson.1953}
C.~R. Henderson.
\newblock Estimation of variance and covariance components.
\newblock \emph{Biometrics. Journal of the International Biometric Society},
  9\penalty0 (2):\penalty0 226--252, 1953.
\newblock ISSN 0006-341X.
\newblock \doi{10.2307/3001853}.

\bibitem[Hoffman and Gelman(2014)]{Hoffman.2014}
M.~D. Hoffman and A.~Gelman.
\newblock The {{No-U-Turn Sampler}}: {{Adaptively Setting Path Lengths}} in
  {{Hamiltonian Monte Carlo}}.
\newblock \emph{Journal of Machine Learning Research}, 15\penalty0
  (47):\penalty0 1593--1623, 2014.
\newblock ISSN 1533-7928.

\bibitem[Hoffman and Sountsov(2022)]{Hoffman.2022}
M.~D. Hoffman and P.~Sountsov.
\newblock Tuning-{{Free Generalized Hamiltonian Monte Carlo}}.
\newblock In \emph{Proceedings of the 25th {{International Conference}} on
  {{Artificial Intelligence}} and {{Statistics}}}, volume 151 of
  \emph{Proceedings of {{Machine Learning Research}}}, pages 7799--7813. PMLR,
  2022.

\bibitem[Hoffman et~al.(2019)Hoffman, Sountsov, Dillon, Langmore, Tran, and
  Vasudevan]{Hoffman.2019a}
M.~D. Hoffman, P.~Sountsov, J.~V. Dillon, I.~Langmore, D.~Tran, and
  S.~Vasudevan.
\newblock {{NeuTra}}-lizing {{Bad Geometry}} in {{Hamiltonian Monte Carlo Using
  Neural Transport}}.
\newblock \emph{arXiv}, Mar. 2019.
\newblock \doi{10.48550/arXiv.1903.03704}.

\bibitem[Hoffman et~al.(2021)Hoffman, Radul, and Sountsov]{Hoffman.2021}
M.~D. Hoffman, A.~Radul, and P.~Sountsov.
\newblock An {{Adaptive-MCMC Scheme}} for {{Setting Trajectory Lengths}} in
  {{Hamiltonian Monte Carlo}}.
\newblock In \emph{Proceedings of the 24th {{International Conference}} on
  {{Artificial Intelligence}} and {{Statistics}}}, volume 130 of
  \emph{Proceedings of {{Machine Learning Research}}}, pages 3907--3915. PMLR,
  2021.

\bibitem[Hollmann et~al.(2025)Hollmann, M{\"u}ller, Purucker, Krishnakumar,
  K{\"o}rfer, Hoo, Schirrmeister, and Hutter]{Hollmann.2025}
N.~Hollmann, S.~M{\"u}ller, L.~Purucker, A.~Krishnakumar, M.~K{\"o}rfer, S.~B.
  Hoo, R.~T. Schirrmeister, and F.~Hutter.
\newblock Accurate predictions on small data with a tabular foundation model.
\newblock \emph{Nature}, 637\penalty0 (8045):\penalty0 319--326, Jan. 2025.
\newblock ISSN 0028-0836, 1476-4687.
\newblock \doi{10.1038/s41586-024-08328-6}.

\bibitem[Huang et~al.(2026)Huang, Schumacher, Stevenson, and Radev]{Huang.2026}
J.~M. Huang, L.~Schumacher, N.~Stevenson, and S.~T. Radev.
\newblock {{CogFormer}}: {{Learn All Your Models Once}}.
\newblock \emph{arXiv}, June 2026.
\newblock \doi{10.48550/arXiv.2603.20520}.

\bibitem[Jeffreys(1961)]{Jeffreys.1961}
H.~Jeffreys.
\newblock \emph{Theory of {{Probability}}}.
\newblock The {{International Series}} of {{Monographs}} on {{Physics}}. Oxford
  University Press, Oxford, 3 edition, 1961.
\newblock ISBN 978-0-19-850368-2.

\bibitem[Kallioinen et~al.(2024)Kallioinen, Paananen, B{\"u}rkner, and
  Vehtari]{Kallioinen.2024}
N.~Kallioinen, T.~Paananen, P.-C. B{\"u}rkner, and A.~Vehtari.
\newblock Detecting and diagnosing prior and likelihood sensitivity with
  power-scaling.
\newblock \emph{Statistics and Computing}, 34\penalty0 (1):\penalty0 57, Feb.
  2024.
\newblock ISSN 0960-3174.
\newblock \doi{10.1007/s11222-023-10366-5}.

\bibitem[Kass and Raftery(1995)]{Kass.1995}
R.~E. Kass and A.~E. Raftery.
\newblock Bayes {{Factors}}.
\newblock \emph{Journal of the American Statistical Association}, 90\penalty0
  (430):\penalty0 773--795, 1995.
\newblock ISSN 0162-1459.
\newblock \doi{10.1080/01621459.1995.10476572}.

\bibitem[Kingma and Dhariwal(2018)]{Kingma.2018}
D.~P. Kingma and P.~Dhariwal.
\newblock Glow: {{Generative Flow}} with {{Invertible}} 1x1 {{Convolutions}}.
\newblock In \emph{Advances in {{Neural Information Processing Systems}}},
  volume~31. Curran Associates, Inc., 2018.

\bibitem[Kucukelbir et~al.(2017)Kucukelbir, Tran, Ranganath, Gelman, and
  Blei]{Kucukelbir.2017}
A.~Kucukelbir, D.~Tran, R.~Ranganath, A.~Gelman, and D.~M. Blei.
\newblock Automatic {{Differentiation Variational Inference}}.
\newblock \emph{Journal of Machine Learning Research}, 18\penalty0
  (14):\penalty0 1--45, 2017.
\newblock ISSN 1533-7928.

\bibitem[K{\"u}hmichel et~al.(2026)K{\"u}hmichel, Huang, Pratz, Arruda,
  Olischl{\"a}ger, Habermann, Kucharsky, Elsem{\"u}ller, Mishra, Bracher,
  Jedhoff, Schmitt, B{\"u}rkner, and Radev]{Kuhmichel.2026}
L.~K{\"u}hmichel, J.~M. Huang, V.~Pratz, J.~Arruda, H.~Olischl{\"a}ger,
  D.~Habermann, S.~Kucharsky, L.~Elsem{\"u}ller, A.~Mishra, N.~Bracher,
  S.~Jedhoff, M.~Schmitt, P.-C. B{\"u}rkner, and S.~T. Radev.
\newblock {{BayesFlow}} 2: {{Multi-Backend Amortized Bayesian Inference}} in
  {{Python}}.
\newblock \emph{arXiv}, Mar. 2026.
\newblock \doi{10.48550/arXiv.2602.07098}.

\bibitem[Kumle et~al.(2021)Kumle, V{\~o}, and Draschkow]{Kumle.2021}
L.~Kumle, M.~L.-H. V{\~o}, and D.~Draschkow.
\newblock Estimating power in (generalized) linear mixed models: {{An}} open
  introduction and tutorial in {{R}}.
\newblock \emph{Behavior Research Methods}, 53\penalty0 (6):\penalty0
  2528--2543, Dec. 2021.
\newblock ISSN 1554-3528.
\newblock \doi{10.3758/s13428-021-01546-0}.

\bibitem[Laird and Ware(1982)]{Laird.1982}
N.~M. Laird and J.~H. Ware.
\newblock Random-effects models for longitudinal data.
\newblock \emph{Biometrics. Journal of the International Biometric Society},
  38\penalty0 (4):\penalty0 963--974, 1982.
\newblock ISSN 0006-341X.
\newblock \doi{10.2307/2529876}.

\bibitem[Lee et~al.(2019)Lee, Lee, Kim, Kosiorek, Choi, and Teh]{Lee.2019}
J.~Lee, Y.~Lee, J.~Kim, A.~Kosiorek, S.~Choi, and Y.~W. Teh.
\newblock Set {{Transformer}}: {{A Framework}} for {{Attention}}-based
  {{Permutation-Invariant Neural Networks}}.
\newblock In \emph{Proceedings of the 36th {{International Conference}} on
  {{Machine Learning}}}, pages 3744--3753. PMLR, May 2019.

\bibitem[Lewandowski et~al.(2009)Lewandowski, Kurowicka, and
  Joe]{Lewandowski.2009}
D.~Lewandowski, D.~Kurowicka, and H.~Joe.
\newblock Generating random correlation matrices based on vines and extended
  onion method.
\newblock \emph{Journal of Multivariate Analysis}, 100\penalty0 (9):\penalty0
  1989--2001, Oct. 2009.
\newblock ISSN 0047259X.
\newblock \doi{10.1016/j.jmva.2009.04.008}.

\bibitem[Li et~al.(2026)Li, Vehtari, B{\"u}rkner, Radev, Acerbi, and
  Schmitt]{Li.2026}
C.~Li, A.~Vehtari, P.-C. B{\"u}rkner, S.~T. Radev, L.~Acerbi, and M.~Schmitt.
\newblock Amortized {{Bayesian Workflow}}.
\newblock \emph{Transactions on Machine Learning Research}, 2026.

\bibitem[Lichtenberg and {\c S}im{\c s}ek(2017)]{Lichtenberg.2017}
J.~M. Lichtenberg and {\"O}.~{\c S}im{\c s}ek.
\newblock Simple {{Regression Models}}.
\newblock In \emph{Proceedings of the {{NIPS}} 2016 {{Workshop}} on {{Imperfect
  Decision Makers}}}, pages 13--25. PMLR, Aug. 2017.

\bibitem[Lindgren and Rue(2015)]{Lindgren.2015}
F.~Lindgren and H.~Rue.
\newblock Bayesian {{Spatial Modelling}} with {{R-INLA}}.
\newblock \emph{Journal of Statistical Software}, 63\penalty0 (19):\penalty0
  1--25, 2015.
\newblock \doi{10.18637/jss.v063.i19}.

\bibitem[Linhart et~al.(2023)Linhart, Gramfort, and Rodrigues]{Linhart.2023}
J.~Linhart, A.~Gramfort, and P.~L.~C. Rodrigues.
\newblock L-{{C2ST}}: {{Local Diagnostics}} for {{Posterior Approximations}} in
  {{Simulation-Based Inference}}.
\newblock In \emph{Advances in {{Neural Information Processing Systems}}},
  volume~36, 2023.

\bibitem[Liu(1996)]{Liu.1996}
J.~S. Liu.
\newblock Metropolized independent sampling with comparisons to rejection
  sampling and importance sampling.
\newblock \emph{Statistics and Computing}, 6\penalty0 (2):\penalty0 113--119,
  June 1996.
\newblock \doi{10.1007/BF00162521}.

\bibitem[Lueckmann et~al.(2017)Lueckmann, Goncalves, Bassetto, {\"O}cal,
  Nonnenmacher, and Macke]{Lueckmann.2017}
J.-M. Lueckmann, P.~J. Goncalves, G.~Bassetto, K.~{\"O}cal, M.~Nonnenmacher,
  and J.~H. Macke.
\newblock Flexible statistical inference for mechanistic models of neural
  dynamics.
\newblock In \emph{Advances in {{Neural Information Processing Systems}}},
  volume~30. Curran Associates, Inc., 2017.

\bibitem[Margossian et~al.(2020)Margossian, Vehtari, Simpson, and
  Agrawal]{Margossian.2020}
C.~C. Margossian, A.~Vehtari, D.~Simpson, and R.~Agrawal.
\newblock Hamiltonian {{Monte Carlo}} using an adjoint-differentiated
  {{Laplace}} approximation: {{Bayesian}} inference for latent {{Gaussian}}
  models and beyond.
\newblock In \emph{Advances in {{Neural Information Processing Systems}}},
  volume~33, 2020.

\bibitem[McCullagh and Nelder(1989)]{McCullagh.1989}
P.~McCullagh and J.~A. Nelder.
\newblock \emph{Generalized {{Linear Models}}}.
\newblock Chapman \& Hall, London, 2 edition, 1989.
\newblock ISBN 978-0-412-31760-6.

\bibitem[Meng and Wong(1996)]{Meng.1996}
X.-L. Meng and W.~H. Wong.
\newblock Simulating ratios of normalizing constants via a simple identity: A
  theoretical exploration.
\newblock \emph{Statistica Sinica}, 6\penalty0 (4):\penalty0 831--860, 1996.

\bibitem[Metropolis et~al.(1953)Metropolis, Rosenbluth, Rosenbluth, Teller, and
  Teller]{Metropolis.1953}
N.~Metropolis, A.~W. Rosenbluth, M.~N. Rosenbluth, A.~H. Teller, and E.~Teller.
\newblock Equation of {{State Calculations}} by {{Fast Computing Machines}}.
\newblock \emph{The Journal of Chemical Physics}, 21\penalty0 (6):\penalty0
  1087--1092, June 1953.
\newblock ISSN 0021-9606.
\newblock \doi{10.1063/1.1699114}.

\bibitem[Mittal et~al.(2025)Mittal, Bracher, Lajoie, Jaini, and
  Brubaker]{Mittal.2025}
S.~Mittal, N.~L. Bracher, G.~Lajoie, P.~Jaini, and M.~Brubaker.
\newblock Amortized {{In-Context Bayesian Posterior Estimation}}.
\newblock \emph{arXiv}, Feb. 2025.
\newblock \doi{10.48550/arXiv.2502.06601}.

\bibitem[M{\"u}ller et~al.(2021)M{\"u}ller, Hollmann, Arango, Grabocka, and
  Hutter]{Muller.2021}
S.~M{\"u}ller, N.~Hollmann, S.~P. Arango, J.~Grabocka, and F.~Hutter.
\newblock Transformers {{Can Do Bayesian Inference}}.
\newblock In \emph{International {{Conference}} on {{Learning
  Representations}}}, Oct. 2021.

\bibitem[M{\"u}ller et~al.(2019)M{\"u}ller, McWilliams, Rousselle, Gross, and
  Nov{\'a}k]{Muller.2019}
T.~M{\"u}ller, B.~McWilliams, F.~Rousselle, M.~Gross, and J.~Nov{\'a}k.
\newblock Neural {{Importance Sampling}}.
\newblock \emph{ACM Transactions on Graphics}, 38\penalty0 (5):\penalty0
  145:1--145:19, 2019.
\newblock \doi{10.1145/3341156}.

\bibitem[Nautiyal et~al.(2026)Nautiyal, Ju, Ernfors, Hagland, Holma,
  S{\"o}derholm, Hellander, and Singh]{Nautiyal.2026}
M.~Nautiyal, L.~Ju, M.~Ernfors, K.~Hagland, V.~Holma, M.~W. S{\"o}derholm,
  A.~Hellander, and P.~Singh.
\newblock {{OneFlowSBI}}: {{One Model}}, {{Many Queries}} for
  {{Simulation-Based Inference}}.
\newblock \emph{arXiv}, Jan. 2026.
\newblock \doi{10.48550/arXiv.2601.22951}.

\bibitem[Oberauer(2022)]{Oberauer.2022}
K.~Oberauer.
\newblock The {{Importance}} of {{Random Slopes}} in {{Mixed Models}} for
  {{Bayesian Hypothesis Testing}}.
\newblock \emph{Psychological Science}, 33\penalty0 (4):\penalty0 648--665,
  Apr. 2022.
\newblock ISSN 0956-7976.
\newblock \doi{10.1177/09567976211046884}.

\bibitem[Papamakarios and Murray(2016)]{Papamakarios.2016}
G.~Papamakarios and I.~Murray.
\newblock Fast \textbackslash epsilon -free {{Inference}} of {{Simulation
  Models}} with {{Bayesian Conditional Density Estimation}}.
\newblock In \emph{Advances in {{Neural Information Processing Systems}}},
  volume~29. Curran Associates, Inc., 2016.

\bibitem[Papamakarios et~al.(2021)Papamakarios, Nalisnick, Rezende, Mohamed,
  and Lakshminarayanan]{Papamakarios.2021}
G.~Papamakarios, E.~Nalisnick, D.~J. Rezende, S.~Mohamed, and
  B.~Lakshminarayanan.
\newblock Normalizing {{Flows}} for {{Probabilistic Modeling}} and
  {{Inference}}.
\newblock \emph{Journal of Machine Learning Research}, 22, 2021.

\bibitem[Papaspiliopoulos et~al.(2007)Papaspiliopoulos, Roberts, and
  Sk{\"o}ld]{Papaspiliopoulos.2007}
O.~Papaspiliopoulos, G.~O. Roberts, and M.~Sk{\"o}ld.
\newblock A {{General Framework}} for the {{Parametrization}} of {{Hierarchical
  Models}}.
\newblock \emph{Statistical Science}, 22\penalty0 (1):\penalty0 59--73, Feb.
  2007.
\newblock ISSN 0883-4237, 2168-8745.
\newblock \doi{10.1214/088342307000000014}.

\bibitem[Paszke et~al.(2019)Paszke, Gross, Massa, Lerer, Bradbury, Chanan,
  Killeen, Lin, Gimelshein, Antiga, Desmaison, Kopf, Yang, DeVito, Raison,
  Tejani, Chilamkurthy, Steiner, Fang, Bai, and Chintala]{Paszke.2019a}
A.~Paszke, S.~Gross, F.~Massa, A.~Lerer, J.~Bradbury, G.~Chanan, T.~Killeen,
  Z.~Lin, N.~Gimelshein, L.~Antiga, A.~Desmaison, A.~Kopf, E.~Yang, Z.~DeVito,
  M.~Raison, A.~Tejani, S.~Chilamkurthy, B.~Steiner, L.~Fang, J.~Bai, and
  S.~Chintala.
\newblock {{PyTorch}}: {{An Imperative Style}}, {{High-Performance Deep
  Learning Library}}.
\newblock In \emph{Advances in {{Neural Information Processing Systems}}},
  volume~32. Curran Associates, Inc., 2019.

\bibitem[Pinheiro et~al.(1999)Pinheiro, Bates, and {R Core
  Team}]{Pinheiro.1999}
J.~Pinheiro, D.~Bates, and {R Core Team}.
\newblock Nlme: {{Linear}} and {{Nonlinear Mixed Effects Models}}.
\newblock \emph{Comprehensive R Archive Network}, Nov. 1999.
\newblock \doi{10.32614/CRAN.package.nlme}.

\bibitem[Priem et~al.(2022)Priem, Piwowar, and Orr]{Priem.2022}
J.~Priem, H.~Piwowar, and R.~Orr.
\newblock {{OpenAlex}}: {{A}} fully-open index of scholarly works, authors,
  venues, institutions, and concepts.
\newblock \emph{arXiv}, May 2022.
\newblock \doi{10.48550/arXiv.2205.01833}.

\bibitem[Qu et~al.(2025)Qu, Holzm{\"u}ller, Varoquaux, and Morvan]{Qu.2025}
J.~Qu, D.~Holzm{\"u}ller, G.~Varoquaux, and M.~L. Morvan.
\newblock {{TabICL}}: {{A Tabular Foundation Model}} for {{In-Context
  Learning}} on {{Large Data}}.
\newblock In \emph{Forty-Second {{International Conference}} on {{Machine
  Learning}}}, June 2025.

\bibitem[Radev et~al.(2022)Radev, Mertens, Voss, Ardizzone, and
  K{\"o}the]{Radev.2022}
S.~T. Radev, U.~K. Mertens, A.~Voss, L.~Ardizzone, and U.~K{\"o}the.
\newblock {{BayesFlow}}: {{Learning Complex Stochastic Models With Invertible
  Neural Networks}}.
\newblock \emph{IEEE Transactions on Neural Networks and Learning Systems},
  33\penalty0 (4):\penalty0 1452--1466, Apr. 2022.
\newblock ISSN 2162-2388.
\newblock \doi{10.1109/TNNLS.2020.3042395}.

\bibitem[Reuter et~al.(2025)Reuter, Rudner, Fortuin, and
  R{\"u}gamer]{Reuter.2025}
A.~Reuter, T.~G.~J. Rudner, V.~Fortuin, and D.~R{\"u}gamer.
\newblock Can {{Transformers Learn Full Bayesian Inference}} in {{Context}}?
\newblock In \emph{Forty-Second {{International Conference}} on {{Machine
  Learning}}}, June 2025.

\bibitem[Rezende and Mohamed(2015)]{Rezende.2015}
D.~Rezende and S.~Mohamed.
\newblock Variational {{Inference}} with {{Normalizing Flows}}.
\newblock In \emph{Proceedings of the 32nd {{International Conference}} on
  {{Machine Learning}}}, pages 1530--1538. PMLR, June 2015.

\bibitem[Roback and Legler(2021)]{Roback.2021}
P.~Roback and J.~Legler.
\newblock \emph{Beyond {{Multiple Linear Regression}}: {{Applied Generalized
  Linear Models}} and {{Multilevel Models}} in {{R}}}.
\newblock {Chapman and Hall/CRC}, New York, 2021.
\newblock ISBN 978-0-429-06666-5.
\newblock \doi{10.1201/9780429066665}.

\bibitem[Robinson(1991)]{Robinson.1991}
G.~K. Robinson.
\newblock That {{BLUP}} is a {{Good Thing}}: {{The Estimation}} of {{Random
  Effects}}.
\newblock \emph{Statistical Science}, 6\penalty0 (1):\penalty0 15--32, 1991.
\newblock ISSN 0883-4237.

\bibitem[Rodrigues et~al.(2021)Rodrigues, Moreau, Louppe, and
  Gramfort]{Rodrigues.2021}
P.~Rodrigues, T.~Moreau, G.~Louppe, and A.~Gramfort.
\newblock {{HNPE}}: {{Leveraging Global Parameters}} for {{Neural Posterior
  Estimation}}.
\newblock In \emph{Advances in {{Neural Information Processing Systems}}},
  volume~34, pages 13432--13443. Curran Associates, Inc., 2021.

\bibitem[Romano et~al.(2022)Romano, Le, La~Cava, Gregg, Goldberg, Chakraborty,
  Ray, Himmelstein, Fu, and Moore]{Romano.2022}
J.~D. Romano, T.~T. Le, W.~La~Cava, J.~T. Gregg, D.~J. Goldberg,
  P.~Chakraborty, N.~L. Ray, D.~Himmelstein, W.~Fu, and J.~H. Moore.
\newblock {{PMLB}} v1.0: An open-source dataset collection for benchmarking
  machine learning methods.
\newblock \emph{Bioinformatics}, 38\penalty0 (3):\penalty0 878--880, Feb. 2022.
\newblock ISSN 1367-4803.
\newblock \doi{10.1093/bioinformatics/btab727}.

\bibitem[Roy(2020)]{Roy.2020}
V.~Roy.
\newblock Convergence {{Diagnostics}} for {{Markov Chain Monte Carlo}}.
\newblock \emph{Annual Review of Statistics and Its Application}, 7:\penalty0
  387--412, 2020.
\newblock ISSN 2326-831X.
\newblock \doi{10.1146/annurev-statistics-031219-041300}.

\bibitem[Rue et~al.(2009)Rue, Martino, and Chopin]{Rue.2009}
H.~Rue, S.~Martino, and N.~Chopin.
\newblock Approximate {{Bayesian}} inference for latent {{Gaussian}} models by
  using integrated nested {{Laplace}} approximations.
\newblock \emph{Journal of the Royal Statistical Society: Series B (Statistical
  Methodology)}, 71\penalty0 (2):\penalty0 319--392, 2009.
\newblock ISSN 1467-9868.
\newblock \doi{10.1111/j.1467-9868.2008.00700.x}.

\bibitem[S{\"a}ilynoja et~al.(2022)S{\"a}ilynoja, B{\"u}rkner, and
  Vehtari]{Sailynoja.2022}
T.~S{\"a}ilynoja, P.-C. B{\"u}rkner, and A.~Vehtari.
\newblock Graphical {{Test}} for {{Discrete Uniformity}} and its
  {{Applications}} in {{Goodness}} of {{Fit Evaluation}} and {{Multiple Sample
  Comparison}}.
\newblock \emph{Statistics and Computing}, 32\penalty0 (2):\penalty0 32, Apr.
  2022.
\newblock ISSN 0960-3174, 1573-1375.
\newblock \doi{10.1007/s11222-022-10090-6}.

\bibitem[S{\"a}ilynoja et~al.(2025)S{\"a}ilynoja, Schmitt, B{\"u}rkner, and
  Vehtari]{Sailynoja.2025}
T.~S{\"a}ilynoja, M.~Schmitt, P.-C. B{\"u}rkner, and A.~Vehtari.
\newblock Posterior {{SBC}}: {{Simulation-Based Calibration Checking
  Conditional}} on {{Data}}.
\newblock \emph{arXiv}, 2025.
\newblock \doi{10.48550/arXiv.2502.03279}.

\bibitem[Schmitt et~al.(2024{\natexlab{a}})Schmitt, B{\"u}rkner, K{\"o}the, and
  Radev]{Schmitt.2024b}
M.~Schmitt, P.-C. B{\"u}rkner, U.~K{\"o}the, and S.~T. Radev.
\newblock Detecting {{Model Misspecification}} in~{{Amortized Bayesian
  Inference}} with~{{Neural Networks}}.
\newblock In U.~K{\"o}the and C.~Rother, editors, \emph{Pattern
  {{Recognition}}: 45th {{DAGM German Conference}}, {{DAGM GCPR}} 2023}, pages
  541--557, Cham, 2024{\natexlab{a}}. Springer Nature Switzerland.
\newblock ISBN 978-3-031-54605-1.
\newblock \doi{10.1007/978-3-031-54605-1_35}.

\bibitem[Schmitt et~al.(2024{\natexlab{b}})Schmitt, Ivanova, Habermann,
  K{\"o}the, B{\"u}rkner, and Radev]{Schmitt.2024a}
M.~Schmitt, D.~R. Ivanova, D.~Habermann, U.~K{\"o}the, P.-C. B{\"u}rkner, and
  S.~T. Radev.
\newblock Leveraging {{Self-Consistency}} for {{Data-Efficient Amortized
  Bayesian Inference}}.
\newblock In \emph{Proceedings of the 41st {{International Conference}} on
  {{Machine Learning}}}, volume 235, pages 43723--43741. PMLR,
  2024{\natexlab{b}}.

\bibitem[Schrimpf et~al.(2021)Schrimpf, Des~Brisay, Johnston, Smith,
  {S{\'a}nchez-Jasso}, Robinson, Warrington, Mahony, Horn, {Strimas-Mackey},
  Fahrig, and Koper]{Schrimpf.2021}
M.~B. Schrimpf, P.~G. Des~Brisay, A.~Johnston, A.~C. Smith,
  J.~{S{\'a}nchez-Jasso}, B.~G. Robinson, M.~H. Warrington, N.~A. Mahony, A.~G.
  Horn, M.~{Strimas-Mackey}, L.~Fahrig, and N.~Koper.
\newblock Reduced human activity during {{COVID-19}} alters avian land use
  across {{North America}}.
\newblock \emph{Science Advances}, 7\penalty0 (39):\penalty0 eabf5073, Sept.
  2021.
\newblock ISSN 2375-2548.
\newblock \doi{10.1126/sciadv.abf5073}.

\bibitem[Searle et~al.(1992)Searle, Casella, and McCulloch]{Searle.1992}
S.~R. Searle, G.~Casella, and C.~E. McCulloch.
\newblock \emph{Variance {{Components}}}.
\newblock Wiley, New York, 1992.
\newblock ISBN 978-0-471-62834-2.
\newblock \doi{10.1002/9780470316856}.

\bibitem[Singmann et~al.(2025)Singmann, Bolker, Westfall, Aust, and
  {Ben-Shachar}]{Singmann.2025}
H.~Singmann, B.~Bolker, J.~Westfall, F.~Aust, and M.~S. {Ben-Shachar}.
\newblock Afex: {{Analysis}} of {{Factorial Experiments}}.
\newblock CRAN, 2025.

\bibitem[Spiegelhalter et~al.(1994)Spiegelhalter, Freedman, and
  Parmar]{Spiegelhalter.1994}
D.~J. Spiegelhalter, L.~S. Freedman, and M.~K.~B. Parmar.
\newblock Bayesian {{Approaches}} to {{Randomized Trials}}.
\newblock \emph{Journal of the Royal Statistical Society. Series A (Statistics
  in Society)}, 157\penalty0 (3):\penalty0 357--416, 1994.
\newblock ISSN 0964-1998.
\newblock \doi{10.2307/2983527}.

\bibitem[Sterne et~al.(2009)Sterne, White, Carlin, Spratt, Royston, Kenward,
  Wood, and Carpenter]{Sterne.2009}
J.~A.~C. Sterne, I.~R. White, J.~B. Carlin, M.~Spratt, P.~Royston, M.~G.
  Kenward, A.~M. Wood, and J.~R. Carpenter.
\newblock Multiple imputation for missing data in epidemiological and clinical
  research: Potential and pitfalls.
\newblock \emph{BMJ}, 338:\penalty0 b2393, June 2009.
\newblock ISSN 0959-8138, 1468-5833.
\newblock \doi{10.1136/bmj.b2393}.

\bibitem[Talts et~al.(2020)Talts, Betancourt, Simpson, Vehtari, and
  Gelman]{Talts.2020}
S.~Talts, M.~Betancourt, D.~Simpson, A.~Vehtari, and A.~Gelman.
\newblock Validating {{Bayesian Inference Algorithms}} with {{Simulation-Based
  Calibration}}.
\newblock \emph{arXiv}, Oct. 2020.
\newblock \doi{10.48550/arXiv.1804.06788}.

\bibitem[Tierney(1994)]{Tierney.1994}
L.~Tierney.
\newblock Markov {{Chains}} for {{Exploring Posterior Distributions}}.
\newblock \emph{The Annals of Statistics}, 22\penalty0 (4):\penalty0
  1701--1728, Dec. 1994.
\newblock ISSN 0090-5364, 2168-8966.
\newblock \doi{10.1214/aos/1176325750}.

\bibitem[Tokdar and Kass(2010)]{Tokdar.2010}
S.~T. Tokdar and R.~E. Kass.
\newblock Importance sampling: A review.
\newblock \emph{WIREs Computational Statistics}, 2\penalty0 (1):\penalty0
  54--60, Jan. 2010.
\newblock ISSN 1939-5108, 1939-0068.
\newblock \doi{10.1002/wics.56}.

\bibitem[Tran et~al.(2013)Tran, Scharth, Pitt, and Kohn]{Tran.2013}
M.-N. Tran, M.~Scharth, M.~K. Pitt, and R.~Kohn.
\newblock Importance {{Sampling Squared}} for {{Bayesian Inference}} in
  {{Latent Variable Models}}.
\newblock \emph{arXiv}, 2013.

\bibitem[Turner and Sahani(2011)]{Turner.2011}
R.~E. Turner and M.~Sahani.
\newblock Two problems with variational expectation maximisation for time
  series models.
\newblock In D.~Barber, A.~T. Cemgil, and S.~Chiappa, editors, \emph{Bayesian
  {{Time Series Models}}}, pages 104--124. Cambridge University Press,
  Cambridge, 2011.
\newblock \doi{10.1017/CBO9780511984679.006}.

\bibitem[Vasishth et~al.(2023)Vasishth, Yadav, Schad, and
  Nicenboim]{Vasishth.2023}
S.~Vasishth, H.~Yadav, D.~J. Schad, and B.~Nicenboim.
\newblock Sample {{Size Determination}} for {{Bayesian Hierarchical Models
  Commonly Used}} in {{Psycholinguistics}}.
\newblock \emph{Computational Brain \& Behavior}, 6\penalty0 (1):\penalty0
  102--126, Mar. 2023.
\newblock ISSN 2522-0861.
\newblock \doi{10.1007/s42113-021-00125-y}.

\bibitem[Vaswani et~al.(2017)Vaswani, Shazeer, Parmar, Uszkoreit, Jones, Gomez,
  {ukasz Kaiser}, and Polosukhin]{Vaswani.2017}
A.~Vaswani, N.~Shazeer, N.~Parmar, J.~Uszkoreit, L.~Jones, A.~N. Gomez,
  {\L}.~{ukasz Kaiser}, and I.~Polosukhin.
\newblock Attention is {{All}} you {{Need}}.
\newblock In \emph{Advances in {{Neural Information Processing Systems}}},
  volume~30. Curran Associates, Inc., 2017.

\bibitem[Vehtari et~al.(2017)Vehtari, Gelman, and Gabry]{Vehtari.2015}
A.~Vehtari, A.~Gelman, and J.~Gabry.
\newblock Practical {{Bayesian}} model evaluation using leave-one-out
  cross-validation and {{WAIC}}.
\newblock \emph{Statistics and Computing}, 27\penalty0 (5):\penalty0
  1413--1432, 2017.
\newblock \doi{10.1007/s11222-016-9696-4}.

\bibitem[Vehtari et~al.(2021)Vehtari, Gelman, Simpson, Carpenter, and
  B{\"u}rkner]{Vehtari.2021}
A.~Vehtari, A.~Gelman, D.~Simpson, B.~Carpenter, and P.-C. B{\"u}rkner.
\newblock Rank-{{Normalization}}, {{Folding}}, and {{Localization}}: {{An
  Improved R\textasciicircum}} for {{Assessing Convergence}} of {{MCMC}} (with
  {{Discussion}}).
\newblock \emph{Bayesian Analysis}, 16\penalty0 (2):\penalty0 667--718, June
  2021.
\newblock ISSN 1936-0975, 1931-6690.
\newblock \doi{10.1214/20-BA1221}.

\bibitem[Wagenmakers et~al.(2010)Wagenmakers, Lodewyckx, Kuriyal, and
  Grasman]{Wagenmakers.2010}
E.-J. Wagenmakers, T.~Lodewyckx, H.~Kuriyal, and R.~Grasman.
\newblock Bayesian hypothesis testing for psychologists: {{A}} tutorial on the
  {{Savage}}--{{Dickey}} method.
\newblock \emph{Cognitive Psychology}, 60\penalty0 (3):\penalty0 158--189,
  2010.
\newblock ISSN 0010-0285.
\newblock \doi{10.1016/j.cogpsych.2009.12.001}.

\bibitem[Wang and Gelfand(2002)]{Wang.2002}
F.~Wang and A.~E. Gelfand.
\newblock A {{Simulation-Based Approach}} to {{Bayesian Sample Size
  Determination}} for {{Performance}} under a {{Given Model}} and for
  {{Separating Models}}.
\newblock \emph{Statistical Science}, 17\penalty0 (2):\penalty0 193--208, May
  2002.
\newblock ISSN 0883-4237.
\newblock \doi{10.1214/ss/1030550861}.

\bibitem[Wang et~al.(2021)Wang, Goldenberg, Dorison, Miller, Uusberg, Lerner,
  Gross, et~al.]{Wang.2021}
K.~Wang, A.~Goldenberg, C.~A. Dorison, J.~K. Miller, A.~Uusberg, J.~S. Lerner,
  J.~J. Gross, et~al.
\newblock A multi-country test of brief reappraisal interventions on emotions
  during the {{COVID-19}} pandemic.
\newblock \emph{Nature Human Behaviour}, 5\penalty0 (8):\penalty0 1089--1110,
  Aug. 2021.
\newblock ISSN 2397-3374.
\newblock \doi{10.1038/s41562-021-01173-x}.

\bibitem[Ward et~al.(2022)Ward, Cannon, Beaumont, Fasiolo, and
  Schmon]{Ward.2022}
D.~Ward, P.~Cannon, M.~Beaumont, M.~Fasiolo, and S.~Schmon.
\newblock Robust {{Neural Posterior Estimation}} and {{Statistical Model
  Criticism}}.
\newblock \emph{Advances in Neural Information Processing Systems},
  35:\penalty0 33845--33859, Dec. 2022.

\bibitem[Wasserman(2010)]{Wasserman.2010}
L.~Wasserman.
\newblock \emph{All of Statistics: A Concise Course in Statistical Inference}.
\newblock Springer Texts in Statistics. Springer, New York Berlin Heidelberg,
  corr. 2. print., [repr.] edition, 2010.
\newblock ISBN 978-1-4419-2322-6.

\bibitem[Whittle et~al.(2026)Whittle, Ziomek, Rawling, and
  Osborne]{Whittle.2026}
G.~Whittle, J.~Ziomek, J.~Rawling, and M.~A. Osborne.
\newblock Distribution {{Transformers}}: {{Fast Approximate Bayesian Inference
  With On-The-Fly Prior Adaptation}}.
\newblock \emph{arXiv}, May 2026.
\newblock \doi{10.48550/arXiv.2502.02463}.

\bibitem[Yang et~al.(2026)Yang, Rissanen, Chang, Loka, Huang, Solin, Heinonen,
  and Acerbi]{Yang.2026}
Y.~Yang, S.~Rissanen, P.~E. Chang, N.~Loka, D.~Huang, A.~Solin, M.~Heinonen,
  and L.~Acerbi.
\newblock {{PriorGuide}}: {{Test-Time Prior Adaptation}} for {{Simulation-Based
  Inference}}.
\newblock In \emph{International {{Conference}} on {{Learning Representations}}
  ({{ICLR}})}, 2026.

\bibitem[Yao et~al.(2018)Yao, Vehtari, Simpson, and Gelman]{Yao.2018}
Y.~Yao, A.~Vehtari, D.~Simpson, and A.~Gelman.
\newblock Yes, but {{Did It Work}}?: {{Evaluating Variational Inference}}.
\newblock In \emph{Proceedings of the 35th {{International Conference}} on
  {{Machine Learning}}}, pages 5581--5590. PMLR, July 2018.

\bibitem[Yu et~al.(2022)Yu, Guindani, Grieco, Chen, Holmes, and Xu]{Yu.2022}
Z.~Yu, M.~Guindani, S.~F. Grieco, L.~Chen, T.~C. Holmes, and X.~Xu.
\newblock Beyond t test and {{ANOVA}}: {{Applications}} of mixed-effects models
  for more rigorous statistical analysis in neuroscience research.
\newblock \emph{Neuron}, 110\penalty0 (1):\penalty0 21--35, Jan. 2022.
\newblock ISSN 1097-4199.
\newblock \doi{10.1016/j.neuron.2021.10.030}.

\bibitem[Zaheer et~al.(2017)Zaheer, Kottur, Ravanbakhsh, Poczos, Salakhutdinov,
  and Smola]{Zaheer.2017}
M.~Zaheer, S.~Kottur, S.~Ravanbakhsh, B.~Poczos, R.~R. Salakhutdinov, and
  A.~Smola.
\newblock Deep {{Sets}}.
\newblock In \emph{Advances in {{Neural Information Processing Systems}}},
  volume~30. Curran Associates, Inc., 2017.

\bibitem[Zhang et~al.(2022)Zhang, Carpenter, Gelman, and Vehtari]{Zhang.2022}
L.~Zhang, B.~Carpenter, A.~Gelman, and A.~Vehtari.
\newblock Pathfinder: {{Parallel}} quasi-newton variational inference.
\newblock \emph{Journal of Machine Learning Research}, 23\penalty0
  (306):\penalty0 1--49, 2022.

\bibitem[Zhao et~al.(2021)Zhao, Dalmasso, Izbicki, and Lee]{Zhao.2021}
D.~Zhao, N.~Dalmasso, R.~Izbicki, and A.~B. Lee.
\newblock Diagnostics for {{Conditional Density Models}} and {{Bayesian
  Inference Algorithms}}.
\newblock In \emph{Proceedings of the {{Thirty-Seventh Conference}} on
  {{Uncertainty}} in {{Artificial Intelligence}}}, volume 161, pages
  1830--1840. PMLR, 2021.

\end{thebibliography}
\end{document}